\documentclass[preprint]{elsarticle}
\usepackage{graphicx}
\usepackage{epstopdf}
\usepackage{lineno,hyperref}
\usepackage{url}
\modulolinenumbers[5]

\journal{Signal Processing: Image Communication}

\begin{document}

\begin{frontmatter}

\title{Saliency maps on image hierarchies \tnoteref{mytitlenote}}

\tnotetext[mytitlenote]{\copyright 2015. This manuscript version is made available under the CC-BY-NC-ND 4.0 license \url{http://creativecommons.org/licenses/by-nc-nd/4.0/}}

\author{Veronica Vilaplana}
\address{Technical University of Catalonia (UPC)\\Department of Signal Theory and Communications\\ Barcelona, Spain}
%\footnote{\copyright 2015. This manuscript version is made available under the CC-BY-NC-ND 4.0 license }
% http://creativecommons.org/licenses/by-nc-nd/4.0/ }


%\ead{veronica.vilaplana@upc.edu}


\begin{abstract}
In this paper we propose two saliency models for salient object segmentation based on a hierarchical image segmentation, a tree-like structure that represents regions at different scales from the details to the whole image (e.g.  gPb-UCM, BPT).
The first model is based on a hierarchy of image partitions. The saliency at each level is computed on a region basis, taking into account the  contrast between regions. The maps obtained for the different partitions are then integrated into a final saliency map.
The second model directly works on the structure created by the segmentation algorithm, computing saliency at each node and integrating these cues in a straightforward manner into a single saliency map. We show that the proposed models produce high quality saliency maps. Objective evaluation demonstrates that the two methods achieve state-of-the-art performance in several benchmark datasets. 
\end{abstract}

\begin{keyword}
region-based saliency map, hierarchical segmentation, salient objects
\end{keyword}

\end{frontmatter}

%\linenumbers

\section{Introduction}

Visual saliency detection on images refers to the ability to select a certain subset of the visual information for further processing.

Most of the works related with saliency in computer vision follow one of following two approaches: prediction of eye fixation or prediction of salient objects.

Methods in the first  group try to predict scene locations where a human observer may fixate. Usually these models generate spotlight saliency maps. Within this group, Itti \emph{et al.}\cite{Itti-98pami} propose a biologically inspired model, using a center-surround operator across different scales, and  generating the saliency maps by normalization and summation of the feature maps. Other saliency models also work with the center-surround scheme using different features, like local contrast based on a fuzzy growth method in \cite{Ma-03icm}, or normalize the maps to highlight conspicuous parts and combine them with other maps using graph algorithms \cite{Harel-07anips}. For more details on fixation prediction the interested reader can refer to these two recent reviews \cite{Borji-13pami,Borji-13tip}.

All these methods are useful for predicting eye fixations and roughly detecting the most salient regions but the boundaries between salient objects and background are usually not accurately preserved.

Methods in the second group are salient object detection models, that aim to predict the most salient object in a scene and may afterwards segment the object. The problem may be understood as a particular segmentation task where the goal is to segment the salient foreground object from the background, as opposed to the generic segmentation problem where the image is partitioned into a set of regions which are homogeneous in some sense (for example in color).

Salient object detection also differs from supervised object detection techniques, which are aimed at finding particular categories like faces, cars, tables, etc. These methods achieve high performance, but the object of interest must be part of a predefined set of classes or categories from which training samples must be available. In salient object detection, there is no a priori information on the type of object. Such techniques are suitable in situations where possible targets and imaging conditions are not known in advance. They can be useful in applications like object detection and recognition, object co-segmentation, content-based image retrieval, photo collage, etc. A review of existing models of salient object segmentation is presented in Section \ref{sec:related_work}.

In this paper we propose two saliency models for salient object segmentation based on a hierarchical segmentation of the image. A hierarchical segmentation is a tree-like structure that represents regions at different scales, from the details to the whole image. We use two state-of-the-art techniques to produce the segmentation, Binary Partition Trees (BPT) \cite{Salembier-00tip,Vilaplana-08tip} and gPb-UCM \cite{Arbelaez-pami11}.  

Our first contribution is a saliency model based on a hierarchy of image partitions, a set of nested partitions where all the regions in a partition are included in the regions of a partition at a higher level. The saliency at each level is computed on a region basis, taking into account the local contrast between each region and its neighboring regions or the global contrast between each region and all the other regions in the partition. The maps obtained for the different partitions are then integrated into a final saliency map.  Our method generalizes previous approaches based on hierarchies \cite{Yan-cvpr13,vilaplana-icip13}, working with different hierarchical segmentation techniques, number of levels, region models and fusion strategies.

Our second contribution is a new model that directly works on the structure created by the hierarchical segmentation, computing saliency on each node and combining the results into a single saliency map. The formulation integrates local saliency cues in a straightforward and efficient manner.

The two models analyze saliency at several levels of detail in the hierarchy. An object that is different from its background will be salient at the scale at which it is represented completely or almost completely with a single region in the hierarchy (either a region in one partition or a node in the segmentation hierarchy). The integration of saliency cues computed at different scales allows the detection of salient objects of different sizes.
Objective and subjective evaluations show that the two proposed models achieve state-of-the-art performance in several benchmark datasets.

The rest of the paper is organized as follows. In Section \ref{sec:related_work} we review salient object detection methods.  In Section \ref{sec:hieseg} we describe the two hierarchies used in this work, BPT and gPb-UCM and the construction of hierarchies of partitions. In Section\ref{sec:hiemodels} we detail the computation of the saliency using the two proposed models. Experimental results, analysis and comparisons with other saliency models are presented in Section \ref{sec:experiments}, and conclusions are given in Section \ref{sec:conclusions}.

\section{Related work}
\label{sec:related_work}

In this section we briefly discuss existing models of salient object segmentation. Methods are organized into three groups taking into account the image representation they use: pixel, flat partition or hierarchy based.

\subsection{Pixel-based methods}
Within this first group, Liu \emph{et al.}\cite{Liu-07cvpr} formulate salient object detection as a foreground-background segmentation problem and learn a conditional random field to combine local, regional and global features for saliency detection: multi-scale contrast, center-surround histogram and color spatial-distribution, producing binary saliency masks. 

Goferman \emph{et al.}\cite{Goferman-10cvpr} model simultaneously local low-level cues, global cues, visual organization rules and high-level features to highlight salient objects along with their contexts. 

Achanta \emph{et al.}\cite{Achanta-09cvpr} propose a frequency-tuned method that defines pixel saliency using color differences from the average image color and a Gaussian blurred version of the image.

A limitation of these methods is that they tend to highlight edges around salient objects but do not obtain high saliency values for the complete objects. A review of other pixel-based models can be found in \cite{Borji-12eccv}.

\subsection{Partition-based methods}
Methods in this group are based on flat segmentation techniques whose output is a single flat partition, that is, a division of the image pixels into subsets (regions). 

Cheng \emph{et al.}\cite{Cheng-11cvpr} propose a method that simultaneously evaluates global contrast differences and spatial coherence.
They first segment the image into regions using a graph-based segmentation technique. The saliency of a region is calculated using a global contrast score measured by the region's contrast and spatial distances to other regions in the image.

Perazzi \emph{et al.}\cite{perazzi-cvpr12} segment the image using an adaptation of SLIC superpixels, and compute two measures of contrast that rate the uniqueness and spatial distribution of superpixels. The two maps are combined and the saliency of each pixel is defined as a weighted combination of the saliency of its surrounding regions.

Zhu \emph{et al.}\cite{Zhu-cvpr14}  construct an undirected graph by connecting adjacent SLIC superpixels and formulate the salient object detection problem as the optimization of the saliency values of the superpixels. The cost function is defined in terms of background and foreground weights that measure the boundary connectivity and the contrast of superpixels.

Margolin \emph{et al.}\cite{margolin-cvpr13} integrate pattern and color distinctness using Principal Component Analysis to find components that best explain the variance in the data. They apply PCA to represent the set of patches of an image and use this representation to estimate distinctness, combined with standard techniques for color uniqueness and organization priors.

Other methods that follow a partition-based approach are \cite{liu-optlet13,YangZLRY13}.

A limitation of using flat partitions is that in order to represent objects at different scales the parameters of the algorithm have to be tuned for the result to be finer or coarser (having more or less regions). It is difficult to determine which resolution is the best, so methods typically use an over-segmentation of the image where objects are represented by several regions. As a consequence, saliency maps may fail to  highlight complete objects.

\subsection{Hierarchy-based methods}
Techniques in this group make use of hierarchies of partitions or hierarchical representations of images or saliency maps.

Yan \emph{et al.}\cite{Yan-cvpr13} use a hierarchy of three levels. They first generate an over-segmentation of the image by a watershed-like method and then apply an iterative merging process based on the scale and the color of the regions. Saliency maps are created for the three levels and merged into a final map by hierarchical inference. This inference is equivalent to applying a weighted average to all single saliency maps with optimal weights for each region.

Liu \emph{et al.}\cite{ZLiu-tip14} propose a framework  termed as saliency tree. They first over-segment the image using gPb-UCM and generate an initial saliency map based on color contrast and spatial sparsity of regions and object prior. Next, a BPT is created using the saliency of the regions to define the merging criterion and merging order. Based on the analysis of the tree structure a final pixel-based saliency map is derived.

\section{Hierarchical segmentation}
\label{sec:hieseg}
Hierarchical segmentation techniques contain partitions of the image at different levels of detail in a single structure. They are usually represented by means of a tree, where the root represents the whole image, the leaves are the regions at the highest level of detail, and a parent node represents the merging
of all their children regions. In this work we use two different types of hierarchies: Binary Partition Trees (BPT) and gPb-UCMs.

\subsection{Binary Partition Tree}
\label{ssec:bpt}
The BPT~\cite{Salembier-00tip} is a structured representation of the image regions that can be obtained from an initial partition using a simple bottom-up merging approach. Starting from a given partition (with any number of regions; we may even assume that each pixel or flat zone is a region), the algorithm proceeds iteratively by (1) computing a similarity measure for all pairs of neighboring regions, (2) selecting the most similar pair of regions and merging them into a new region and (3) updating the neighborhood and the similarity measures. The algorithm iterates steps (2) and (3) until all regions are merged into a single region. The BPT stores the whole merging sequence from the initial partition to the one-single region representation. The leaves in the tree are the regions in the initial partition. A merging is represented by creating a parent node (the new region resulting from the merging) and linking it to its two children nodes (the pair of regions that are merged).
In this work, the similarity measure between two regions is the Euclidean distance between the mean colors of the regions, weighted by the size of the regions. Area weighting is used to encourage the merging of small and semantically unimportant regions before larger regions are merged \cite{Vilaplana-08tip}. The color space used is \emph{CIE Lab}, because of the perceptual nature of color metrics in this space. 

Single partitions ares commonly created by sampling the BPT merging sequence at different points.

\subsection{gPb-UCM} 
\label{ssec:ucm}
This techique \cite{Arbelaez-pami11} builds an Ultrametric Contour Map (UCM) on the globalized Probability of boundary (gPb) contour detector. The gPb method starts with a local edge extraction procedure which has been optimized using learning techniques. The results of this edge extraction step are then used as input to a spectral partitioning procedure which globalizes the results using Normalized Cuts. This globalization stage helps to focus attention on the most salient edges in the scene. Then, the Oriented Watershed Transform is used for constructing a set of initial regions from an oriented contour signal. Finally, using an agglomerative clustering procedure, a hierarchy is created as a result of the iterative fusion of the most similar regions. Here the similarity is measured by the weakness of the contour between the regions. The hierarchy is represented by an Ultrametric Contour Map, a real-valued image obtained by weighting each boundary by its scale of disappearance.

Single partitions are commonly created by thresholding the UCM at a certain value or scale. The higher the threshold, the coarser the partition.

\subsection{Hierarchies of image partitions}
\label{ssec:hierarchy}

In this section we introduce some notation and describe how the hierarchies of image partitions used in our first saliency model are constructed.

Formally, a partition $P$ of an image $I$ is a set of disjoint regions $P=\{{R_i^P}\}_{i=1...N}$ so that the union of the regions is the whole image $I=\cup_{i=1}^{N}R_i^P$. An ordering relation between two partitions can be defined: a partition $P$ is included in a partition $Q$ if every region $R_j^P$ in $P$ is completely included in a region $R_i^{Q}$ in $Q$.

Let $H$ be a set of partitions of an image. $H$ is a hierarchy of nested partitions if it is possible to define an inclusion order between any pair of elements in $H$. Formally, a hierarchy of nested partitions is a set $H=\{P_1,P_2,...,P_l\}$ so that all the regions of a partition $P_m=\{R_i^m\}_{i=1,...,N_m}$ are included in the regions of the partition $P_n=\{R_j^n\}_{j=1,...,N_n}$ , for $n>m$, with $N_n<N_m$ and $R_i^m \subseteq R_j^n$ or $R_i^m \cap R_j^n=\emptyset$.

The hierarchy can be created with any segmentation algorithm capable of generating nested partitions. In this work we use two techniques: Binary Partition Trees (BPT) and gPb-UCMs.

\begin{figure}[h]
\centering
\begin{tabular}{ccccc}
\includegraphics[width=1.6cm]{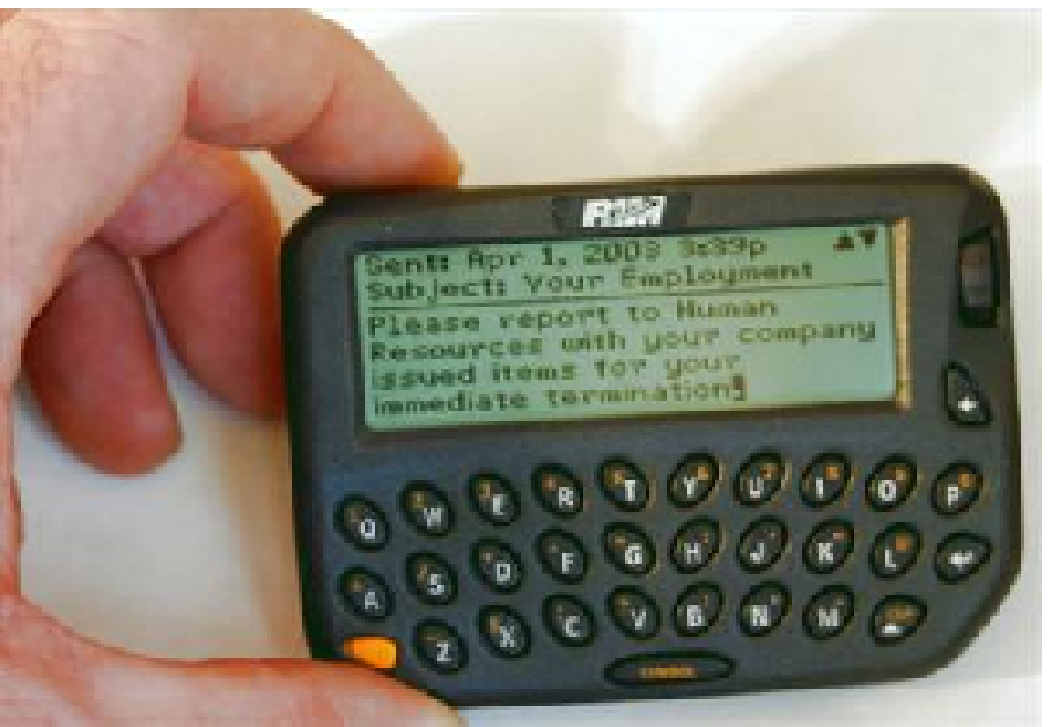} \\
\includegraphics[width=1.6cm]{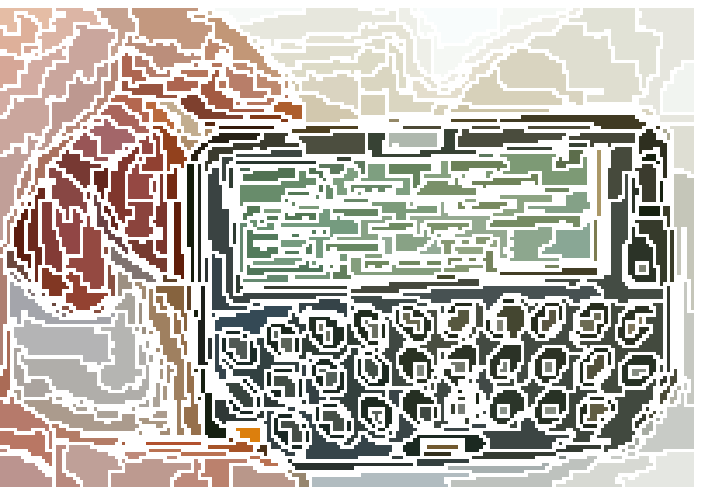} &
\includegraphics[width=1.6cm]{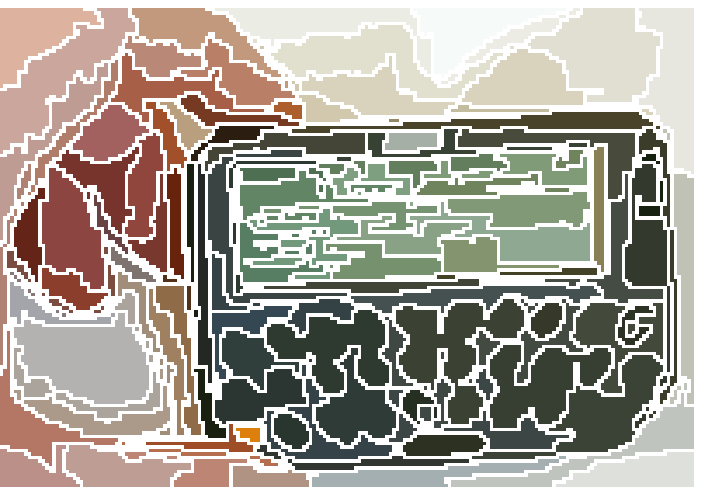} &
\includegraphics[width=1.6cm]{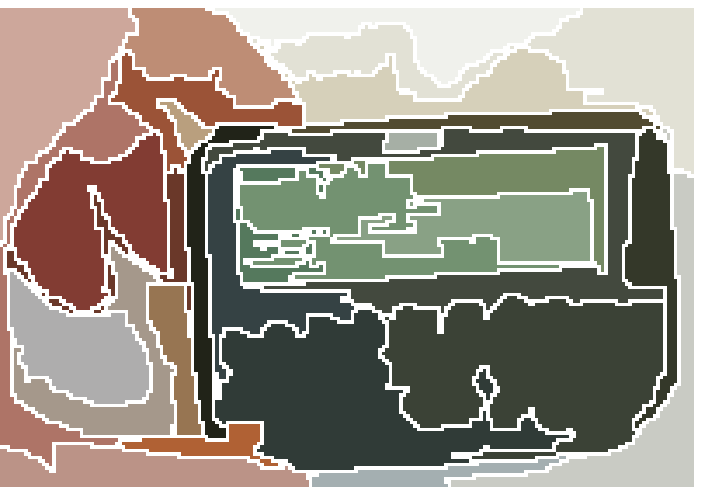} &
\includegraphics[width=1.6cm]{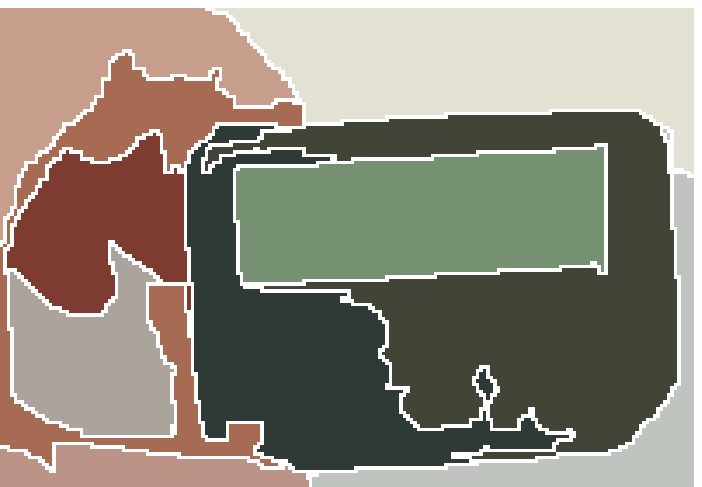} &
\includegraphics[width=1.6cm]{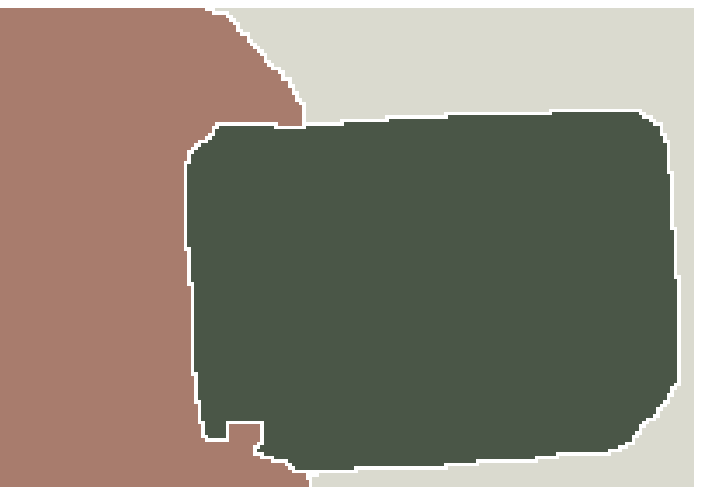} \\
\includegraphics[width=1.6cm]{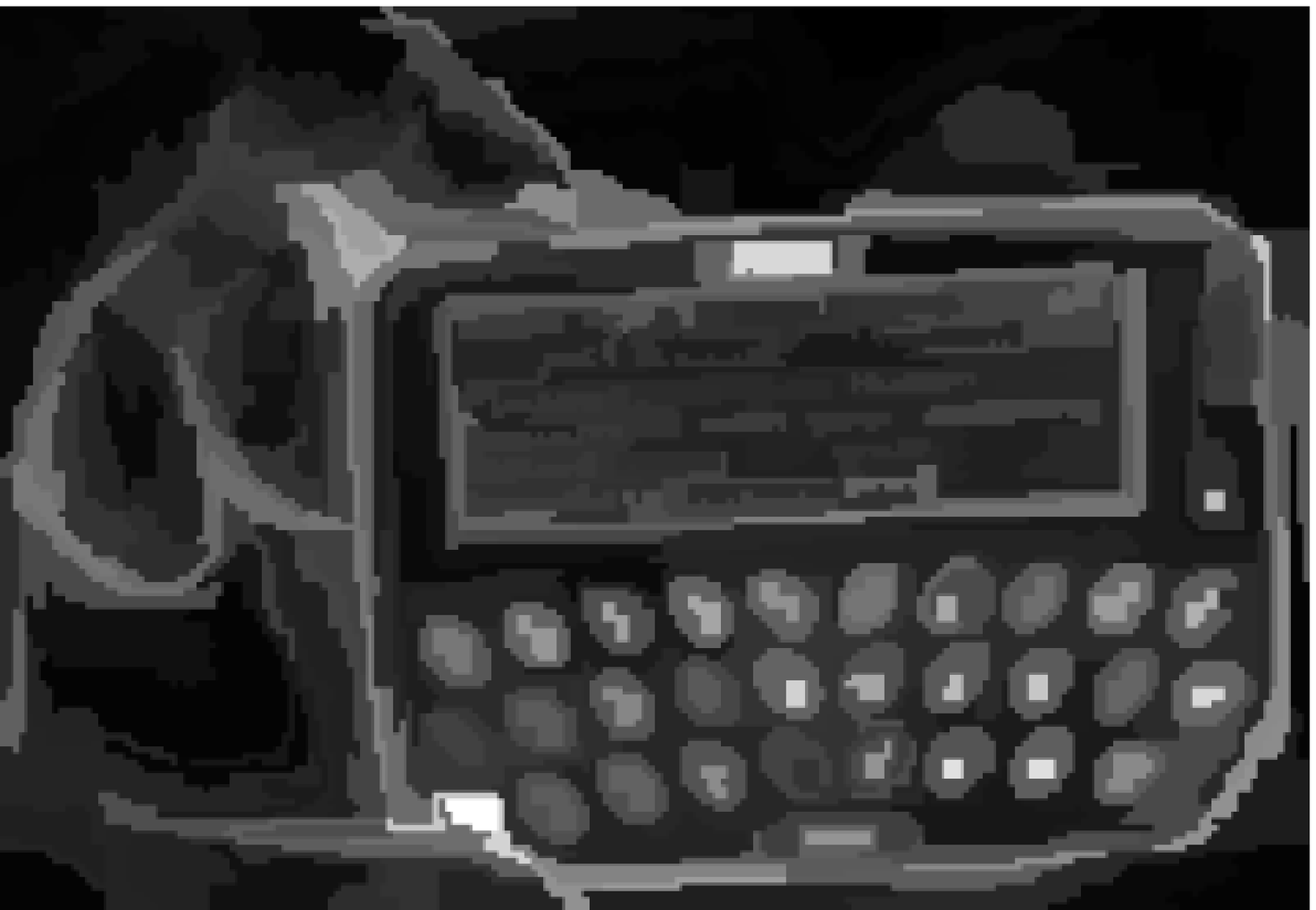} &
\includegraphics[width=1.6cm]{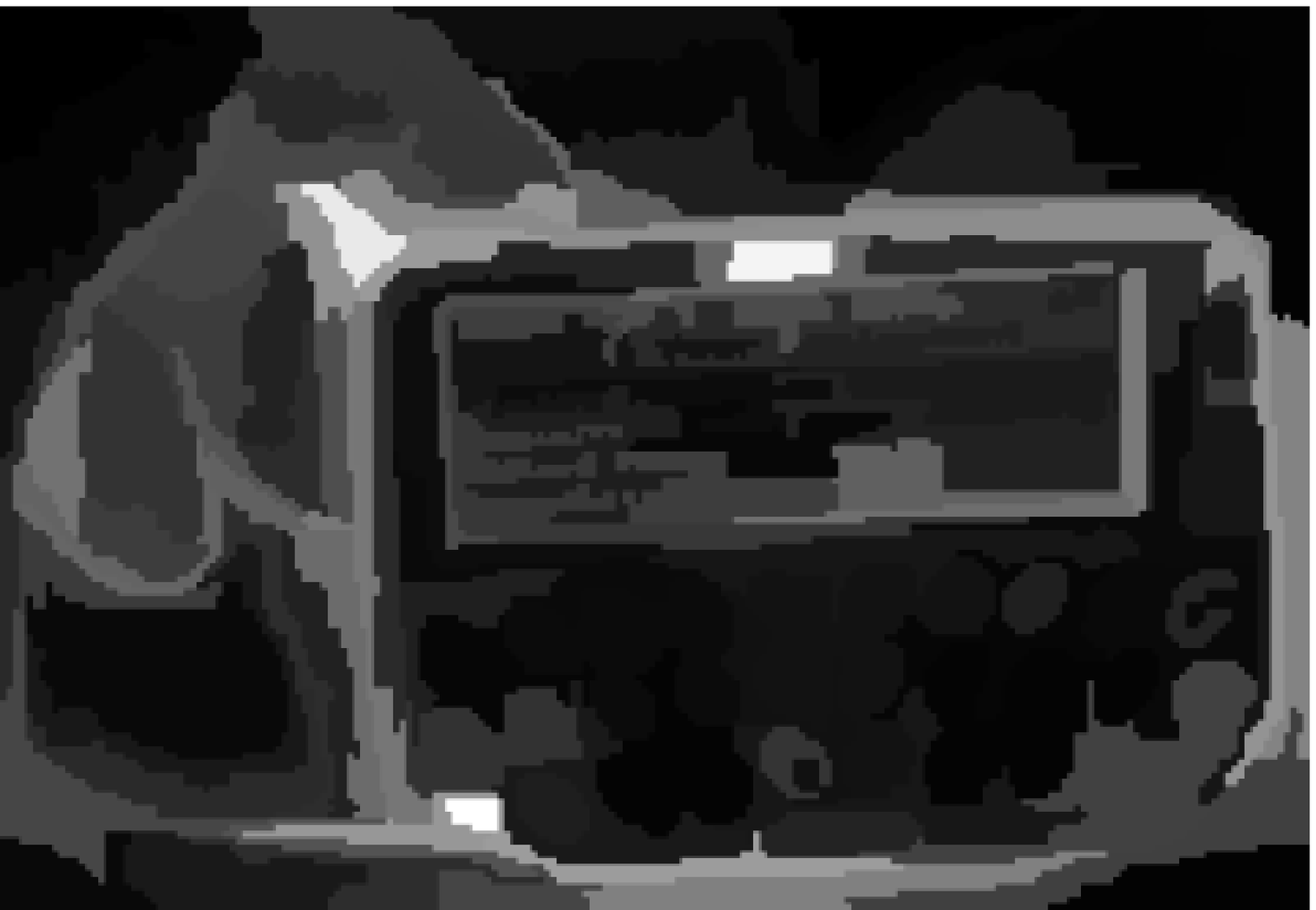} &
\includegraphics[width=1.6cm]{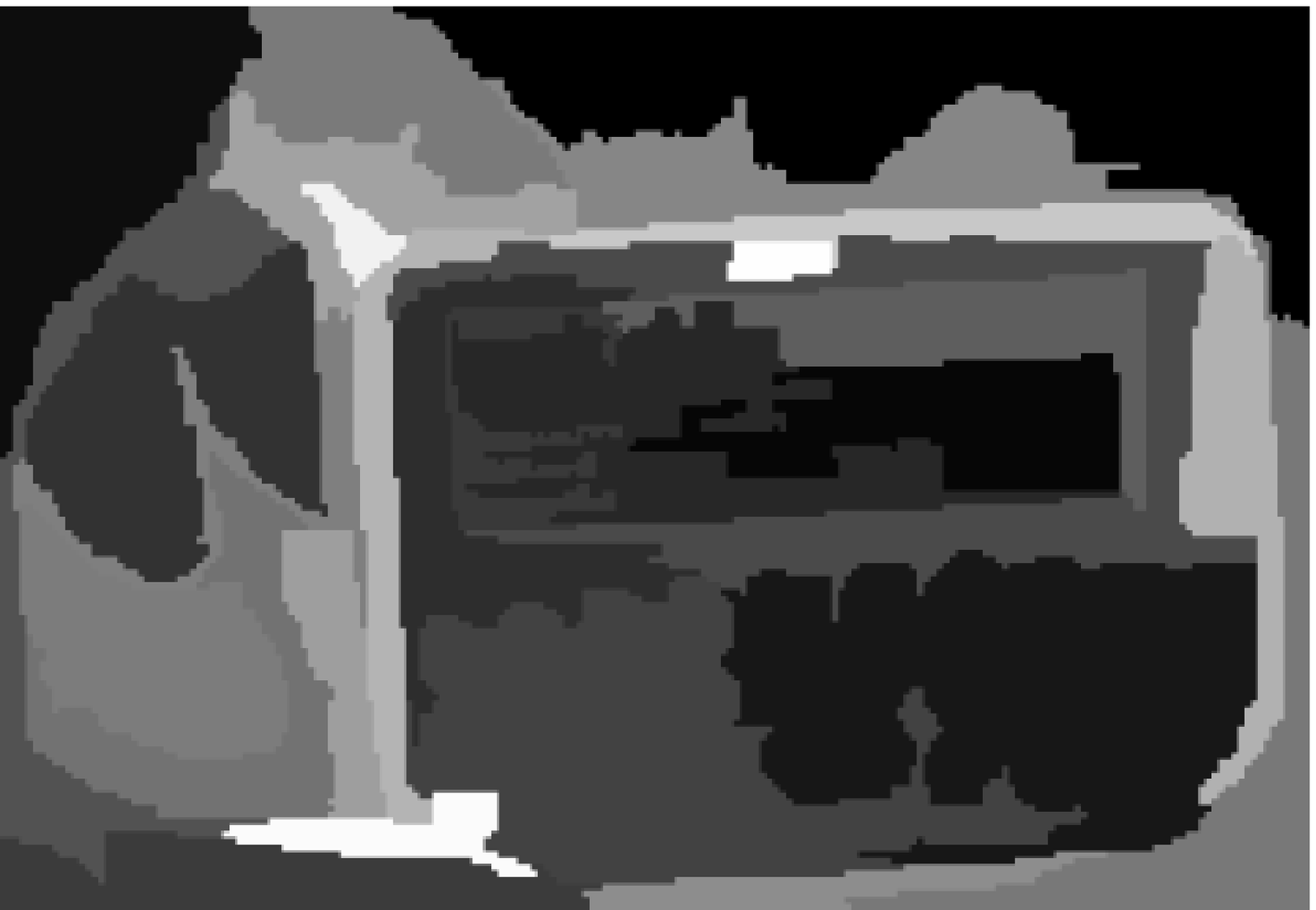} &
\includegraphics[width=1.6cm]{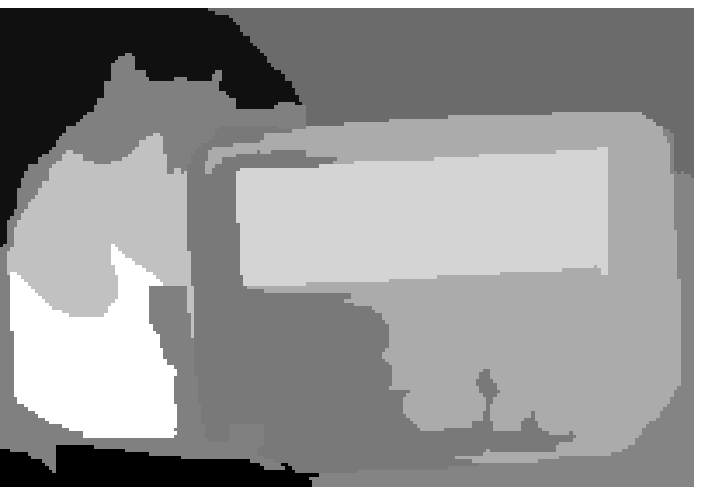} &
\includegraphics[width=1.6cm]{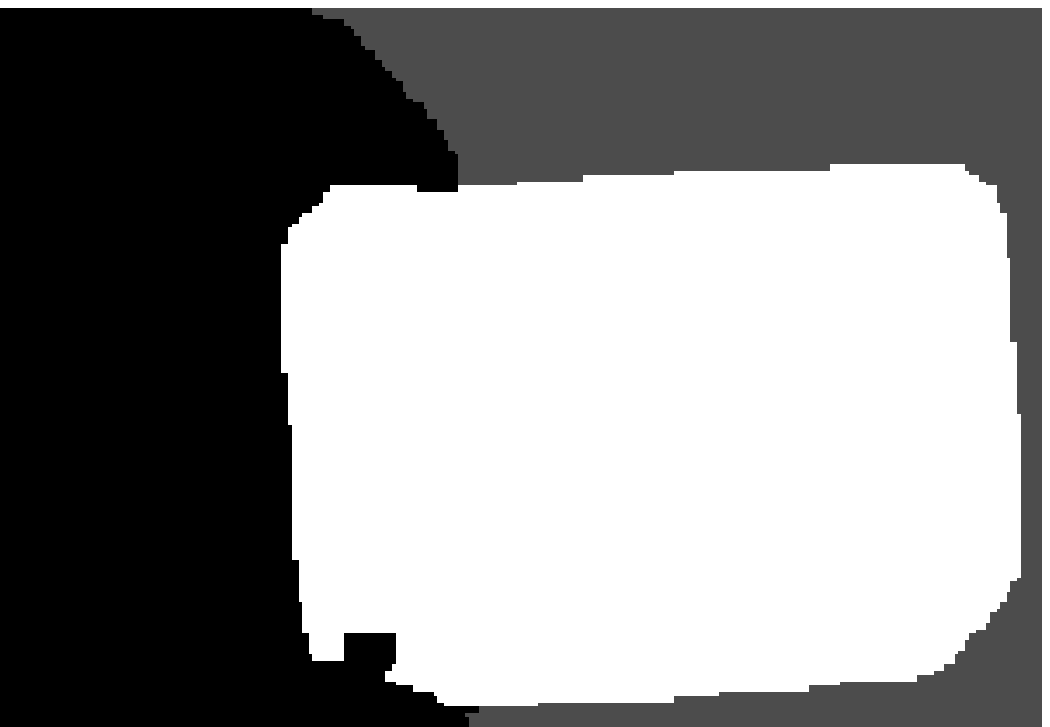} \\
(a) 300 & (b) 100 & (c) 30 & (d) 10 & (e) 3\\
\end{tabular}
\caption{Hierarchy of nested partitions for the original image (first row) generated with BPT. Partitions with 300 (a), 100 (b), 30 (c), 10 (d) and 3 (e) regions (second row) and their saliency maps (third row) based on local contrast. The regions in the partitions are represented with their mean color, and the saliency maps are linearly normalized to [0,255] for visualization purposes.} \label{fig:hierarchyBPT}
\end{figure}

When using Binary Partition Trees, the set of nested partitions is created by sampling the BPT merging sequence at different points. The top row in Figure~\ref{fig:hierarchyBPT} shows an example of a hierarchy of five partitions, where each region is represented by the mean color computed on the region pixels, with contours in white. The number of regions in the five partitions, from left to right, is 300, 100, 30, 10 and 3, respectively\cite{vilaplana-icip13}. The example shows how small objects like the calculator keys or letters in the keys are represented as complete regions in the lower levels whereas larger objects like the hand or the calculator appear as single or as a few regions in the upper levels.

For gPb-UCM, the set of nested partitions used for building the saliency maps is created by extracting the partitions at different contour strengths. The base level of the set corresponds to an over-segmentation of the image, while the upper levels are coarser partitions. 

\begin{figure}[h]
\centering
\begin{tabular}{ccccc}
\includegraphics[width=1.6cm]{Figures/hierarchy/0_16_16940} \\
\includegraphics[width=1.6cm]{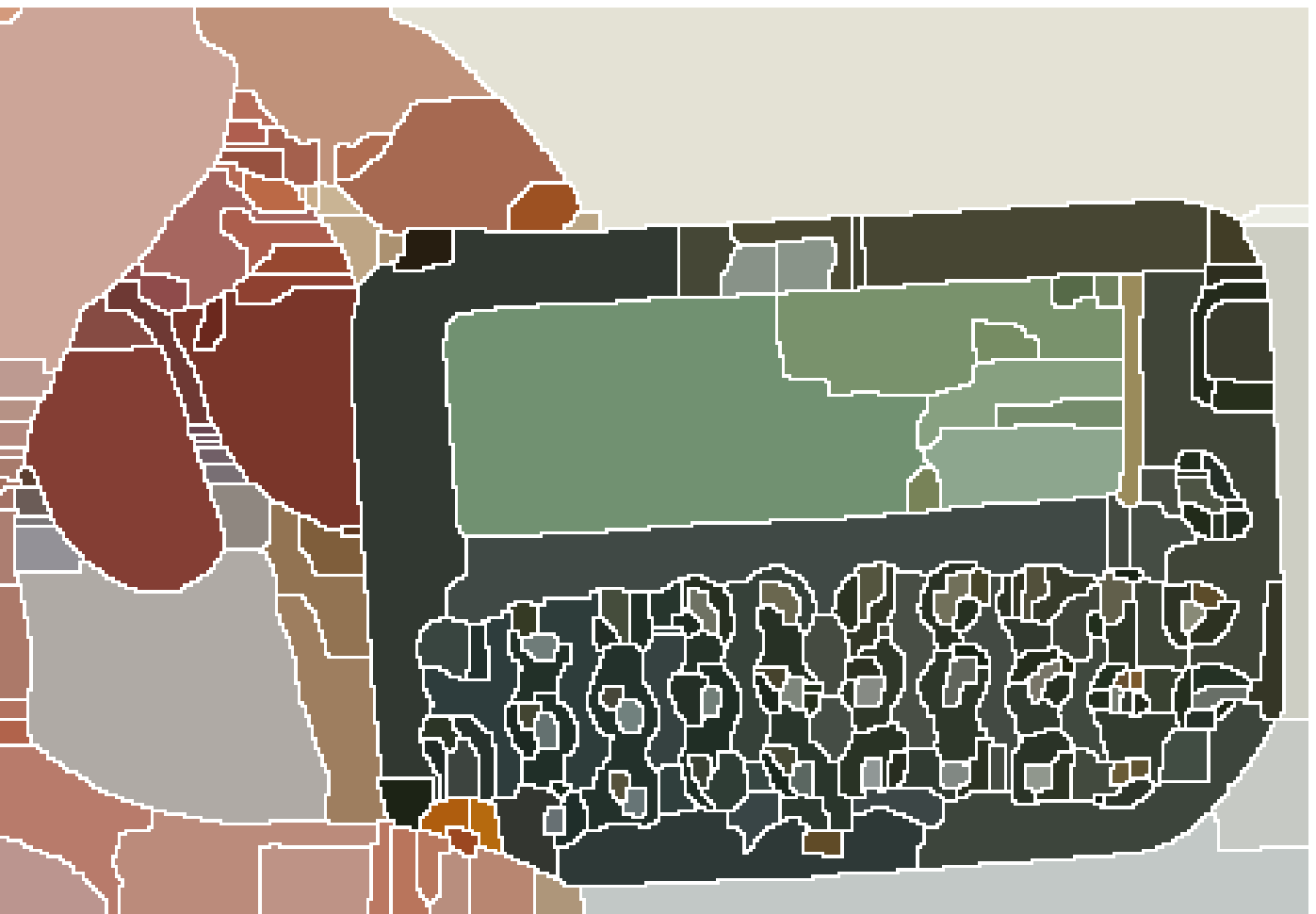} &
\includegraphics[width=1.6cm]{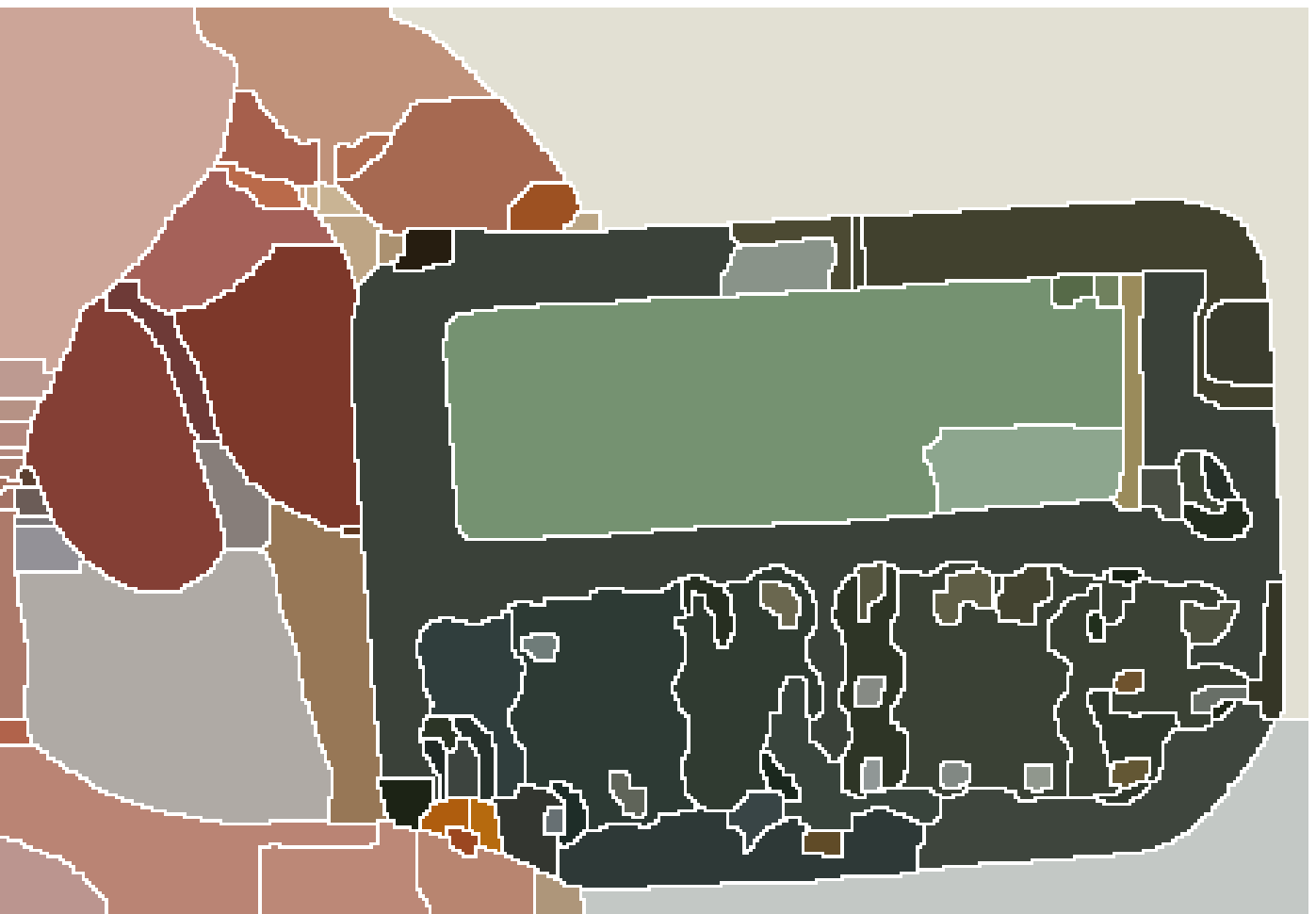} &
\includegraphics[width=1.6cm]{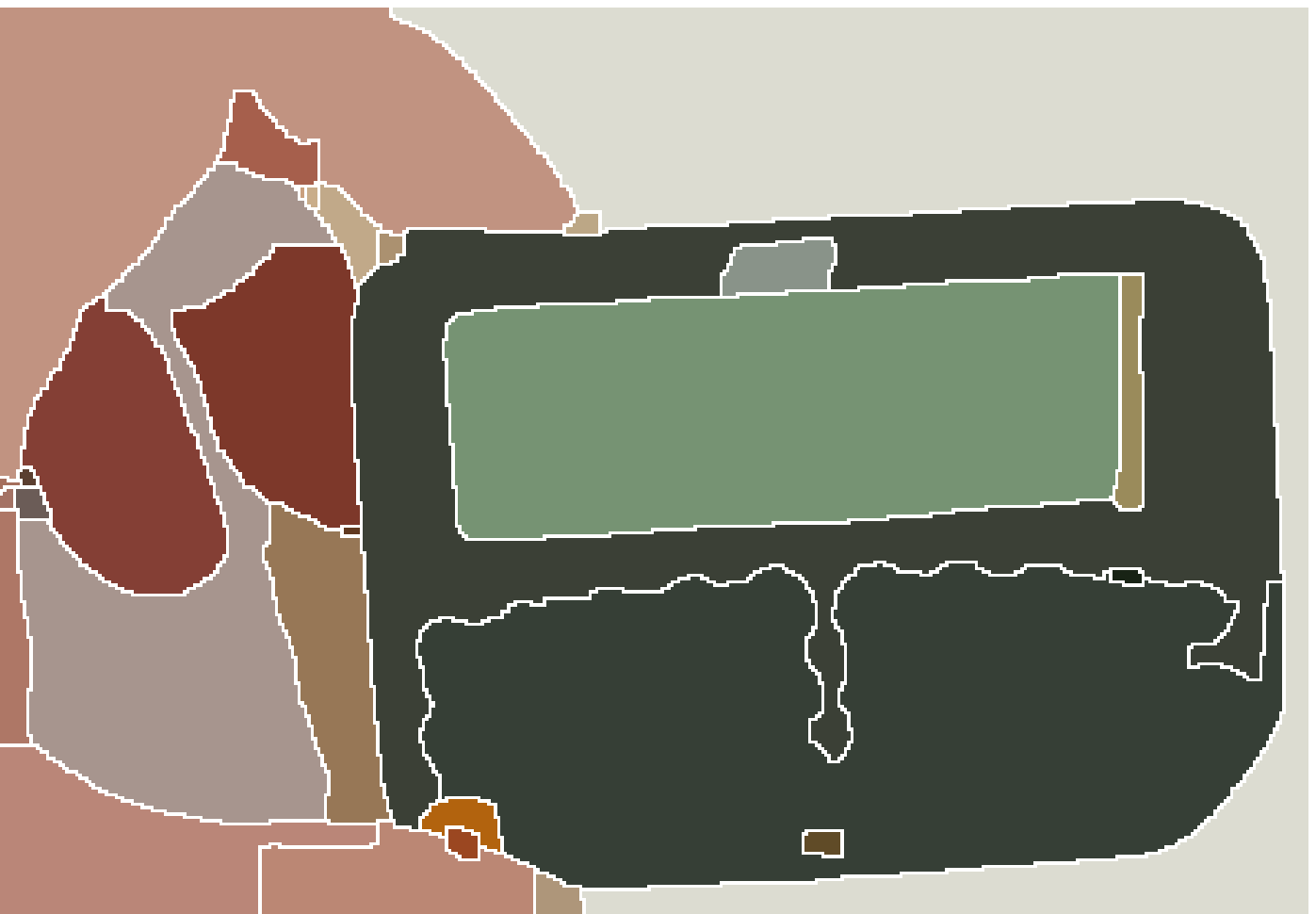} &
\includegraphics[width=1.6cm]{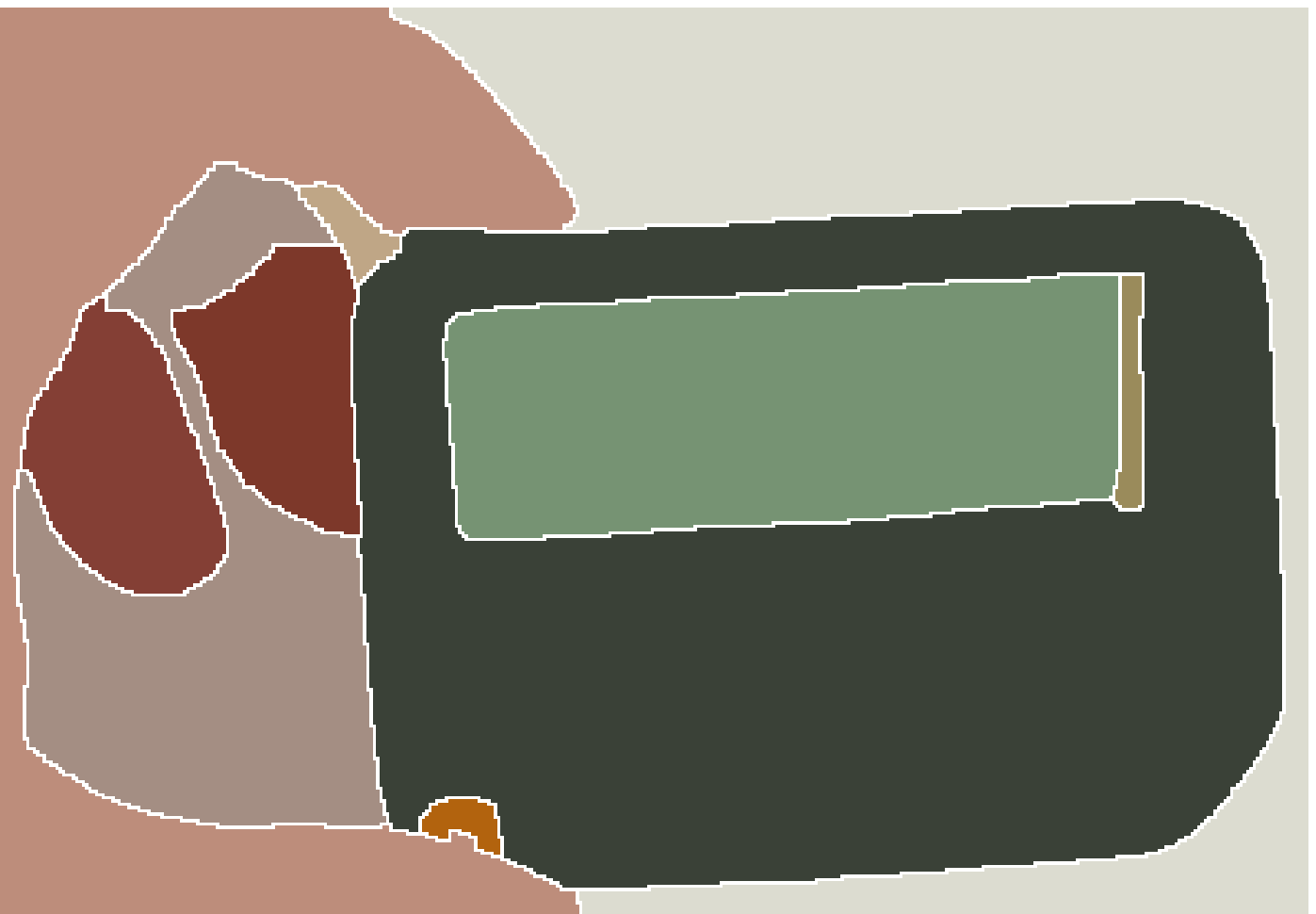} &
\includegraphics[width=1.6cm]{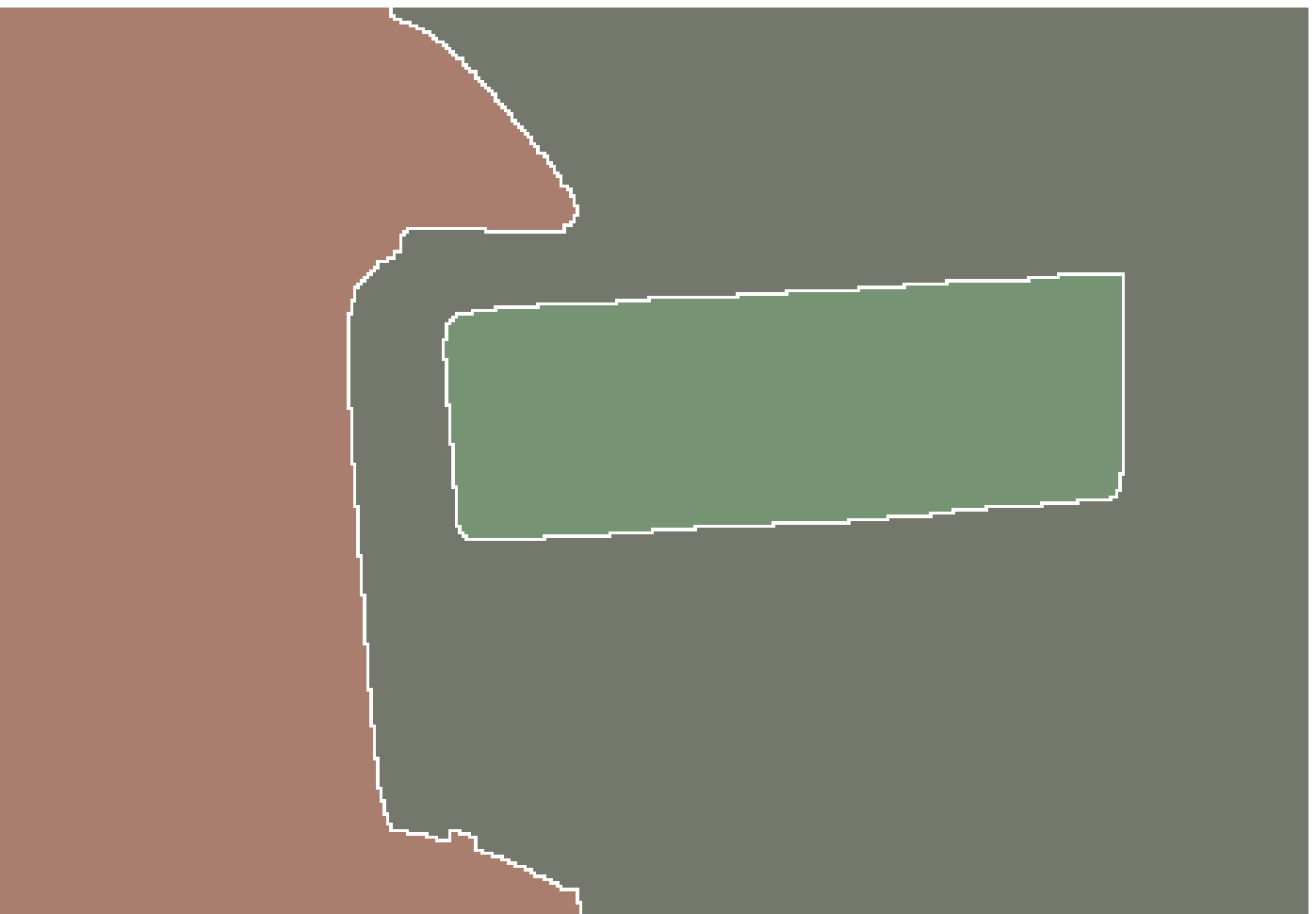} \\
\includegraphics[width=1.6cm]{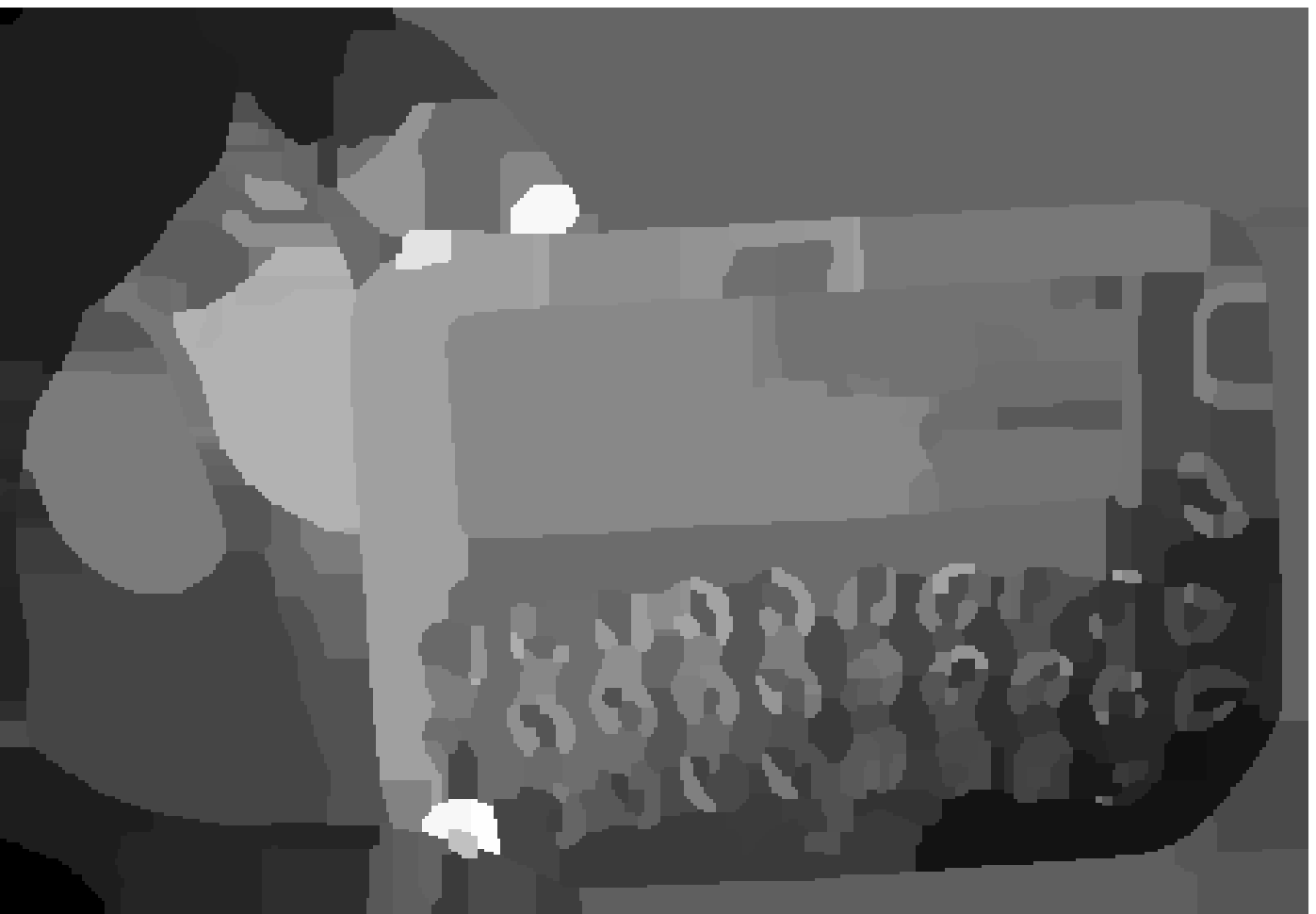} &
\includegraphics[width=1.6cm]{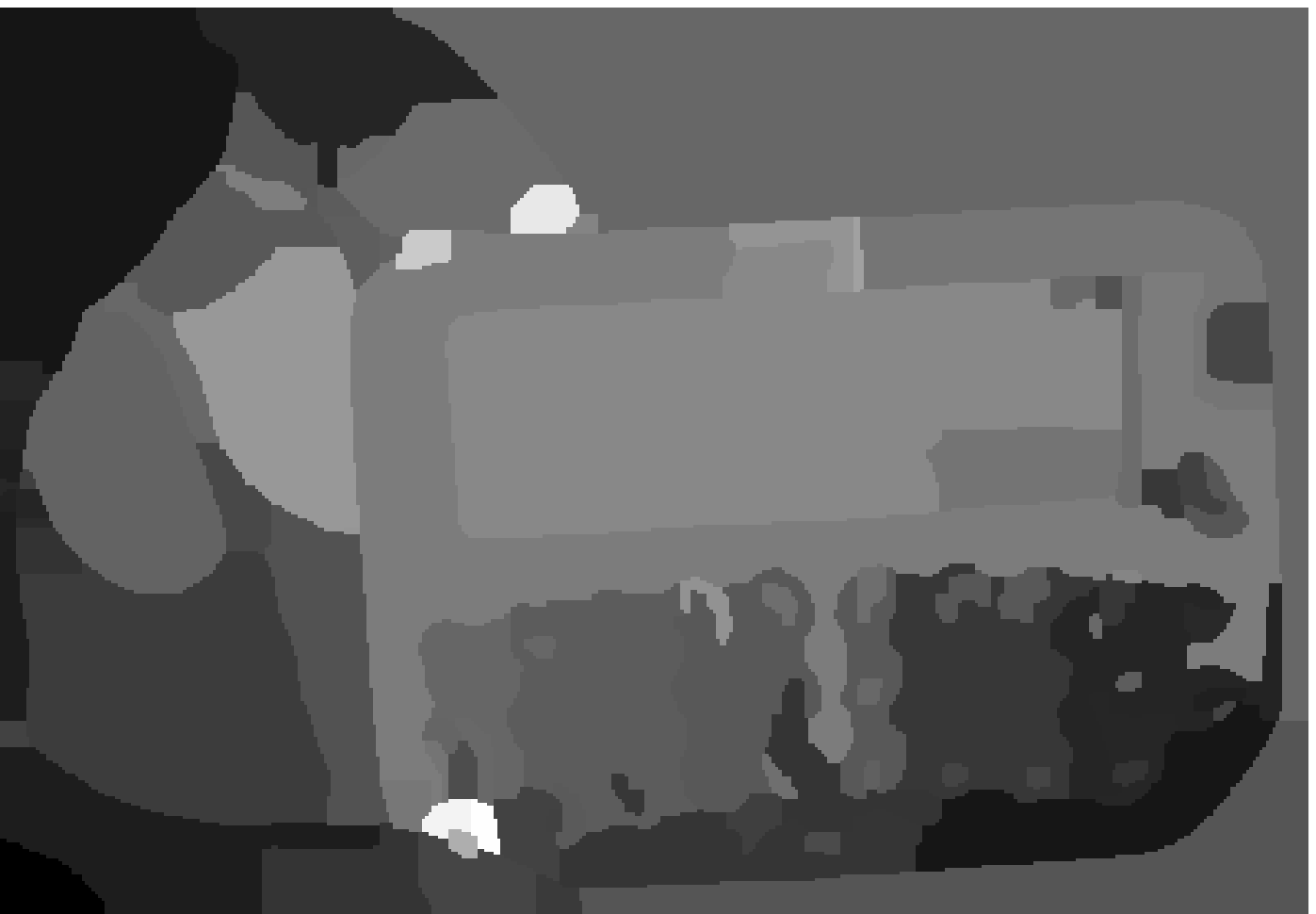} &
\includegraphics[width=1.6cm]{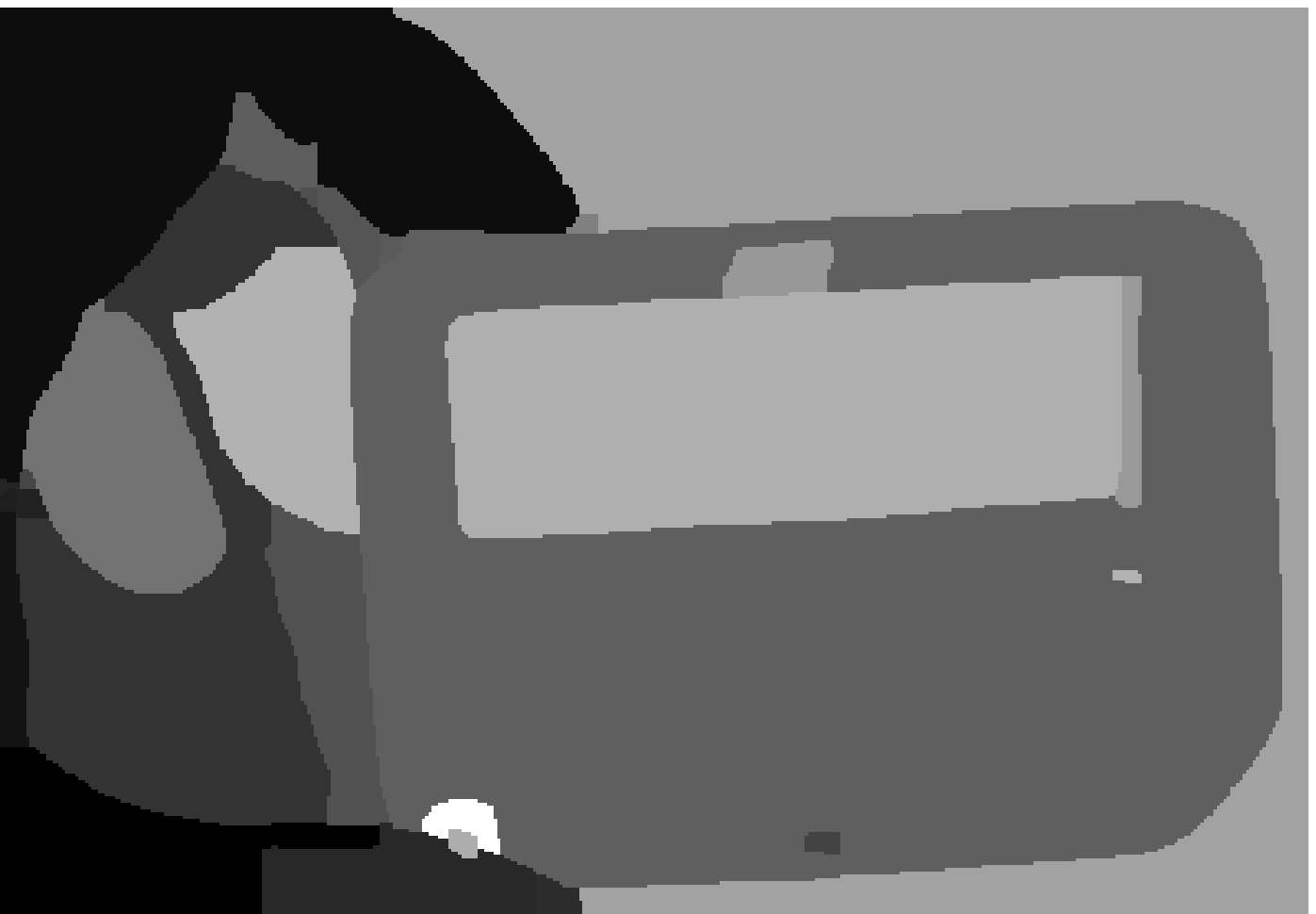} &
\includegraphics[width=1.6cm]{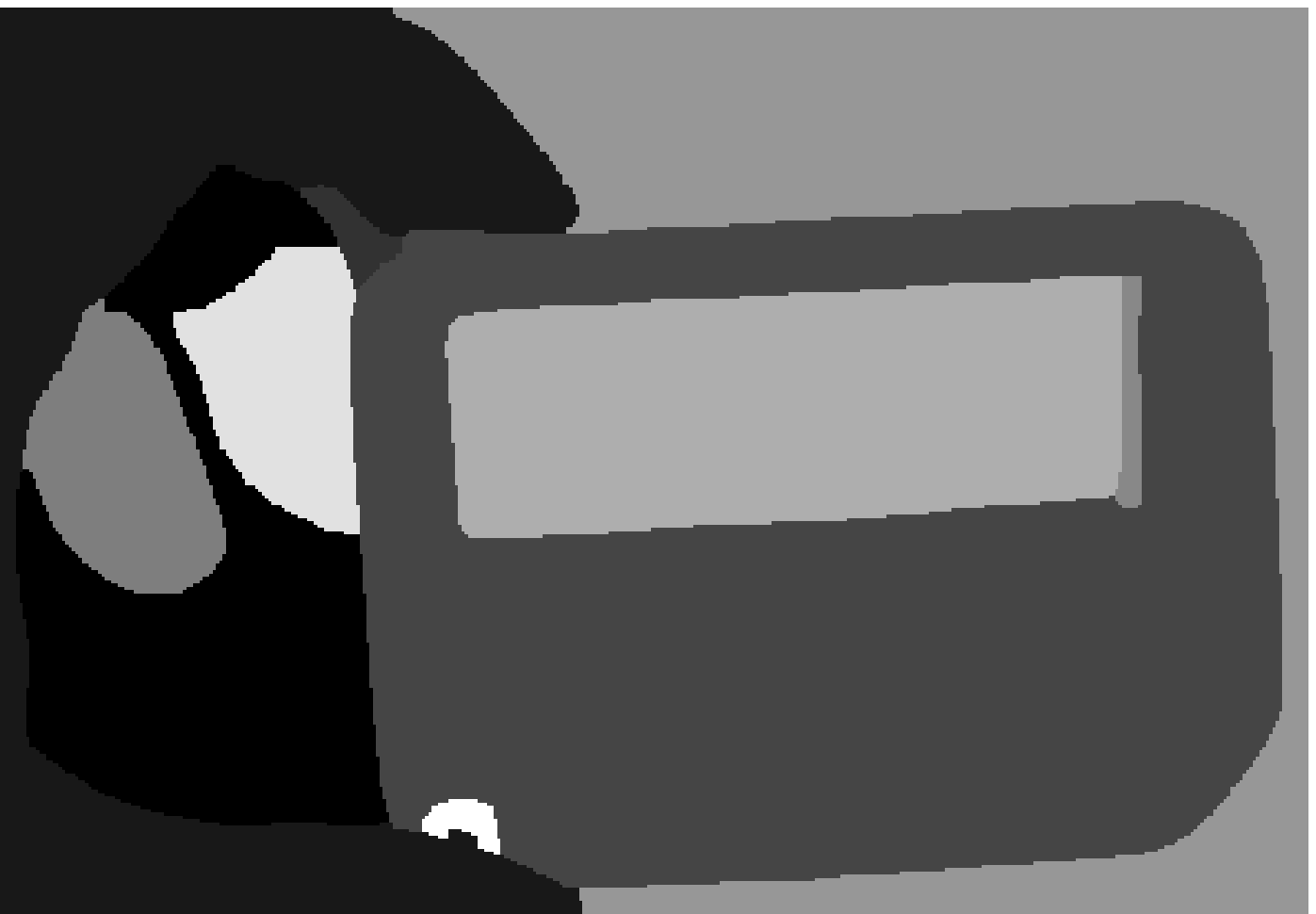} &
\includegraphics[width=1.6cm]{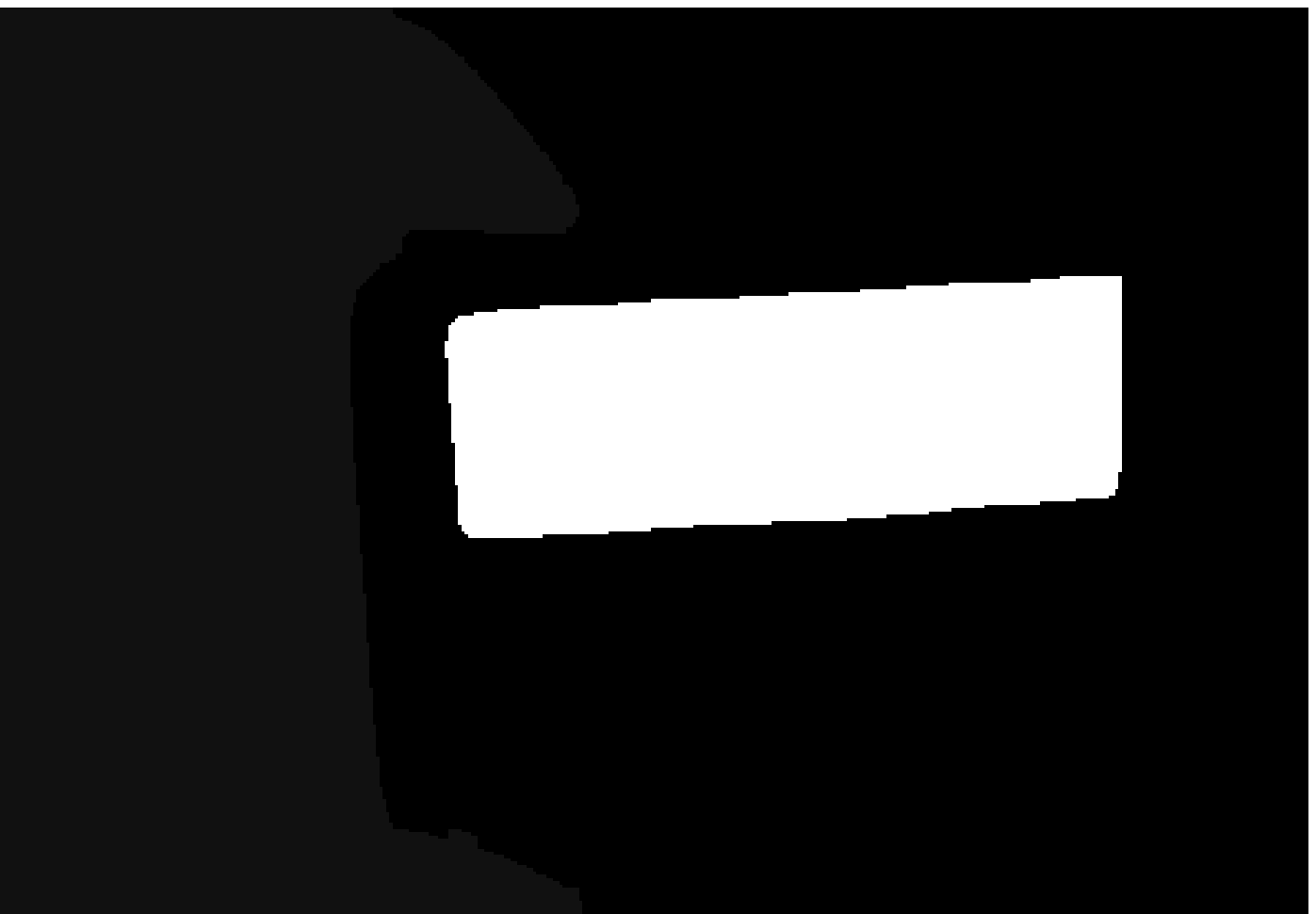} \\
\includegraphics[width=1.6cm]{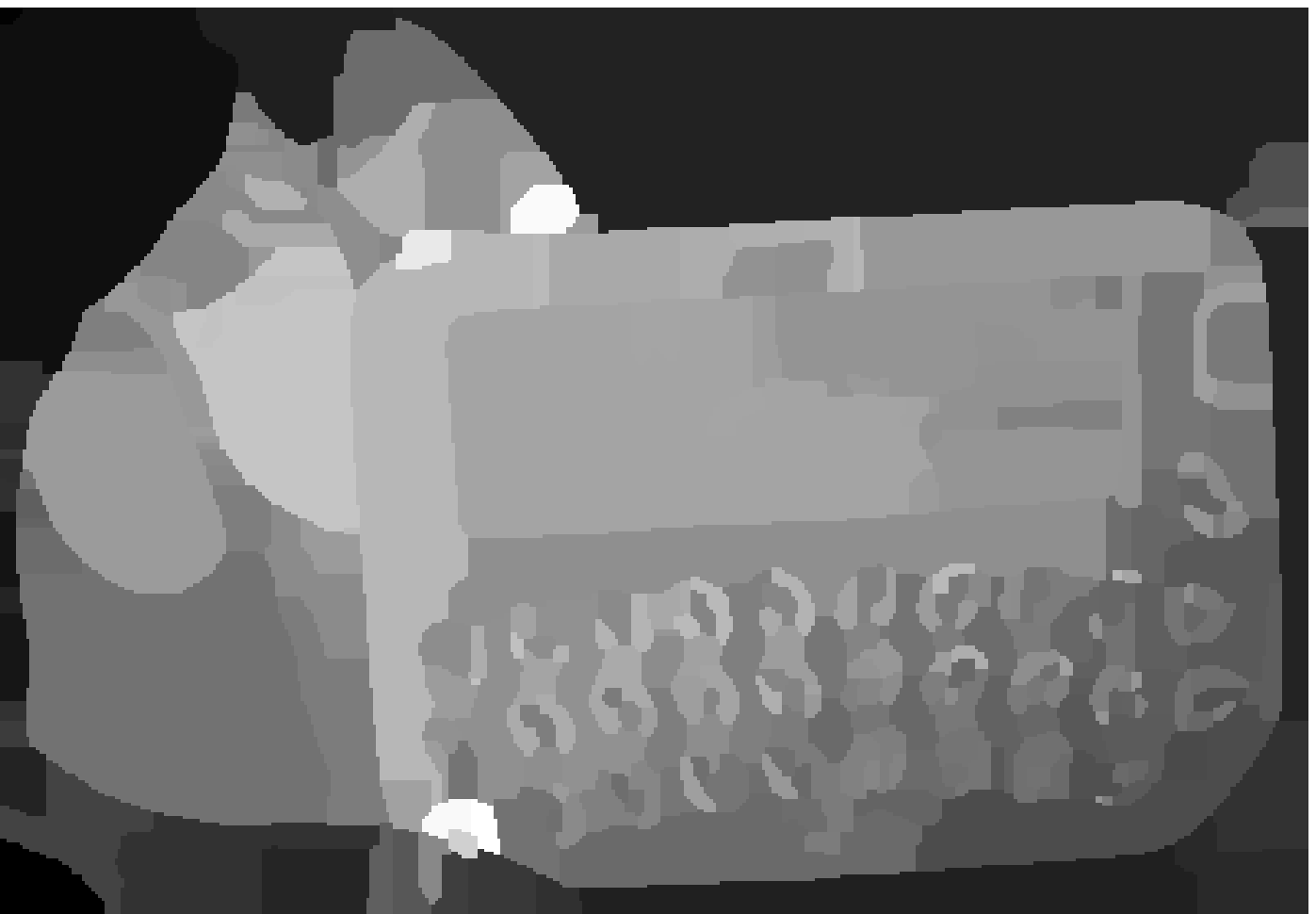} &
\includegraphics[width=1.6cm]{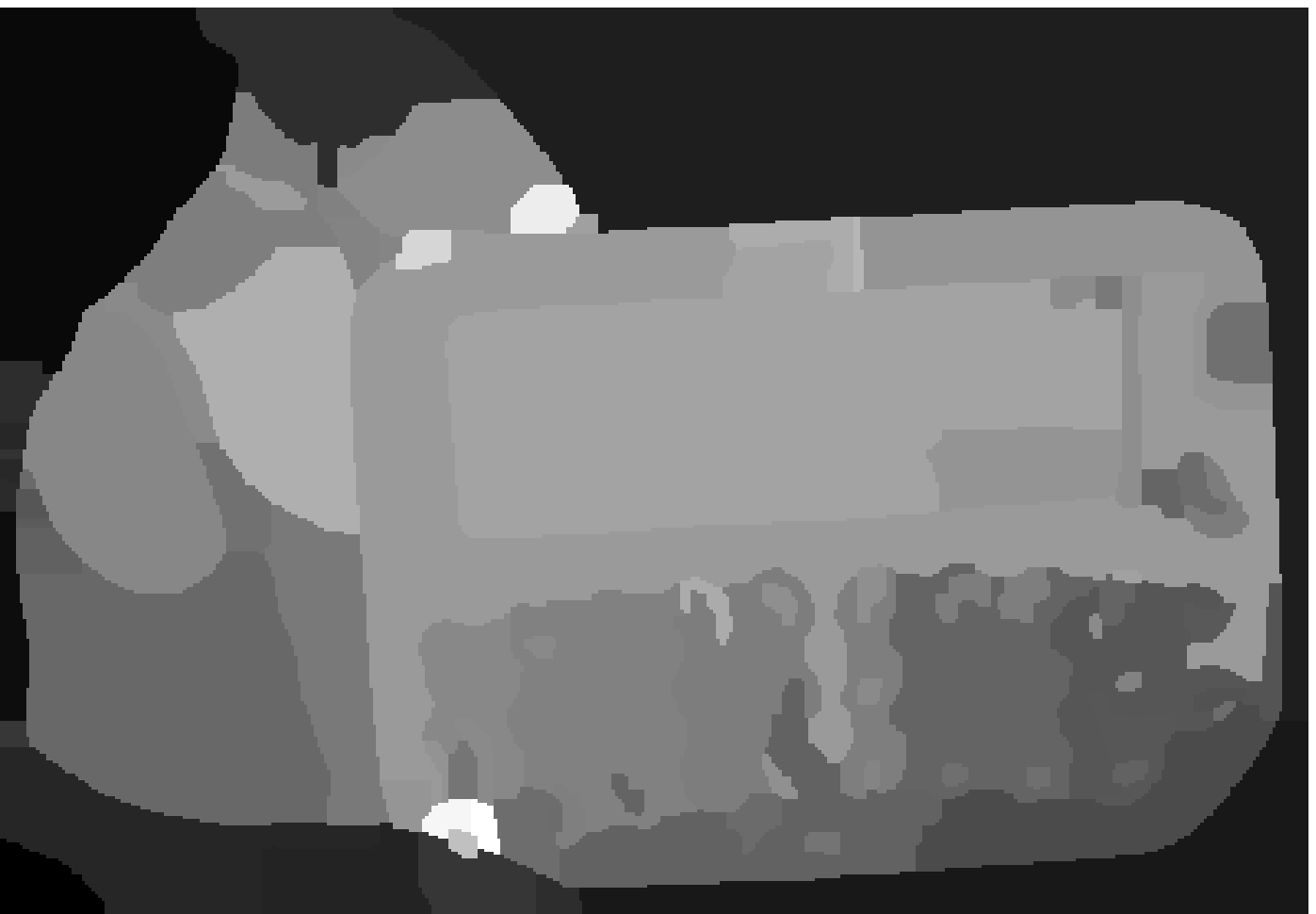} &
\includegraphics[width=1.6cm]{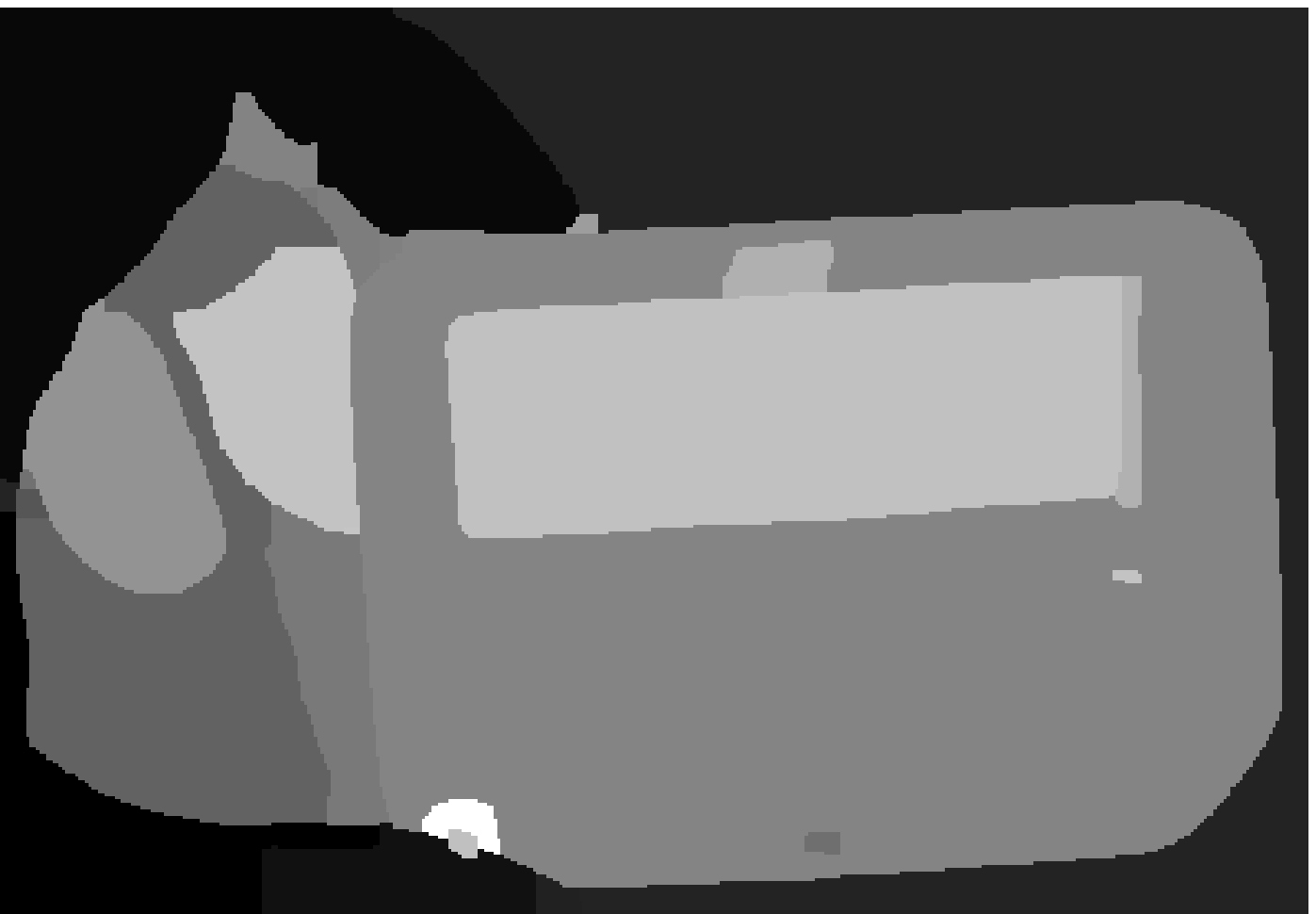} &
\includegraphics[width=1.6cm]{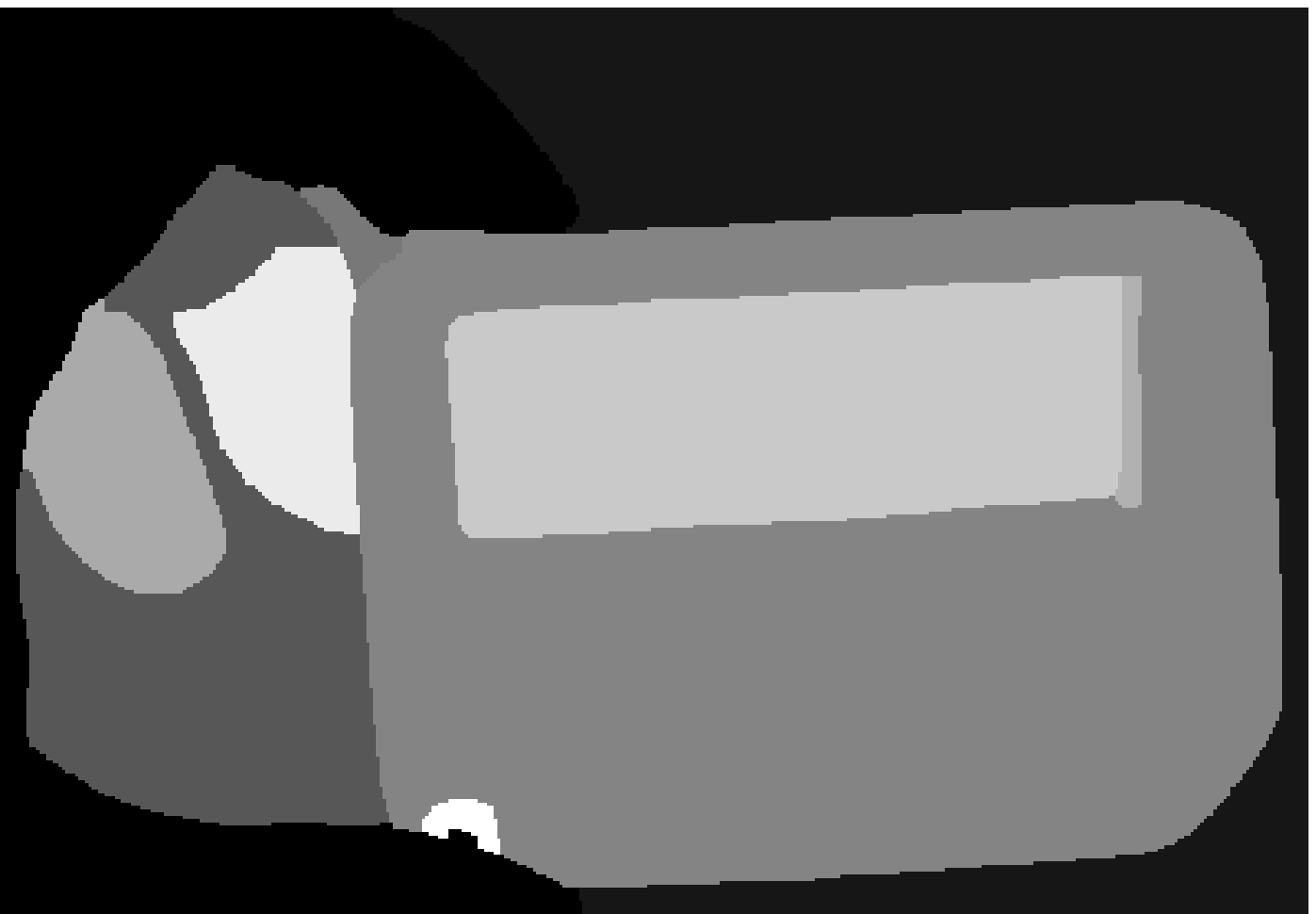} &
\includegraphics[width=1.6cm]{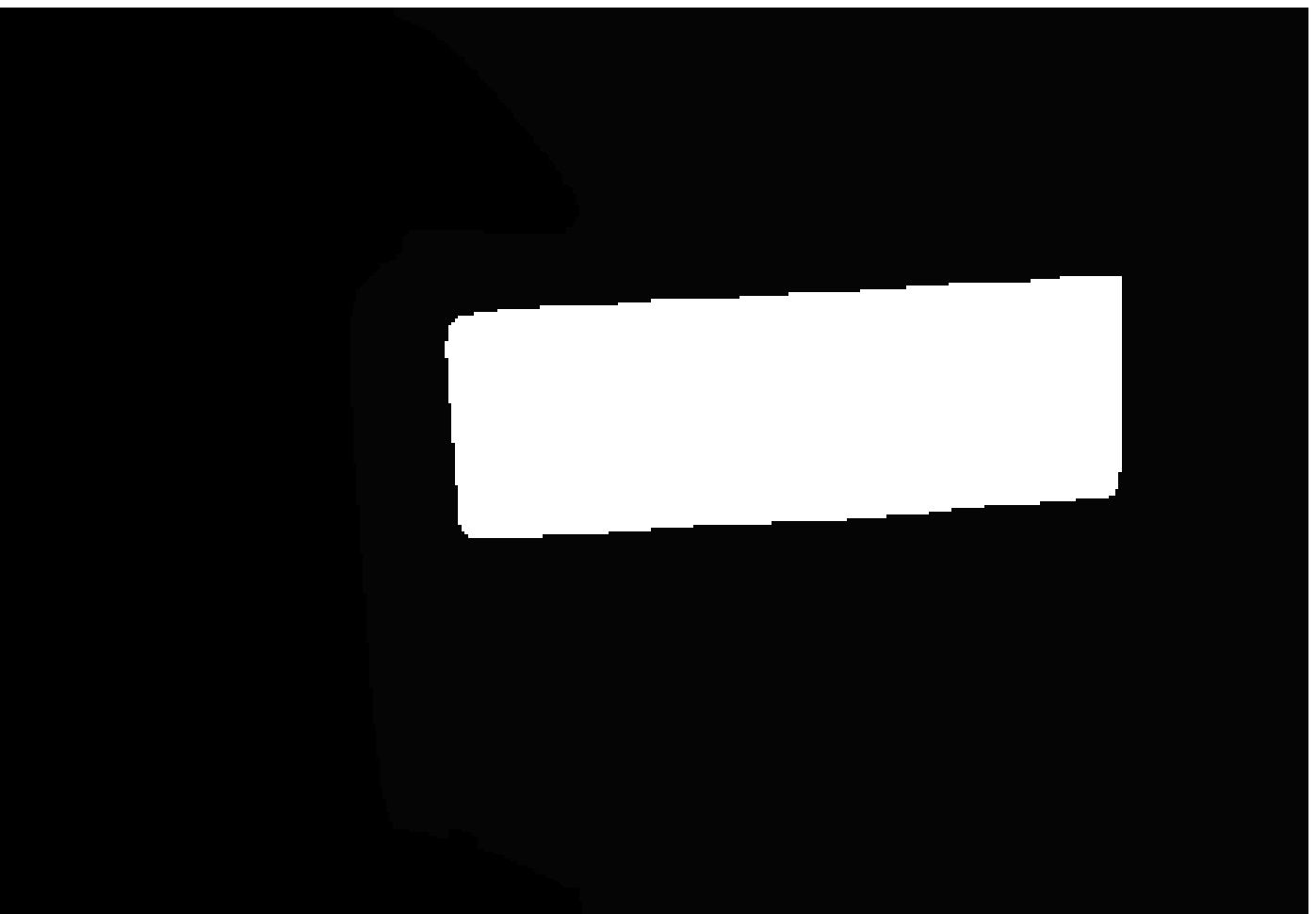} \\
(a) 300 & (b) 100 & (c) 30 & (d) 10 & (e) 3\\
\end{tabular}
\caption{Hierarchy of nested partitions for the original image (first row) generated with gPb-UCM. Partitions with 300 (a), 100 (b), 30 (c), 10 (d) and 3 (e) regions (second row) and their saliency maps generated using global contrast (third row) and global contrast with spatial prior (fourth row). The regions in the partitions are represented with their mean color, and the saliency maps are linearly normalized to [0,255] for visualization purposes.} \label{fig:hierarchyUCM}
\end{figure}

Figure \ref{fig:hierarchyUCM} shows an example of hierarchy of five partitions generated with this algorithm, with $300$, $100$, $30$, $10$ and $3$ regions, respectively.

\subsection{Nested partitions}
\label{ssec:nested}

Figures \ref{fig:hierarchyBPT} and \ref{fig:hierarchyUCM} illustrate an important property of the partitions generated with hierarchical methods: there is an inclusion relationship between regions at different levels which simplifies their joint processing or the merging of information from the different levels. On the contrary, if partitions are generated with flat partition methods, regions at different levels of detail cannot be directly related and thus have to be processed separately. 

Another issue to consider is that regions in the hierarchies aim at representing the objects in the image, from the smallest ones or the finest detail, to the largest ones. However, there may be no single region representing the object of interest, that is, hierarchies can also suffer some degree of over-segmentation.

Finally, if we compare the set of partitions obtained using BPT and UCM, we see the different behavior of the two algorithms. Partitions generated with UCM are smoother and small objects may not be correctly represented. In turn, BPT accurately segments small objects (the small key letters) but may also produce noisy contours in flat zones (the over-segmentation of the calculator in the partition with 10 regions).

\section{Hierarchical saliency models}
\label{sec:hiemodels}

In the following we describe the two proposed saliency models. The first model is based on a hierarchy of partitions and the second one works directly on the  nodes created by the hierarchical segmentation. As all saliency methods, we rely on priors or assumptions on the properties of objects and backgrounds \cite{Wei-eccv12}. The most important one is the \emph{contrast prior}, the assumption that a salient pixel or region presents high contrast within a certain context, either local or global. Other commonly used priors are the \emph{boundary prior} and the \emph{center prior}. The first one assumes that objects seldom touch the image boundary, so the image boundary is mostly background. The center prior assumes that pixels or regions located near the image center are more important.

\subsection{Saliency on a hierarchy of partitions}
\label{ssec:smaps}
We rely on the contrast and boundary priors to extract saliency maps for each level in the hierarchy, and then combine the results into a final map.
The following subsections detail the two processes.

\subsubsection{Saliency from a single partition}
\label{ssec:saliency_one}

We propose two different methods to compute the saliency at each level in the hierarchy. The first one is based on local contrast, and measures the contrast between each region and its neighboring regions in the partition. The second is based on global contrast, measuring the contrast between each region and all the other regions in the partition.

\medskip
\textbf{Local contrast: } Given a partition $P_k=\{R_i^k\}_i$, let us denote the set of neighbors of a region $R_i^k$ as $N(R_i^k)$.

The saliency at level $k$ is defined in a region $R_i^k$ (that is, it has the same value for all the pixels in the region) as:
\begin{equation}
S(R_i^k)=\sum_{R_j^k\in N(R_i^k)}w_{cc}(R_i^k,R_j^k)d(M_{R_i^k},M_{R_j^k})
\end{equation}
where $d(M_{R_i^k},M_{R_j^k})$ is a distance measure between region $R_i^k$ and a neighboring region $R_j^k$ in the partition, $M(R_i^k)$ denotes the model used to characterize the region.

The factors $w_{cc}(R_i^k,R_j^k)$ weight the contribution of each neighbor $R_j^k$ to the saliency of region $R_i^k$. The weight is proportional to the common perimeter between the regions:
\begin{equation}
w_{cc}(R_i^k,R_j^k)=\frac{|\partial{R_i^k}\cap \partial{R_j^k}|}{|\partial{R_i^k}|}
\end{equation}
where $\partial{R_i^k}$ denotes the contour of region $R_i^k$. The larger the common contour between two regions, the higher the contribution of their color distance to the saliency map.

Many different models could be used to represent regions. In this paper we work with a 3D color histogram in the $CIE Lab$ color space. The distance used to compare histograms is the Earth Mover Distance (EMD) \cite{Rubner-98iccv}. In order to compute the histogram of a region, the region is first uniformly quantized using $64$ bins per channel. Next, the most frequent colors are selected to build 256 signatures which are then used to compute the EMD.

The bottom row in Figure \ref{fig:hierarchyBPT} shows the saliency maps built for the partitions in the top row using this local contrast method.  If a region is surrounded by regions with similar color, the saliency value of the region is low, whereas if a region is very different from its neighbors, its saliency value is high. If an object is represented by a single region at a particular scale, and it differs from the background (from its neighboring background regions), then the saliency value for the object is high at that scale.

In general, the behavior of the saliency maps at the different scales is the following. At lower scales, where we find partitions with a large number of small regions, saliency maps present high values for regions in the boundary of the objects, and very low values in the inner part of homogeneous objects (see Figure~\ref{fig:hierarchyBPT} b). There are also high values for very small salient objects. At higher scales, partitions contain a fewer number of larger regions. Objects represented by a single region that is different from the background present high values of saliency (see Figure~\ref{fig:hierarchyBPT} e, f). Small objects appear with high saliency values at lower scales, while larger objects are salient at higher scales.

\medskip
\textbf{Global contrast: }  The previous method uses only local information and may fail to correctly highlight salient large objects if they are not represented by a single region or by a few regions in one of the levels in the hierarchy. This second method estimates the saliency of a region taking into account all the other regions in the partition. However, since the saliency of a region depends mainly on the contrast to nearby regions, and the contrast to distant regions is less important, the contribution of each region is weighted by a factor that depends on the distance between regions.

The saliency at level $k$ is defined in a region $R_i^k$  as:
\begin{equation}
S(R_i^k)=\sum_{R_j^k\neq R_i^k} |R_j^k|w_s(R_i^k,R_j^k) d(M_{R_i^k},M_{R_j^k})
 \label{eq:sal}
\end{equation}
where, as before,  $d(M_{R_i^k},M_{R_j^k})$ is a distance measure between regions $R_i^k$ and $R_j^k$. Note that this time the summation involves all the regions in the partition, not only the neighboring regions. 
$| R_j^k |$ is the number of pixels in $R_j^k$ and $w_s(R_i^k,R_j^k)$ is a weight based on the distance between regions.
\begin{equation}
w_s(R_i^k,R_j^k)=exp(-\| c_i^k - c_k^k \|_2 / \sigma_s^2)
\label{eq:sweight}
\end{equation}
where $\|c_i^k - c_j^k \|_2$ is the Euclidean distance between region centroids $c_i^k$ and $c_i^k$, and $\sigma_s$ controls the strength of the weighting \cite{Cheng-11cvpr}. Large values of $\sigma_s$ increase the contribution of farther regions to the saliency of region $R_i^k$. In our experiments, best results are obtained for $\sigma_s^2=4$.

The third row in Figure \ref{fig:hierarchyUCM} shows the saliency maps built for the partitions in the top row using the global contrast method. 
Now, for large and homogeneous objects represented by several regions at lower levels, all the regions receive similar saliency values.

\medskip
Relying on the boundary prior (\cite{Cheng-11cvpr,Zou-bmvc13,ZLiu-tip14}) we can assume that salient objects do not touch the image border, or that their intersection with the image border is lower than the intersection of background regions with the image border. Under this assumption, we add an optional weight factor to measure the saliency of a region $S(R_i^k)$ in eq.\ref{eq:sal}
\begin{equation}
w_b(R_i^k)=exp(- (| \partial{R_i^k} \cap \partial I| / | \partial{R_i^k} | ) / \sigma_b^2)
\label{eq:wb}
\end{equation}
The factor depends on the fraction of the region contour $\partial{R_i^k}$ that intersects the image border $\partial I$. In the experiments we use $\sigma_b^2=0.5$. This prior gives a small weight to regions touching the image border.

The last row in Figure \ref{fig:hierarchyUCM} shows the saliency maps built for the UCM partitions in the top row using the global contrast method  and using the spatial prior. We can observe that regions that intersect the image border (both background regions or regions from foreground objects like the hand) have a lower saliency value.

If we compare these maps with the saliency maps generated for BPT partitions using local contrast in Figure \ref{fig:hierarchyBPT}, we see that the local method already reduces the importance of background regions, hence there is no need to apply the boundary weight. The reason is that the local method uses as weighting factor the fraction of the region contour that intersects the contour of neighboring regions. For regions touching the image border, the region contour includes pixels in the image border, so weights are smaller than weights of regions that do not intersect the border.

\begin{figure}[ht]
\begin{center}
\begin{tabular}{ccc}
\includegraphics[width=3cm]{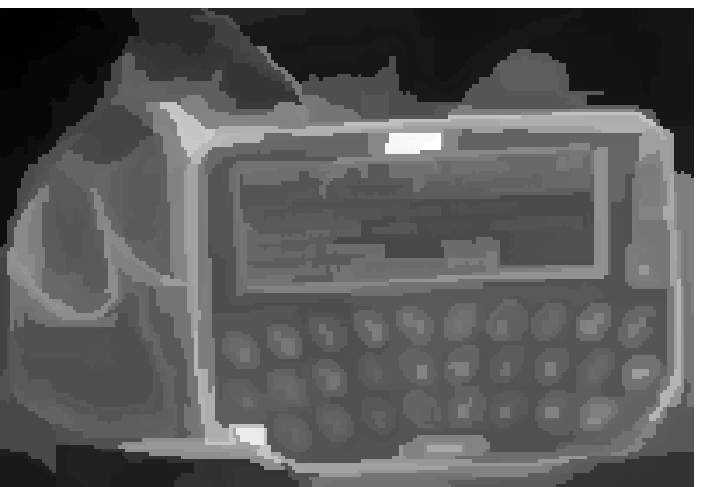} &
\includegraphics[width=3cm]{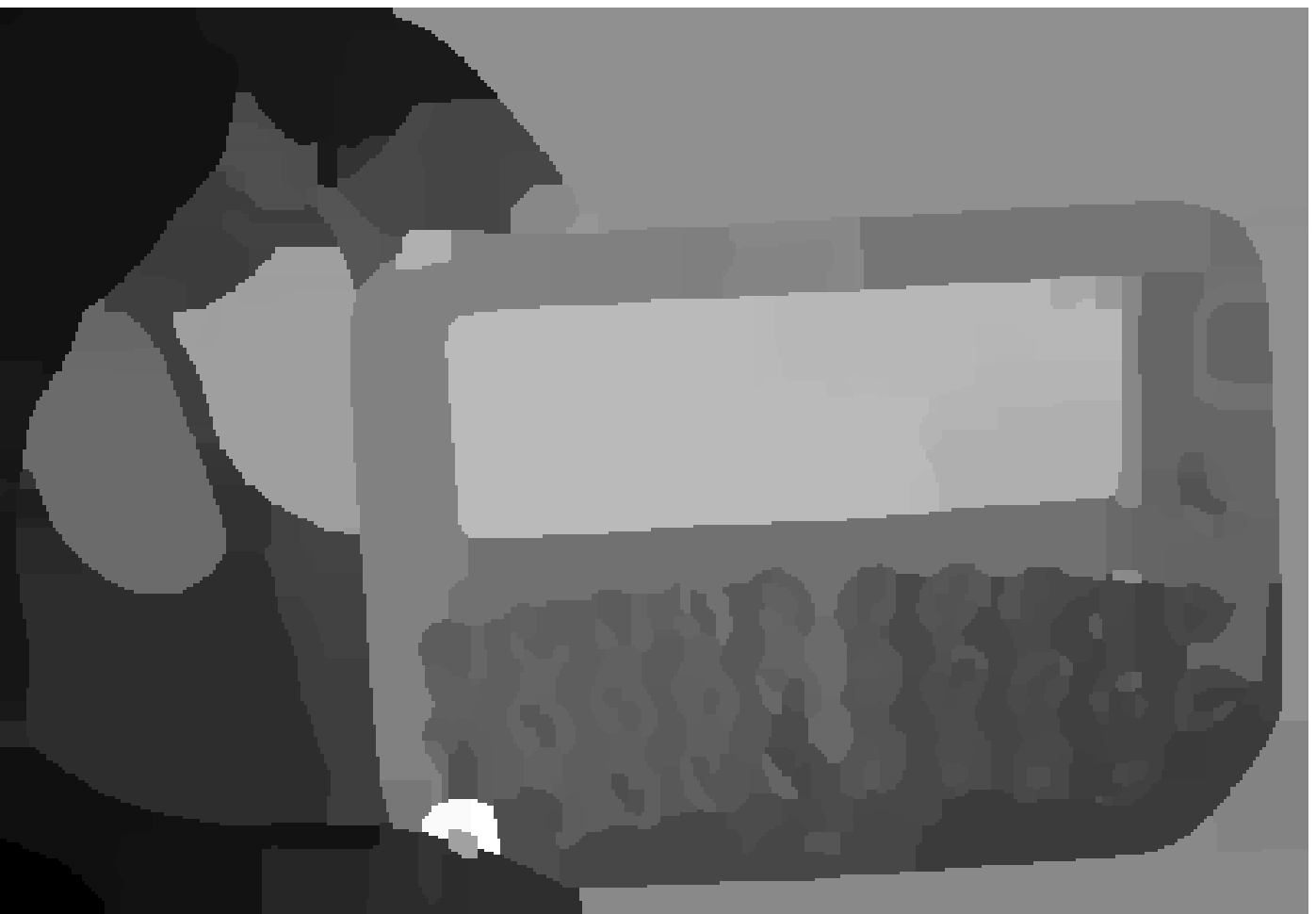} &
\includegraphics[width=3cm]{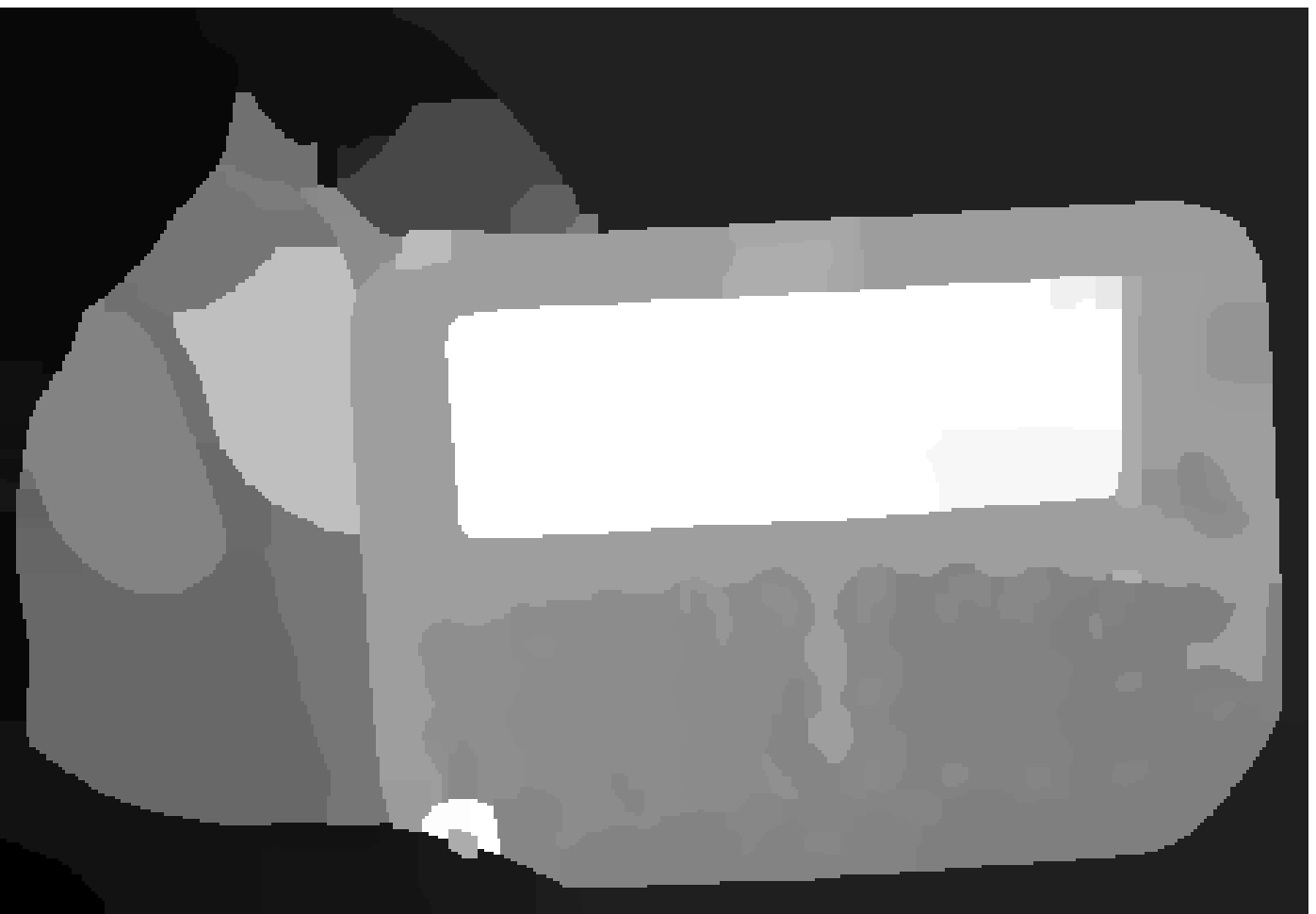} \\
(a) BPT & (b) UCM & (c) UCM boundary \\
\includegraphics[width=3cm]{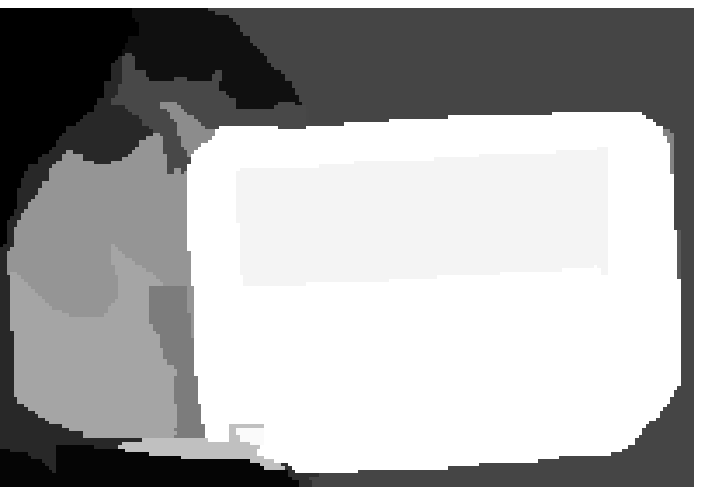} &
\includegraphics[width=3cm]{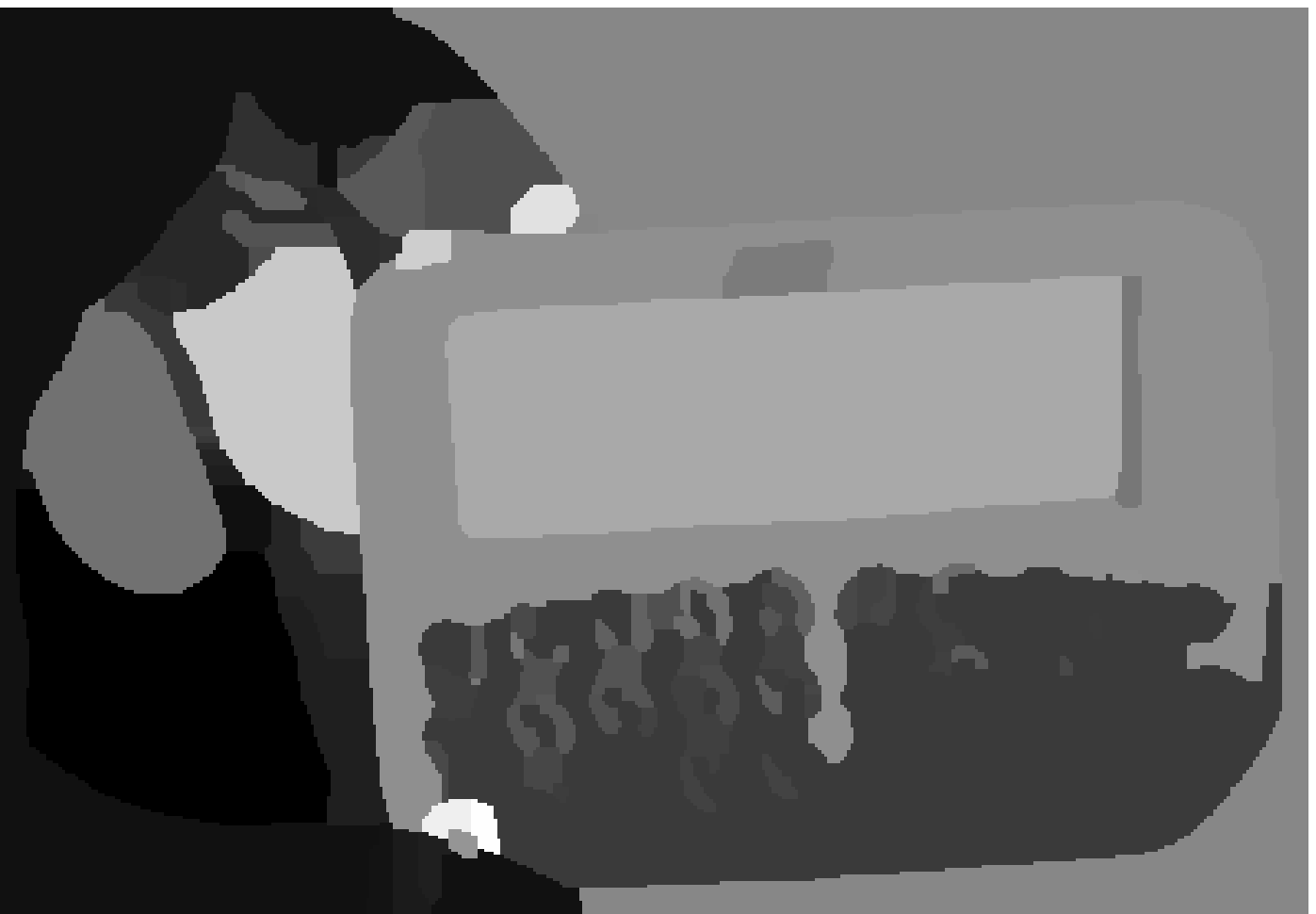} &
\includegraphics[width=3cm]{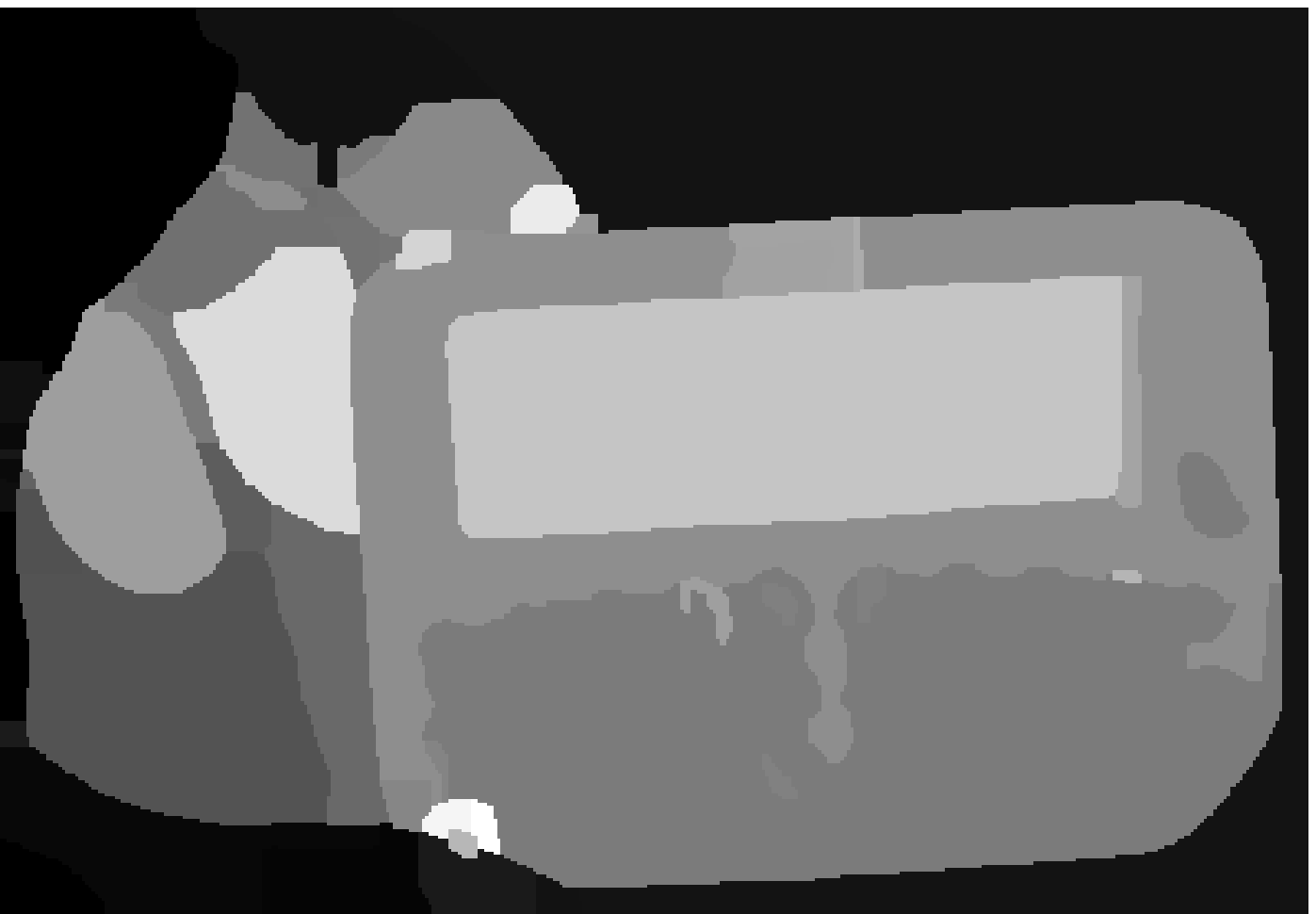} \\
(d) BPT & (e) UCM & (f) UCM boundary\\
\end{tabular}
\caption{First row: final maps obtained using the \emph{mean} combination of the saliency maps constructed from BPT partitions (Figure~\ref{fig:hierarchyBPT}) (a), from UCM partitions (Figure~\ref{fig:hierarchyUCM}) (b) and UCM partitions with boundary prior (c).
Second row: the fusion with  the \emph{max} criterion for the same sets of partitions (d),(e) and (f), respectively.} \label{fig:combined_maps}.
\end{center}
\end{figure}

\subsubsection{Final saliency map}
\label{ssec:saliency_final}

The saliency maps obtained at the different levels in the hierarchy are fused into a compound map. Before this step, maps are globally normalized considering the minimum and maximum saliency values in all the levels. 

The multi-scale approach and the measure used to compute similarity between regions result in comparable saliency values at all levels. 
A small but very salient object (very different from its neighboring regions) will have high saliency values at fine scales. At coarser scales the region will not be merged with its neighbors, so the saliency value at higher levels will be high too. On the other hand, a small non salient region in a fine partition (very similar to its neighboring regions) will still be non salient at coarser partitions or it will be merged with some neighbors forming a larger salient object (with high saliency value) at coarser level.

Local normalization is not performed because for the fusion we want to compare original values, scaled to cover the maximum possible range by global normalization.

We test three combination rules: \emph{mean}, \emph{max} and \emph{hierarchical inference}, the last one with two different graphical models.

\medskip
\textbf{Max and mean:}  For a pixel $x$ in the image, let $l_k = P^k(x)$ be the label of the pixel in the partition at level $k$. Using this notation, the saliency of the pixel at this level is  $S(R_{l_k}^k)$.

For the \emph{mean} combination, the saliency of each pixel $x$ is the average of the saliency values of the pixel for all levels. 
\begin{equation}
S(x)=\sum_{k=1}^{K}{S(R_{l_k}^k)}
\end{equation}

Analogously, the $max$ combination computes the maximum value of the saliency maps at each pixel:
\begin{equation}
S(x)=\max_{k=1,...,K}{S(R_{l_k}^k)}
\end{equation}

Figure \ref{fig:combined_maps} shows the maps generated using the \emph{mean} (top row) and \emph{max} (bottom row) combination methods for the hierarchy of partitions presented in Figures \ref{fig:hierarchyBPT} and \ref{fig:hierarchyUCM} .

 When using the $mean$ combination, all  maps contribute equally to the final saliency, and saliency maps tend to be smoother. With the $max$ combination, regions that are very salient in one level are also very salient in the final map.

\medskip
\textbf{Hierarchical inference:}  The last two fusion methods, following the approach proposed in \cite{Yan-cvpr13,Zhu-cvpr14},  use an optimization framework to integrate the multiple level maps.
The hierarchy of nested partitions naturally induces a graph structure where hierarchical inference can be applied.
In this graph nodes correspond to regions in the partitions while edges reflect relationships existent between regions. We add a root node to represent the entire image and edges between this node and all regions in the highest partition.

We define two different graphical models. In the first one we only create edges between nodes corresponding to regions in different partitions. The connections reflect inclusion relationships existent among regions in the nested partitions. The goal is that large scale regions in the upper levels influence the assignment in lower levels. In the second model, we also add connections between nodes in the same level (regions in the same partition), taking into account neighborhood relationships between regions. These connections enforce consistency between neighboring regions \cite{Yan-cvpr13}.

Examples of the two graphs are shown in Figure~\ref{fig:inference}, for a 3-level hierarchy of partitions with 9, 6 and 3 regions. In graph (a) edges only link regions in different partitions, while in graph (b) we add edges between neighboring regions.

The energy function for the first model is the following, where $s_i^k$  is the saliency value of the node associated with region $R_i^k$ at level $k$. 
\begin{equation}
E=  \sum_{k} \sum_{i} D (s_i^k)+ \sum_{k} \sum_{(i,j)} {V(s_i^k,s_j^{k+1})}
\end{equation}

The data term $D(s_i^k)$ is the cost of assigning label $s_i^k$ to region $R_i^k$ , and $s_{i0}^k$ is the initial saliency value obtained from the computation of saliency at the different partitions.
\begin{equation}
D(s_i^k) = \| s_i^k - s_{i0}^k  \|_2^2
\end{equation}

The smoothness term enforces consistency between connected nodes. Since we only encode inclusion relationships, the smoothness term enforces consistency between saliency values of corresponding regions in different levels, that is, at different scales of resolution. 
\begin{equation}
V(s_i^k,s_j^{k+1})=\|s_i^k - s_j^{k+1}\|_2^2
\end{equation}

In the second model we add edges between neighboring regions at each level. The energy function now is:
\begin{equation}
E=  \sum_{k} \sum_{i} D (s_i^k)+ \sum_{k} \sum_{(i,j)} {V(s_i^k,s_j^{k+1})}+\sum_{k} \sum_{(i,j)} {V(s_i^k,s_j^k)}
\end{equation}

Data terms $D (s_i^k)$ and smoothness terms between levels  ${V(s_i^k,s_j^k)}$ are the same as before. The smoothness term for the edges between neighboring regions at level $k$ are
\begin{equation}
V(s_i^k,s_j^k)= exp(-d(M_{R_i^k},M_{R_j^k})^2 /\sigma_c^2) \|s_i^k - s_j^{k+1}\|_2^2
\end{equation}
enforcing consistency between neighbors at each level. $d(M_{R_i^k},M_{R_j^k})$ is a distance between region models (see section \ref{ssec:saliency_one}).

The objective function can be minimized using belief propagation for the first model, and loopy belief propagation for the second model, since in this case there are loops at each level. After convergence, the saliency values in the first level of the hierarchy are used to compute the final saliency map.
Our implementation of hierarchical inference is based on the Matlab package UGM \cite{UGM}.

\begin{figure}[htbp]
\centering
\begin{tabular}{cccccc}
\scriptsize{IM}&
\includegraphics[height=1.6cm]{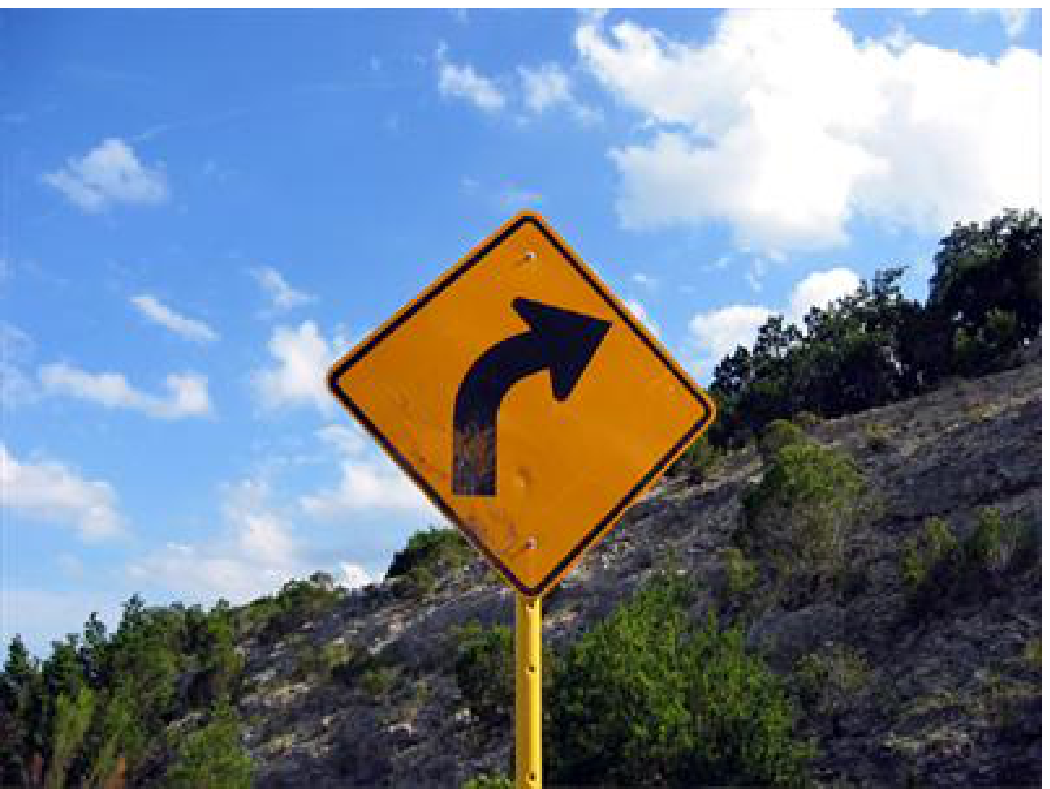} &
\includegraphics[height=1.6cm]{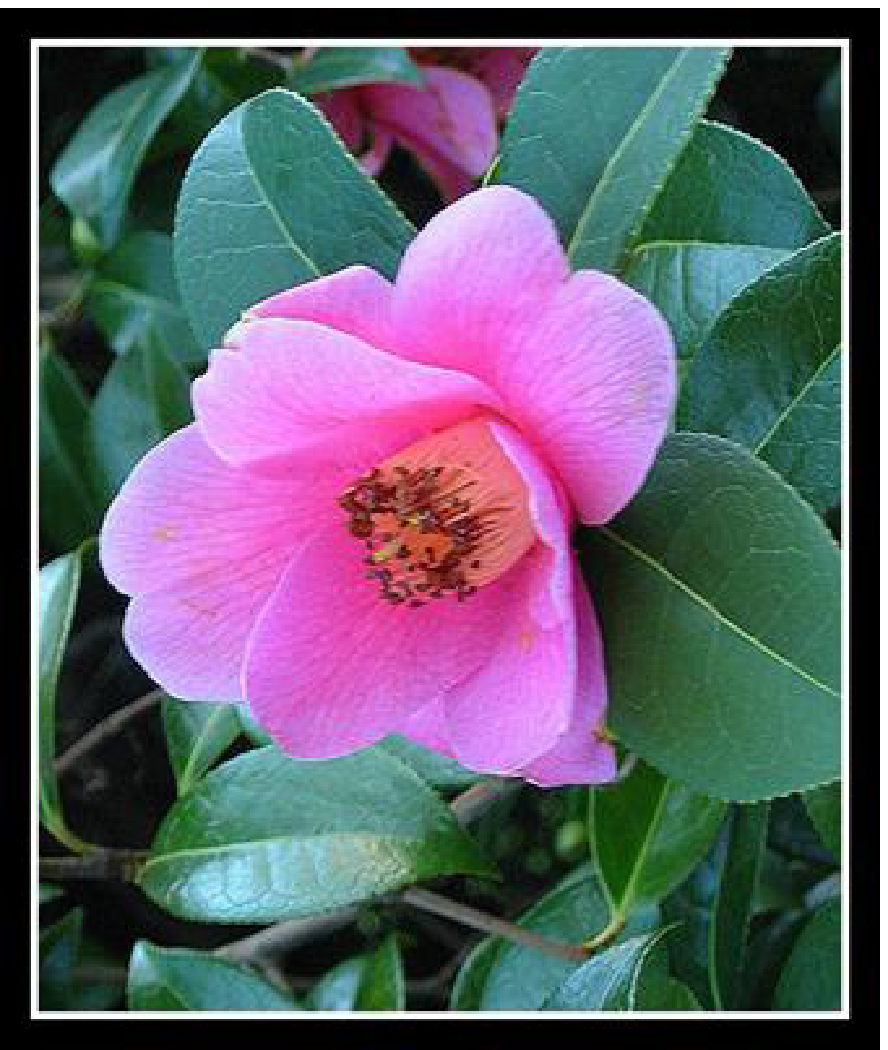} &
\includegraphics[height=1.6cm]{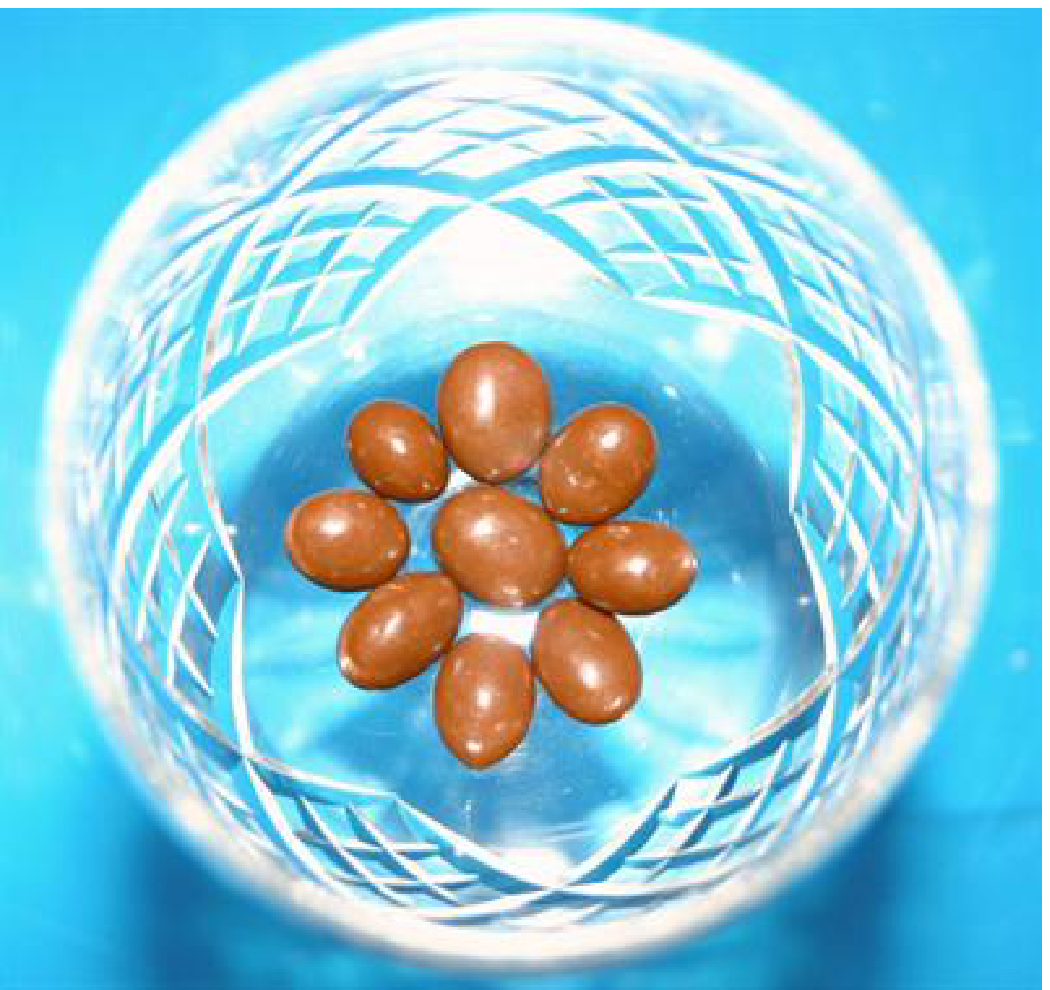} &
\includegraphics[height=1.6cm]{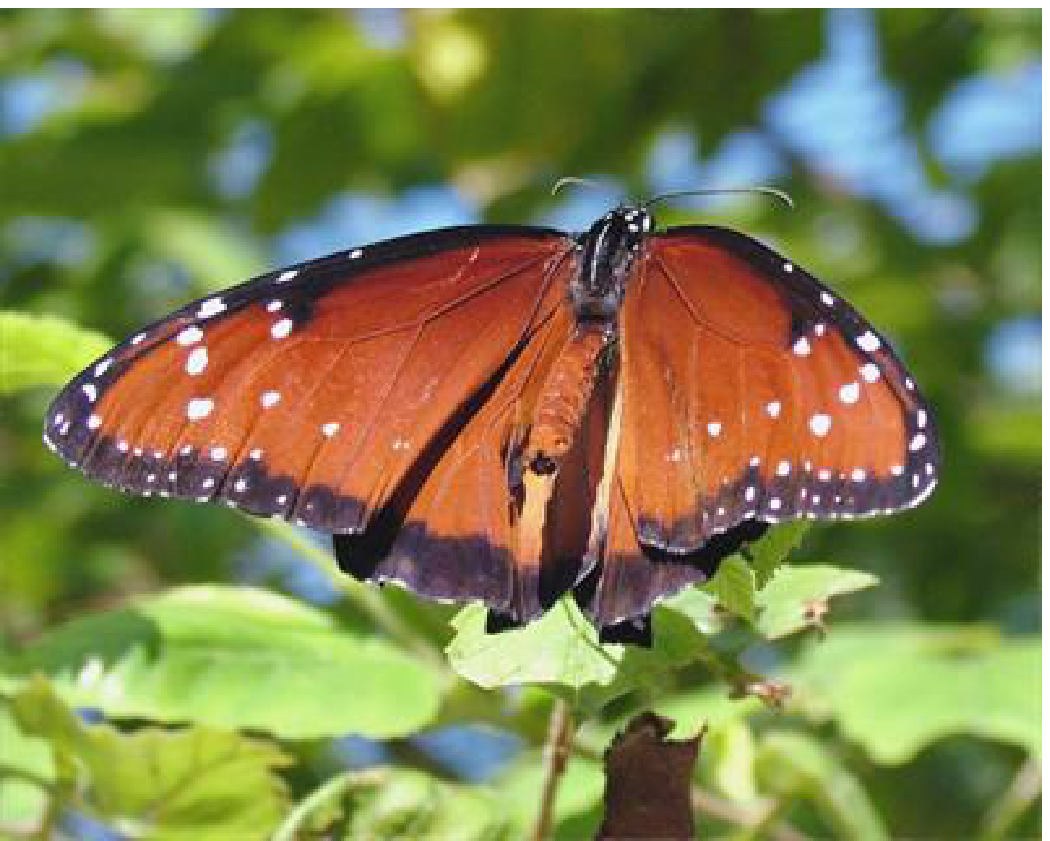} &
\includegraphics[height=1.6cm]{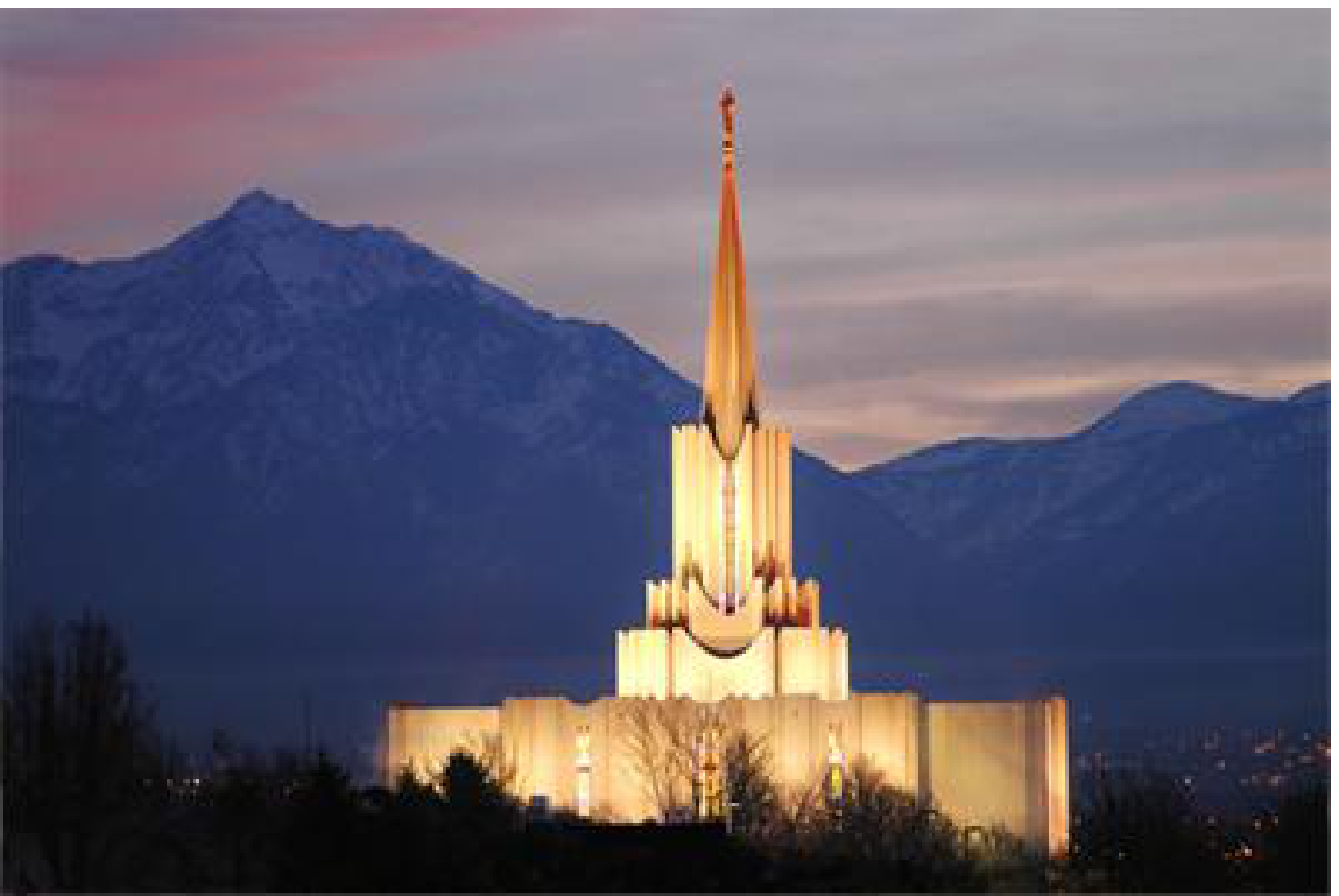} \\

\scriptsize{HP-BP}&
\includegraphics[height=1.6cm]{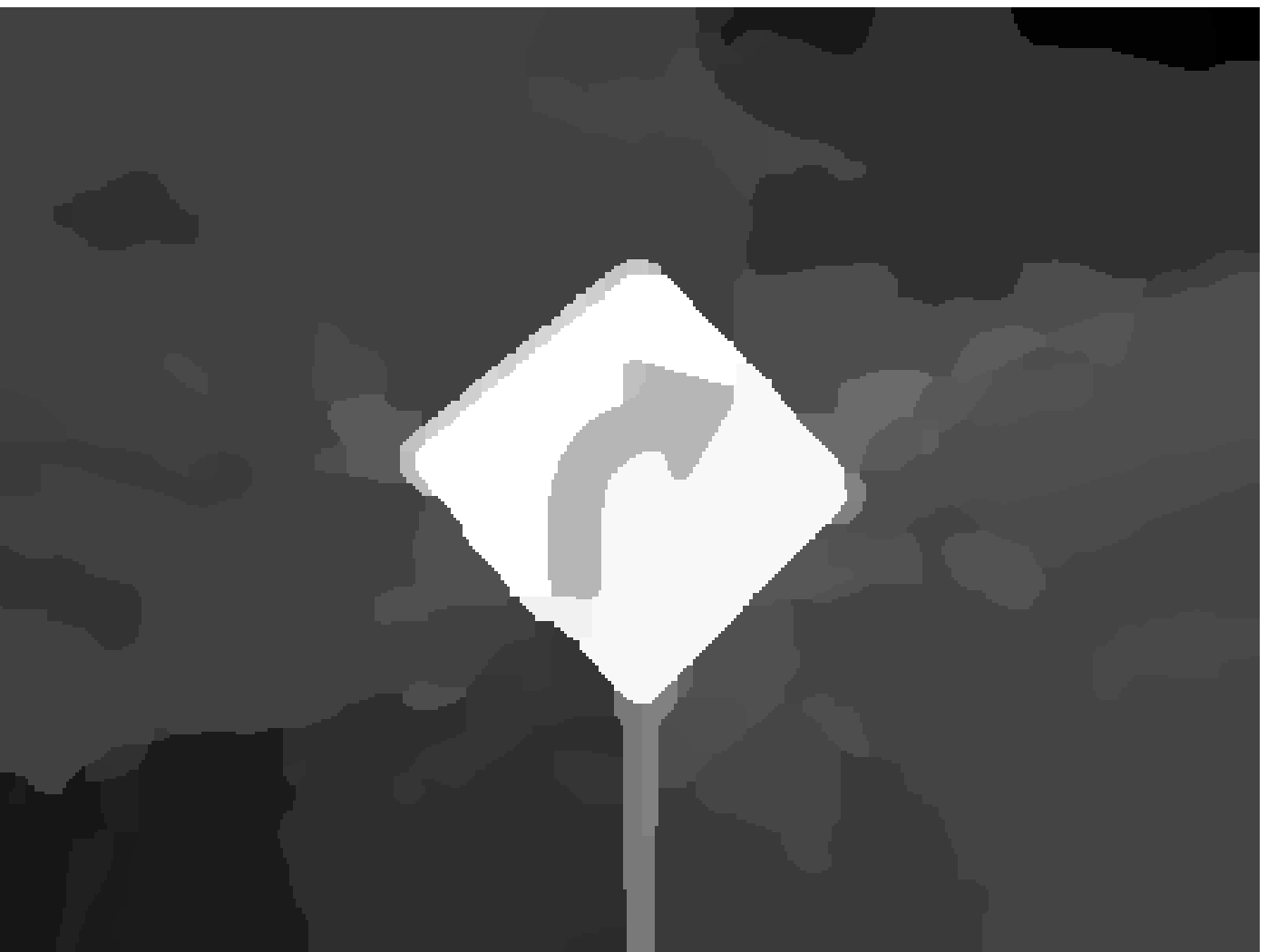} &
\includegraphics[height=1.6cm]{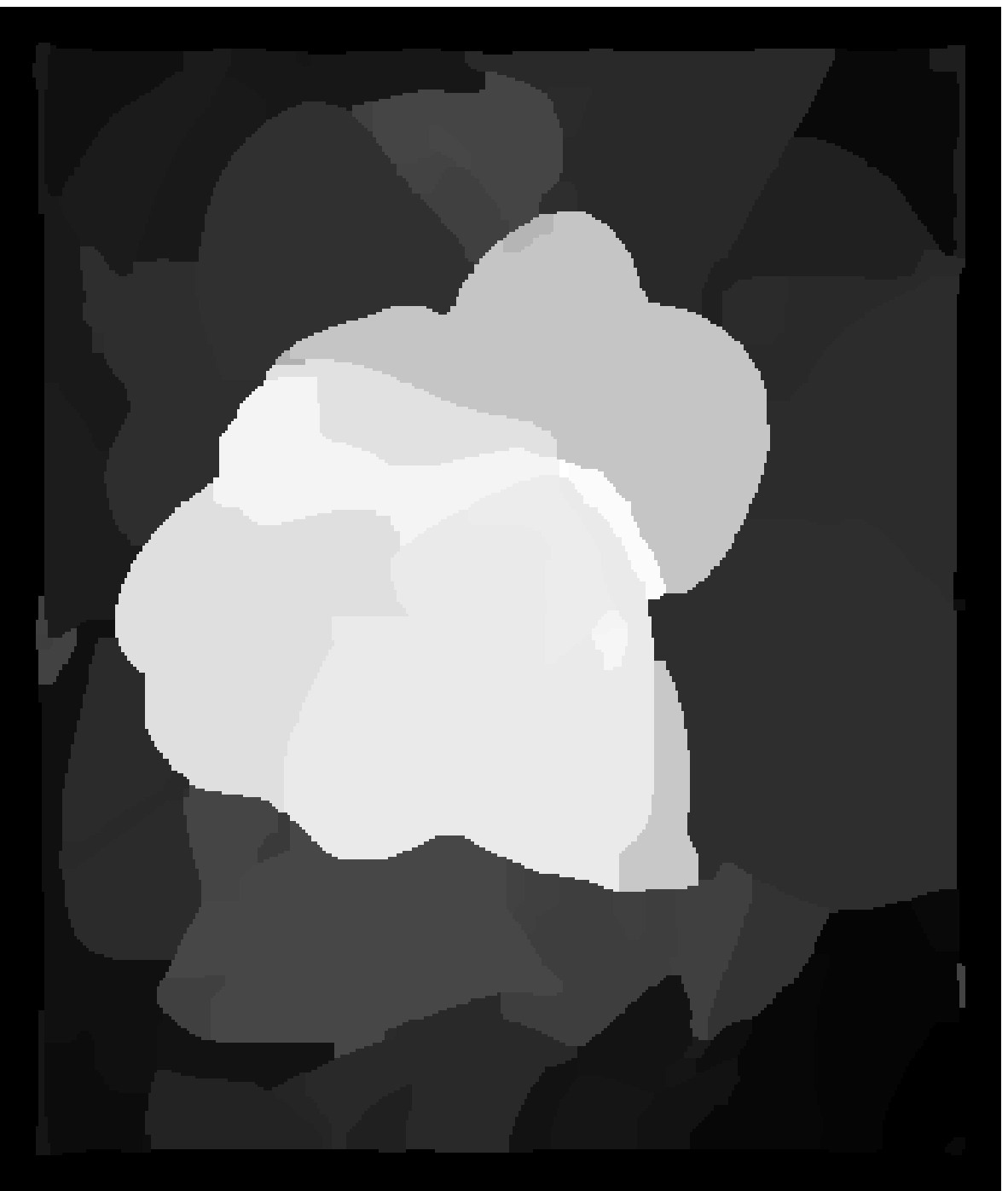} &
\includegraphics[height=1.6cm]{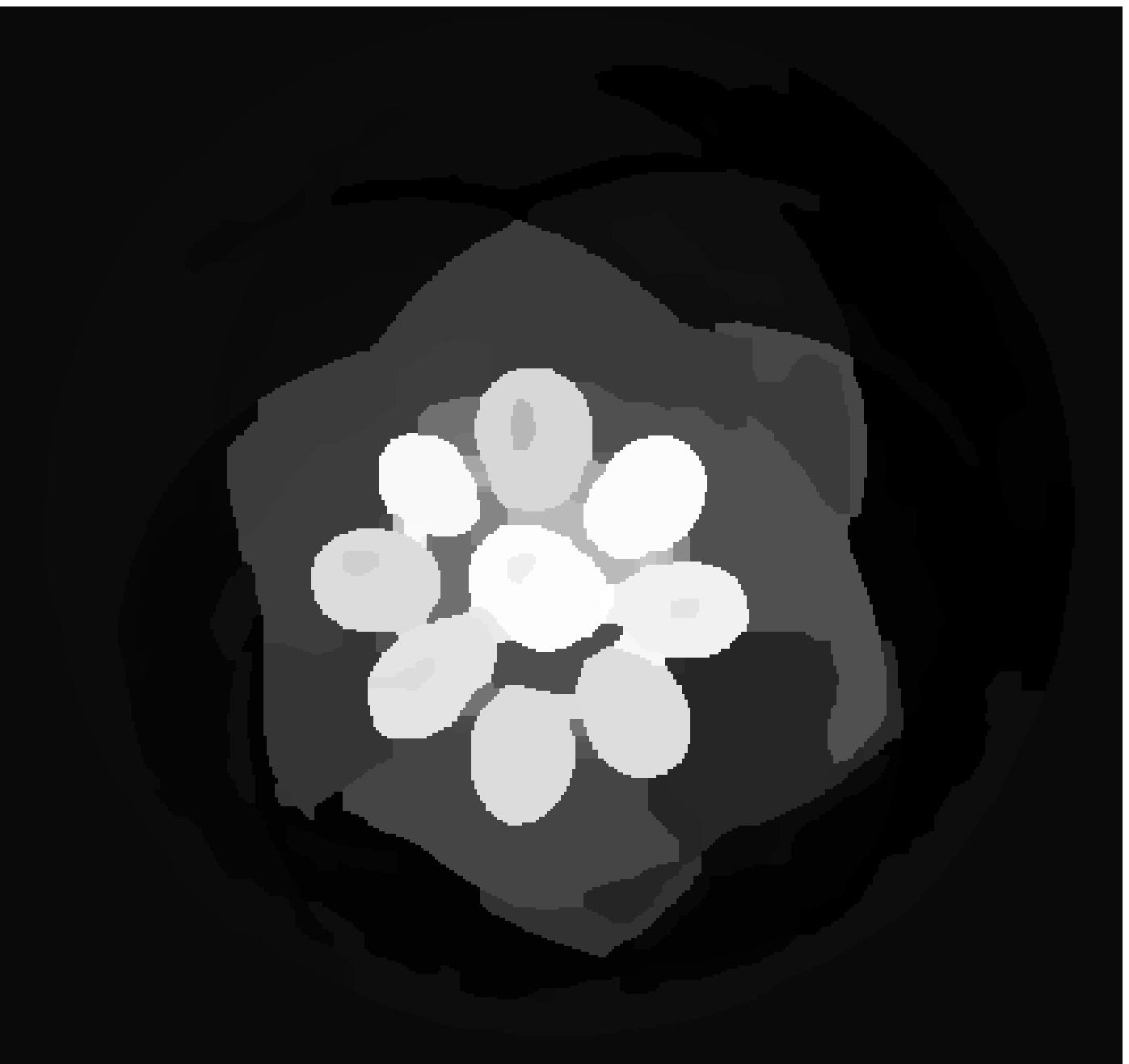} &
\includegraphics[height=1.6cm]{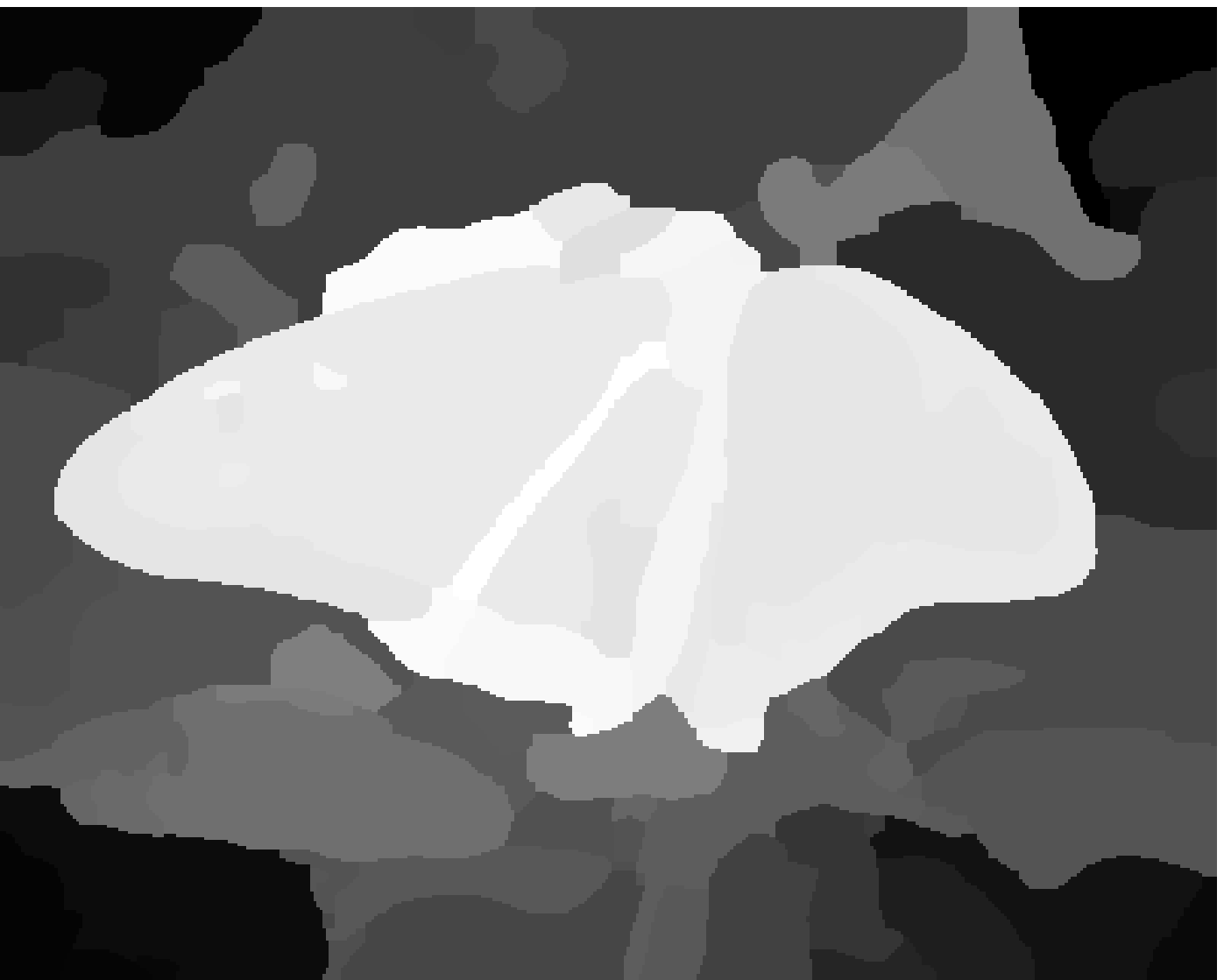} &
\includegraphics[height=1.6cm]{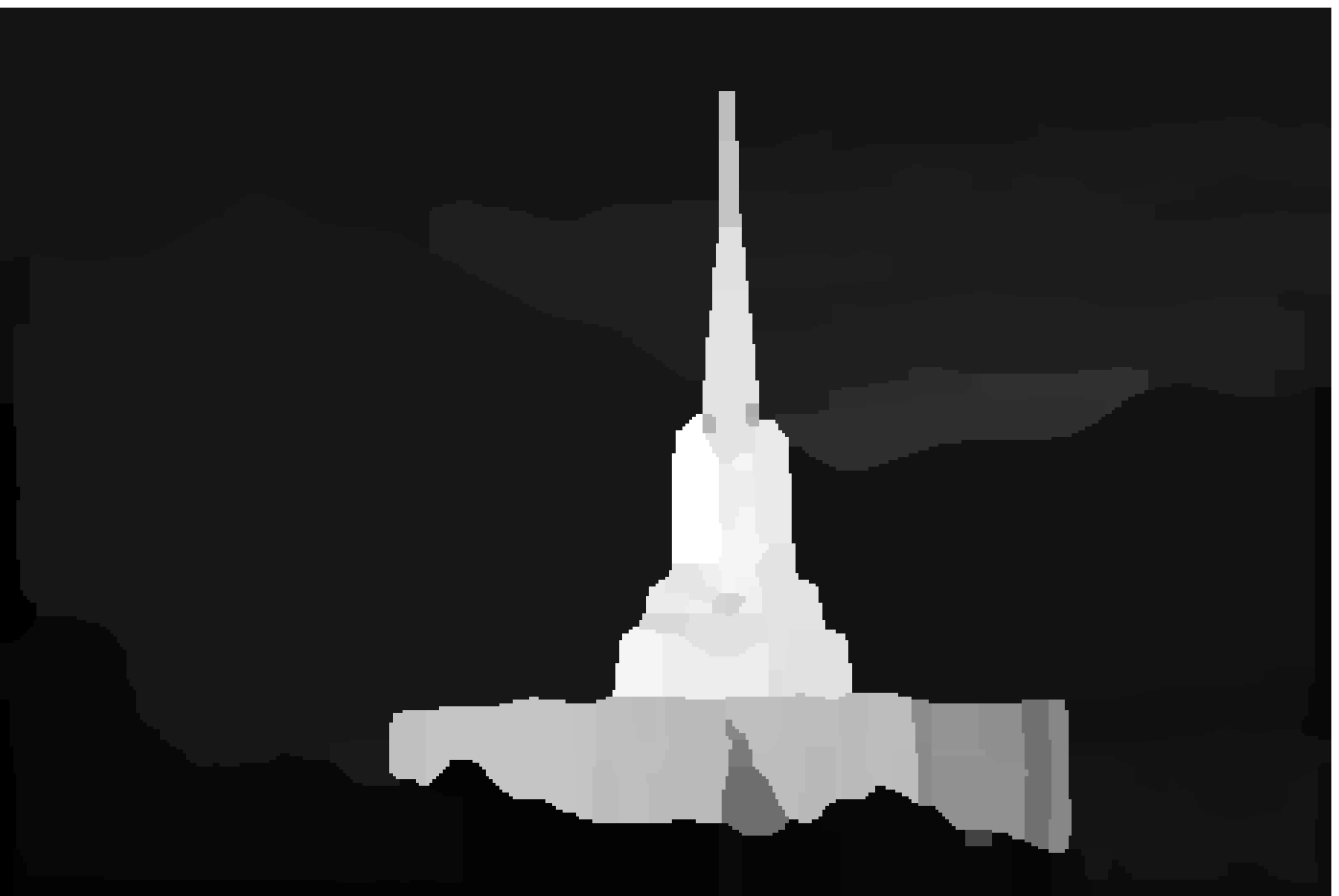} \\

\scriptsize{HP-LBP}&
\includegraphics[height=1.6cm]{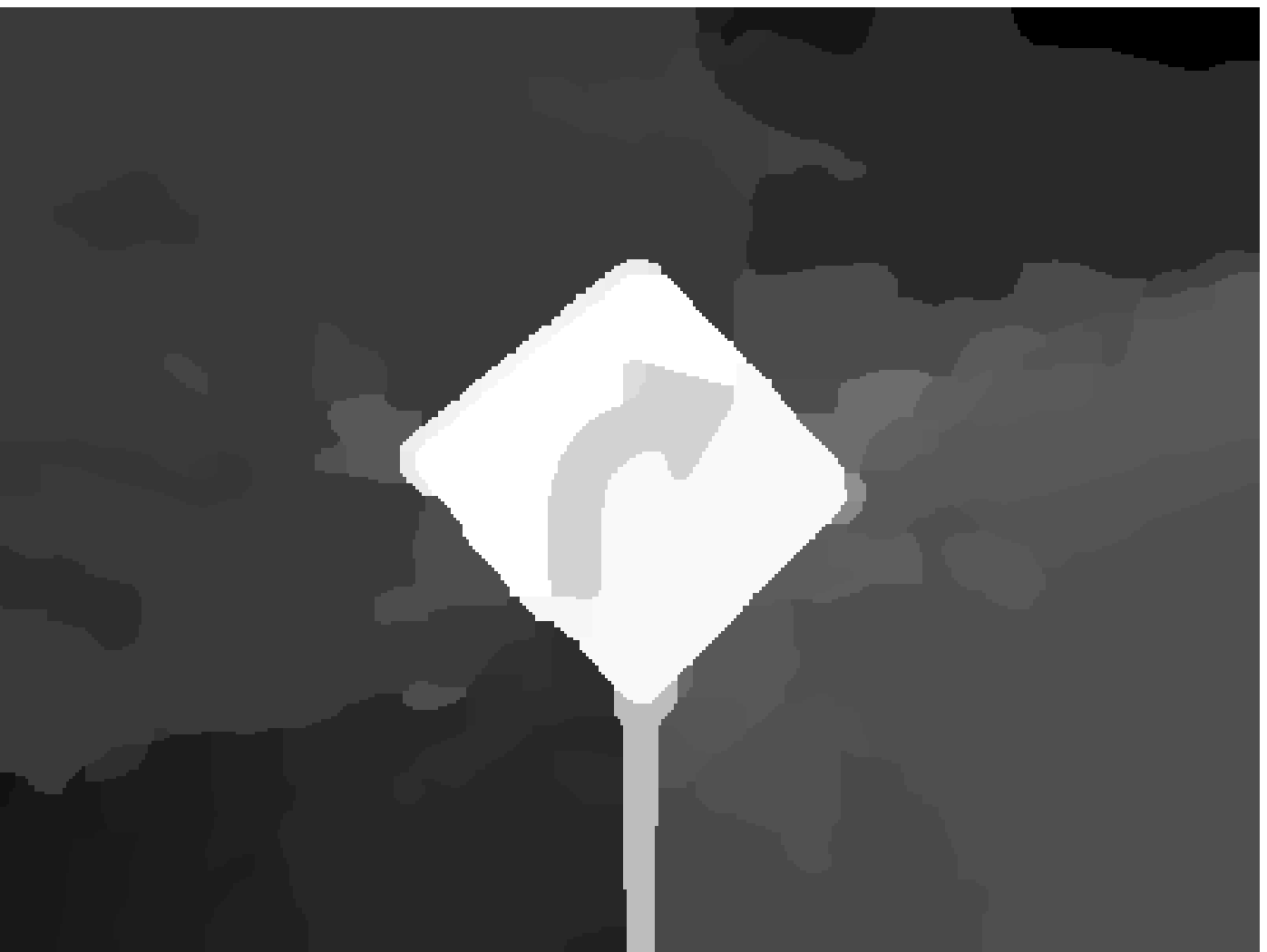} &
\includegraphics[height=1.6cm]{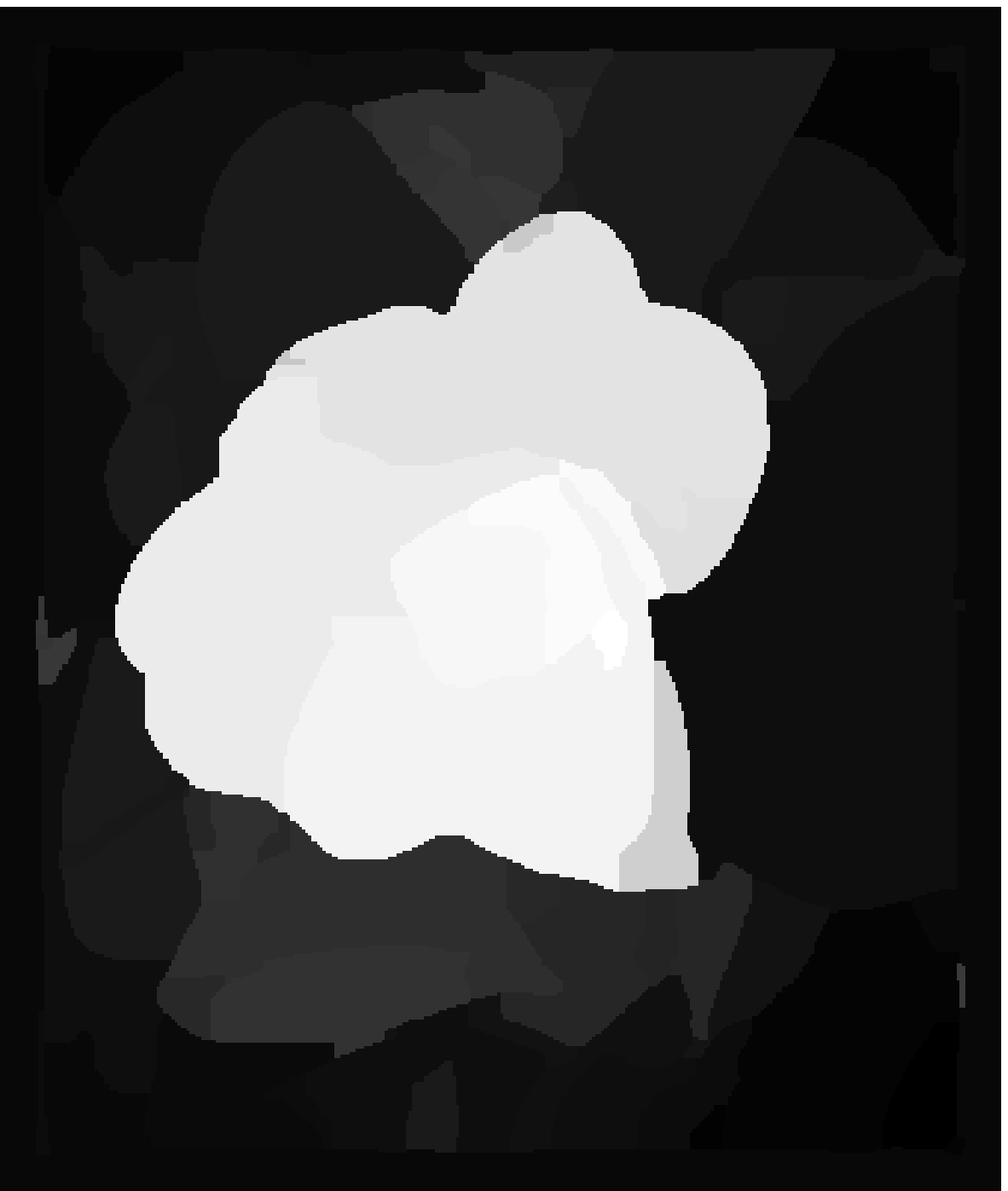} &
\includegraphics[height=1.6cm]{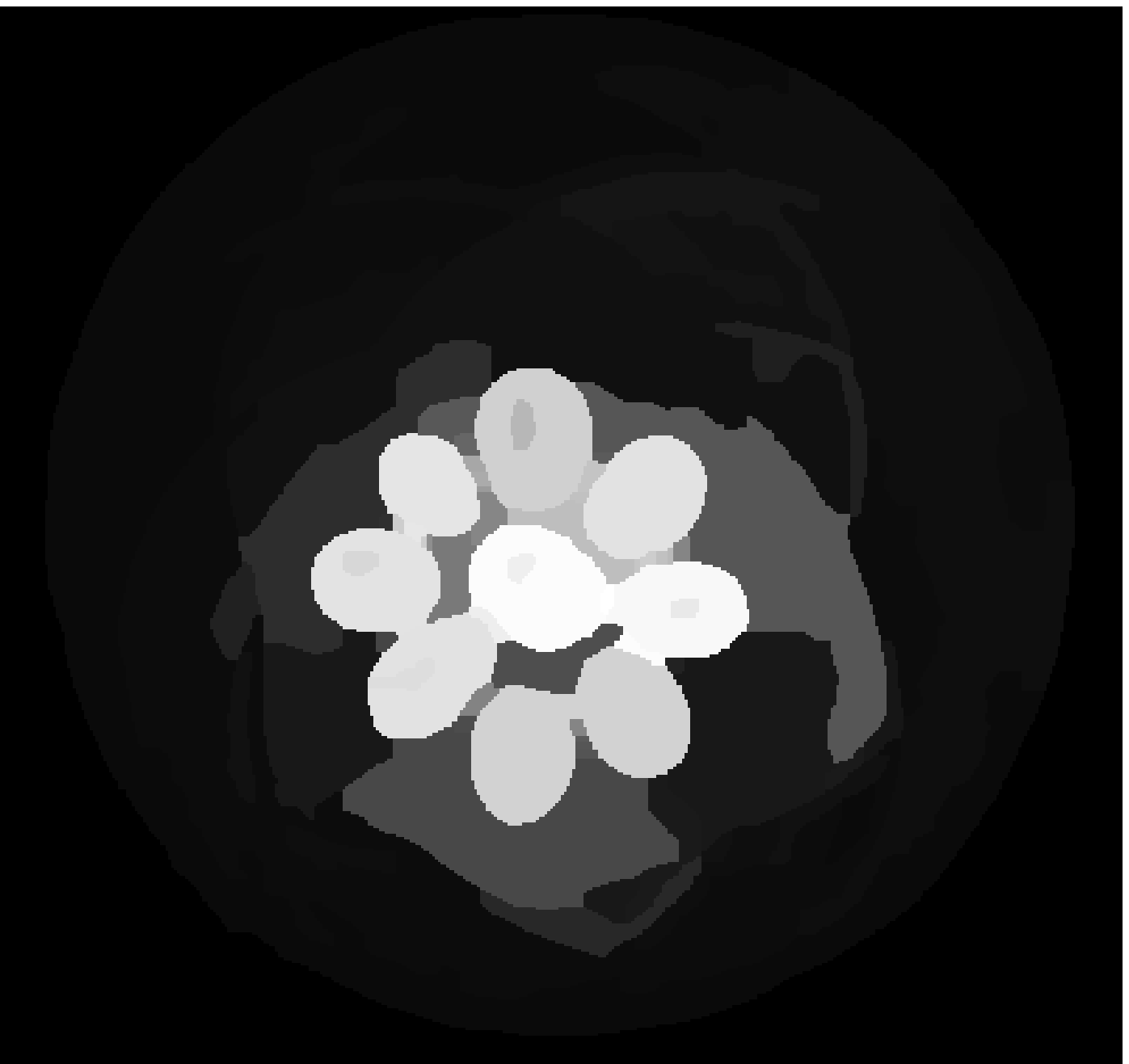} &
\includegraphics[height=1.6cm]{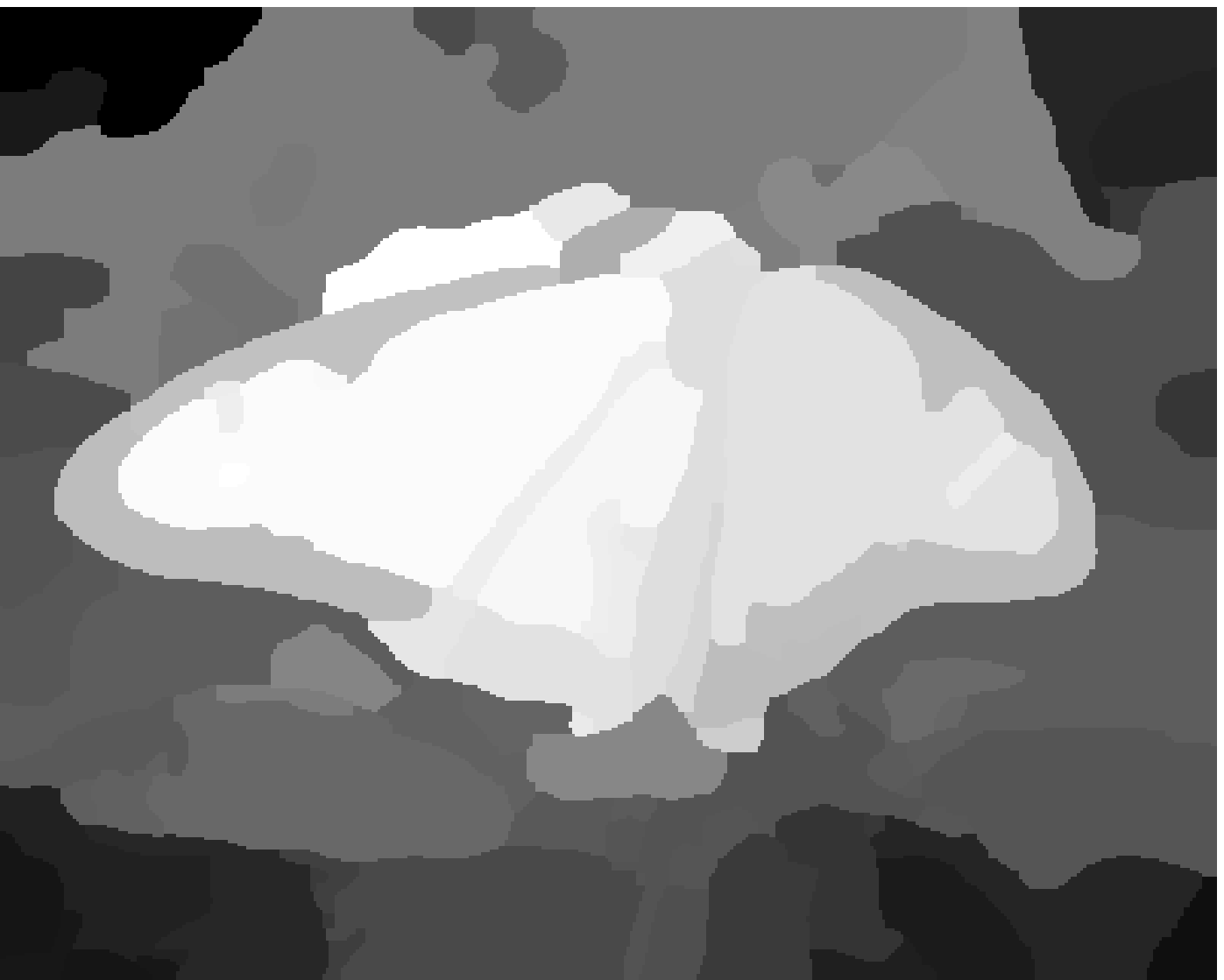} &
\includegraphics[height=1.6cm]{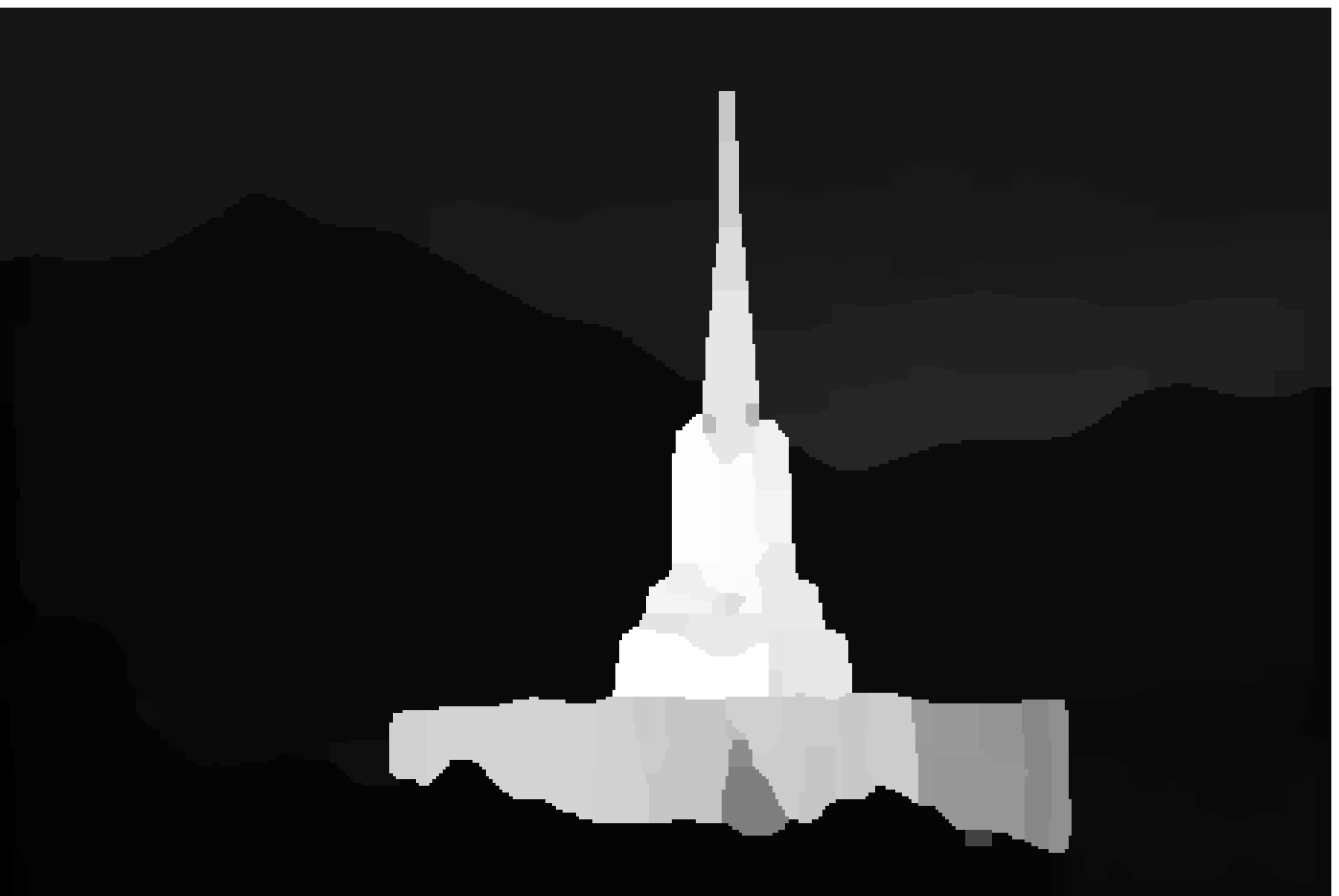} \\

\scriptsize{HP-MEAN}&
\includegraphics[height=1.6cm]{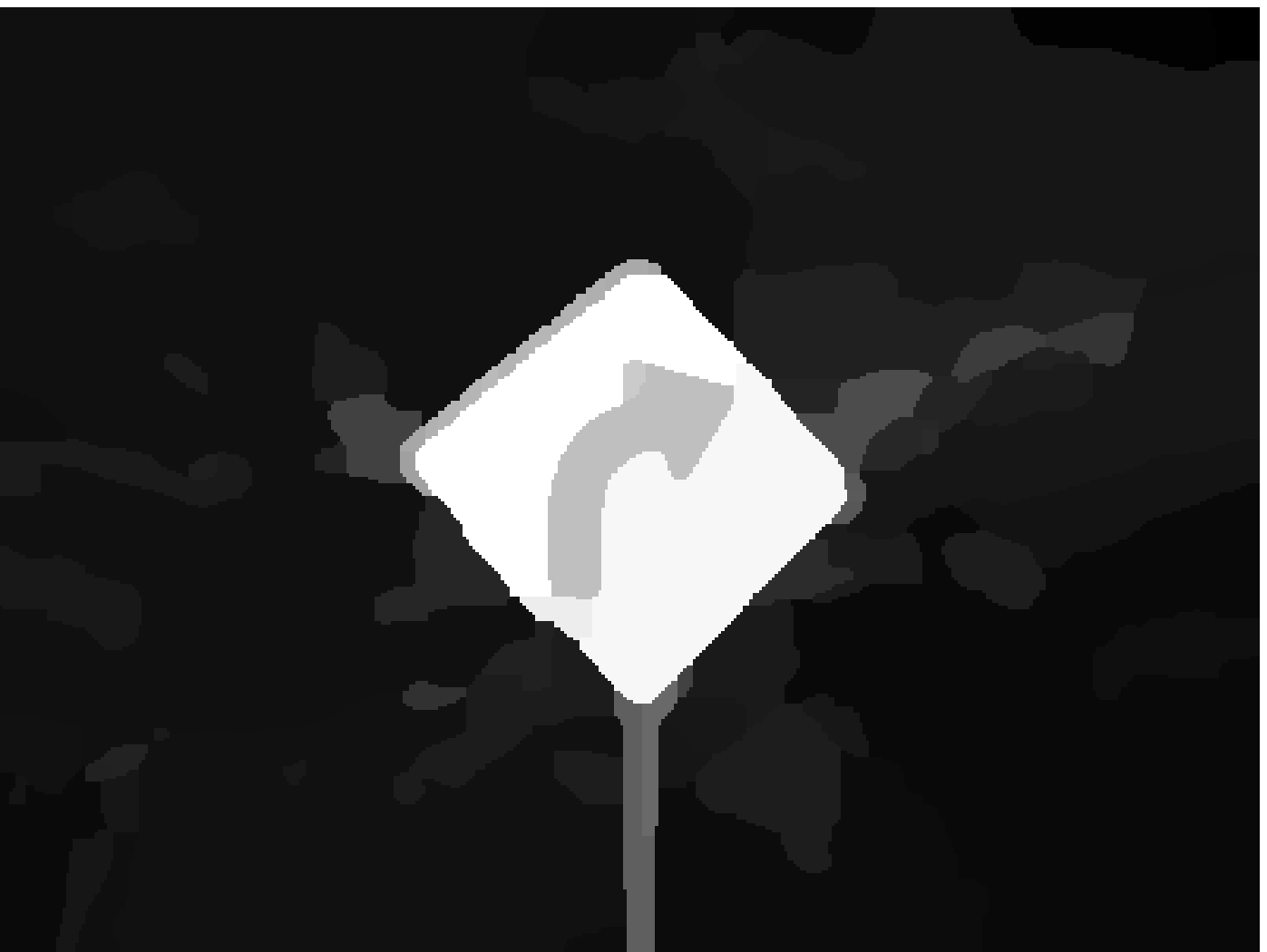} &
\includegraphics[height=1.6cm]{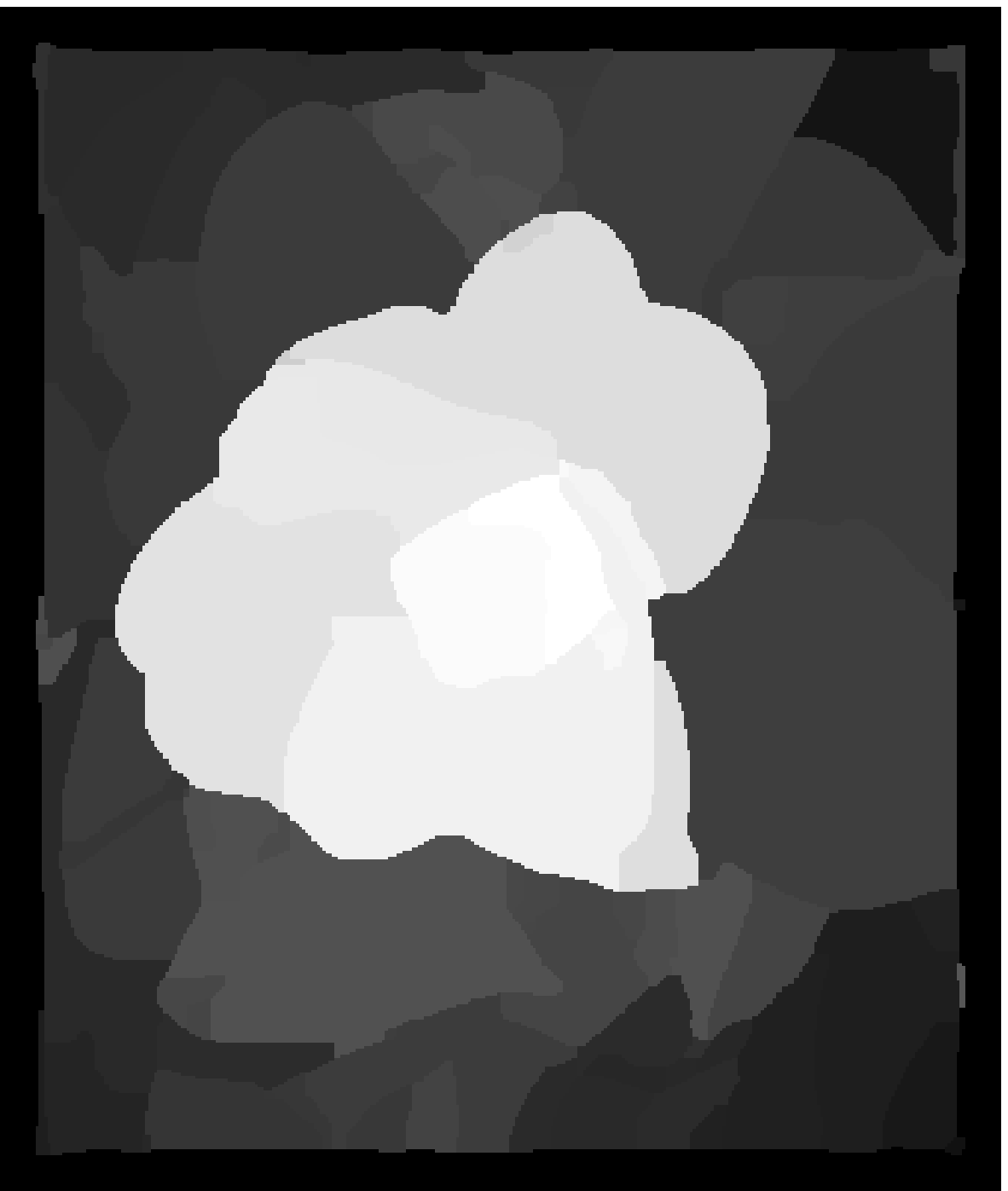} &
\includegraphics[height=1.6cm]{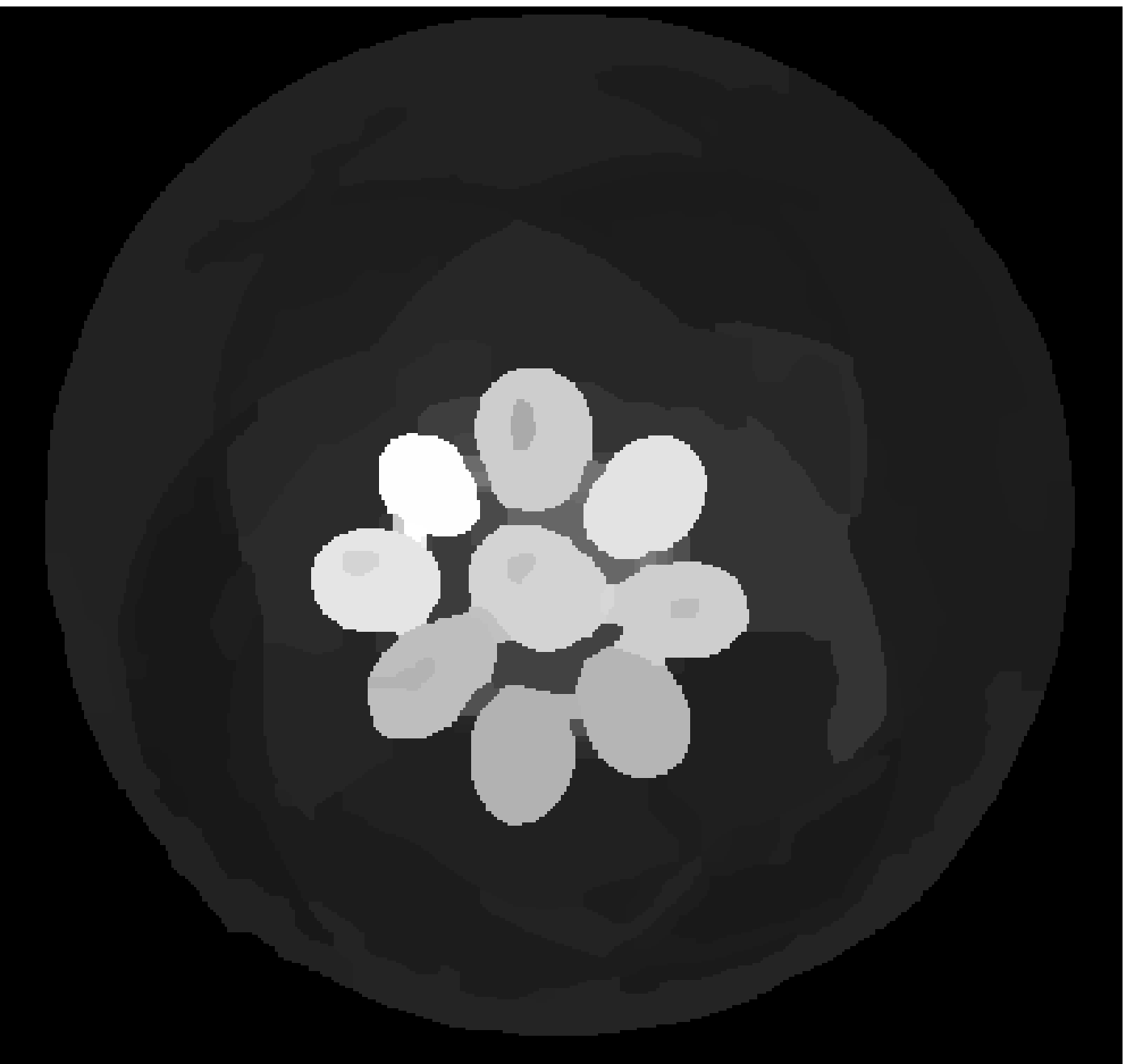} &
\includegraphics[height=1.6cm]{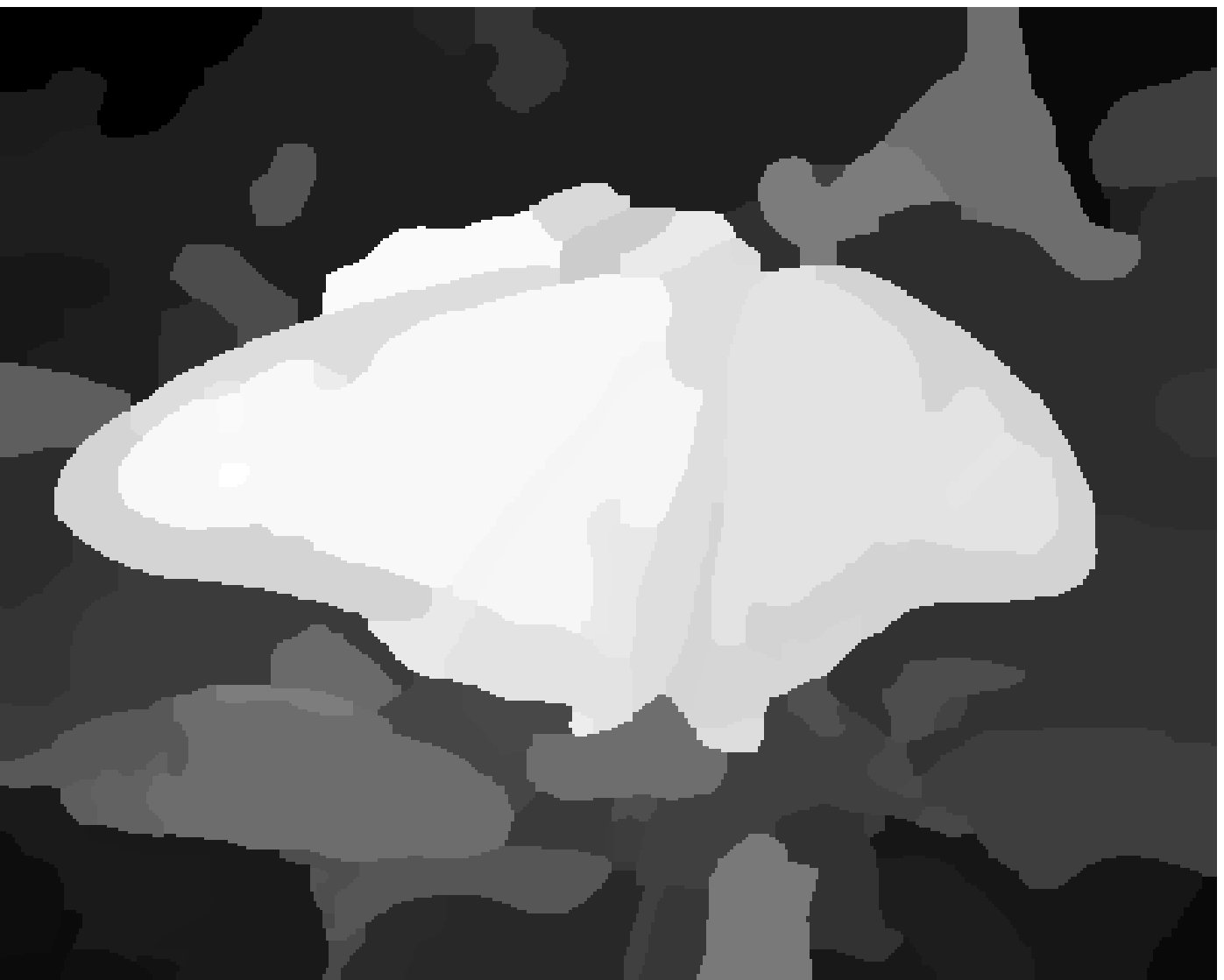} &
\includegraphics[height=1.6cm]{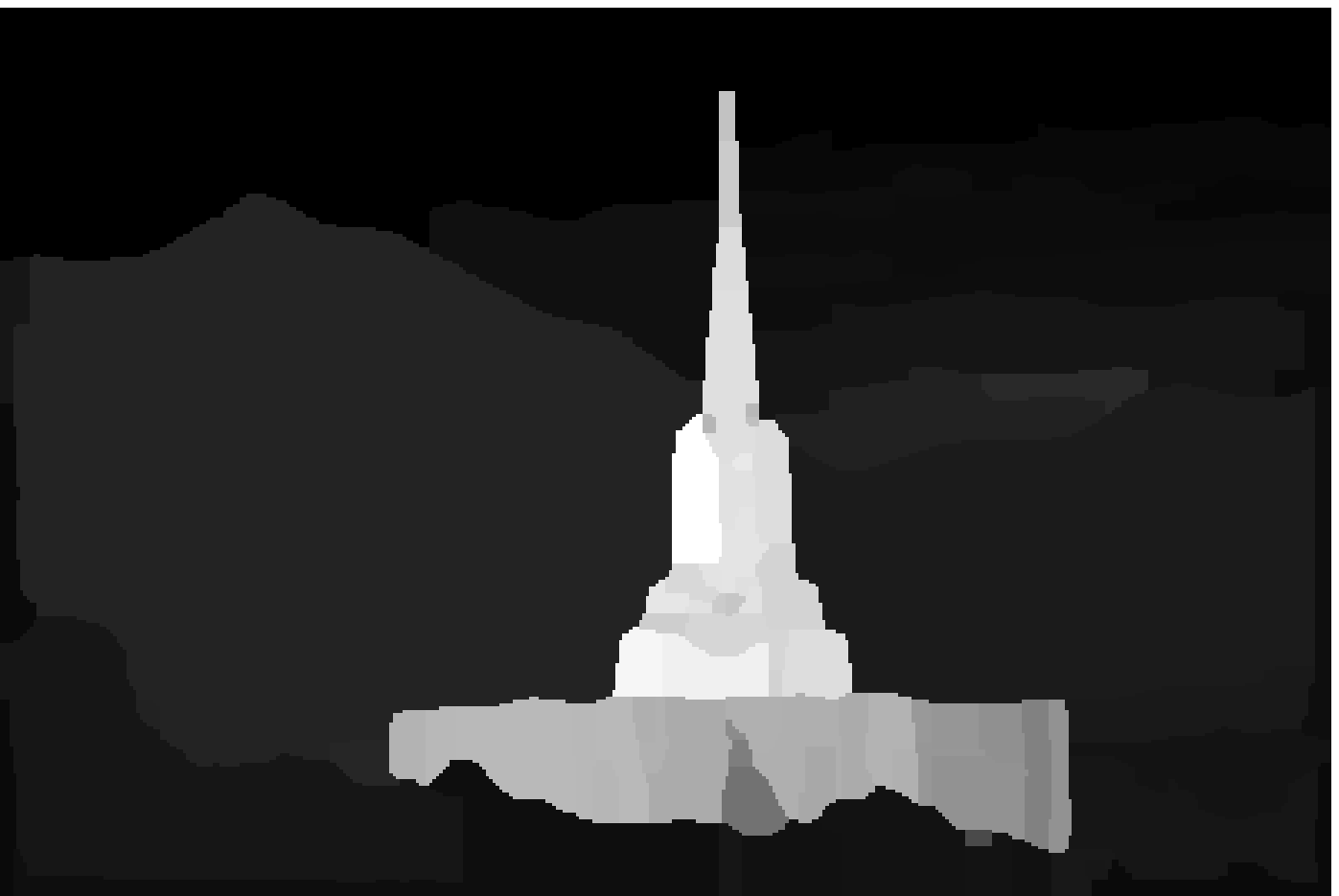} \\

\scriptsize{HP-MAX}&
\includegraphics[height=1.6cm]{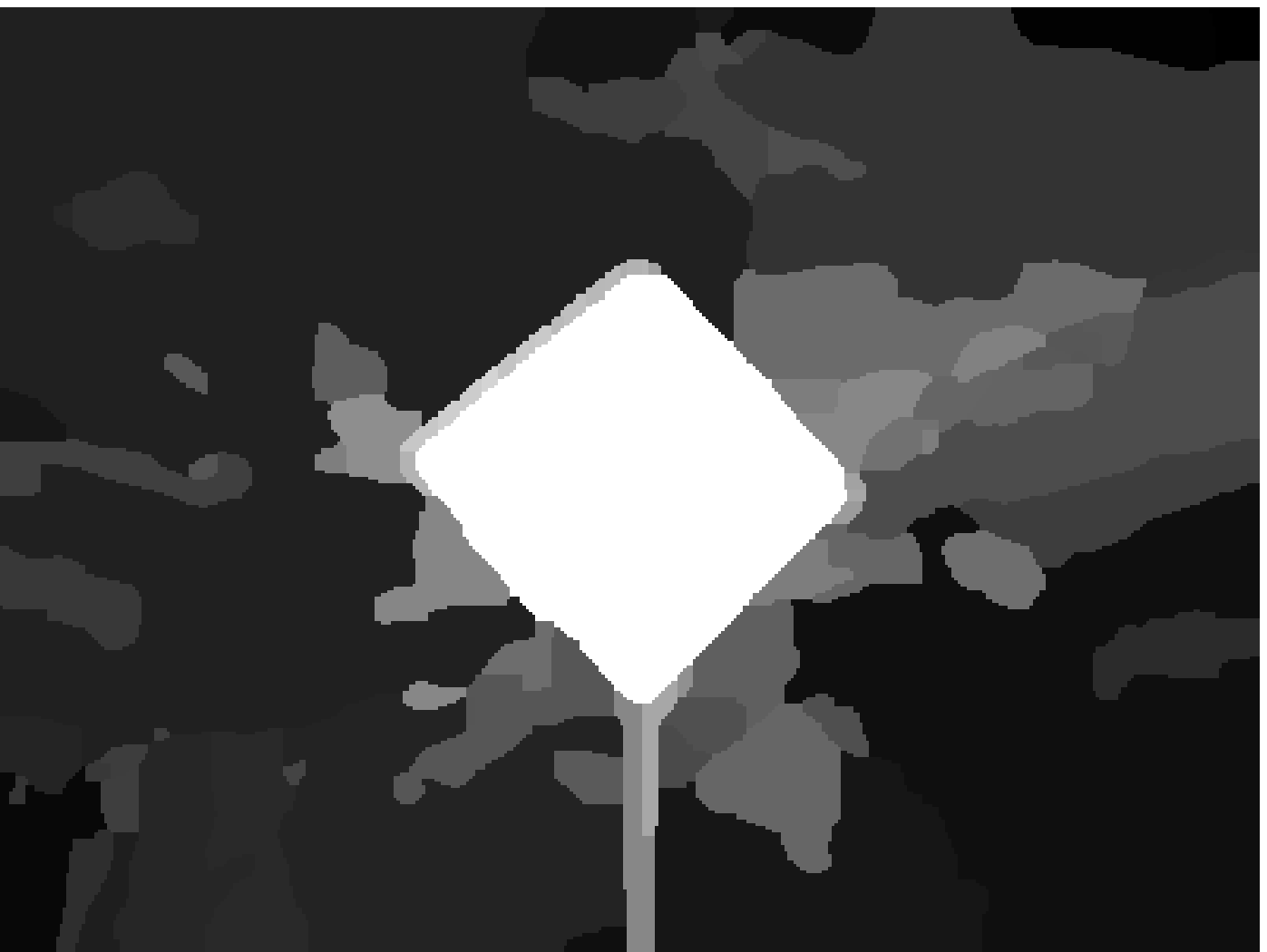} &
\includegraphics[height=1.6cm]{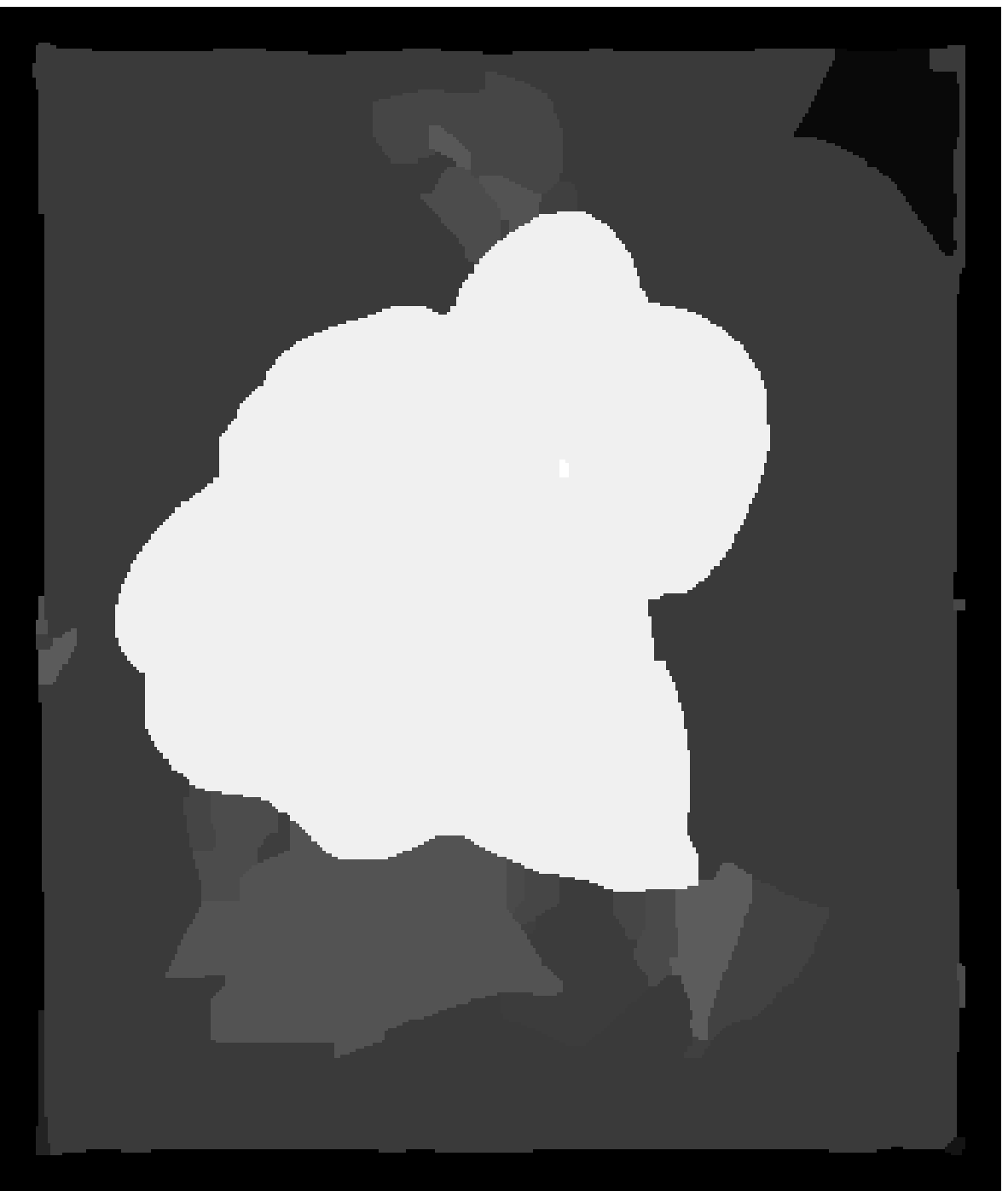} &
\includegraphics[height=1.6cm]{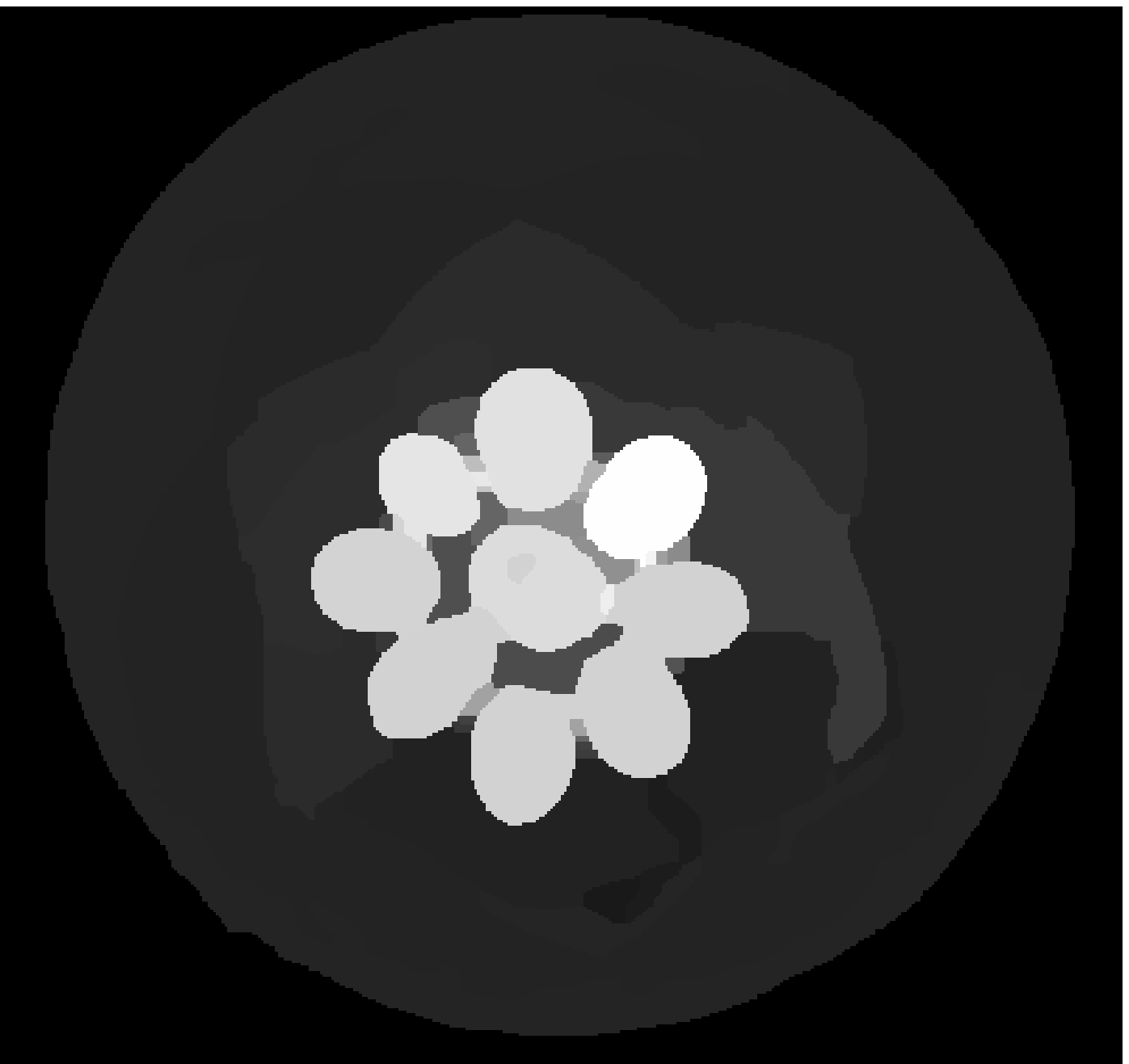} &
\includegraphics[height=1.6cm]{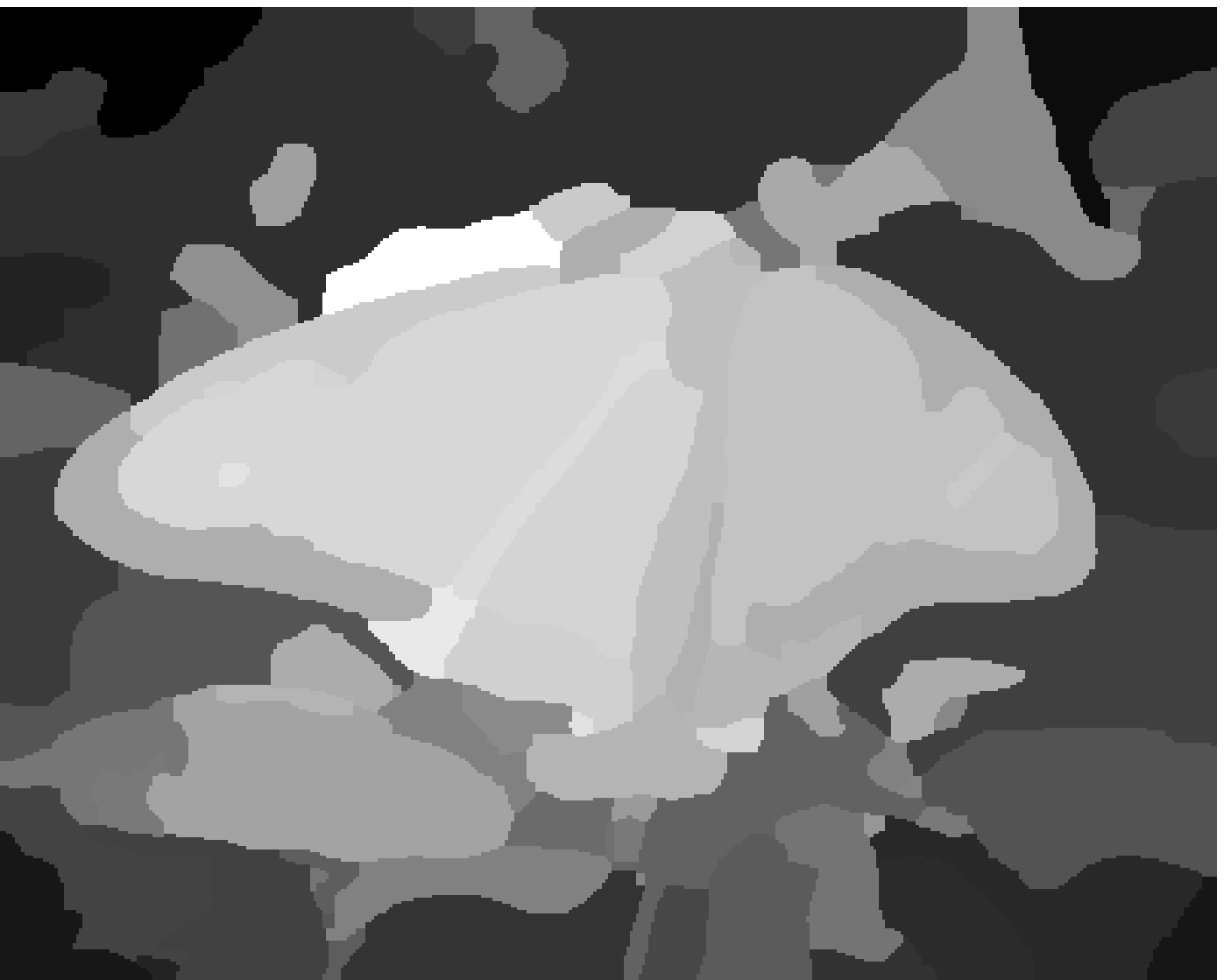} &
\includegraphics[height=1.6cm]{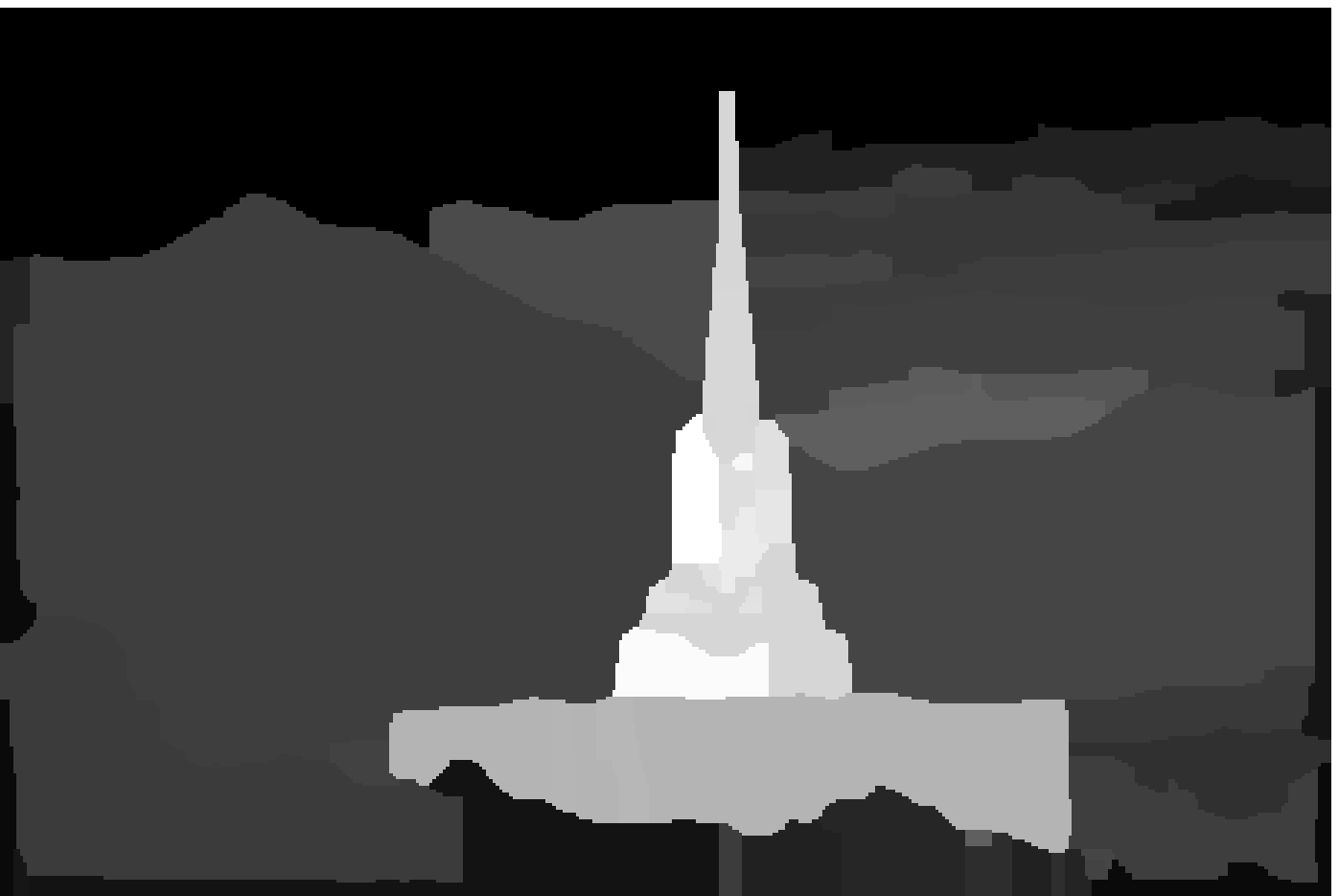} \\

\scriptsize{SOH}&
\includegraphics[height=1.6cm]{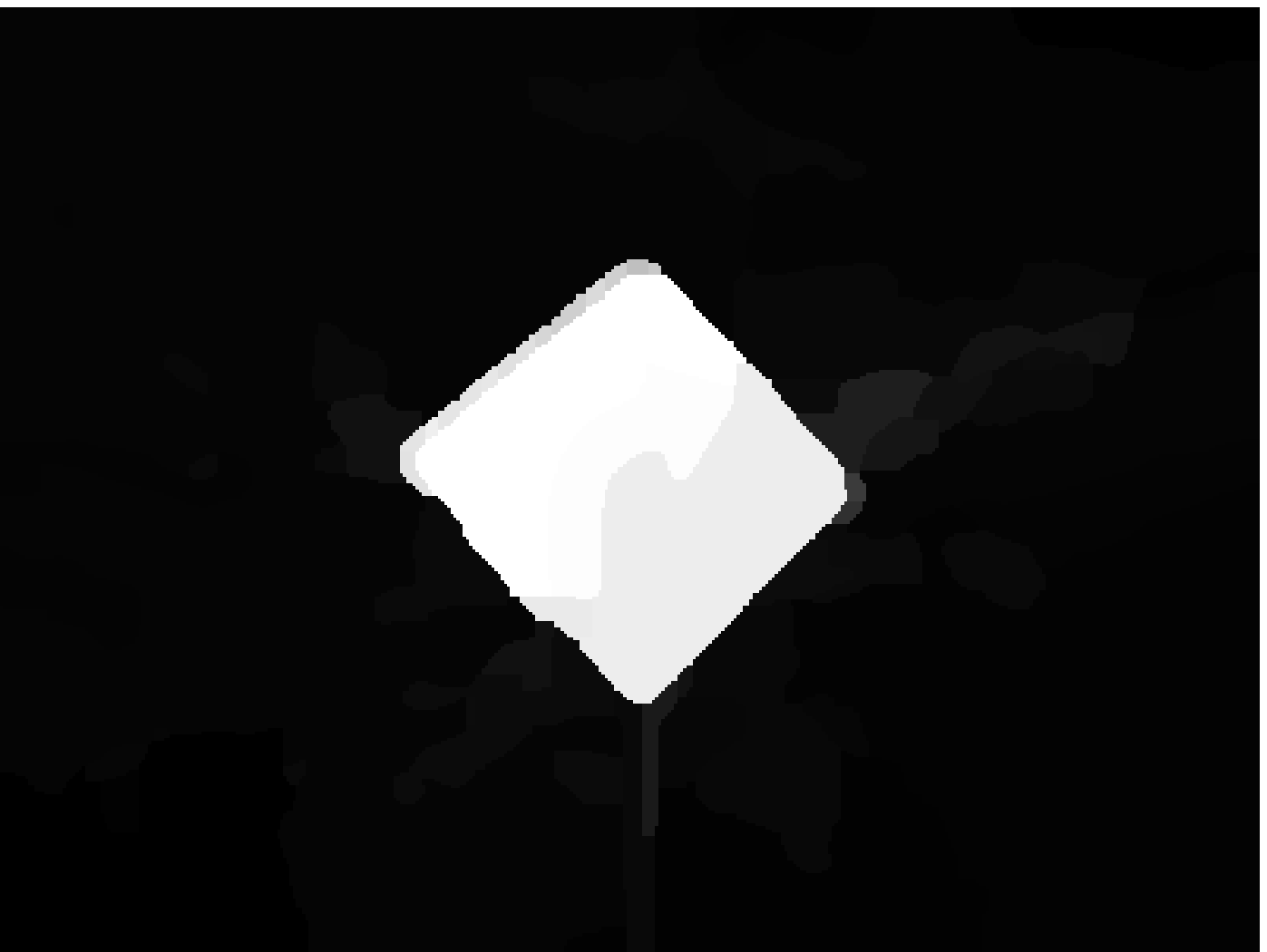} &
\includegraphics[height=1.6cm]{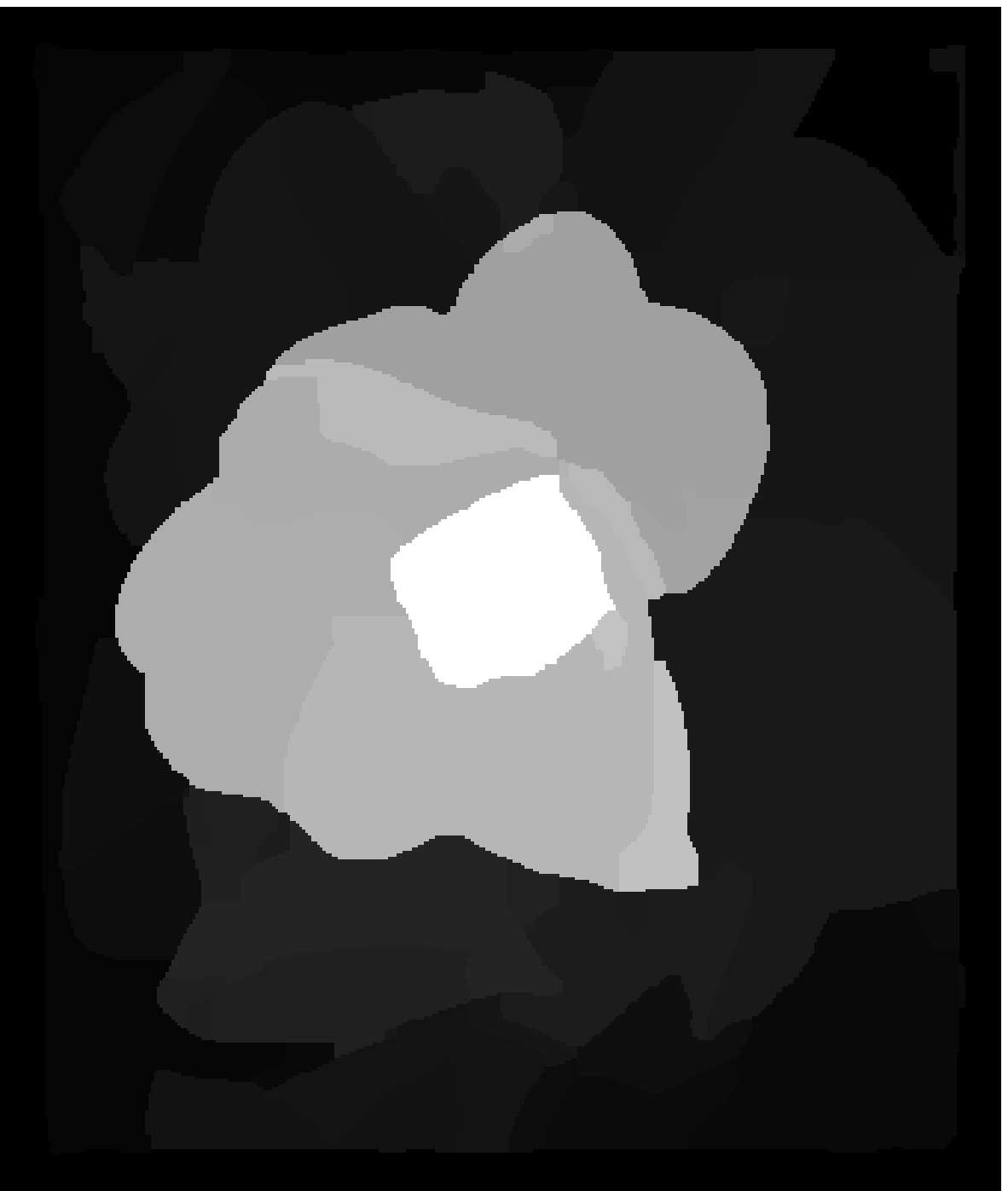} &
\includegraphics[height=1.6cm]{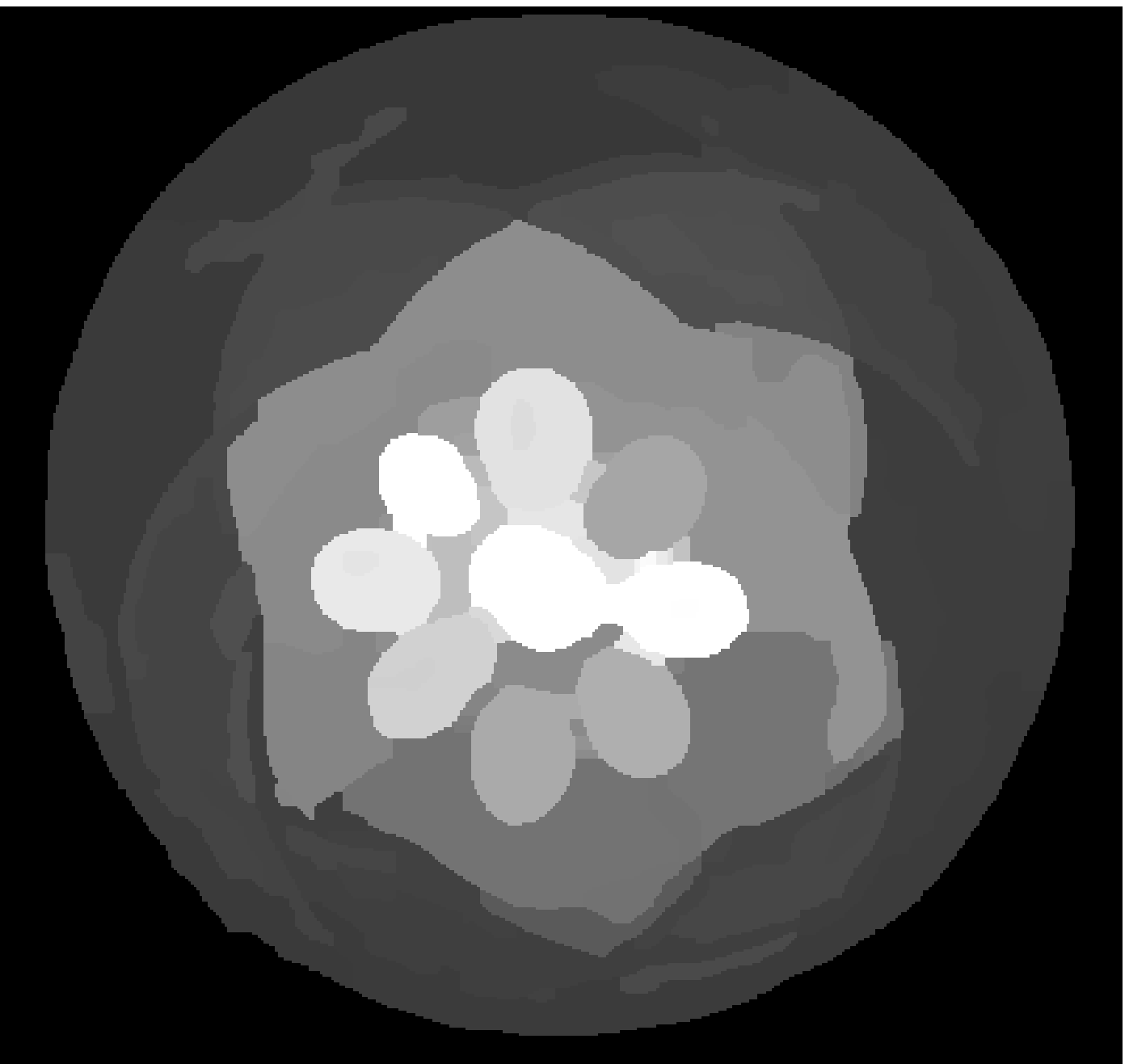} &
\includegraphics[height=1.6cm]{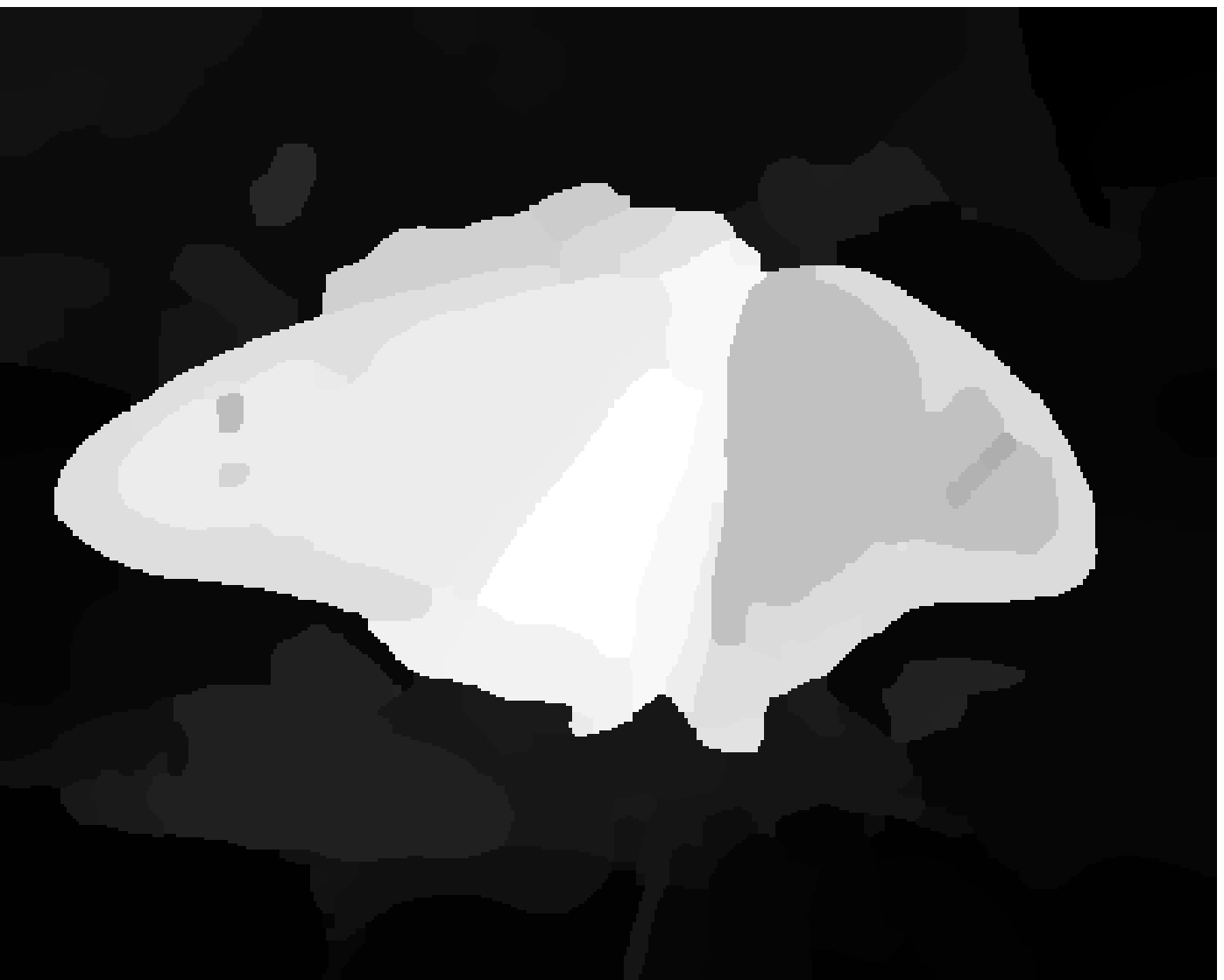} &
\includegraphics[height=1.6cm]{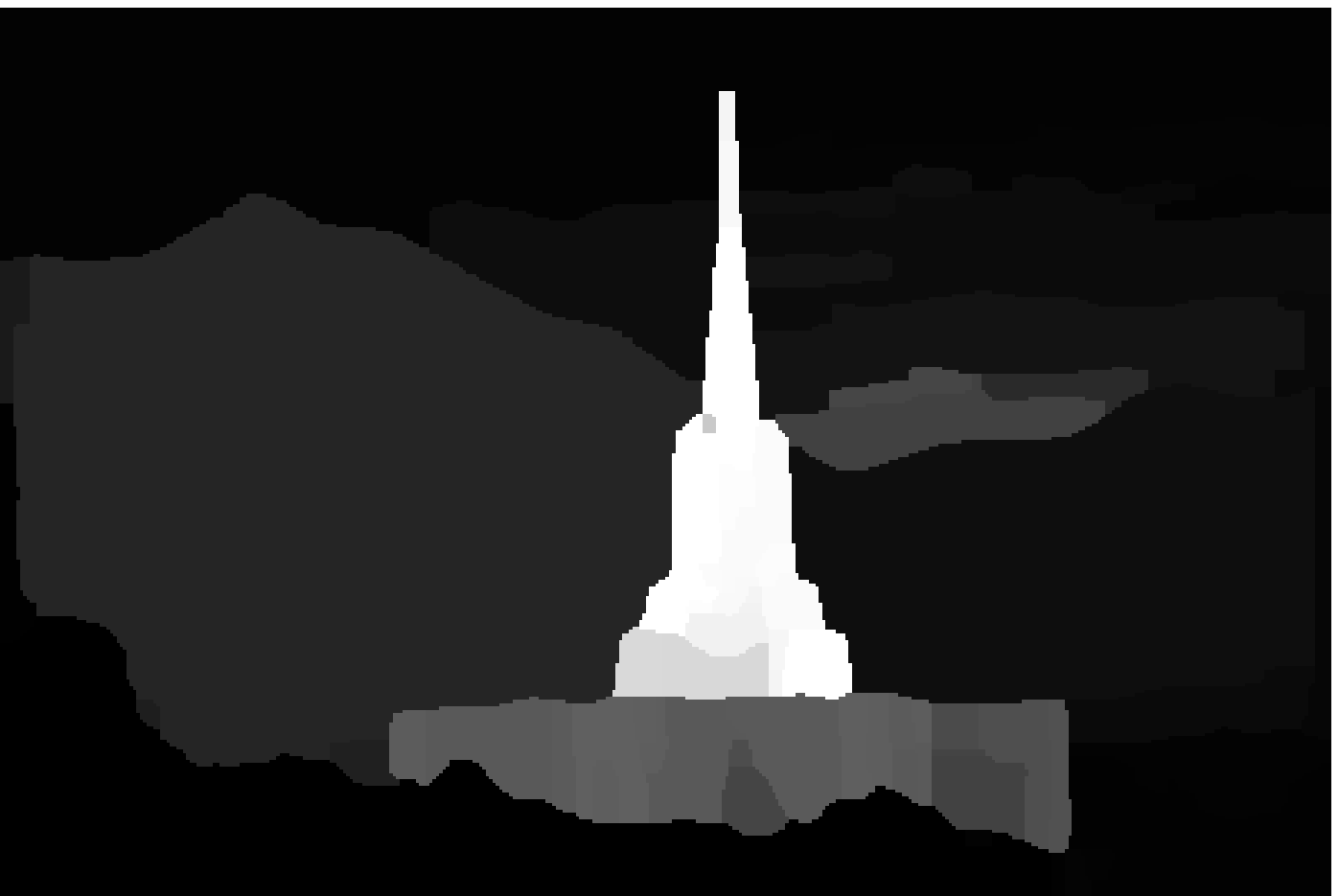} \\

\scriptsize{GT}&
\includegraphics[height=1.6cm]{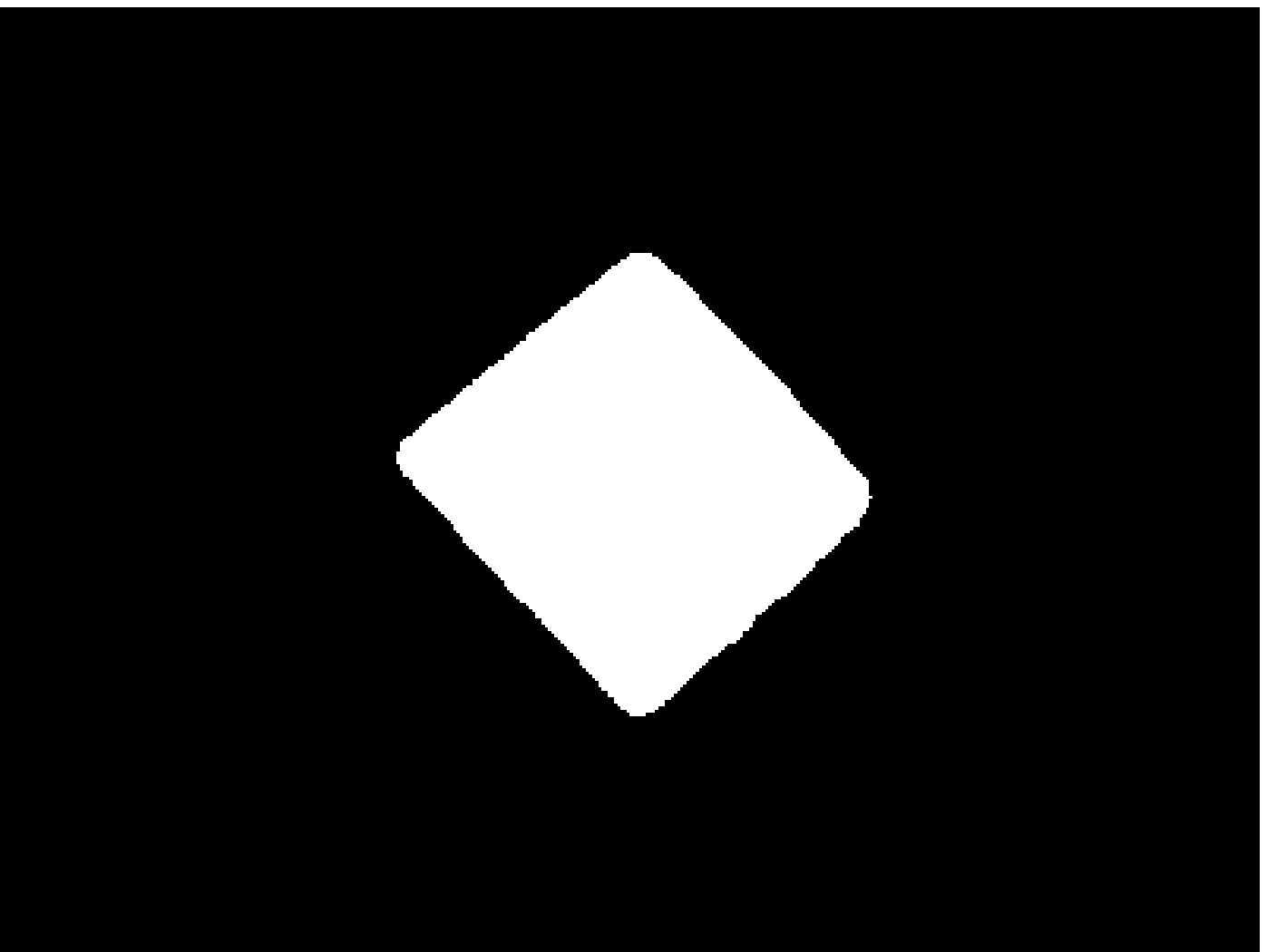} &
\includegraphics[height=1.6cm]{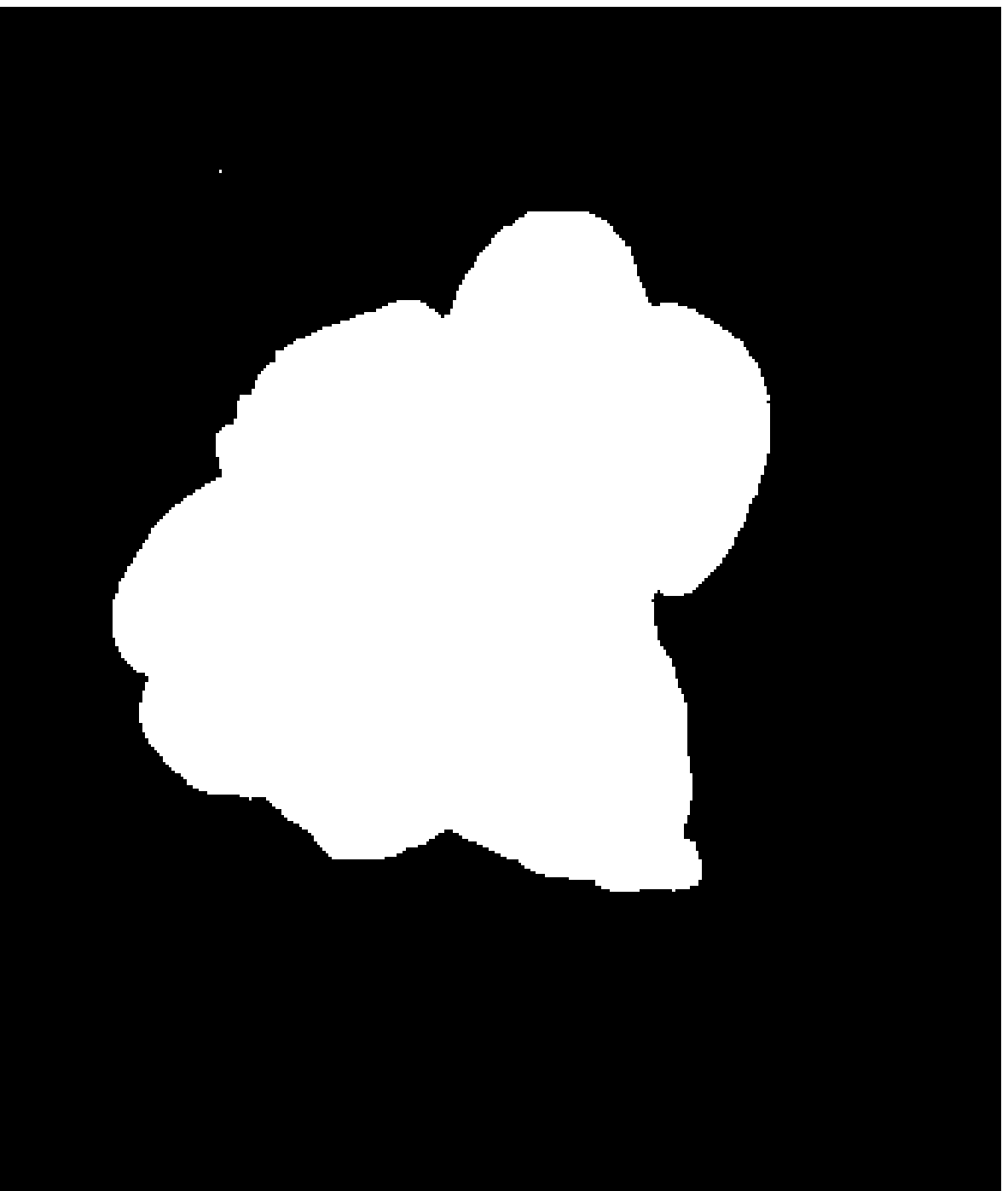} &
\includegraphics[height=1.6cm]{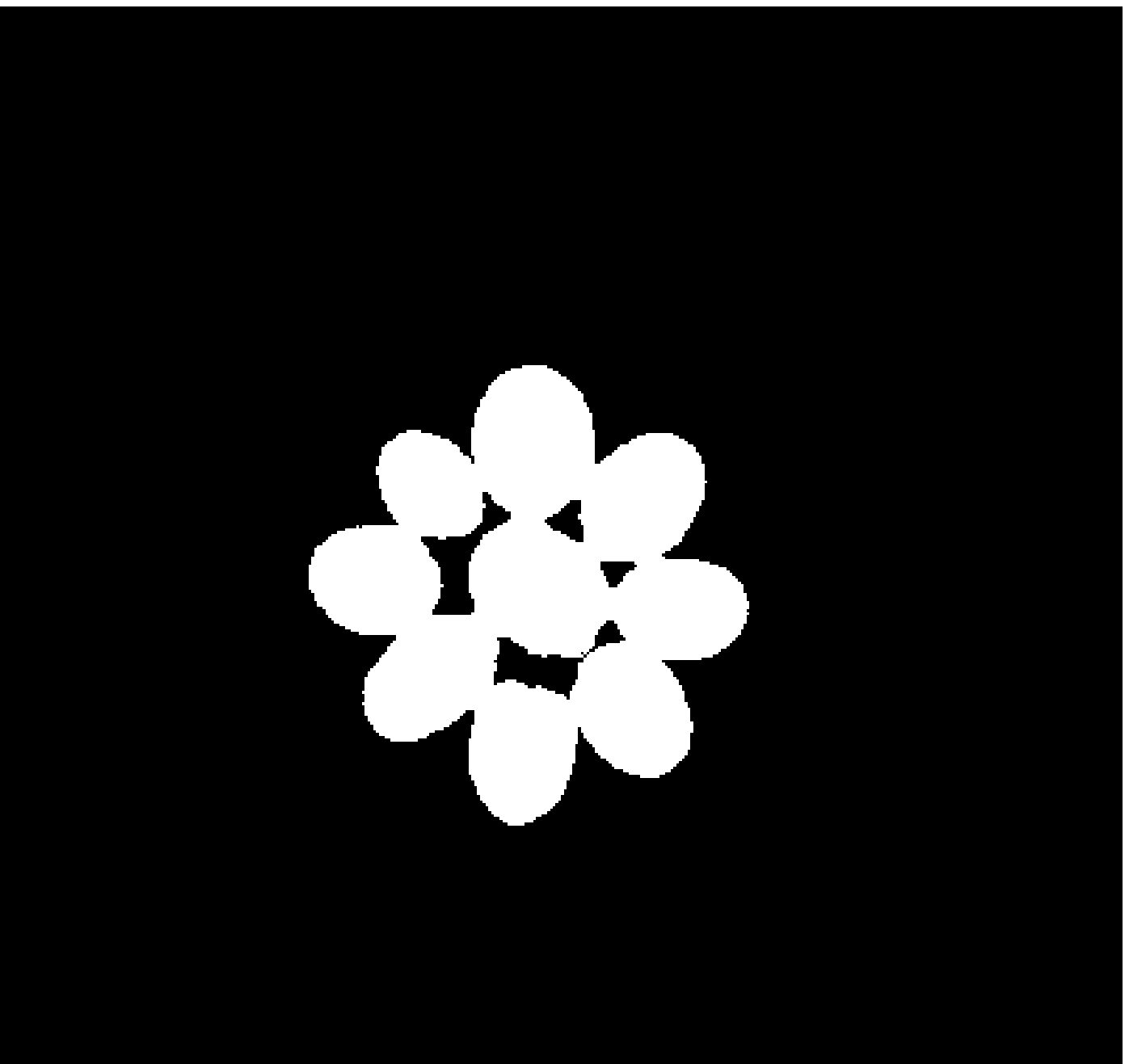} &
\includegraphics[height=1.6cm]{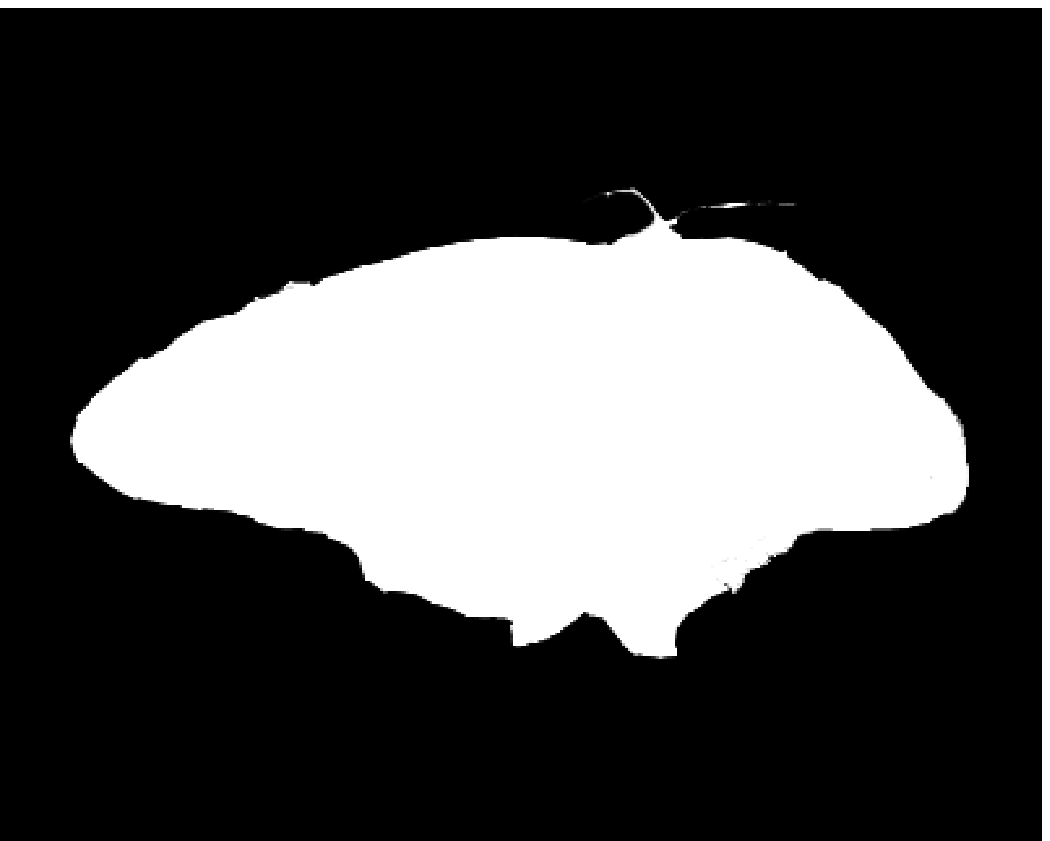} &
\includegraphics[height=1.6cm]{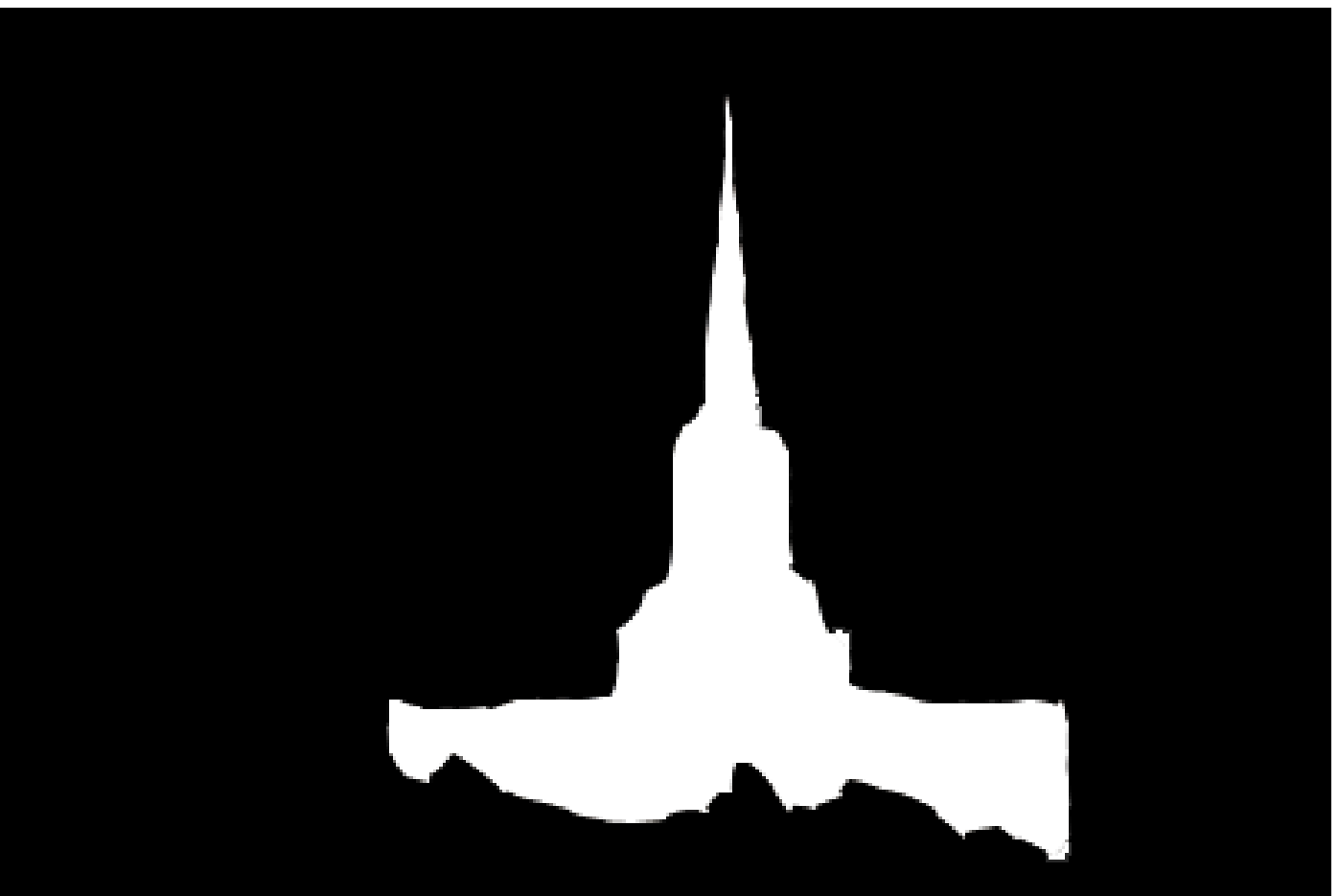} \\
\end{tabular}
\caption{Saliency maps generated using the model based on hierarchies of partitions with different fusion criteria: hierarchical inference with belief propagation (HP-BP) and loopy belief propagation (HP-LBP), mean (HP-MEAN), and max (H-MAX), and using the saliency over the hierarchy (SOH) model. The first and last rows show the original images and ground truth masks, respectively.} \label{fig:fusion_hp_soh}
\end{figure}

Figure \ref{fig:fusion_hp_soh} provides a visual comparison of  maps generated using the two fusion criteria based on  hierarchical inference (BP for Belief Propagation and LBP for Loopy Belief Propagation), and maps obtained using $mean$ and $max$.
The two criteria based on hierarchical inference produce high quality saliency maps, which are very similar to the maps obtained using the $mean$. An evaluation in terms of precision-recall curves over the entire ASD dataset, for both BPT and UCM hierarchies, is presented in the Experiments section.

\setlength{\tabcolsep}{1pt}.

\subsection{Saliency over the hierarchy}
\label{ssec:snodes}

Our second saliency model works directly on the structure created by the hierarchical segmentation. 
We calculate a saliency value for each node in the hierarchy and these values are integrated into one single saliency map.

When using the previous model based on hierarchies of partitions we need to choose the number of partitions and the scales or number of regions in each partition. With the direct model we only need to fix one partition, the finest one, and saliency is measured for all the regions created by the iterative fusion of regions in this partition.

Moreover, saliency is computed at all possible scales while methods based on hierarchies of partitions work at a subset of such scales, that is, they use only some points in the merging sequence.

\begin{figure}[ht]
\begin{center}
\begin{tabular}{ccc}
\includegraphics[width=3cm]{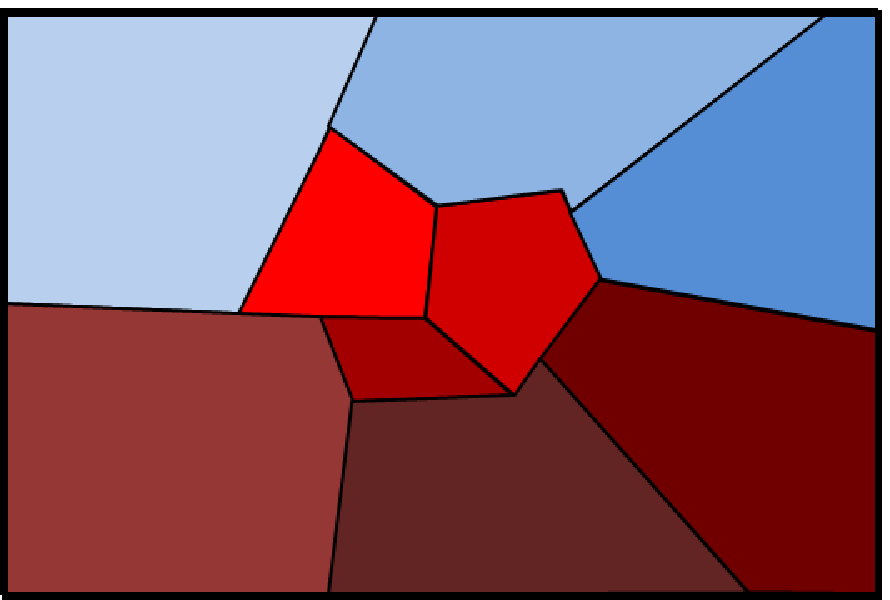} &
\includegraphics[width=3cm]{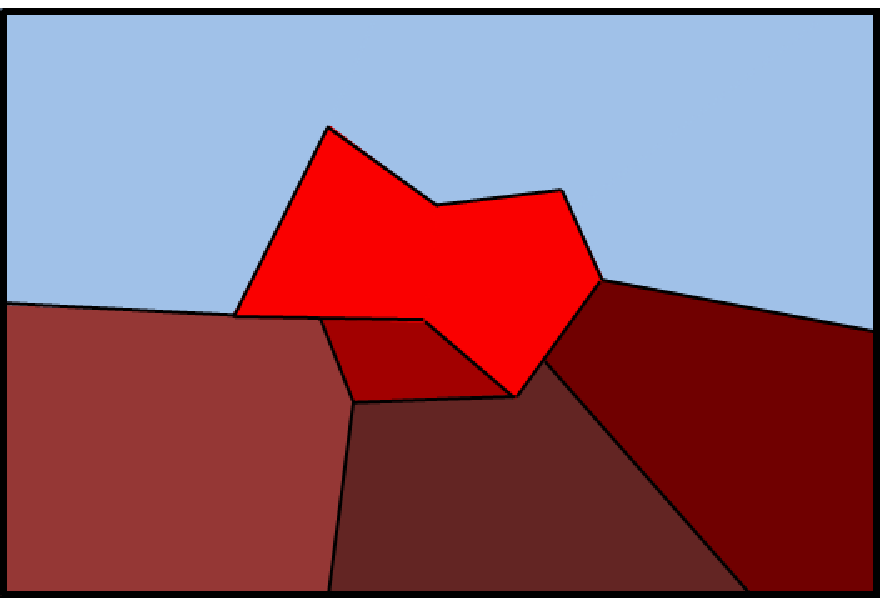} &
\includegraphics[width=3cm]{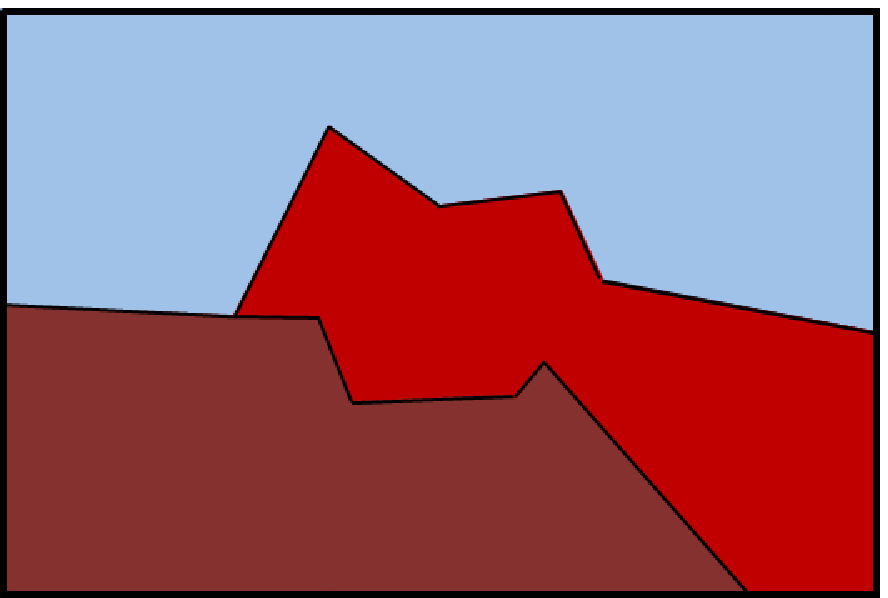} \\
(a) & (b) & (c)\\
\includegraphics[width=3.5cm]{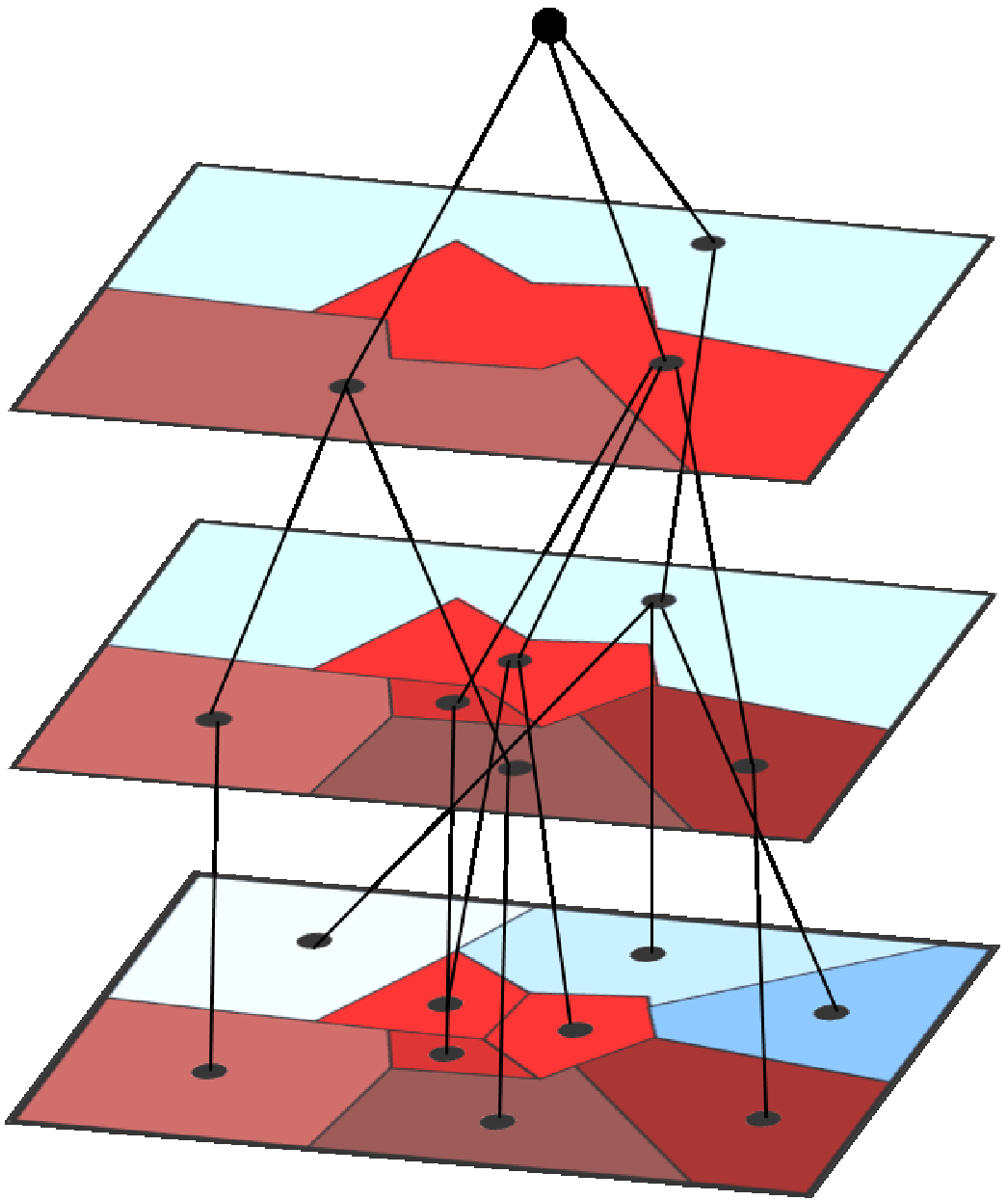} &
\includegraphics[width=3.5cm]{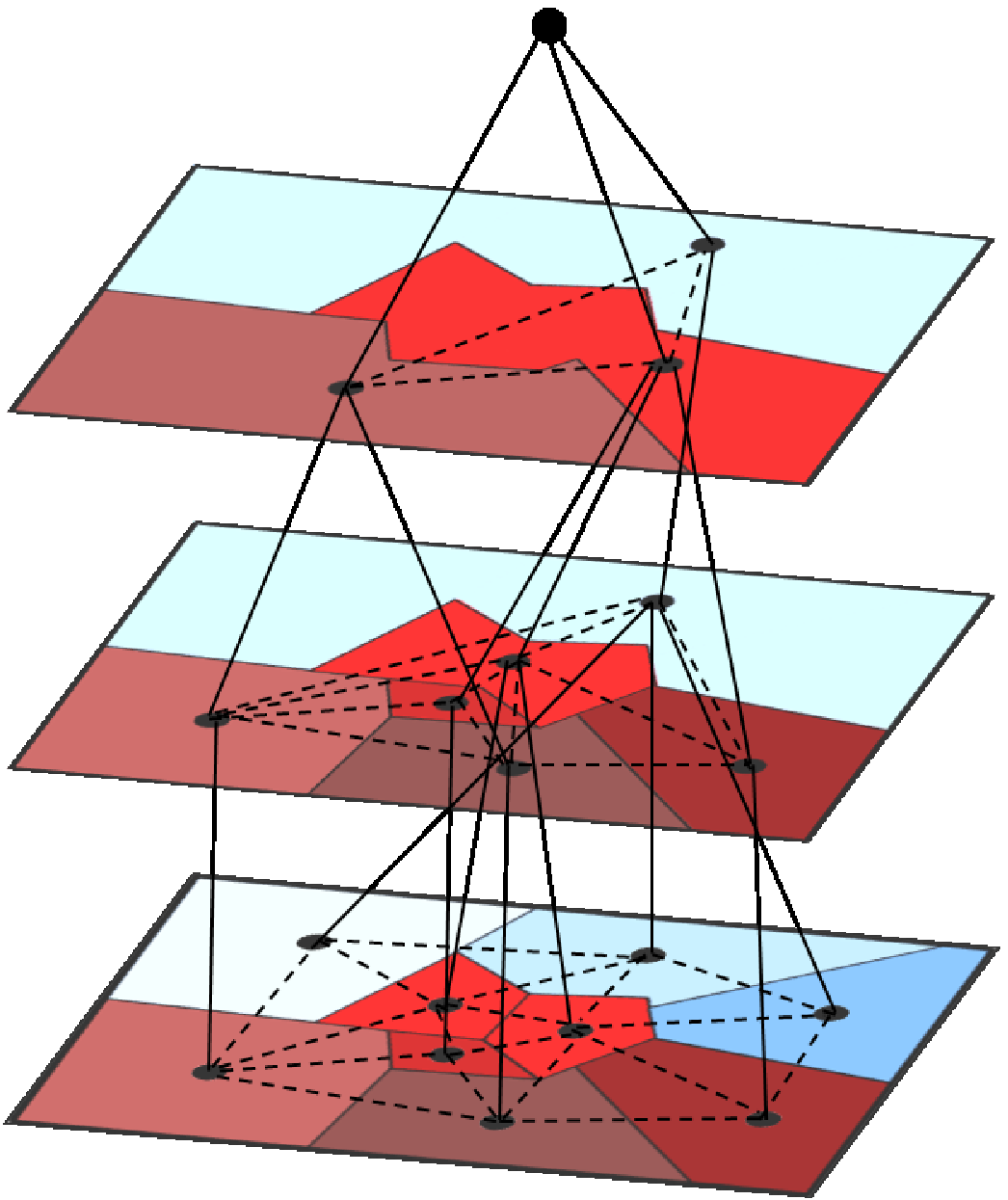} &
\includegraphics[width=3.5cm]{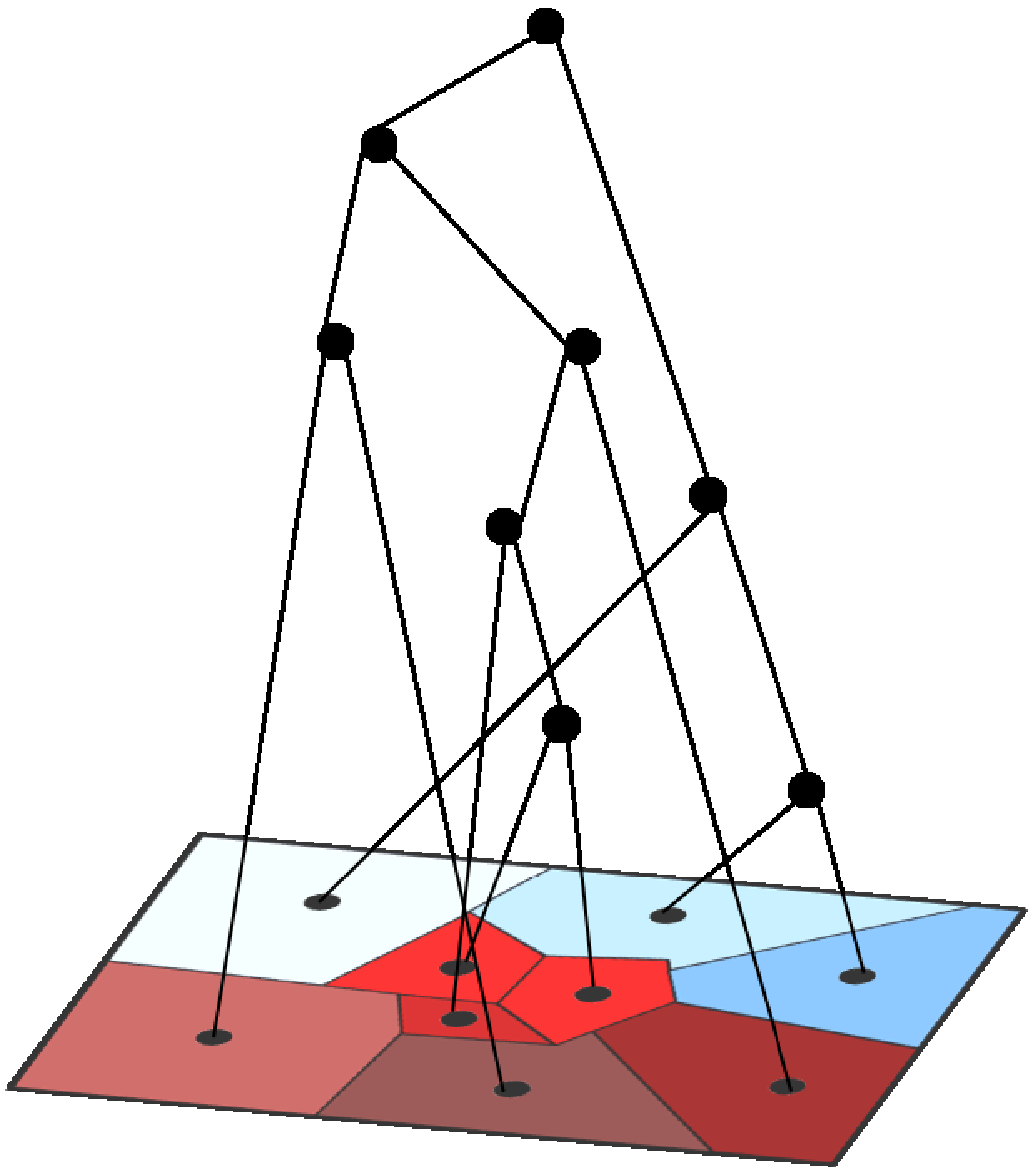} \\
(d) & (e) & (f)\\
\end{tabular}
\caption{Hierarchy of 3 partitions with 9 (a), 6 (b) and 3 (c) regions. Graph models used for hierarchical inference without (d) and with  (e) connections between neighboring regions at each level. Hierarchical structure representing the sequence of mergings starting from the finest partition (f).}
\label{fig:inference}
\end{center}
\end{figure}

This finer analysis is performed, however, without increasing the computational complexity, because only $2N-1$ saliency values are computed for an initial partition with $N$ regions. This number may be lower than the number of saliency values measured when using a hierarchy of partitions (see example in Figure \ref{fig:inference}).

The saliency of a region $R_i$, relying on contrast, boundary and center priors, is:
\begin{equation}
S(R_i)=w_b(R_i) w_c(R_i) \sum_j  w_s(R_i,R_j) |R_j| w_b(R_j)d(M_{R_i},M_{R_j})
 \label{eq:hrsal}
\end{equation}
where, as before,  $d(M_{R_i},M_{R_j})$ is a distance measure between regions, $w_s(R_i,R_j)$ is a weight based on the distance between regions (eq.\ref{eq:sweight}), $w_b(R_j)$ is the boundary weight factor that depends on the fraction of region that intersects the image boundary (eq.\ref{eq:wb}), $w_c(R_i)$ is a center weight factor estimated as the average distance between each region pixel and the image center.

The computation of the saliency $S(R_i)$ for a leaf node (a region in the initial partition) $R_i$ is direct: the  sum is over all the regions in the initial partition.

For a non-leaf node, we need to keep track of the sequence of mergings. A non-leave node (region) is created at some point in the hierarchical segmentation algorithm by merging two children nodes. The saliency of the new region $R_i$ is measured using color and spatial differences with all the other regions $R_j$ in the partition defined by this point in the merging sequence. Note, however, that there is no need to explicitly find this partition, we only need to keep track of the regions that are merged at each step, and update the region descriptors (color histogram, region centroid, size) that are used in the saliency computation.

The integration of the saliency values into a saliency map is straightforward. For each pixel $x$, we  average the saliency values of all the regions containing the pixel: 
\begin{equation}
S(x)=\sum_{i / x \in R_i} S(R_i) / N_x
\end{equation}
where $N_x=|\{ i / x \in R_i \}|$ . 

Note that the integration is different from the \emph{mean} method used with the hierarchy of partitions. $N_x$ is the number of regions in the hierarchy that contain $x$, so it may change for different values of $x$.

We also tried alternative methods (max, hierarchical inference) to integrate the local saliency values but performance was not so good.
Some examples of saliency maps created directly over the hierarchy are provided in Figure \ref{fig:fusion_hp_soh} (SOH row).

\section{Experiments}
\label{sec:experiments}

This section explains the experiments conducted to analyze the performance of our models for different choices of the  parameters and fusion criteria. We first describe datasets and evaluation metrics.

\subsection{Datasets and metrics}
\label{ssec:datasets}
The datasets used in the experiments are the following. All of them include ground truth pixel-wise binary masks of salient objects.
\begin{enumerate}
\item{MSRA} \cite{Achanta-09cvpr} contains 5000 natural images with large variety in content.
\item{ASD} \cite{Achanta-09cvpr} contains 1000 natural images selected from the MSRA set. Images present large variety in the content but the background structures are simple. It is the most commonly used set for evaluation. 
\item {ECSSD} \cite{Yan-cvpr13}, the Extended Complex Saliency Dataset, contains 1000 images with diverse patterns in both foreground and background, collected from Berkeley Segmentation Dataset BSD300, PASCAL VOC2012 segmentation challenge dataset and internet.
\end{enumerate}
MSRA and ASD are the sets where the notion of saliency is much less ambiguous. ECSSD contains images that where not collected or annotated specifically for saliency evaluation, but for other tasks: object detection, segmentation and recognition. Therefore, in many cases pixel-accurate masks are created only for instances of the object categories analyzed (e.g. in PASCAL VOC) but not for other salient objects in the image. Moreover, in many cases, masks correspond to objects that are not visually salient. 

\bigskip

For objective evaluation we use precision and recall analysis. Precision measures the percentage of salient pixels correctly detected, while recall measures the percentage of salient pixels detected. We follow the usual approach and binarize the saliency maps using each possible fixed threshold in $[0,255]$. Next, we compute precision and recall values for each threshold between 0 and 255, average them over the number of images in the dataset and plot precision-recall curves.

To evaluate the overall performance we also use an image dependent adaptive threshold \cite{otsu79} to obtain one single binary mask for each saliency  map. Then we calculate the average precision, recall and $F_\beta$-measure for $\beta=1$.

Recently, Margolin \emph{et al.}\cite{margolinEval14} showed that the previous measures do not always provide a reliable evaluation and identified the causes of inaccurate evaluation. They proposed a new measure based on the extension of the basic TP, TN, FP, FN quantities to non-binary values, and weighting errors according to their location and neighborhood. The measure is an extension of the $F_{\beta}$ measure named weighted $F_{\beta}^w$  measure, and we also use it to evaluate the proposed models. Our implementation of this measure is based on Matlab code provided by the authors.

Finally, for a more balanced comparison of results, that is, to avoid favoring methods that correctly find salient pixels but fail to detect non-salient regions over methods with the opposite behavior, we also use the mean absolute error (MAE) measure proposed by Perazzi \emph{et al.}\cite{perazzi-cvpr12}
\begin{equation}
MAE=\frac{1}{| I |}\sum_x{|S(x)-G(x)|}
\end{equation}
where $S$ is a continuous-valued saliency map, $G$ is the ground truth image, $|I|$ is the number of pixels in the image $I$.

\subsection{Models and parameters}

The first experiments compare the performance of our first model, based on hierarchies of partitions, for the two segmentation techniques, BPT and gPb-UCM, and for different configurations: local or global contrast, region model based on mean color or histogram, and use of the boundary prior, on the ASD dataset.

\begin{figure}[htbp]
\centering
\begin{tabular}{cc}
\includegraphics[height=4.5cm]{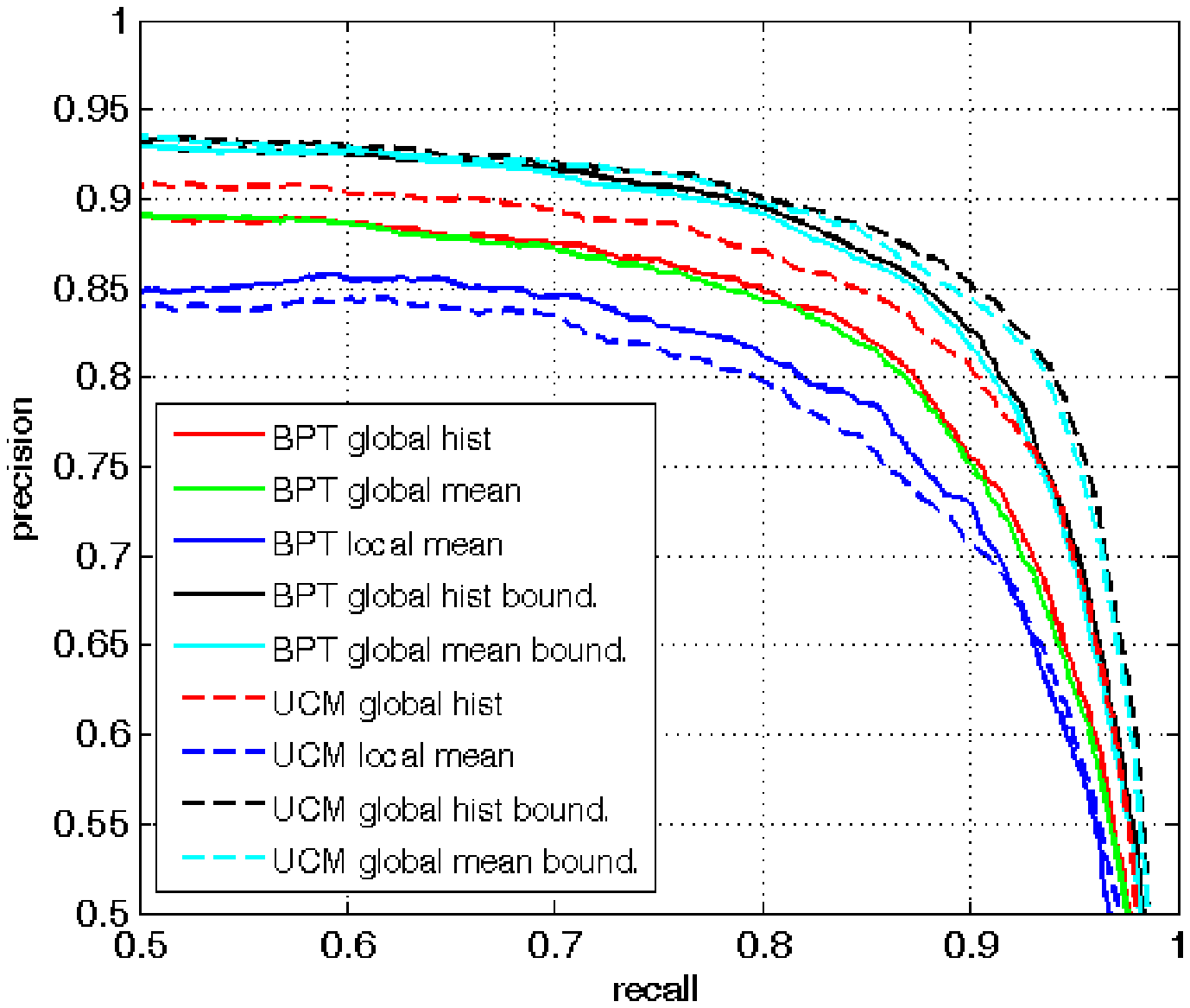}&
\includegraphics[height=4.5cm]{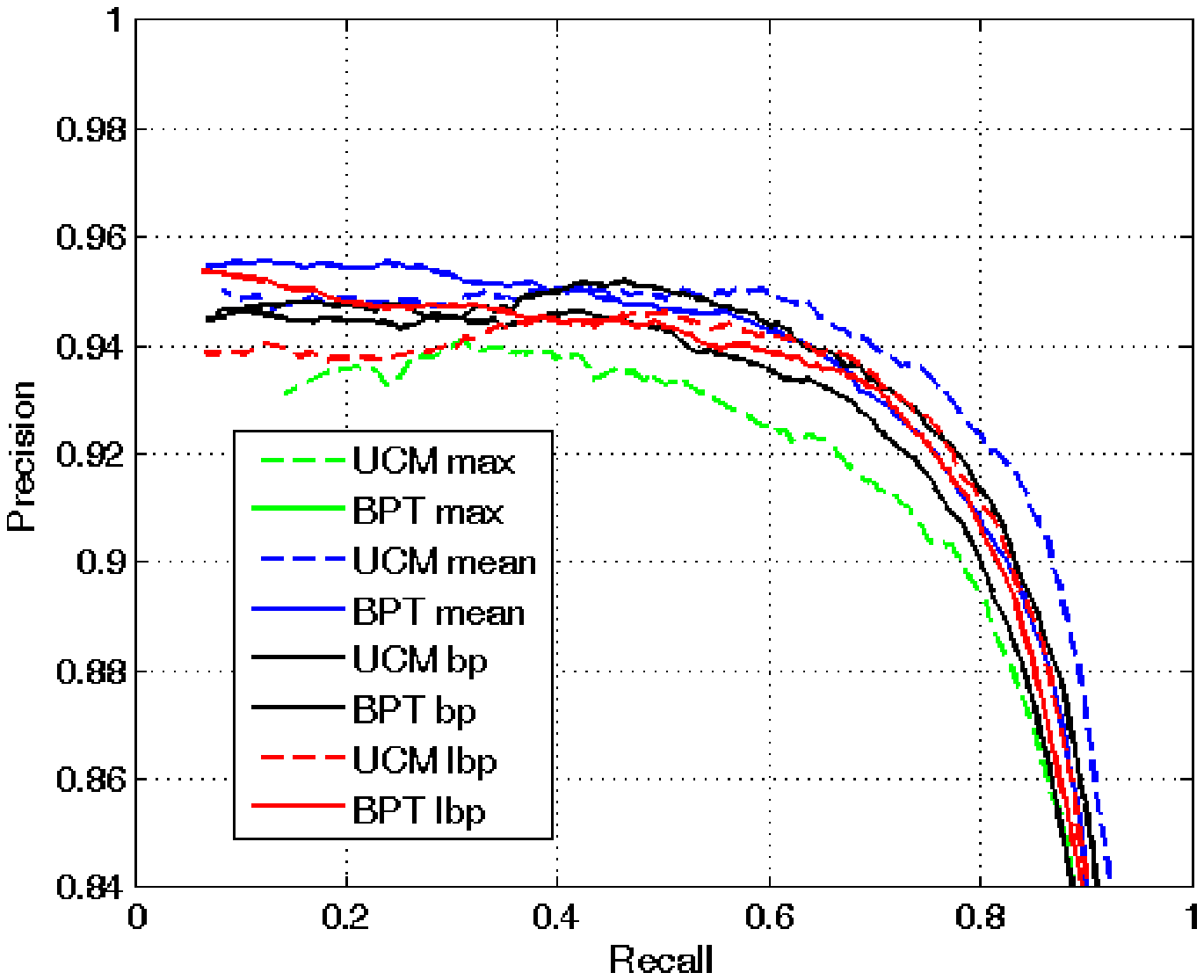}\\
(a) & (b)\\
\end{tabular}
\caption{Precision-recall curves for the model based on hierarchies of partitions. Left (a):  'hist' and 'mean' denote the region model, 'global' or 'local' is the type of contrast, 'bound.' means that we use the boundary prior. The fusion rule is always the $mean$. Right (b): BPT and UCM segmentation with different fusion criteria: mean, max, hierarchical inference with or without loops.}\label{fig:pr_param}
\end{figure}

For space reasons we only present some of the results. In all the experiments we use a hierarchy of 10 levels, with the $max$ fusion criterion.
The number of regions in each level is defined as follows: we fix the number of regions for the initial and the final partition (first and last levels in the hierarchy) and the intermediate numbers are calculated following a geometric progression. For the experiments, we use 100 and 3 regions for the first and the last partitions, respectively. Using more regions in the first level does not improve the results for the ASD dataset.

Figure~\ref{fig:pr_param} (a) shows precision-recall curves for different configurations. UCMs generally perform slightly better than BPTs, and all the configurations improve when using the boundary prior. The best results are obtained for UCMs with global contrast, the histogram model and boundary prior.
We observed a similar behavior for these parameters changing the of number of levels and fusion criteria.

\begin{figure}[htbp]
\centering
\begin{tabular}{cc}
\includegraphics[height=4.5cm]{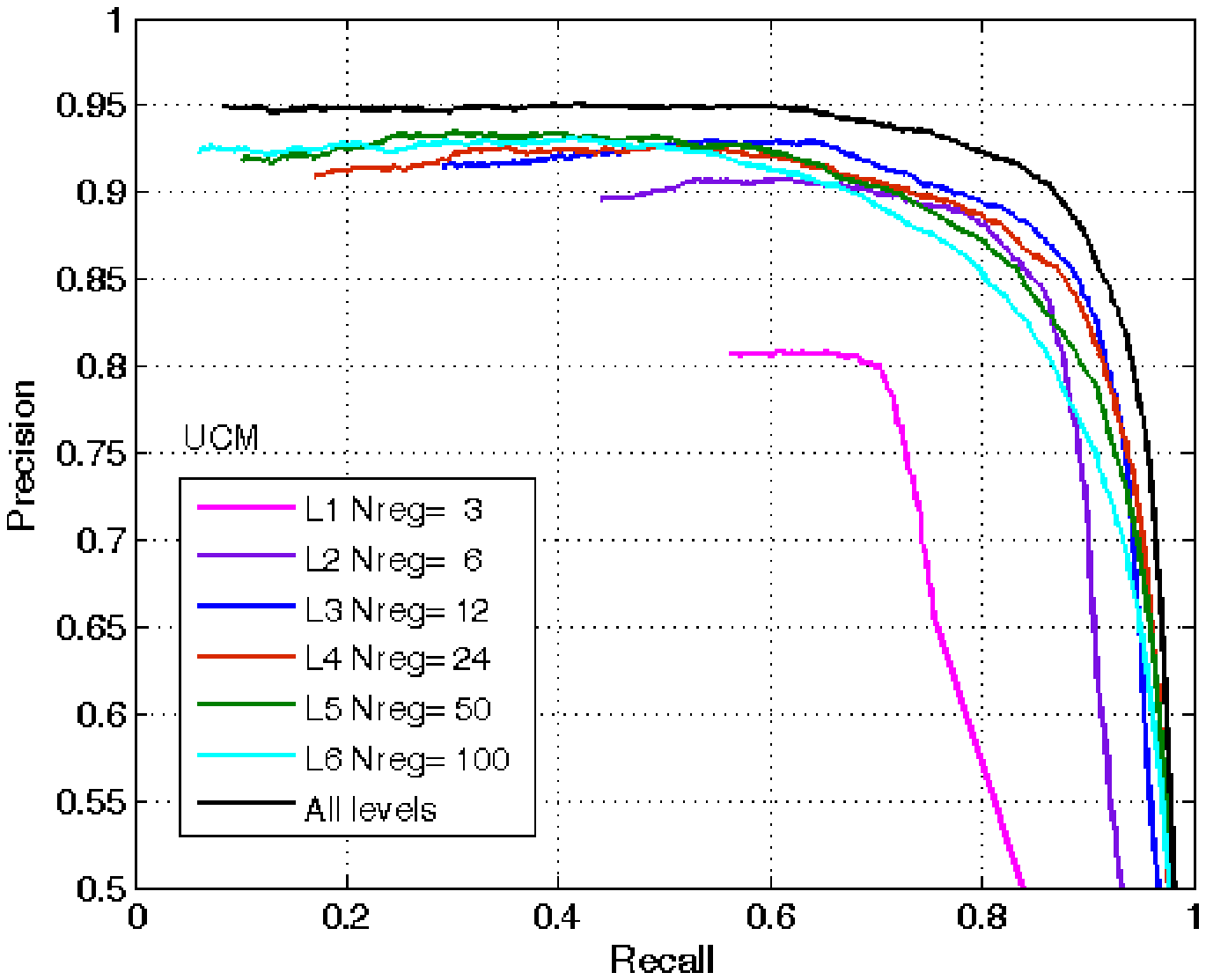} &
\includegraphics[height=4.5cm]{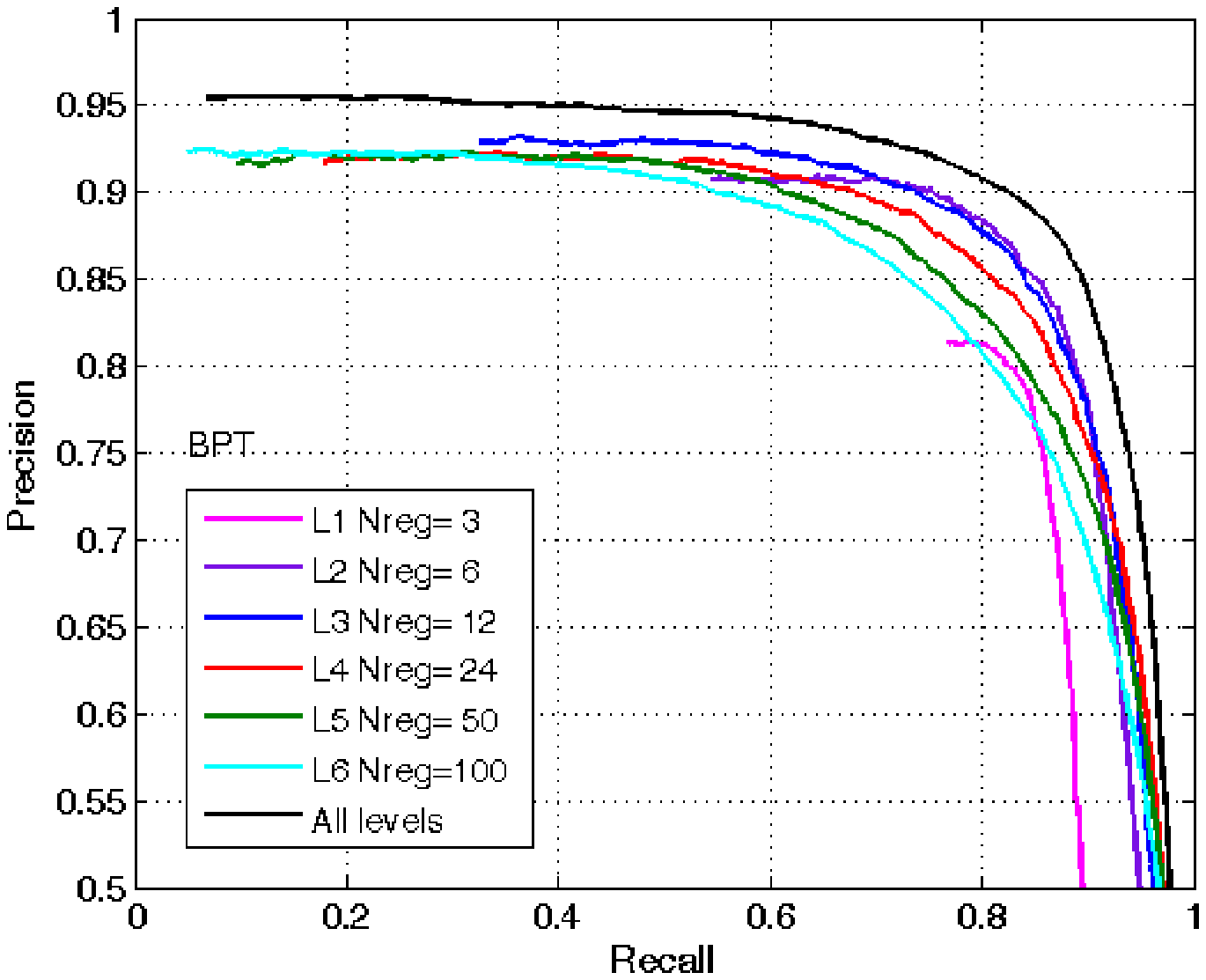} \\
\end{tabular}
\caption{Precision-recall curves for saliency maps generated for different levels and for the final map, UCM (left) and BPT (right).}\label{fig:pr_levels}
\end{figure}

The second experiments compare the performance of  the model based on hierarchical partitions for the two segmentation techniques, BPT and gPb-UCM, and for different fusion criteria: max, mean, and hierarchical inference using belief propagation or looping belief propagation. Here the number of levels is fixed to 6, and the number of regions again follows a geometrical progression, from 100 to 3.

Precision-recall curves presented in  Figure \ref{fig:pr_param}(b) show a slightly better performance for UCM and the $mean$ criterion over hierarchical inference and $max$. It has been argued in favor of inference-based fusion criteria that averaging maps is not a good choice considering possible complex background or foreground \cite{Yan-cvpr13, Zhu-cvpr14}. However in our experiments we obtain similar results for the two strategies, being the $mean$ method much faster than hierarchical inference.

\begin{figure}[htbp]
\centering
\begin{tabular}{c}
\includegraphics[height=4.5cm]{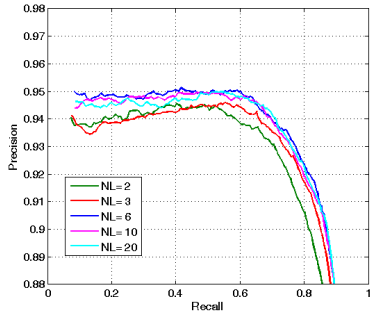} 
\end{tabular}
\caption{Precision-recall curves for saliency maps generated using different number of levels in the hierarchy.}\label{fig:pr_nlevels}
\end{figure}

\bigskip
Our model based on hierarchies of partitions utilizes information from multiple levels to obtain the final saliency map. Figure \ref{fig:pr_levels}(a) presents the precision-recall curves for each level in the hierarchy, and for the final map, and illustrates the usefulness of the approach. The final map achieves the highest precision in the entire recall range.
For this experiment we use global contrast on a hierarchy of 6 partitions generated with UCM and regions modeled with the histogram, and the $mean$ criterion for the integrated map.

We also observe that the behavior is not monotonic. The performance does not always increase or decrease with the number of regions in the partition. The best results for a single level are obtained for level 3 which corresponds to partitions with 12 regions. Note that level 3 is the best single level for the ASD dataset, but may not be the best for other datasets. The result is probably related to the number and size of salient objects in ASD images.

The same behavior is found when using BPT for the segmentation. Precision-recall curves are shown in Figure \ref{fig:pr_levels} (b) .

Finally, the last experiment compares precision-recall curves for hierarchies of partitions created with different number of partitions. We present some of the results, using hierarchies with 20, 10, 6, 3 an 2 levels in Figure \ref{fig:pr_nlevels}. In all these hierarchies we include a partition with 12 regions (the best performing single-level result obtained in the previous experiment). Best results are obtained for partitions with 6 or more levels. Since we do not observe a significant improvement  using hierarchies with more than 6 levels, all the experiments in the following subsection, comparing the performance of our saliency models with other approaches, will use 6 levels.

\subsection{Experimental comparisons}

\begin{figure}[thbp]
\centering
\begin{tabular}{cc}
\includegraphics[height=4.5cm]{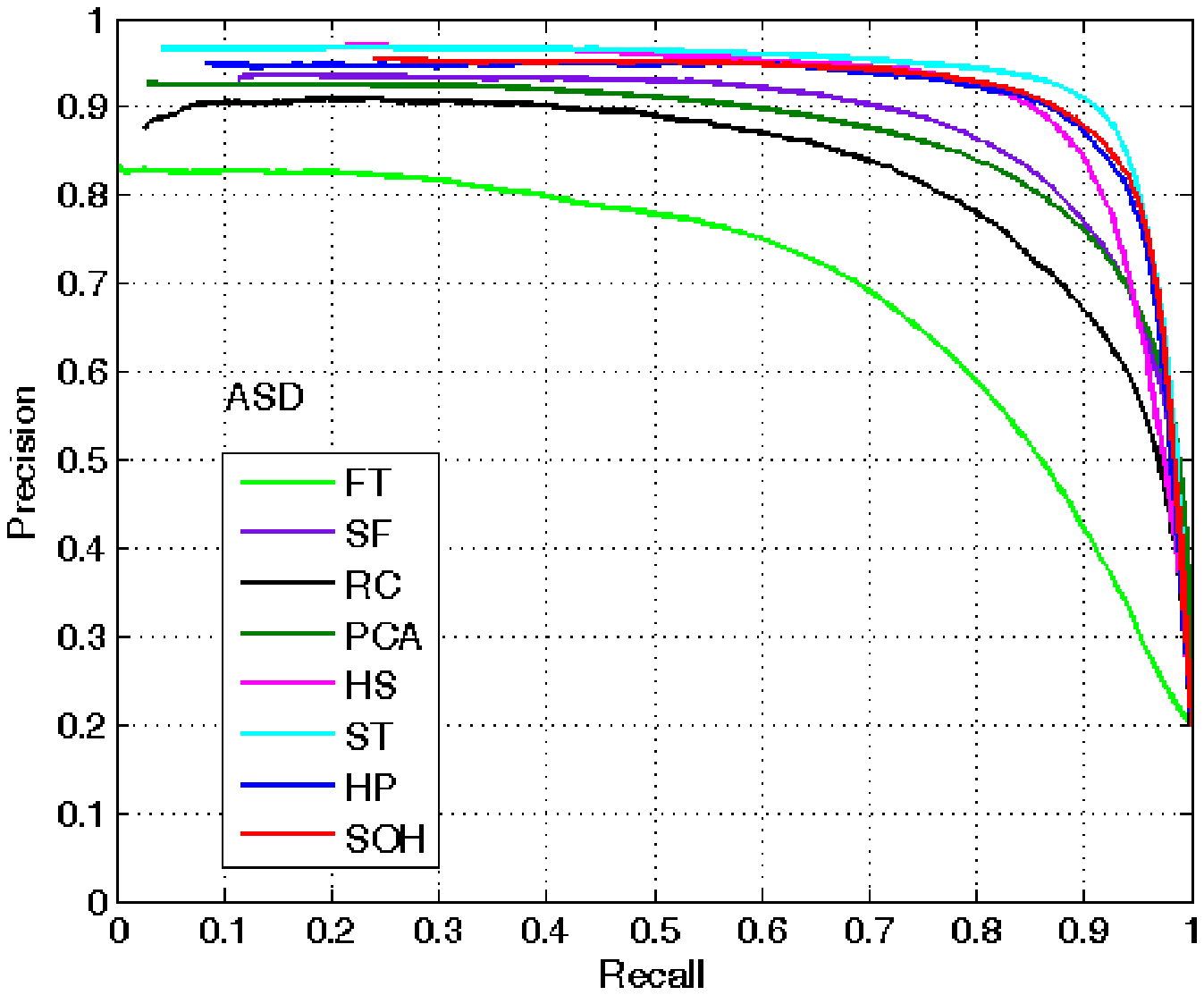} &
\raisebox{.2\height}{\includegraphics[height=3.5cm]{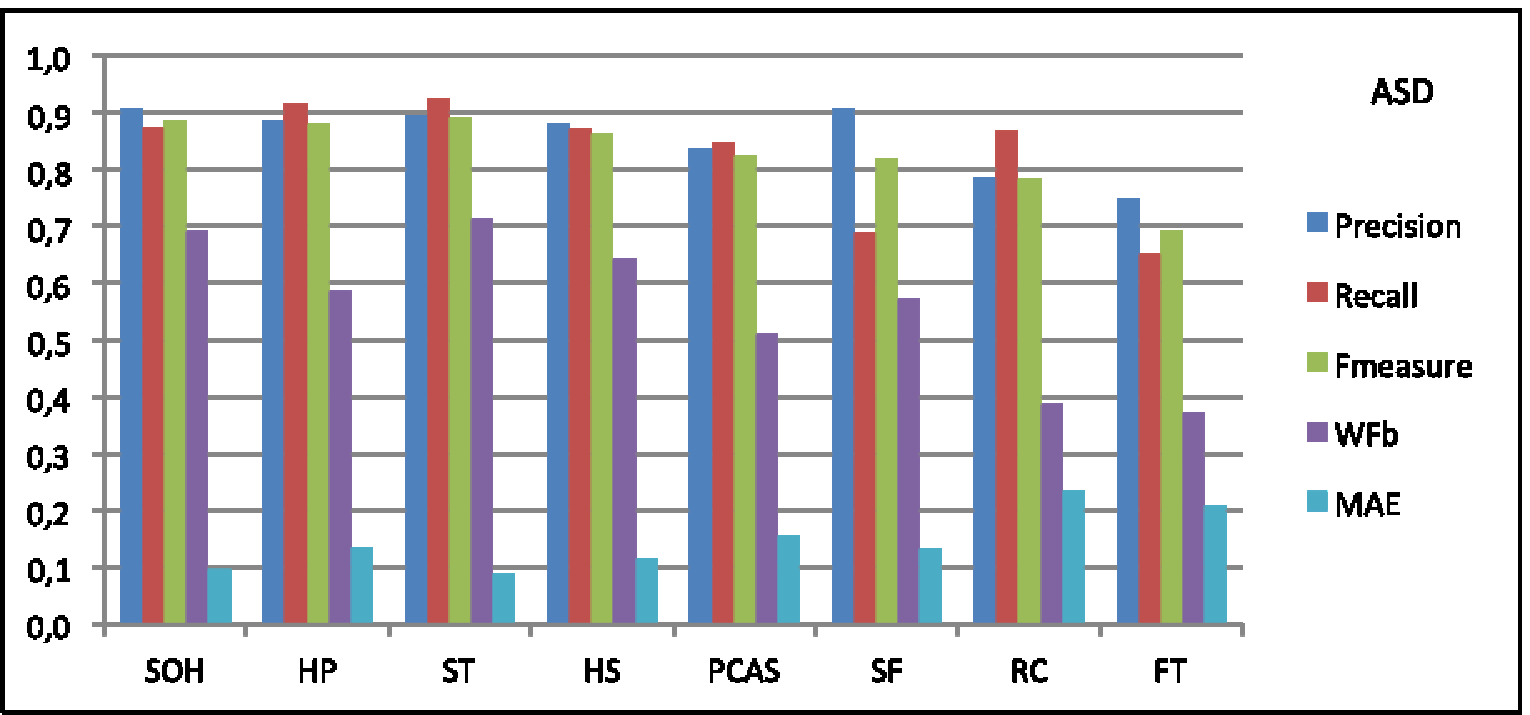}} \\
(a) & (b)\\
\end{tabular}
\caption{Precision-recall curves (left) and precision, recall, $F_{\beta}$-measure, $Wf_{\beta}$ and Mae values (right) for ASD (our methods SOH and HP, ST \cite{ZLiu-tip14}, HS \cite{Yan-cvpr13}, FT \cite{Achanta-09cvpr}, RC \cite{Cheng-11cvpr}, SF \cite{perazzi-cvpr12}, PCAS \cite{margolin-cvpr13})}\label{fig:prfm_asd}
\end{figure}

First we compare the performance of our methods (denoted HP for Hierarchies of Partitions and SOH for Saliency Over the Hierarchy) on the ASD dataset with several state-of-the-art approaches for salient object segmentation, where we include pixel-based methods (FT \cite{Achanta-09cvpr}), partition-based methods (RC \cite{Cheng-11cvpr}, SF \cite{perazzi-cvpr12}, PCAS \cite{margolin-cvpr13}) and hierachy-based methods (HS \cite{Yan-cvpr13}, ST \cite{ZLiu-tip14}). Precision-recall curves are shown in Figure~\ref{fig:prfm_asd}(a). Next, we perform adaptive thresholding \cite{otsu79} and calculate precision, recall, F-measure, weighted F-measure and mean absolute error for all the models. Results are presented in Figure~\ref{fig:prfm_asd}(b).

For ASD, the best results in terms of P-R curves are obtained for hierarchy-based techniques. Saliency Tree (ST) is the best performing method, closely followed by our two models (SOH and HP) and Hierarchical Saliency (HS).

For the adaptive threshold,  $F_1$-measures are $0.891$ (ST), $0.884$ (SOH), $0.878$ (HP) and $0.863$ (HS), $F_{\beta}^w$  values are $0.711$ (ST), $0.692$ (SOH), $0.588$ (HP), and $0.642$ (HS). MAE values are $0.088$ (ST), $0.096$ (SOH), $0.136$ (HP) and $0.114$ (HS).

\begin{figure}[thbp]
\centering
\begin{tabular}{cc}
\includegraphics[height=4.5cm]{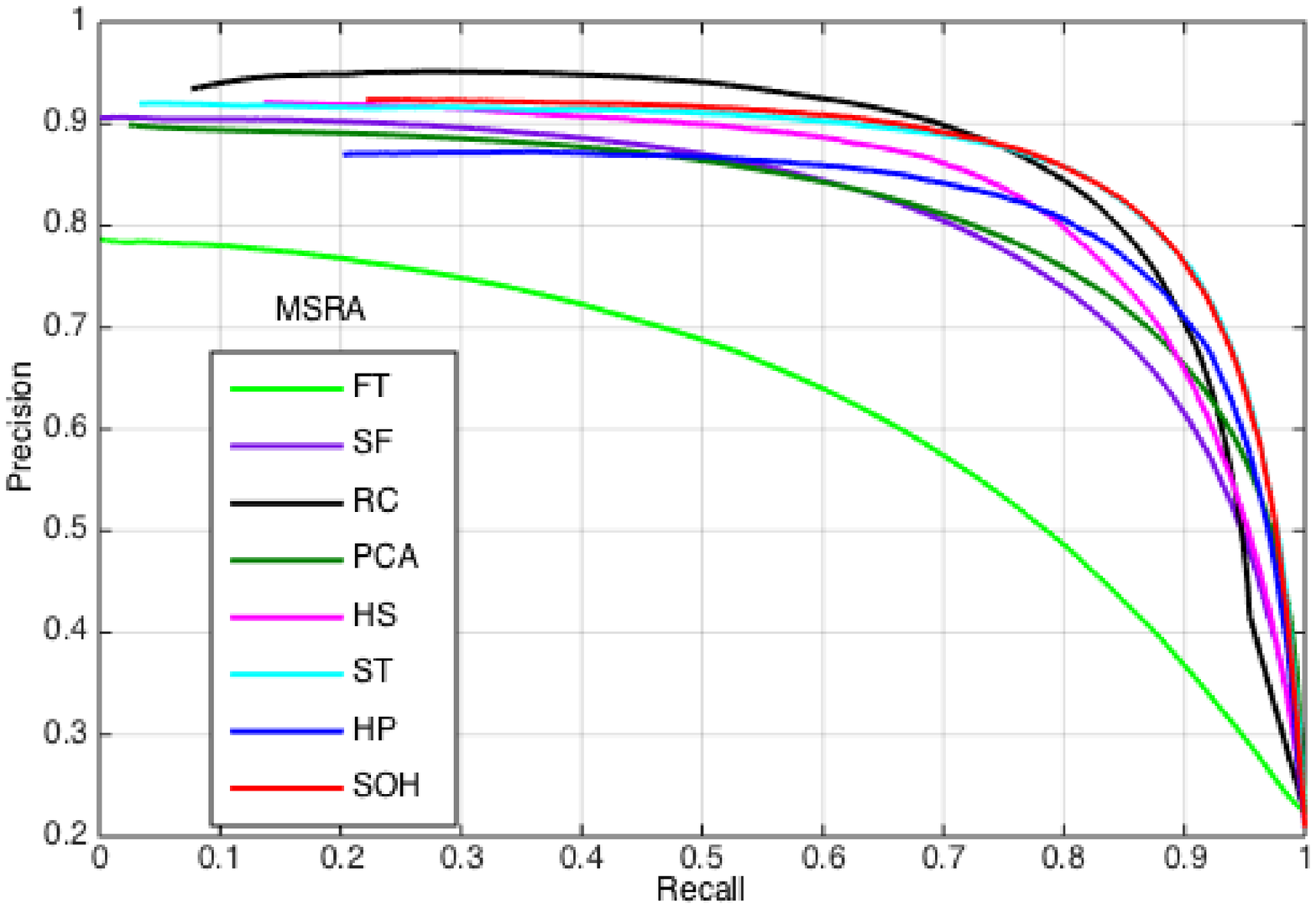} &
\raisebox{.2\height}{\includegraphics[height=3.5cm]{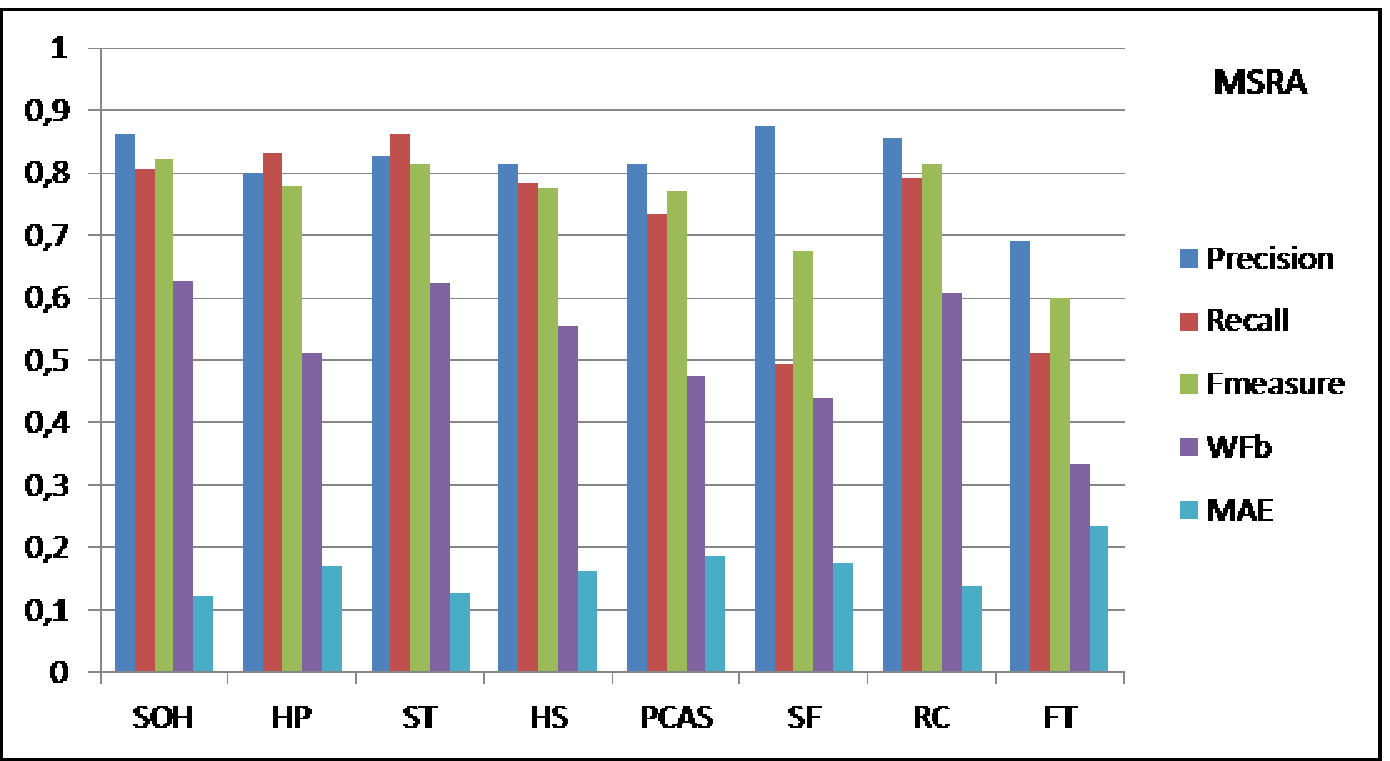}} \\
(a) & (b)\\
\end{tabular}
\caption{Precision-recall curves (left) and precision, recall, $F_{\beta}$-measure, $Wf_{\beta}$ and Mae values (right) for MSRA (our methods SOH and HP, ST \cite{ZLiu-tip14}, HS \cite{Yan-cvpr13}, FT \cite{Achanta-09cvpr}, RC \cite{Cheng-11cvpr}, SF \cite{perazzi-cvpr12}, PCAS \cite{margolin-cvpr13})}
\label{fig:prfm_msra}
\end{figure}

\begin{figure}[thbp]
\centering
\begin{tabular}{cccccc}
IM&
\includegraphics[height=1.6cm]{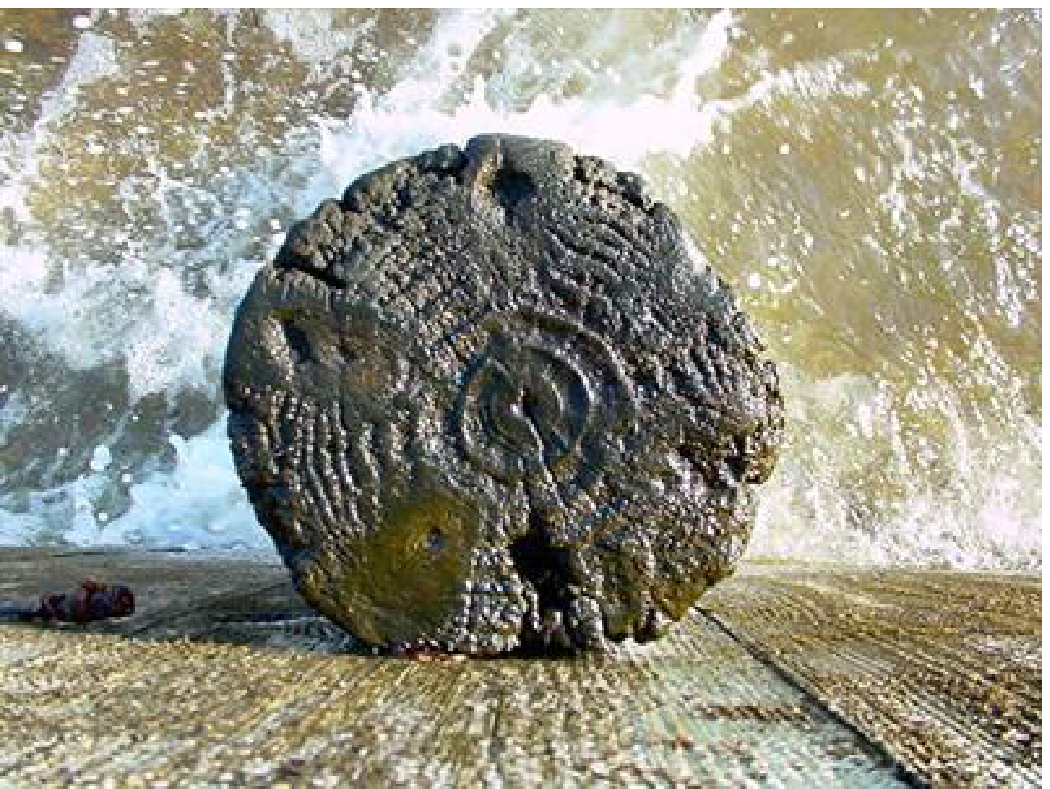} &
\includegraphics[height=1.6cm]{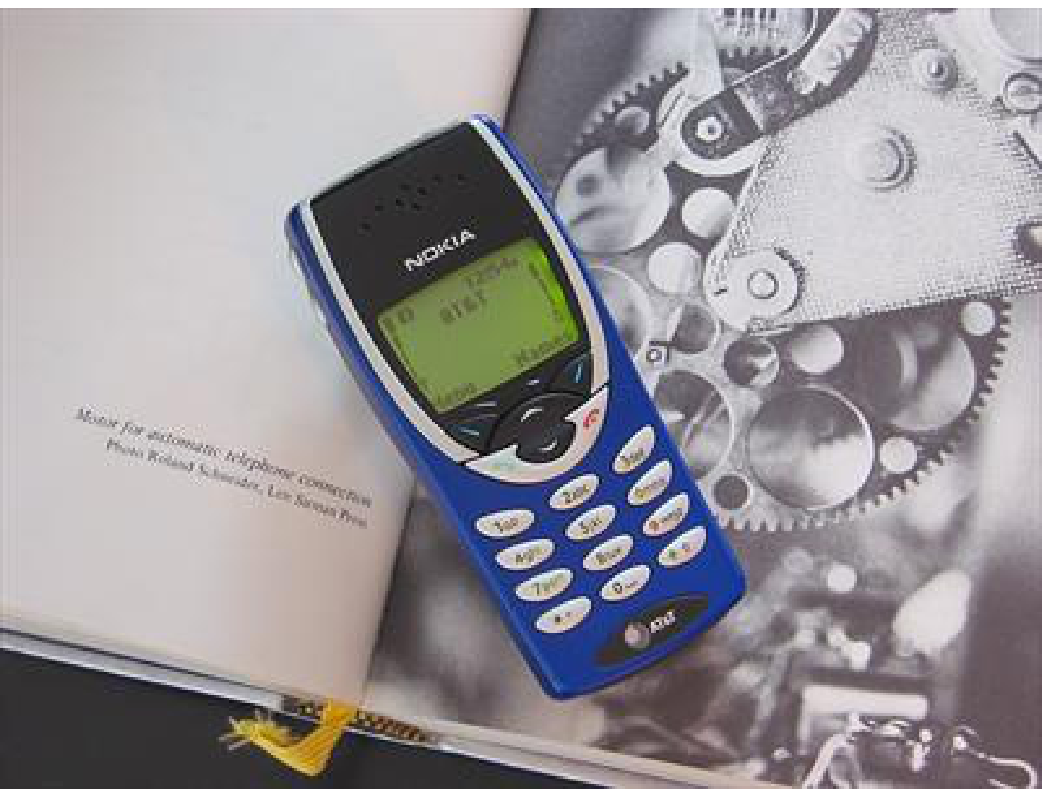} &
\includegraphics[height=1.6cm]{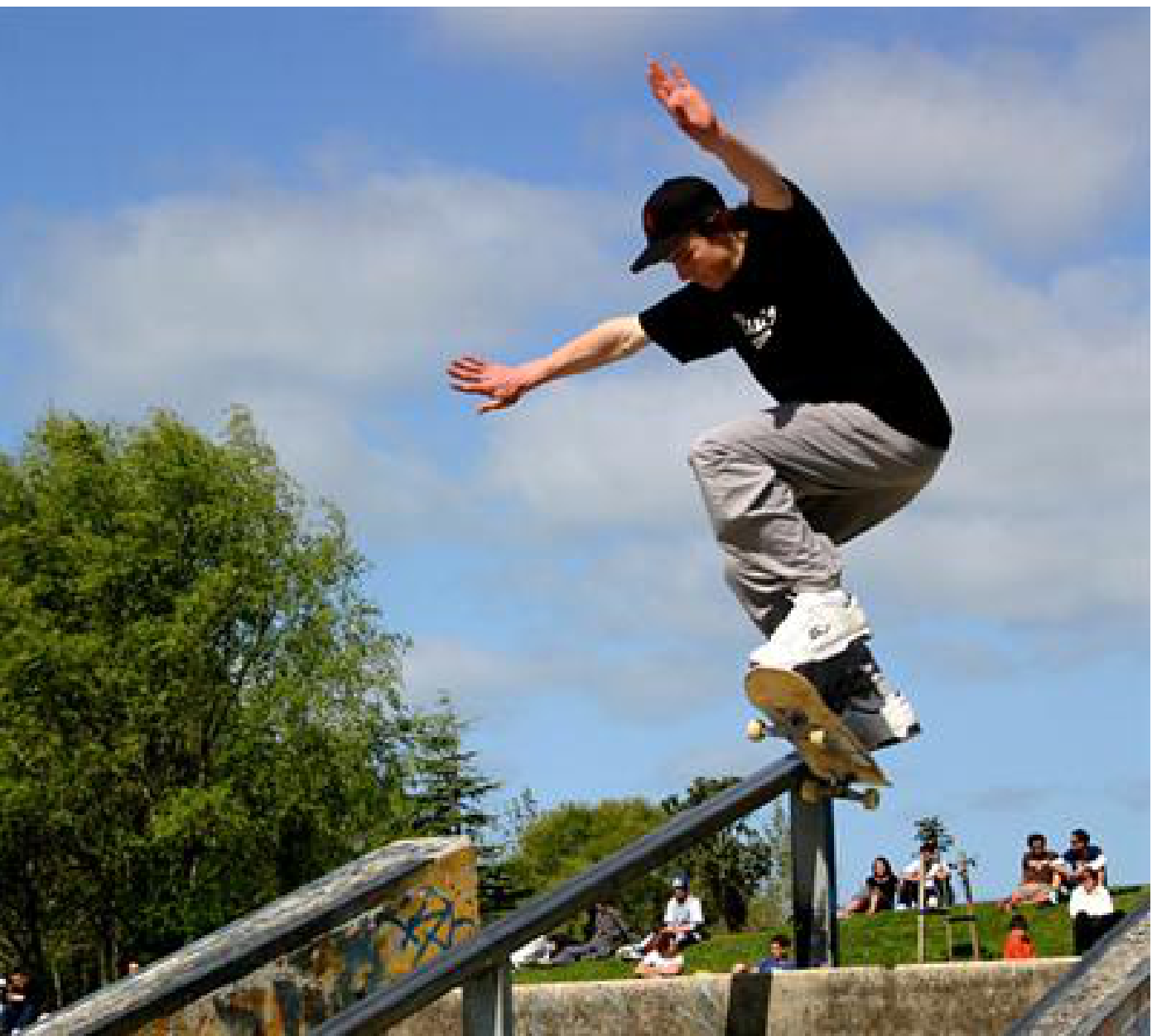} &
\includegraphics[height=1.6cm]{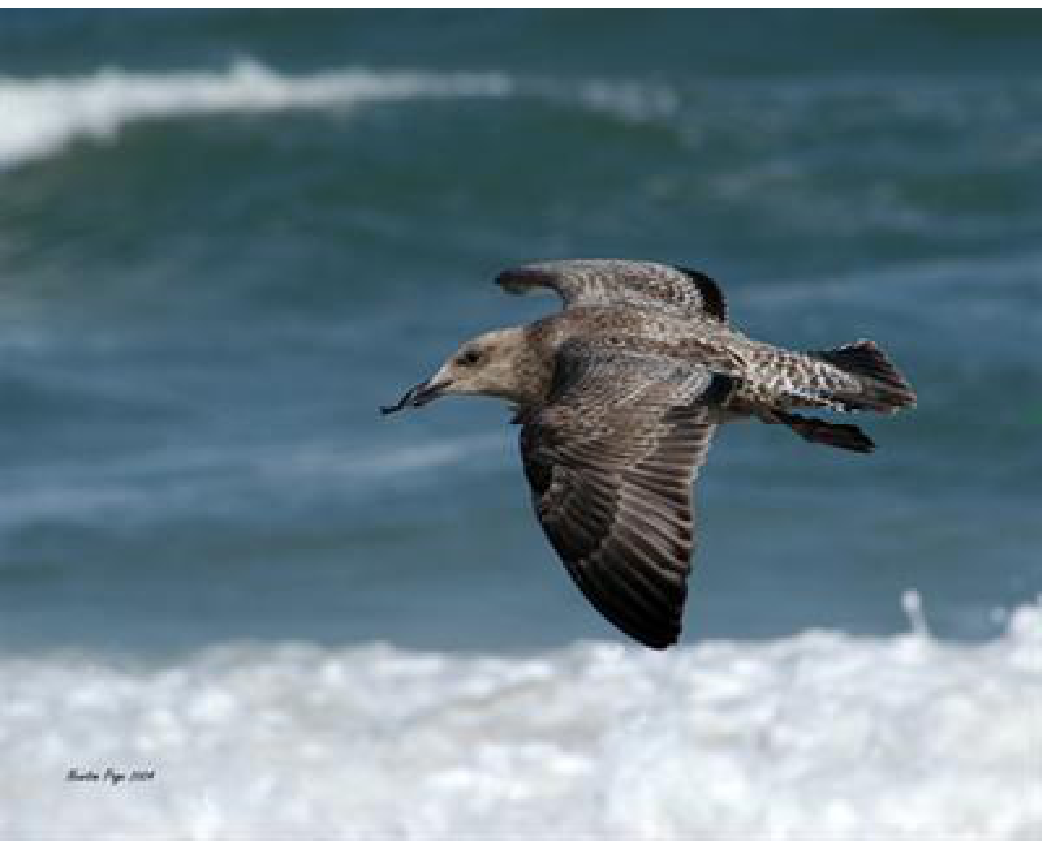} &
\includegraphics[height=1.6cm]{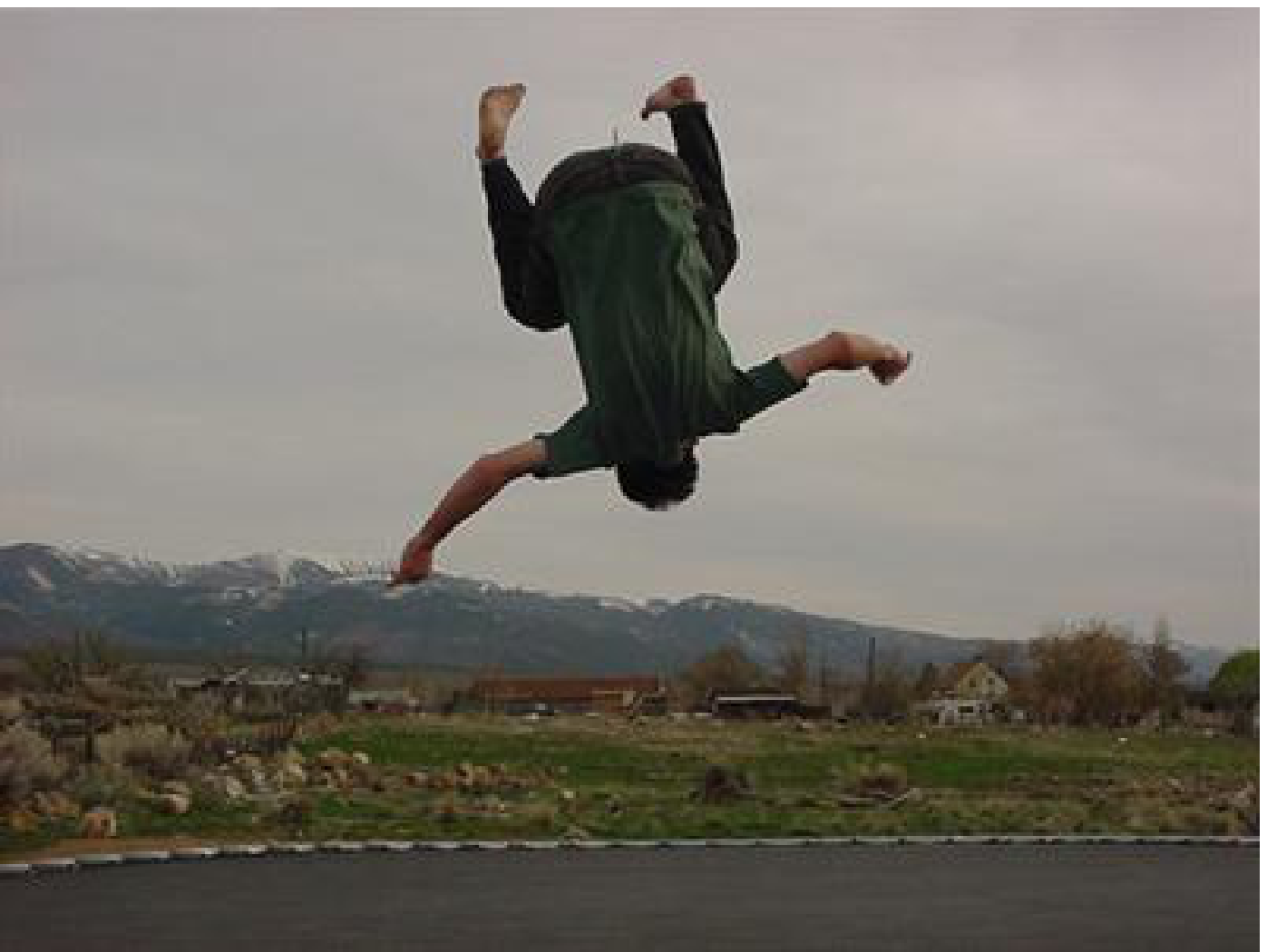} \\
FT&
\includegraphics[height=1.6cm]{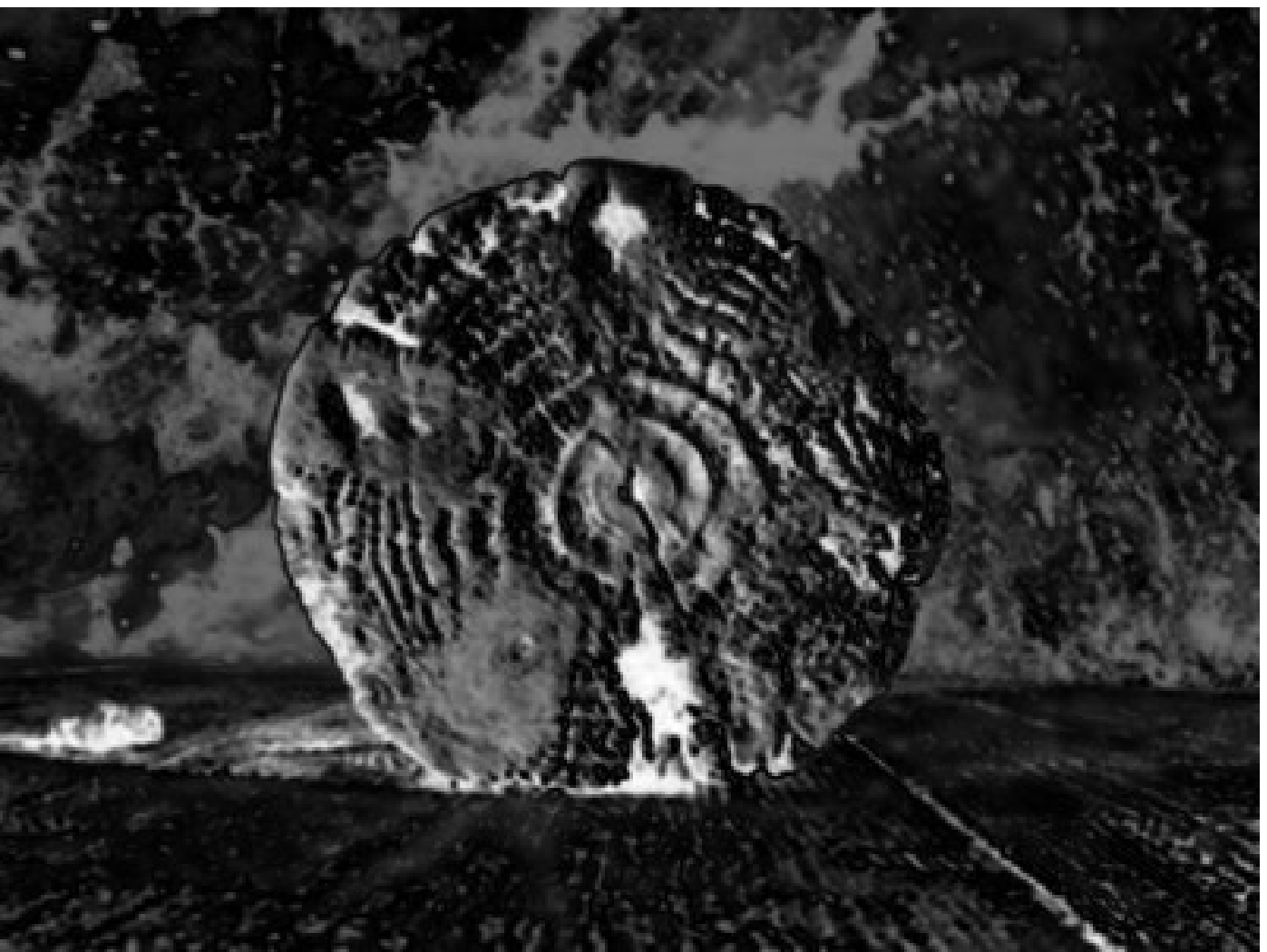} &
\includegraphics[height=1.6cm]{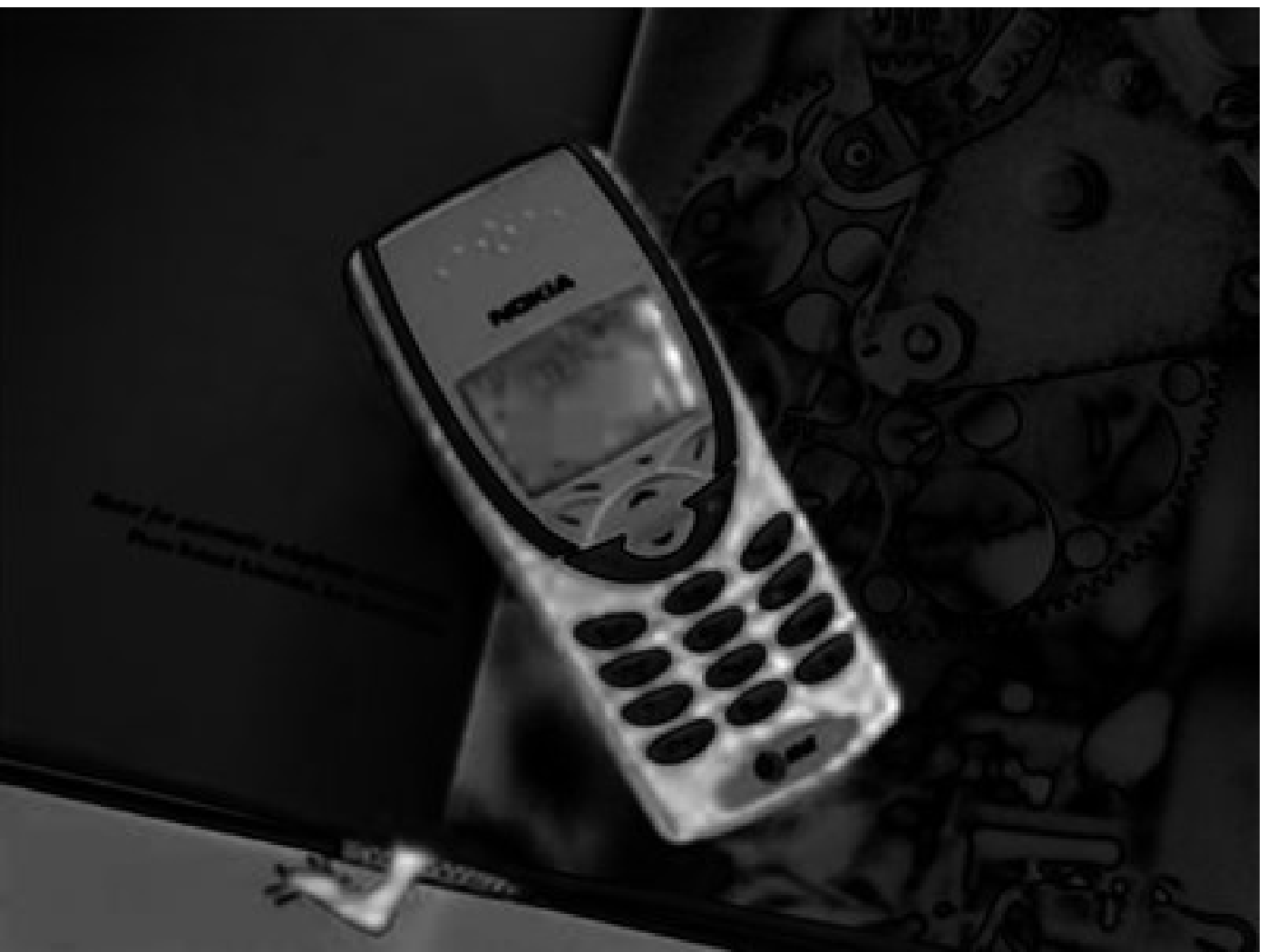} &
\includegraphics[height=1.6cm]{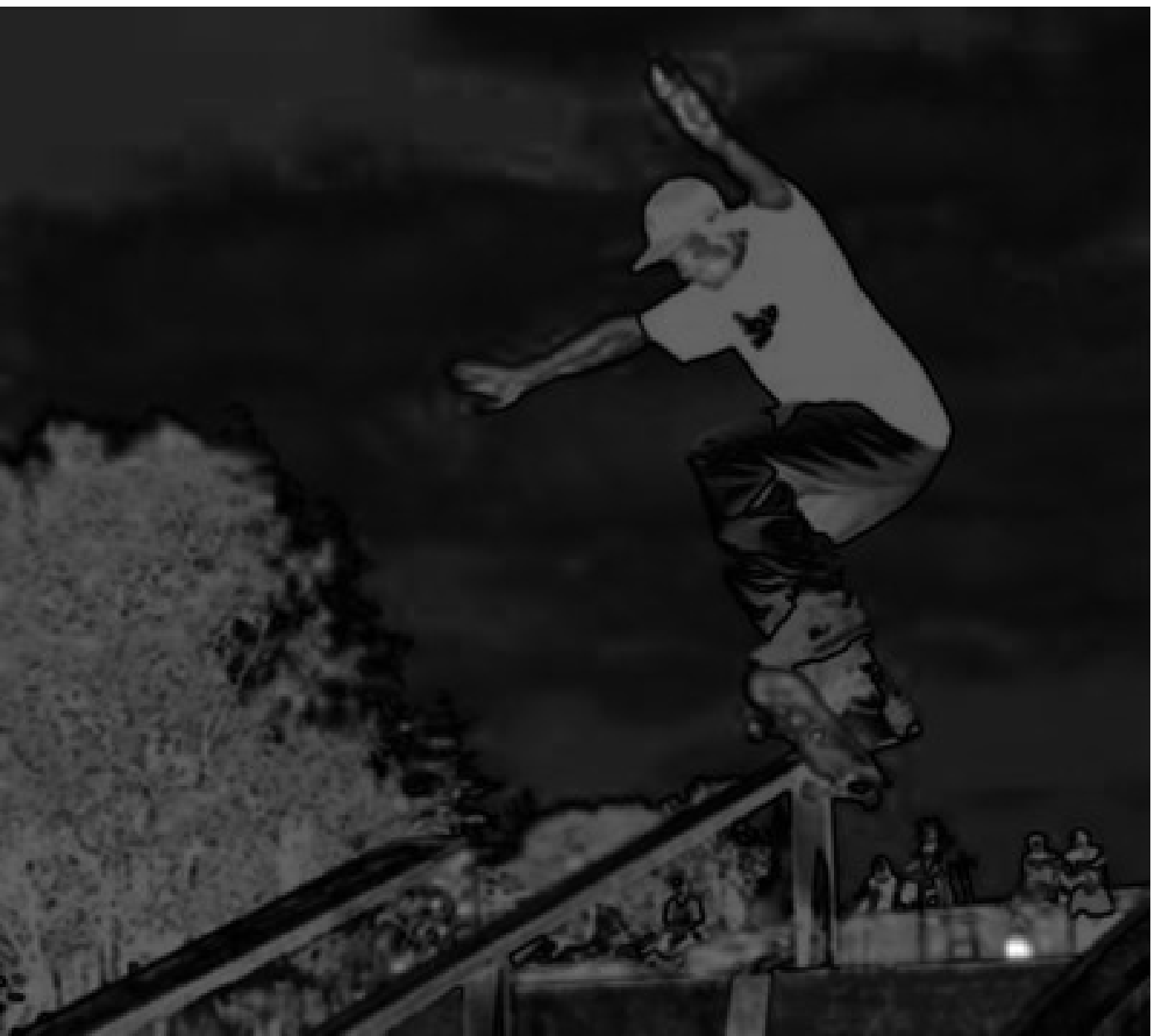} &
\includegraphics[height=1.6cm]{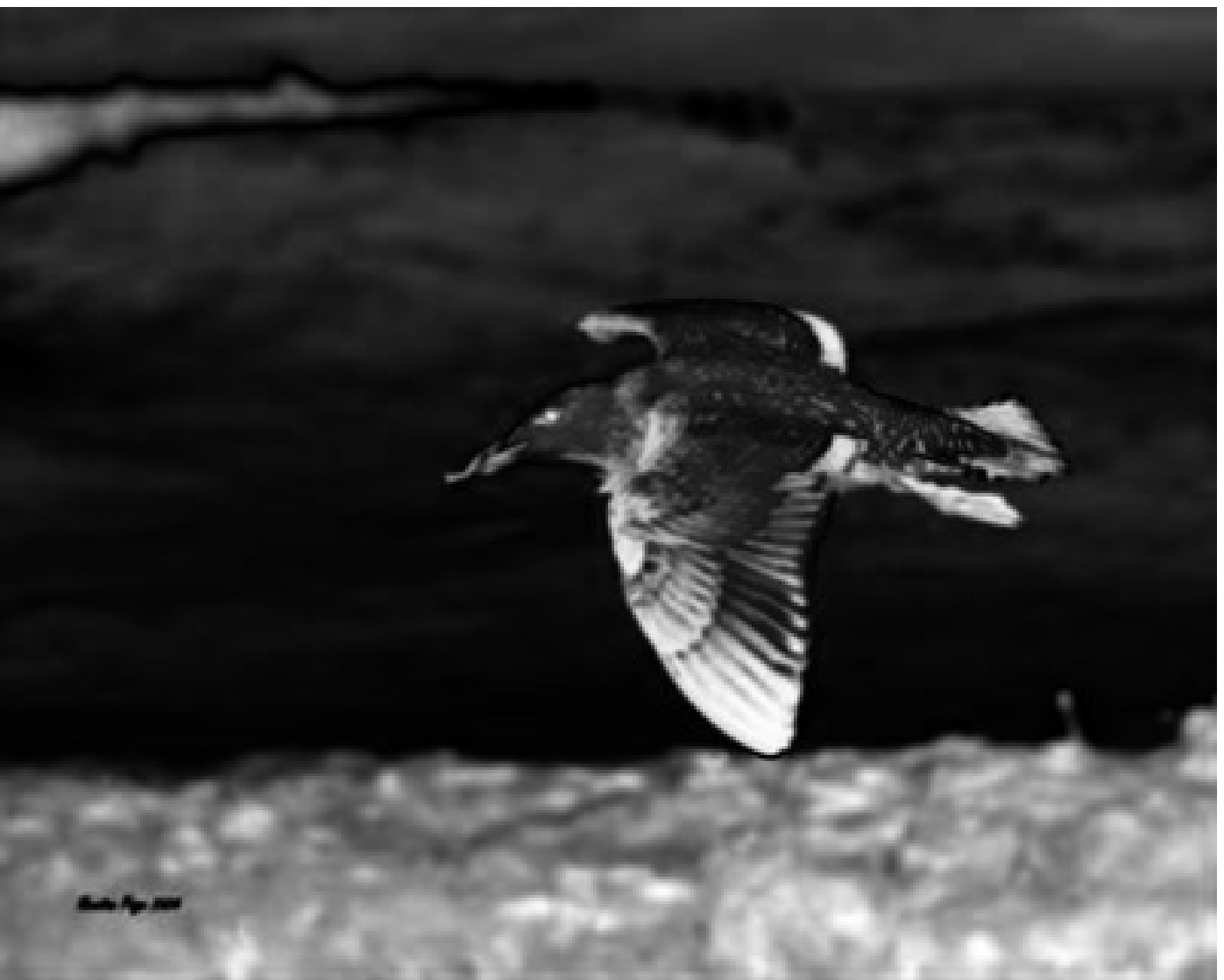} &
\includegraphics[height=1.6cm]{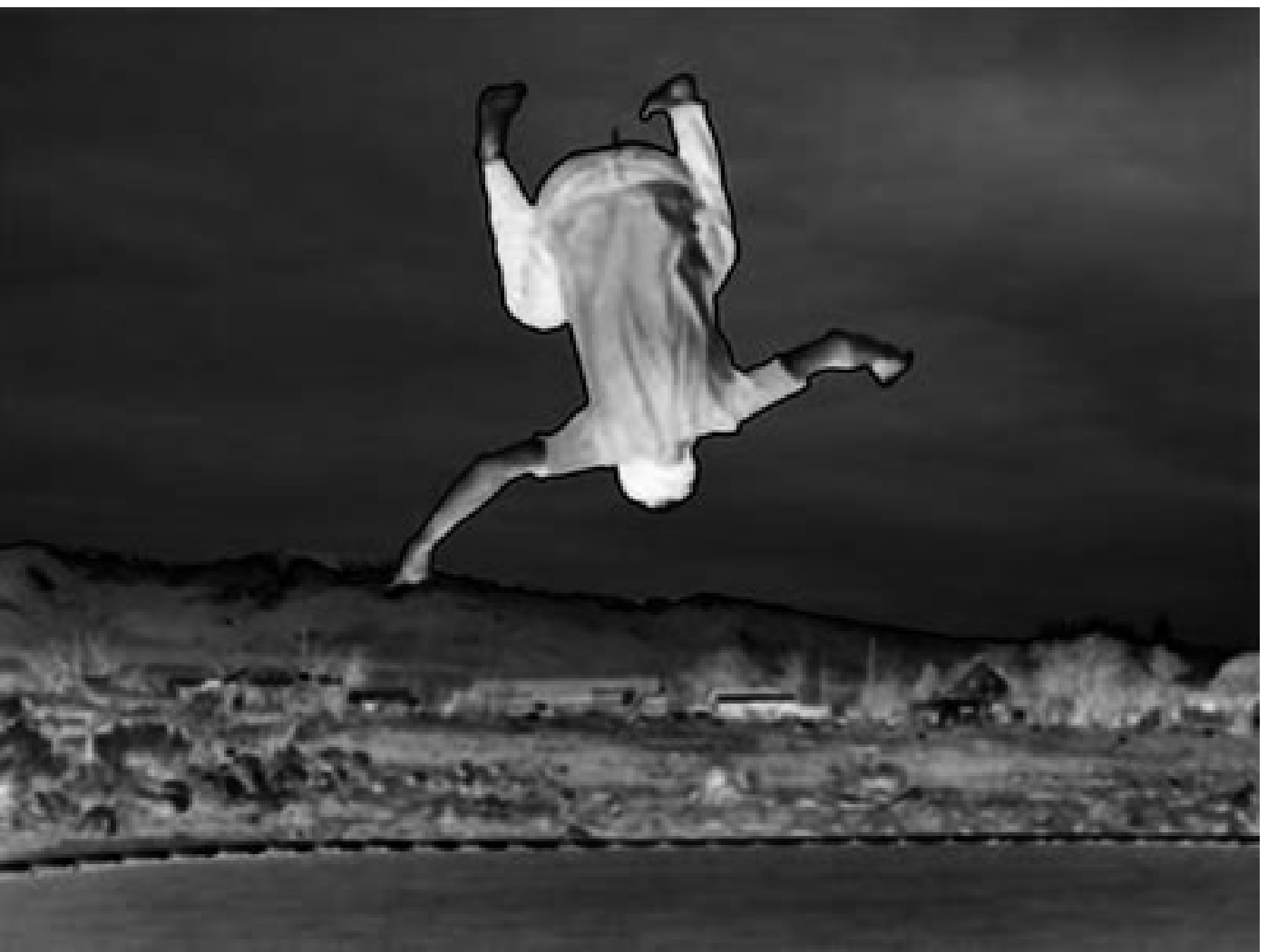} \\
RC&
\includegraphics[height=1.6cm]{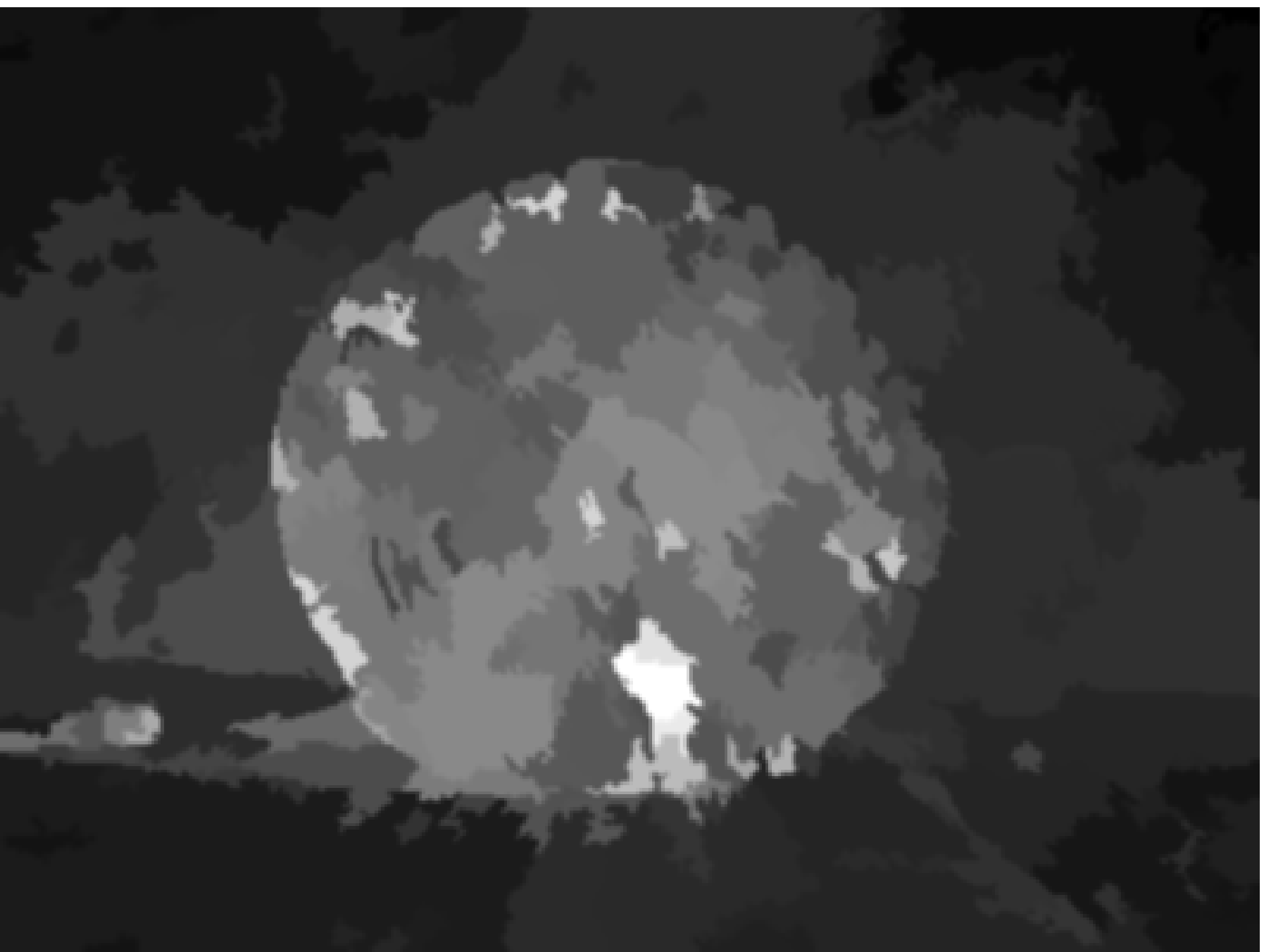} &
\includegraphics[height=1.6cm]{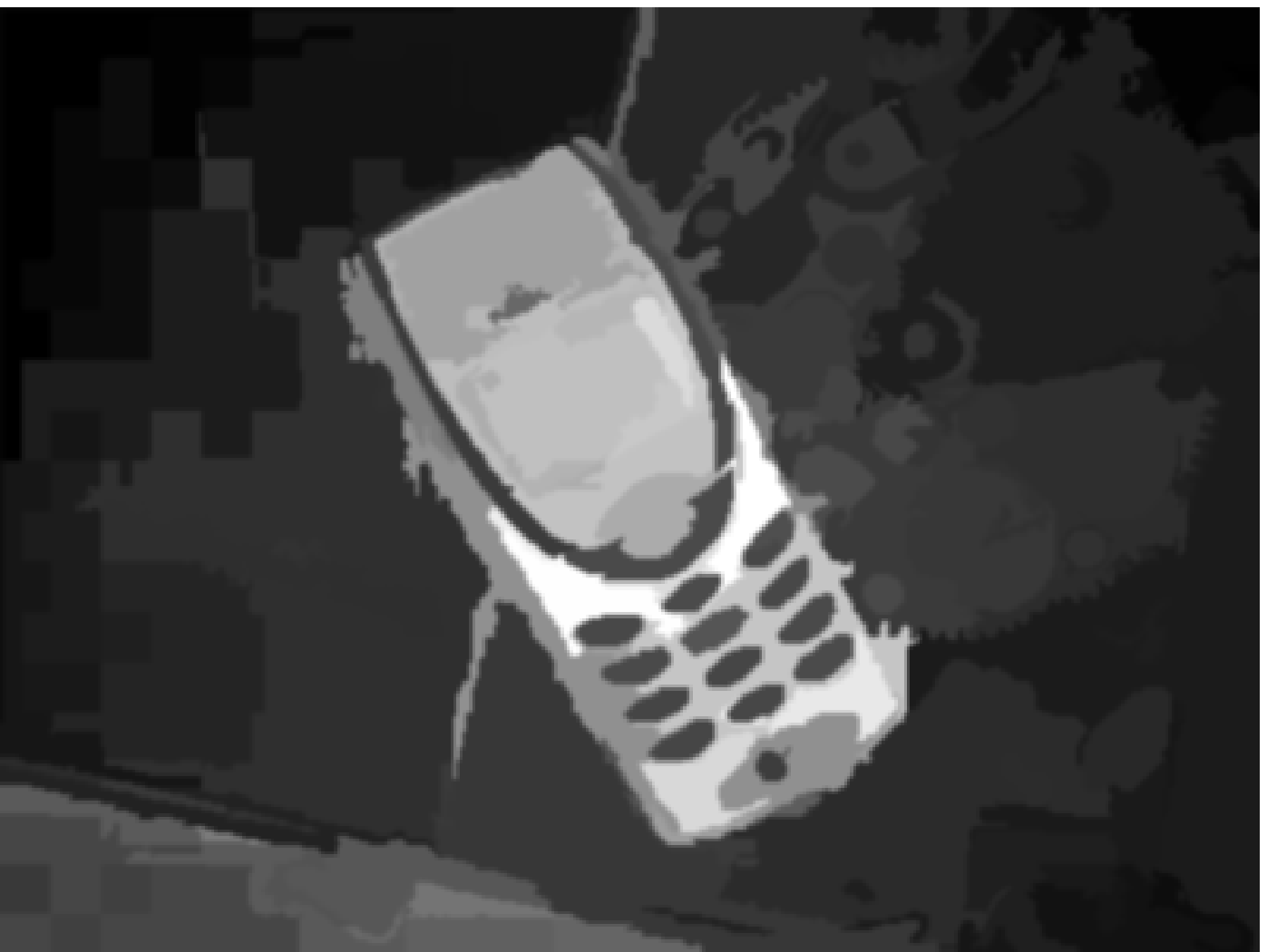} &
\includegraphics[height=1.6cm]{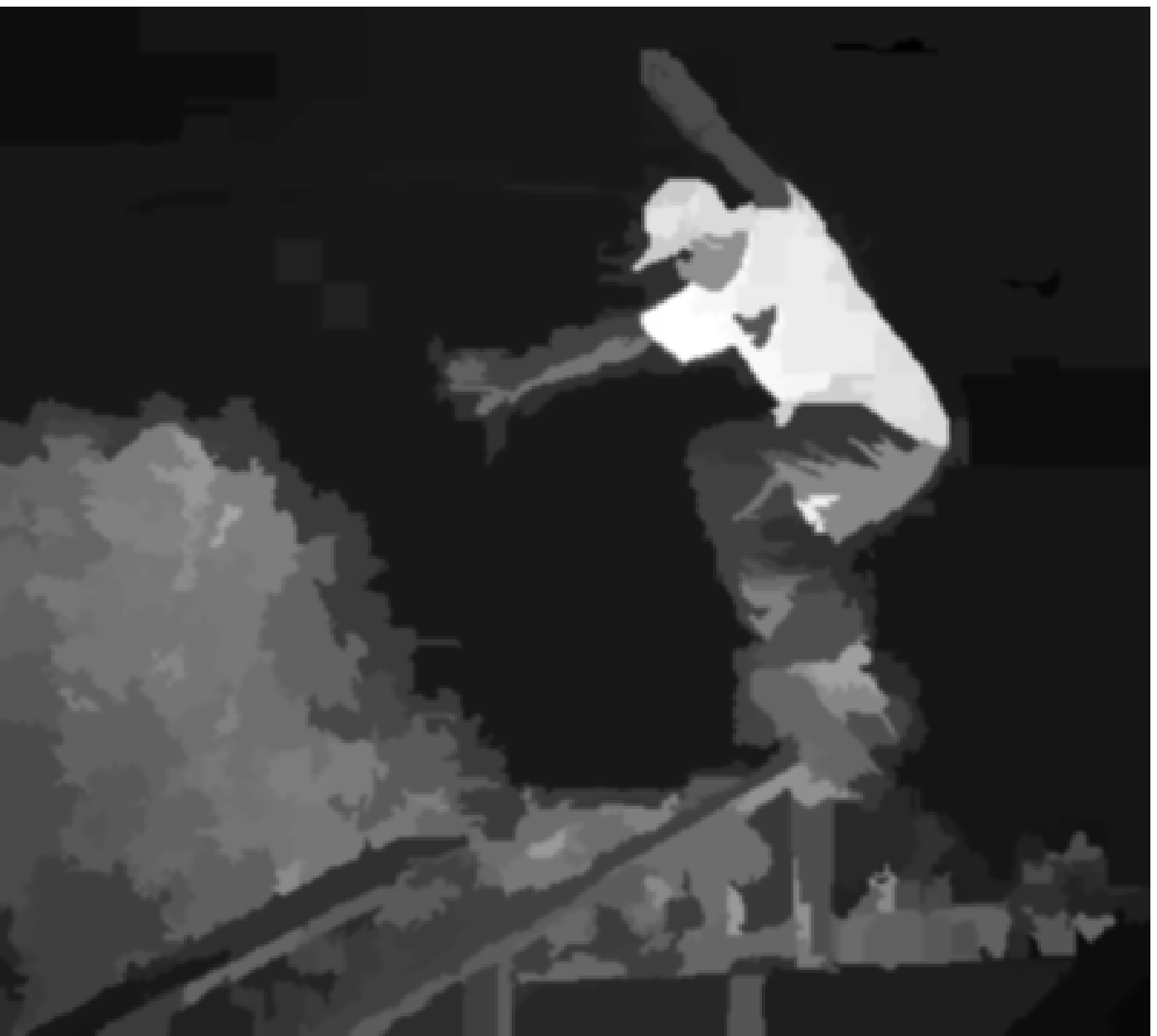} &
\includegraphics[height=1.6cm]{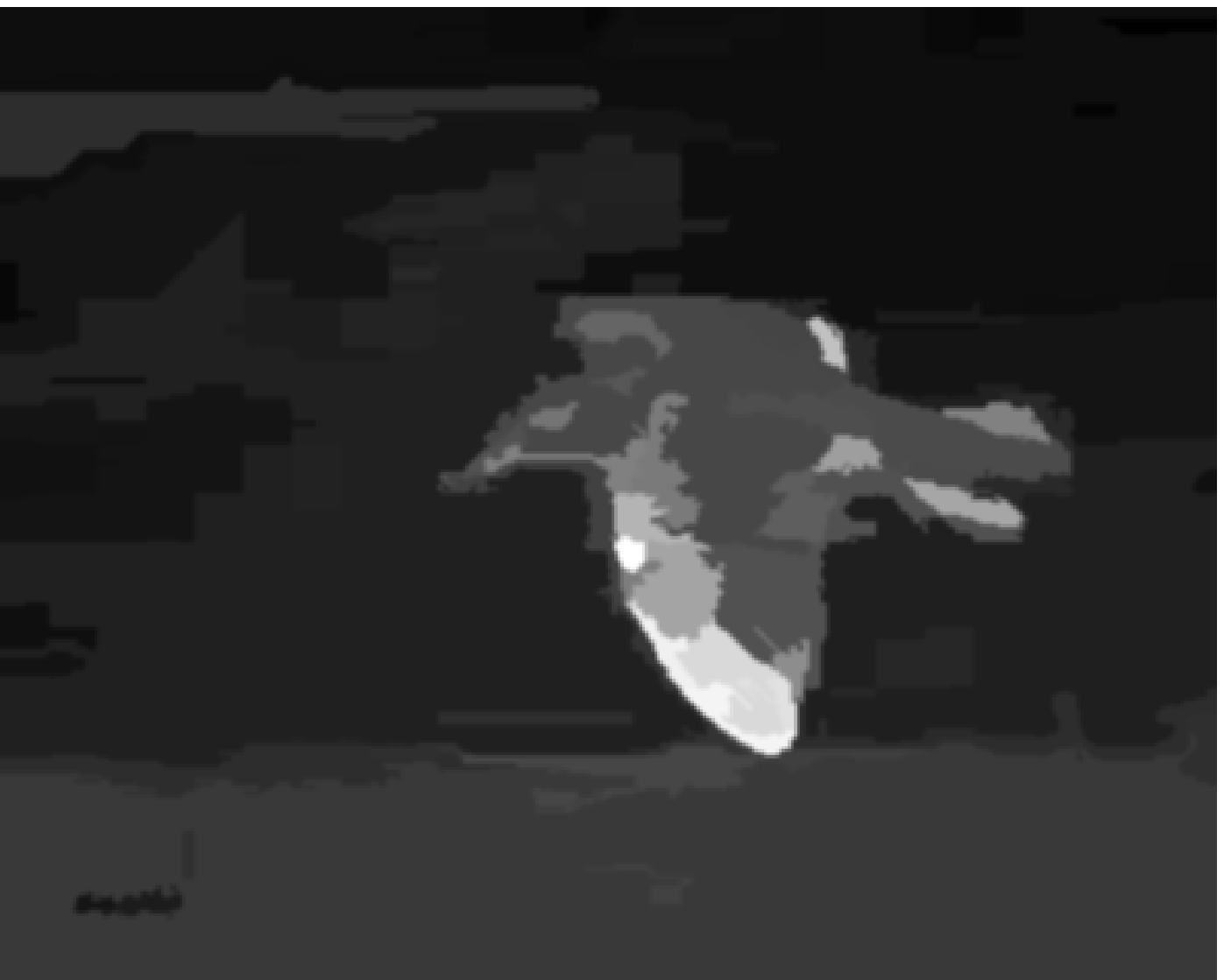} &
\includegraphics[height=1.6cm]{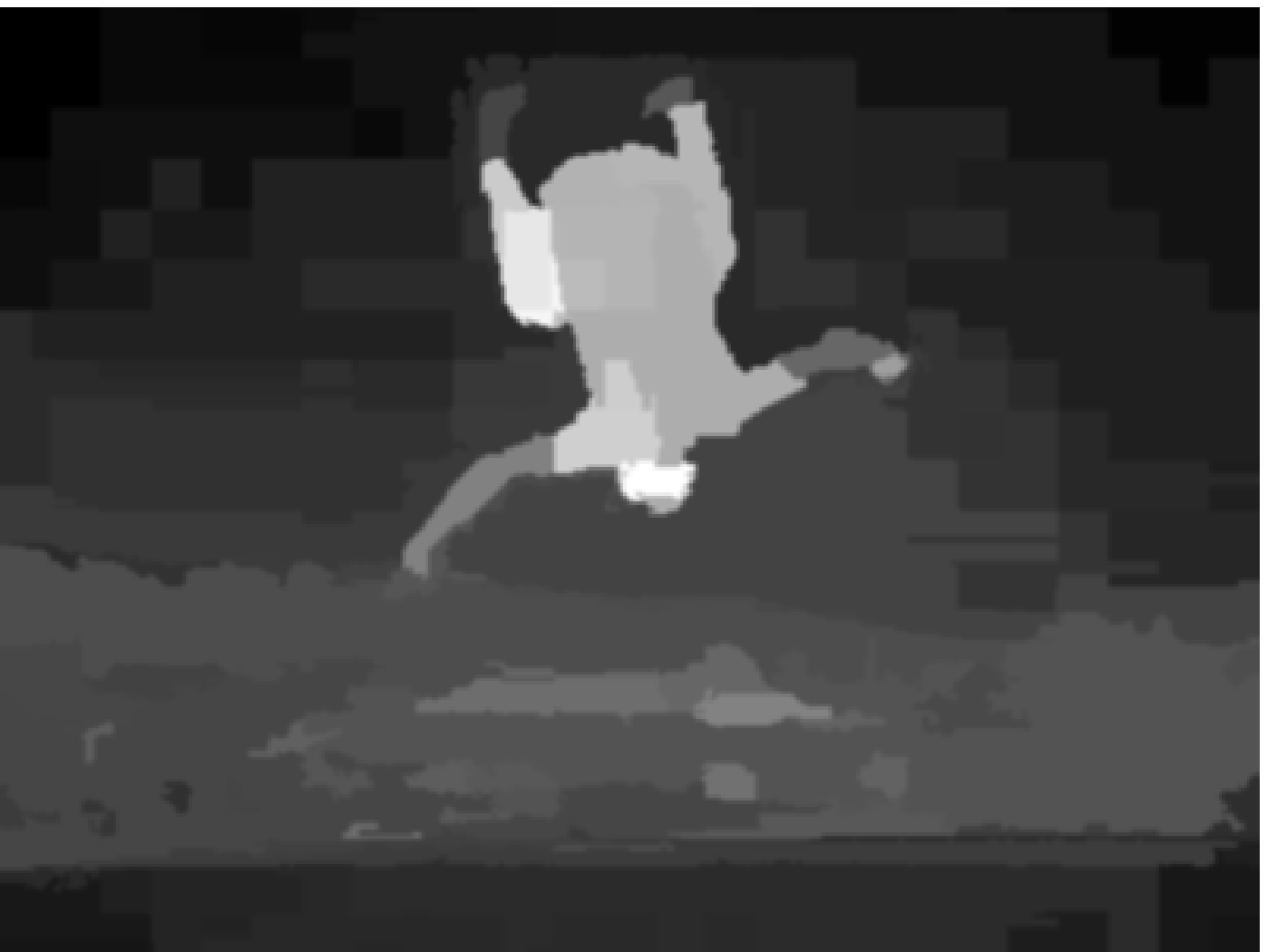} \\
SF&
\includegraphics[height=1.6cm]{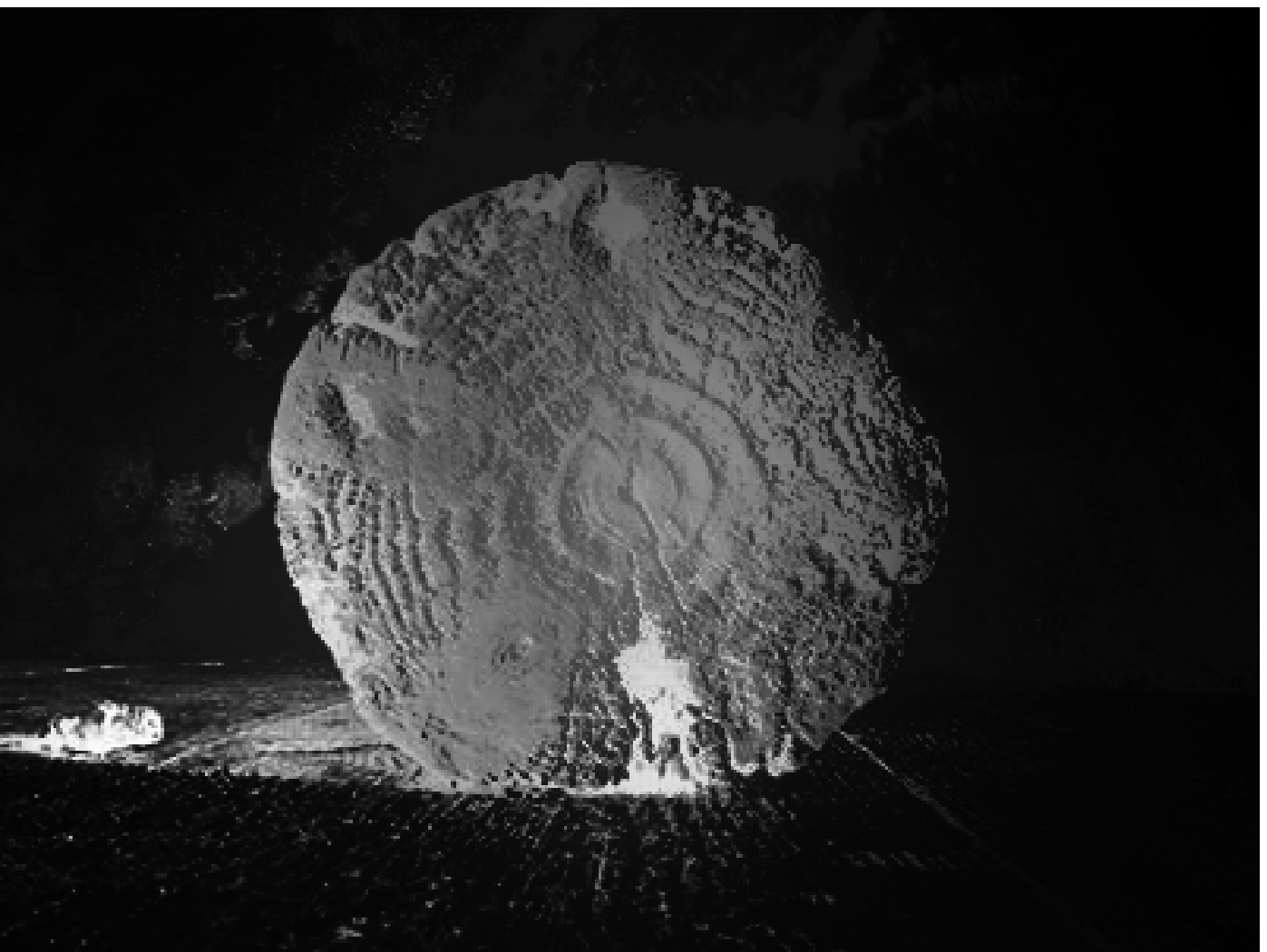} &
\includegraphics[height=1.6cm]{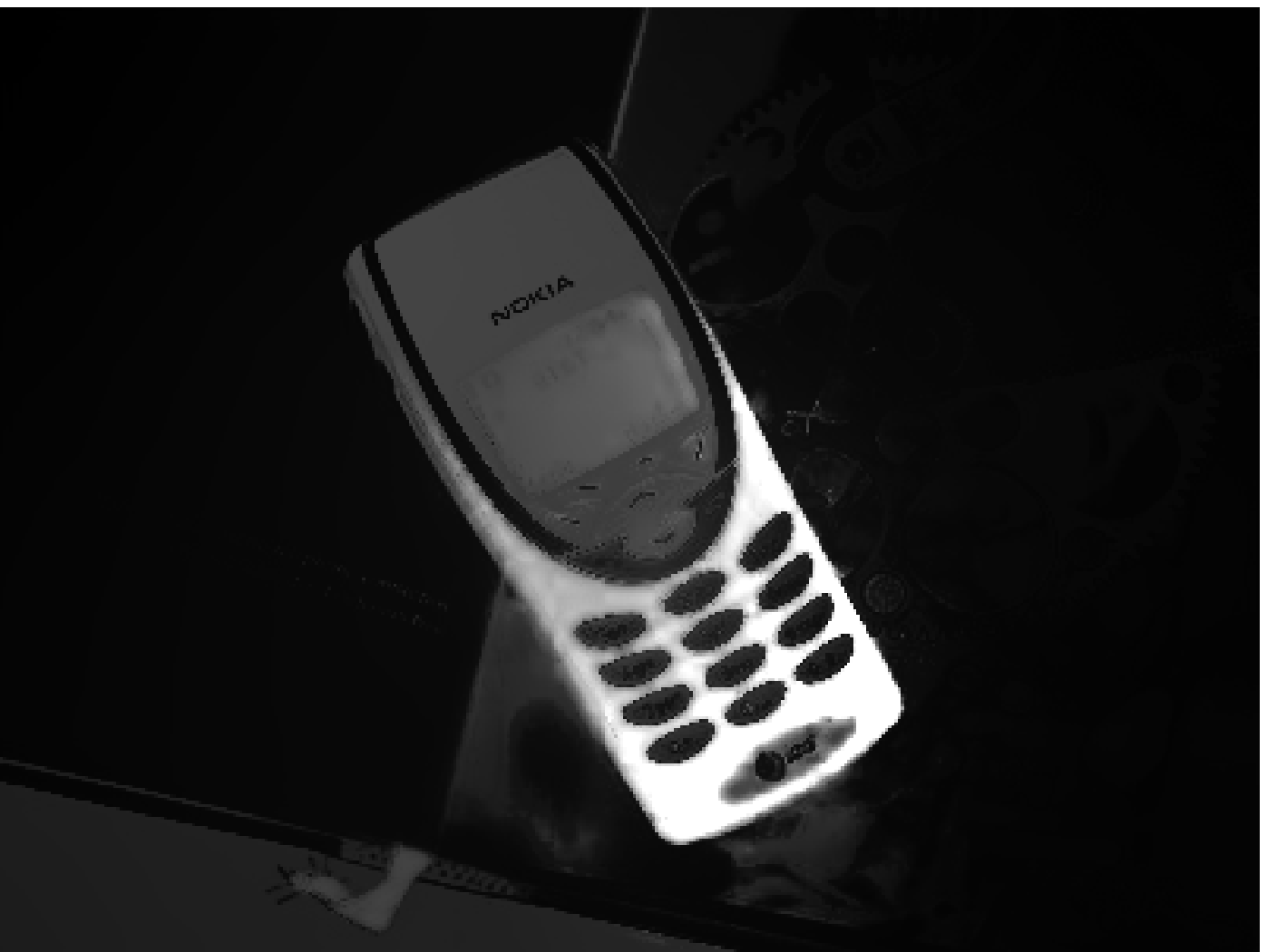} &
\includegraphics[height=1.6cm]{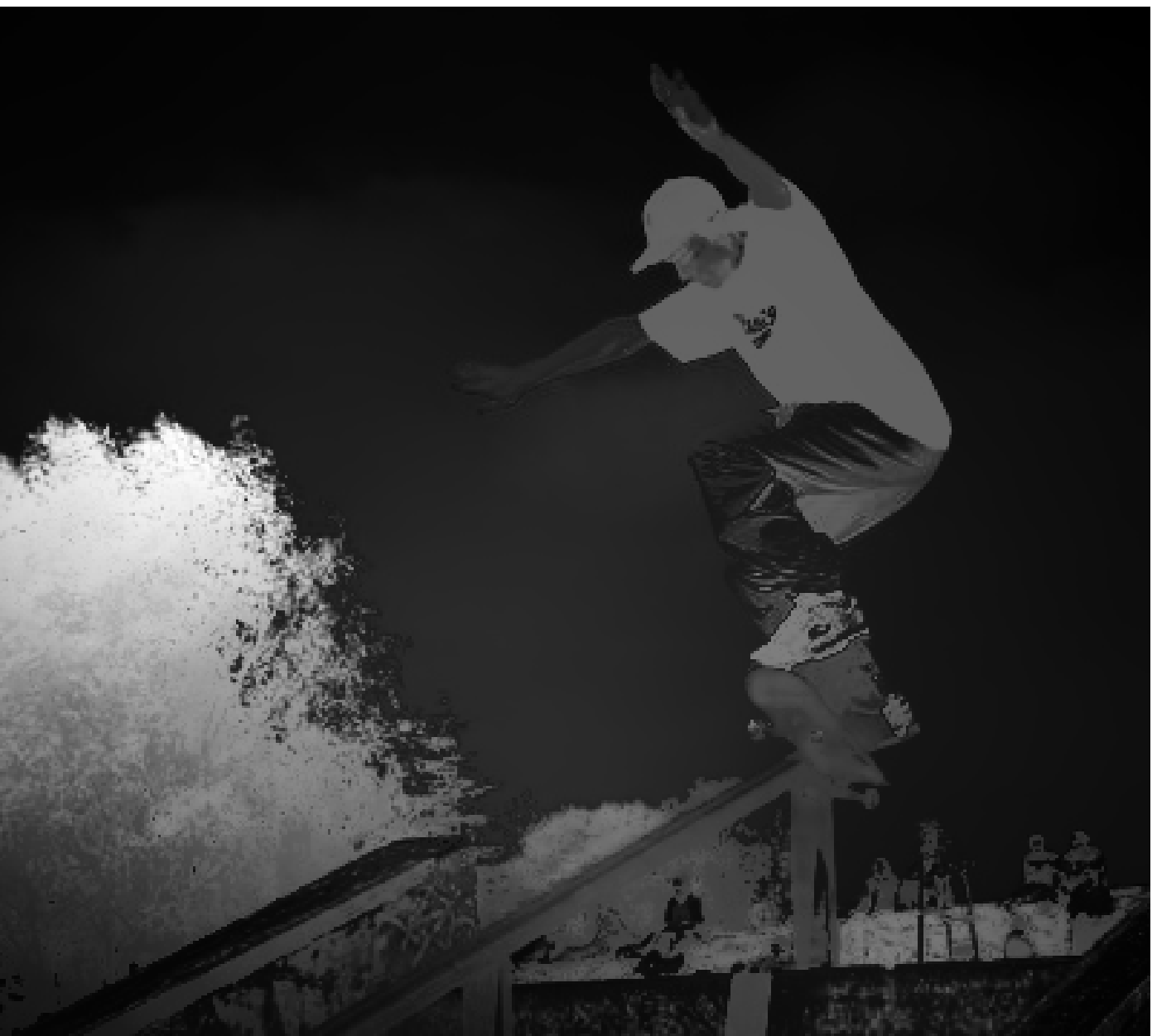} &
\includegraphics[height=1.6cm]{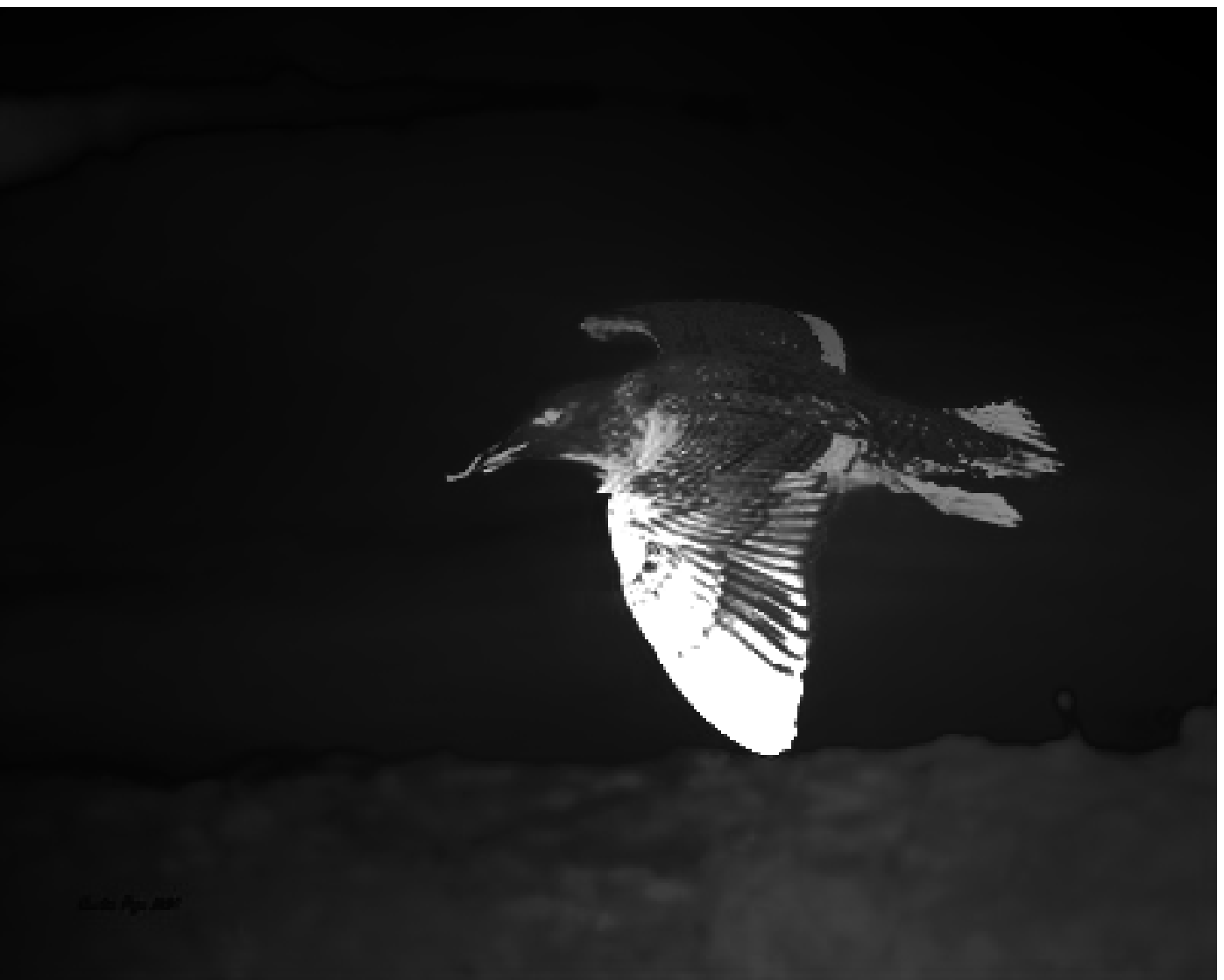} &
\includegraphics[height=1.6cm]{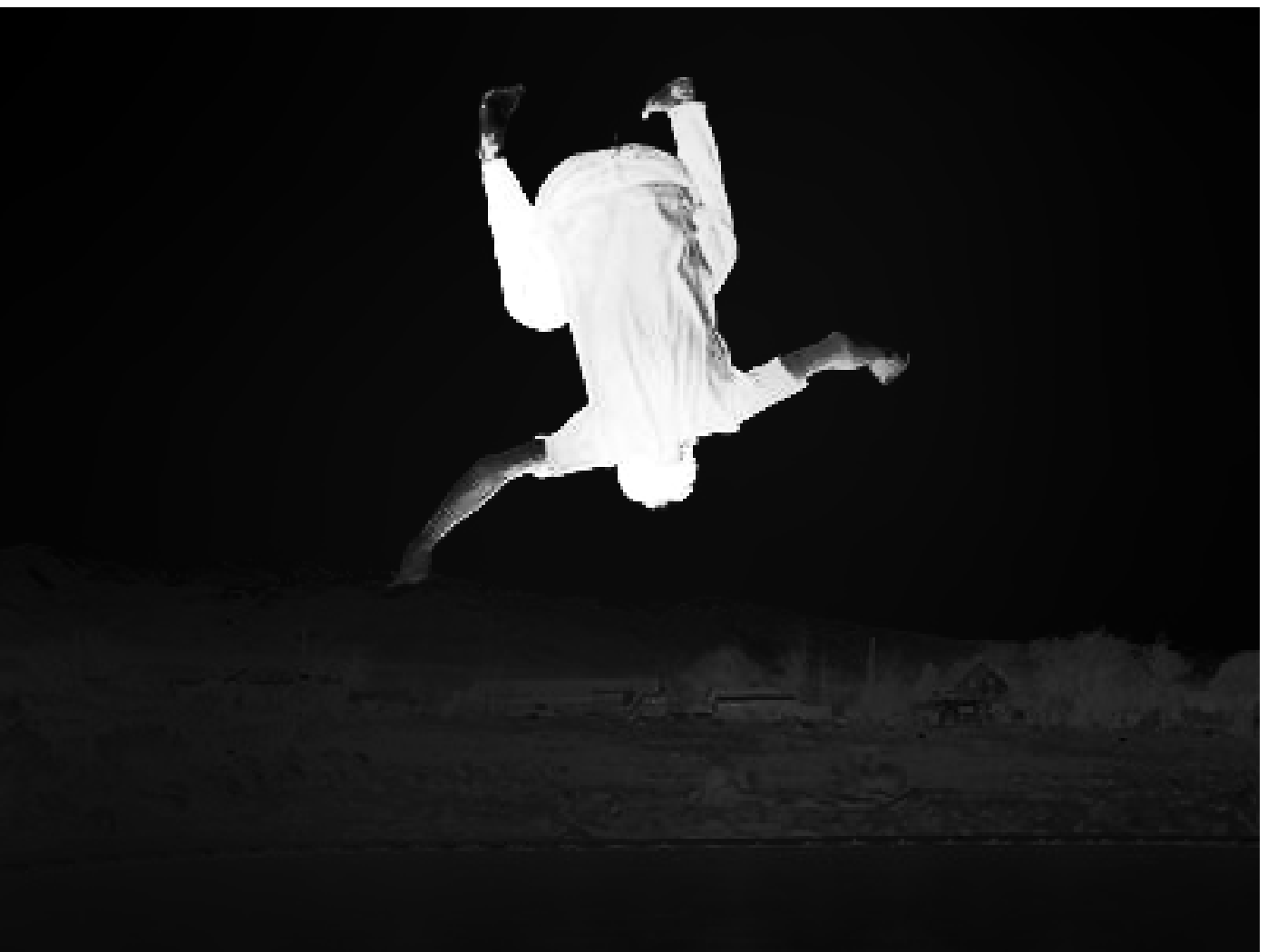} \\
PCAS&
\includegraphics[height=1.6cm]{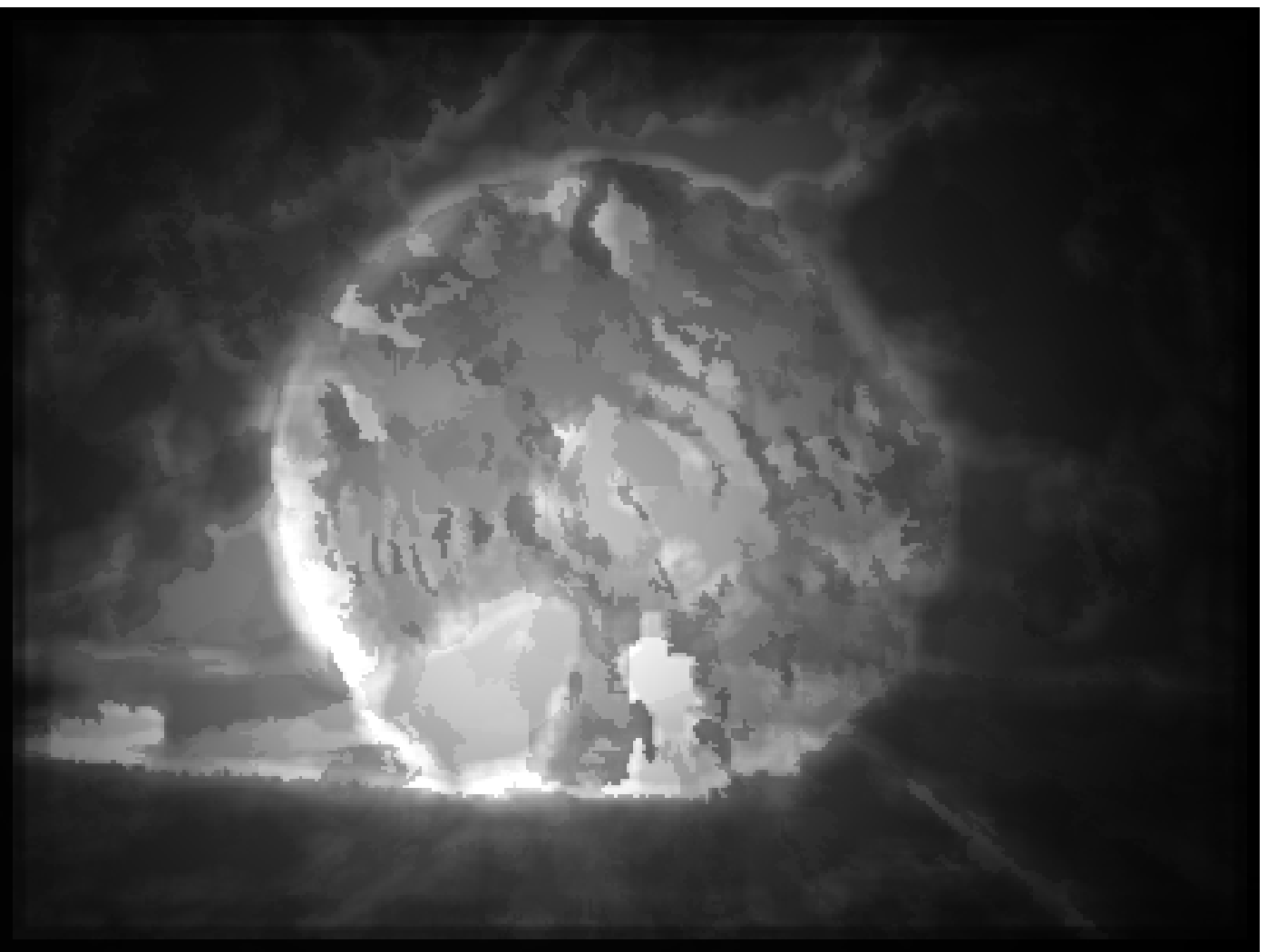} &
\includegraphics[height=1.6cm]{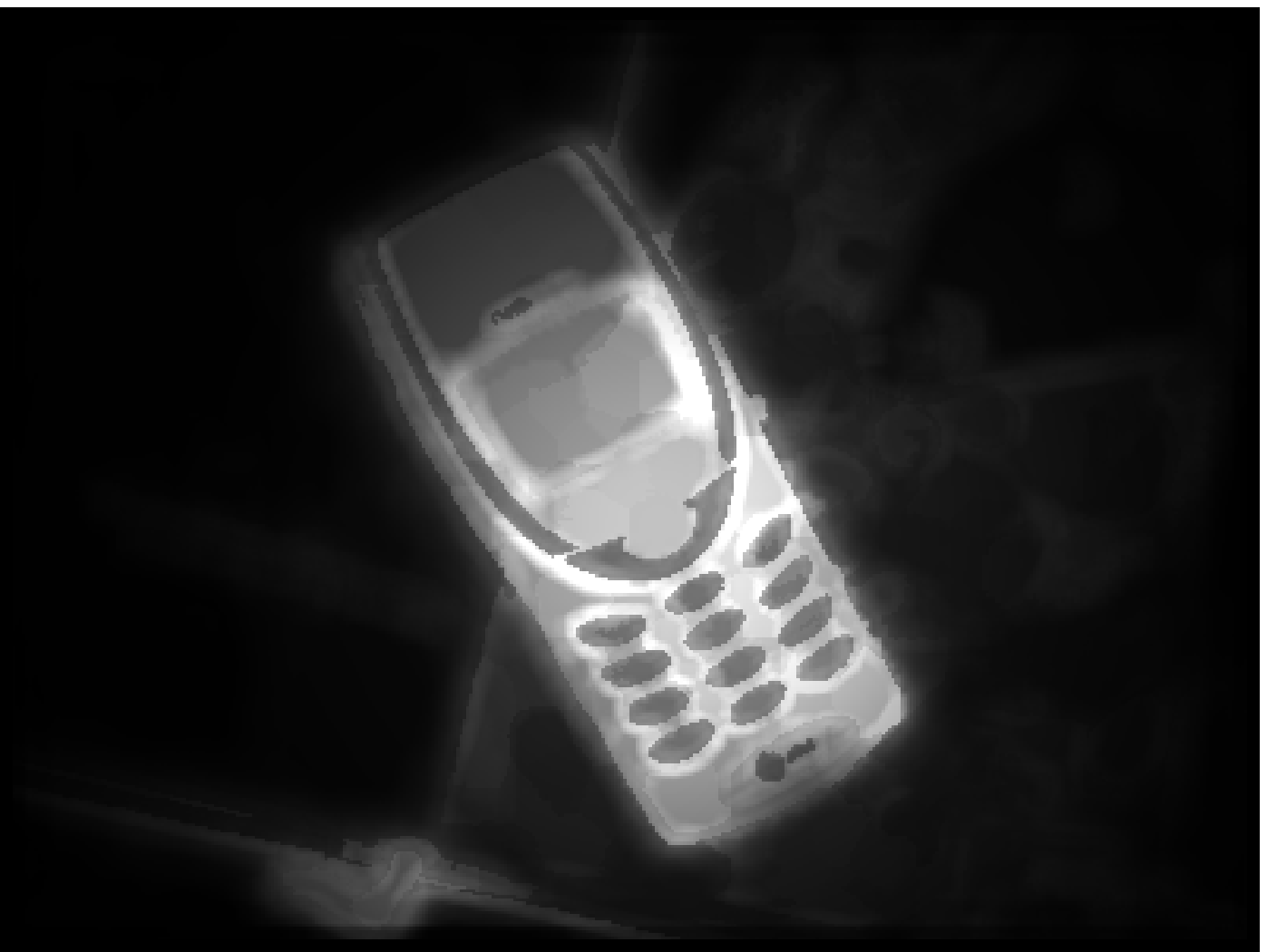} &
\includegraphics[height=1.6cm]{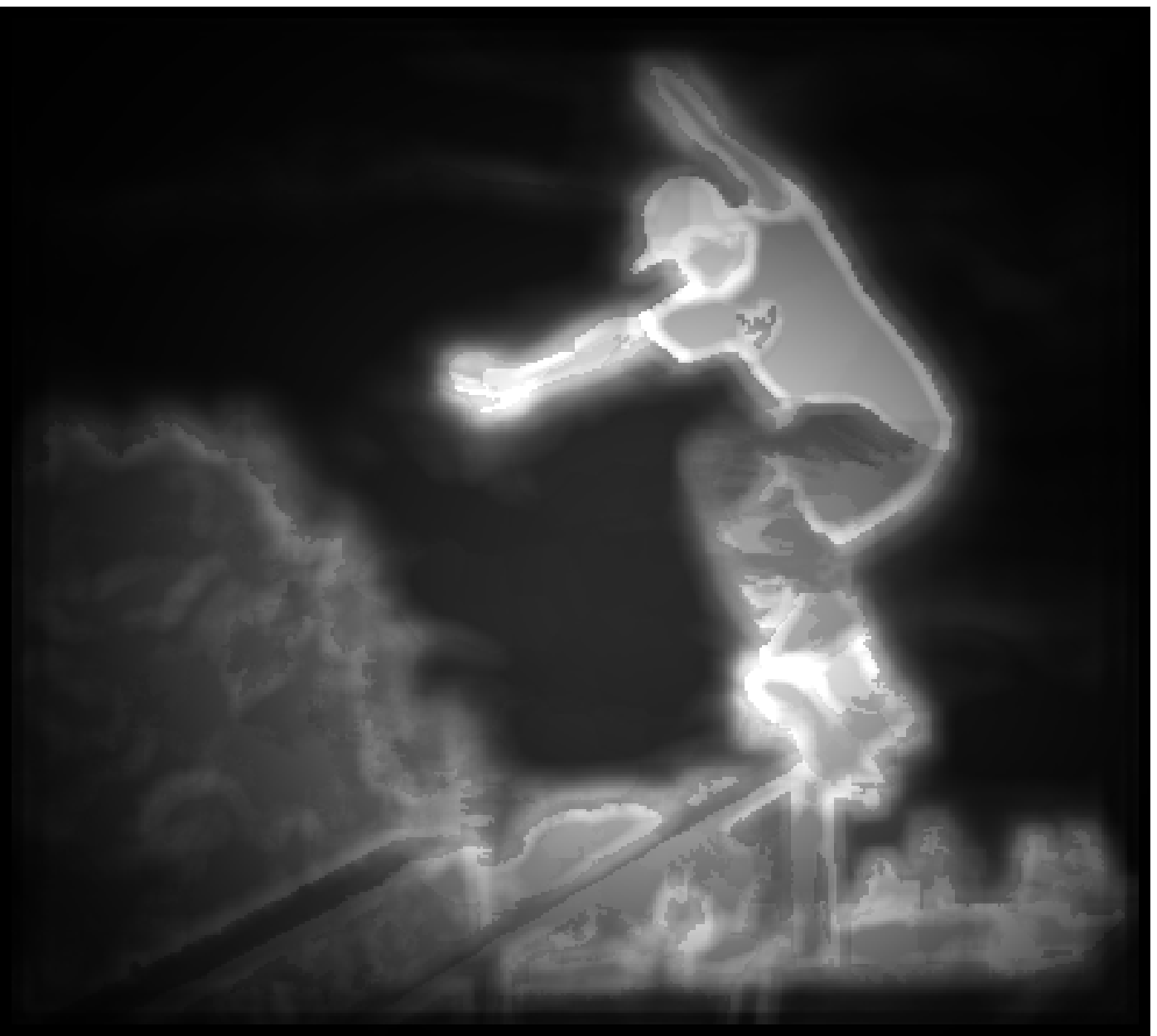} &
\includegraphics[height=1.6cm]{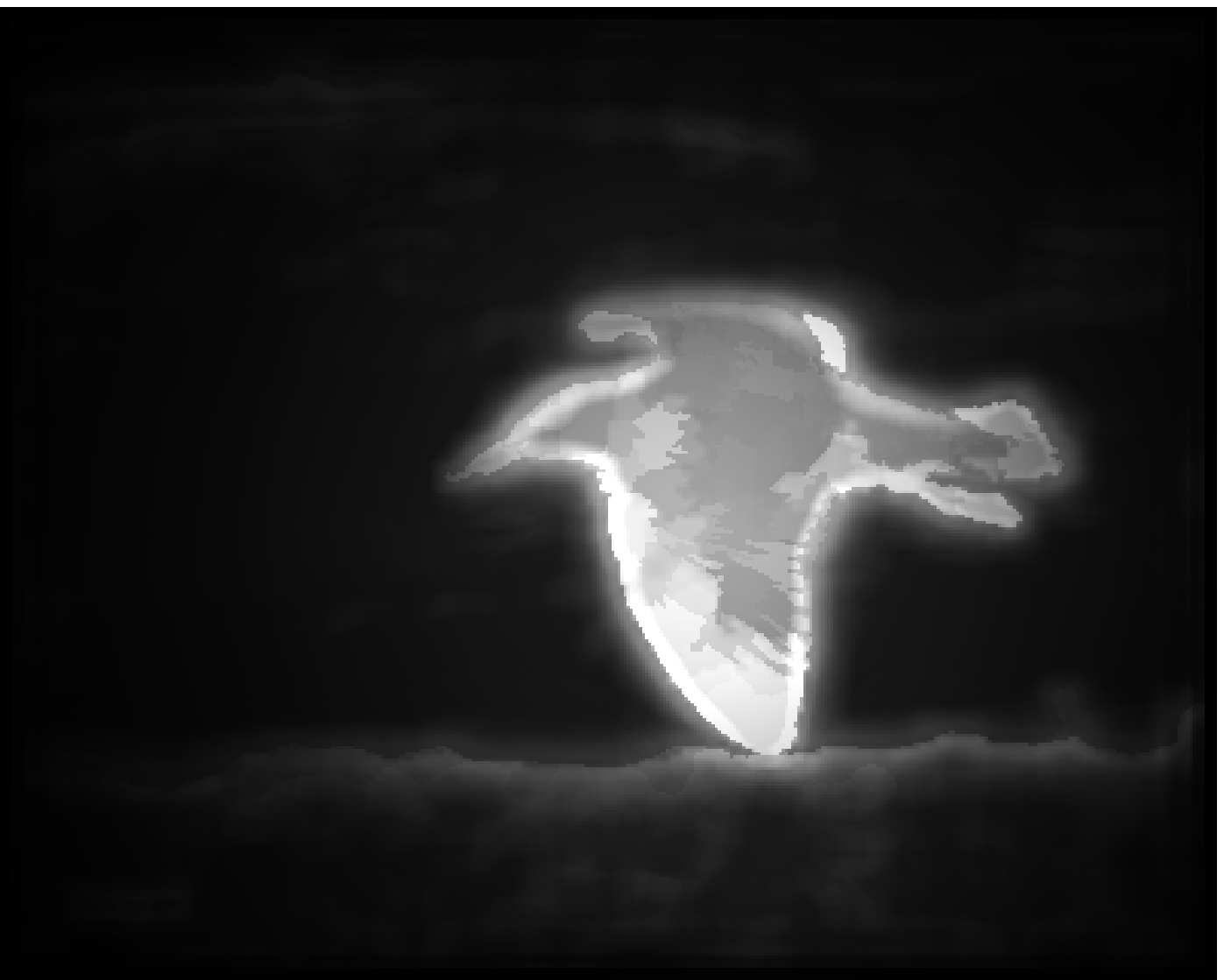} &
\includegraphics[height=1.6cm]{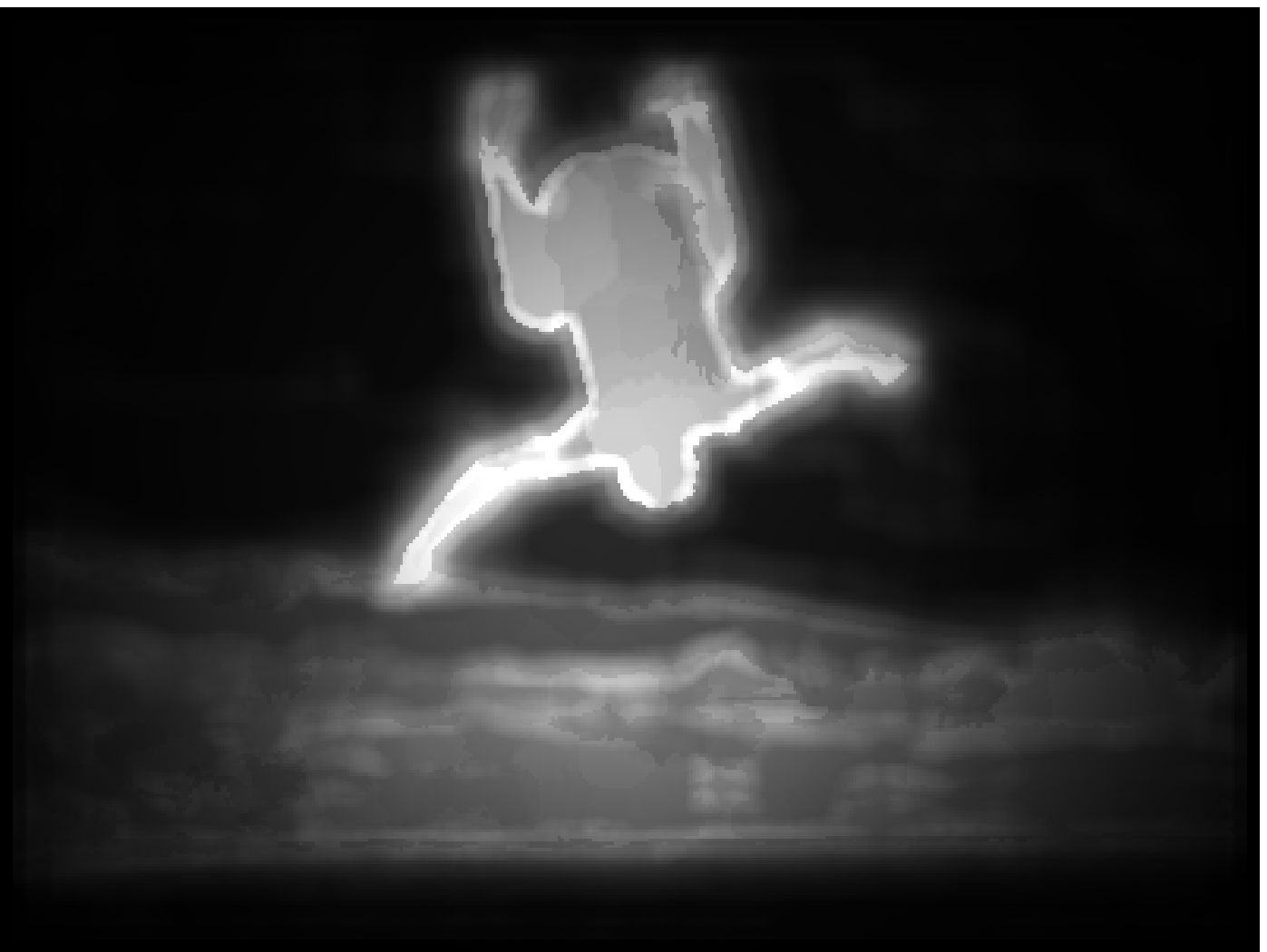} \\
HS&
\includegraphics[height=1.6cm]{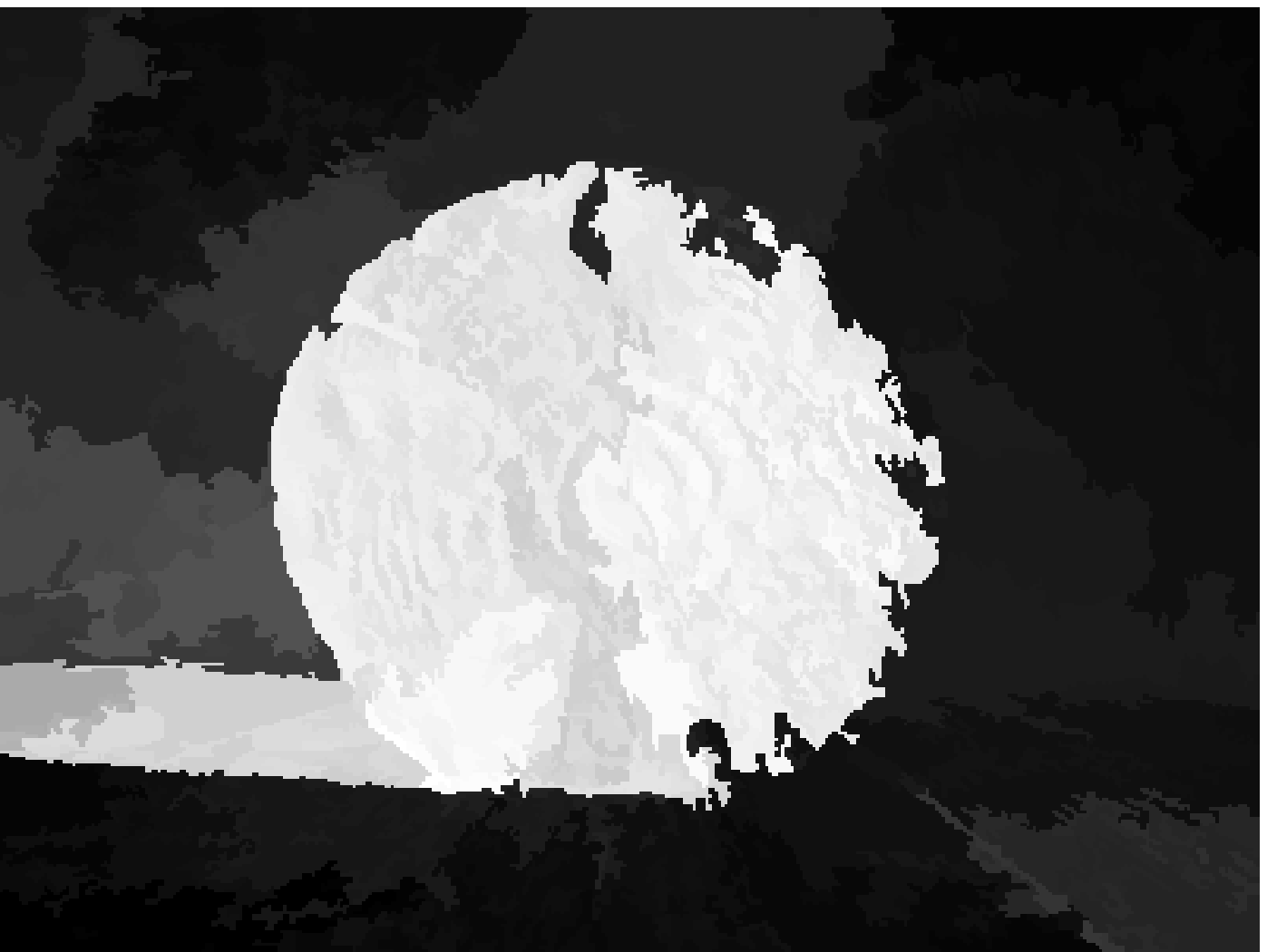} &
\includegraphics[height=1.6cm]{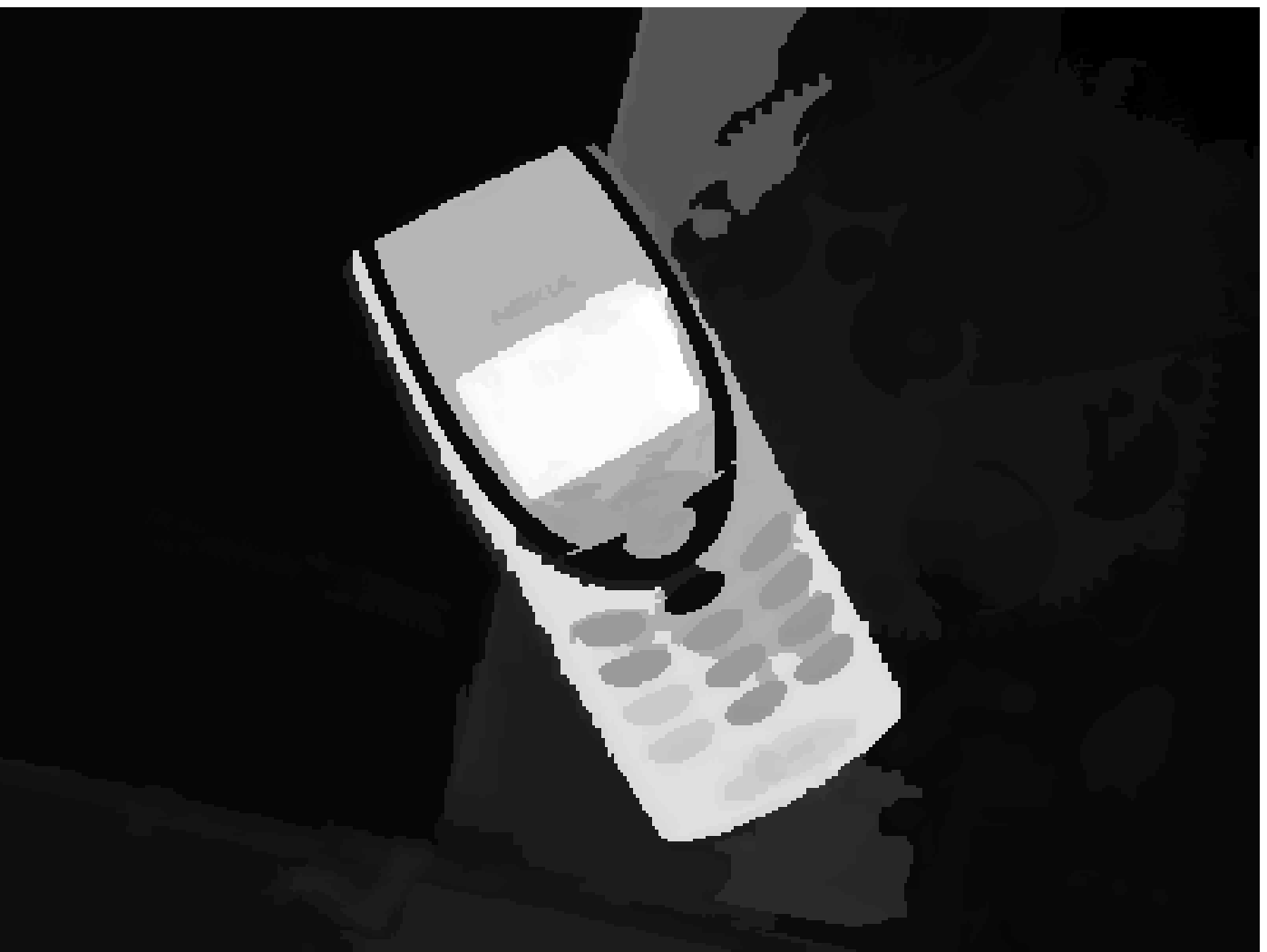} &
\includegraphics[height=1.6cm]{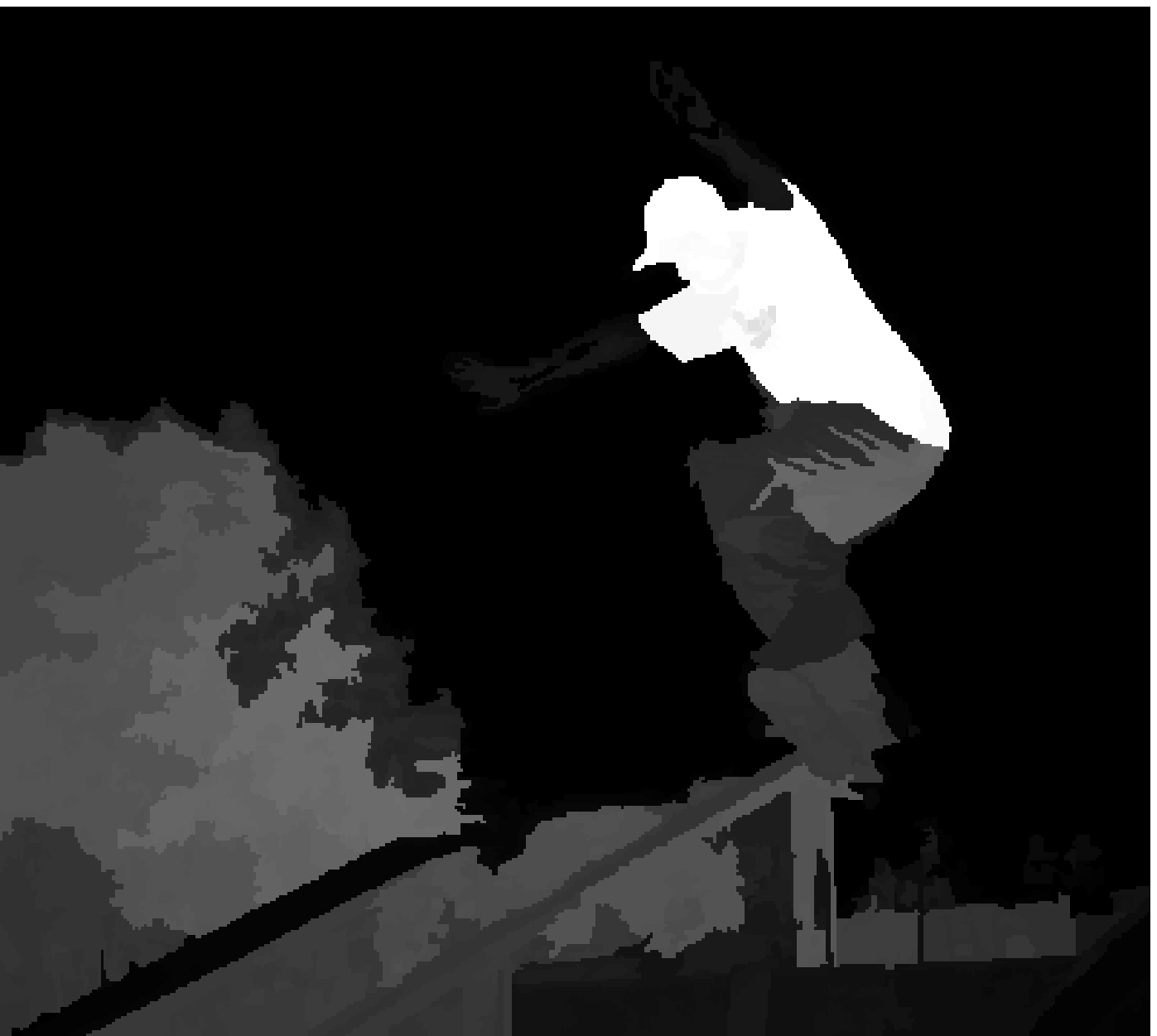} &
\includegraphics[height=1.6cm]{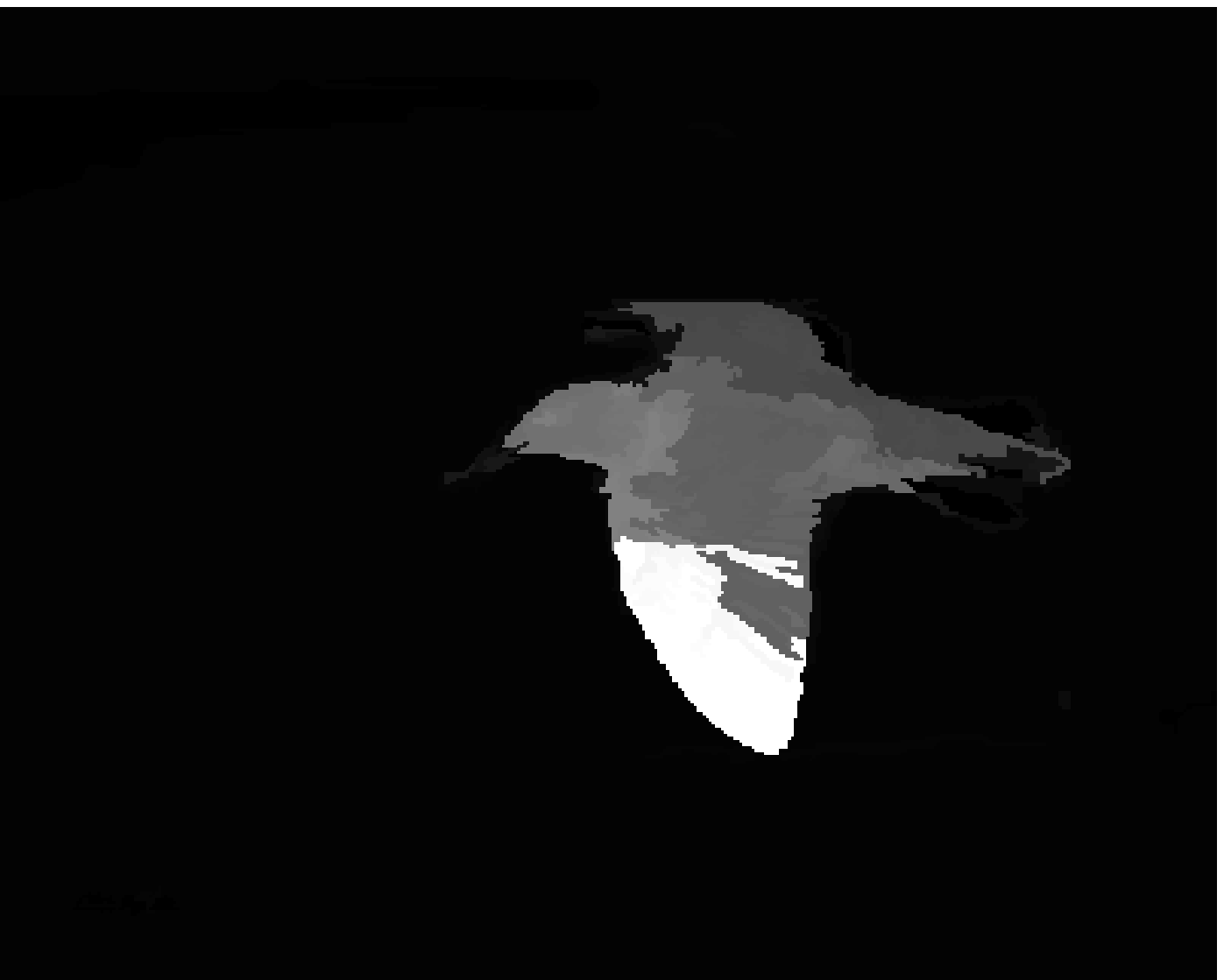} &
\includegraphics[height=1.6cm]{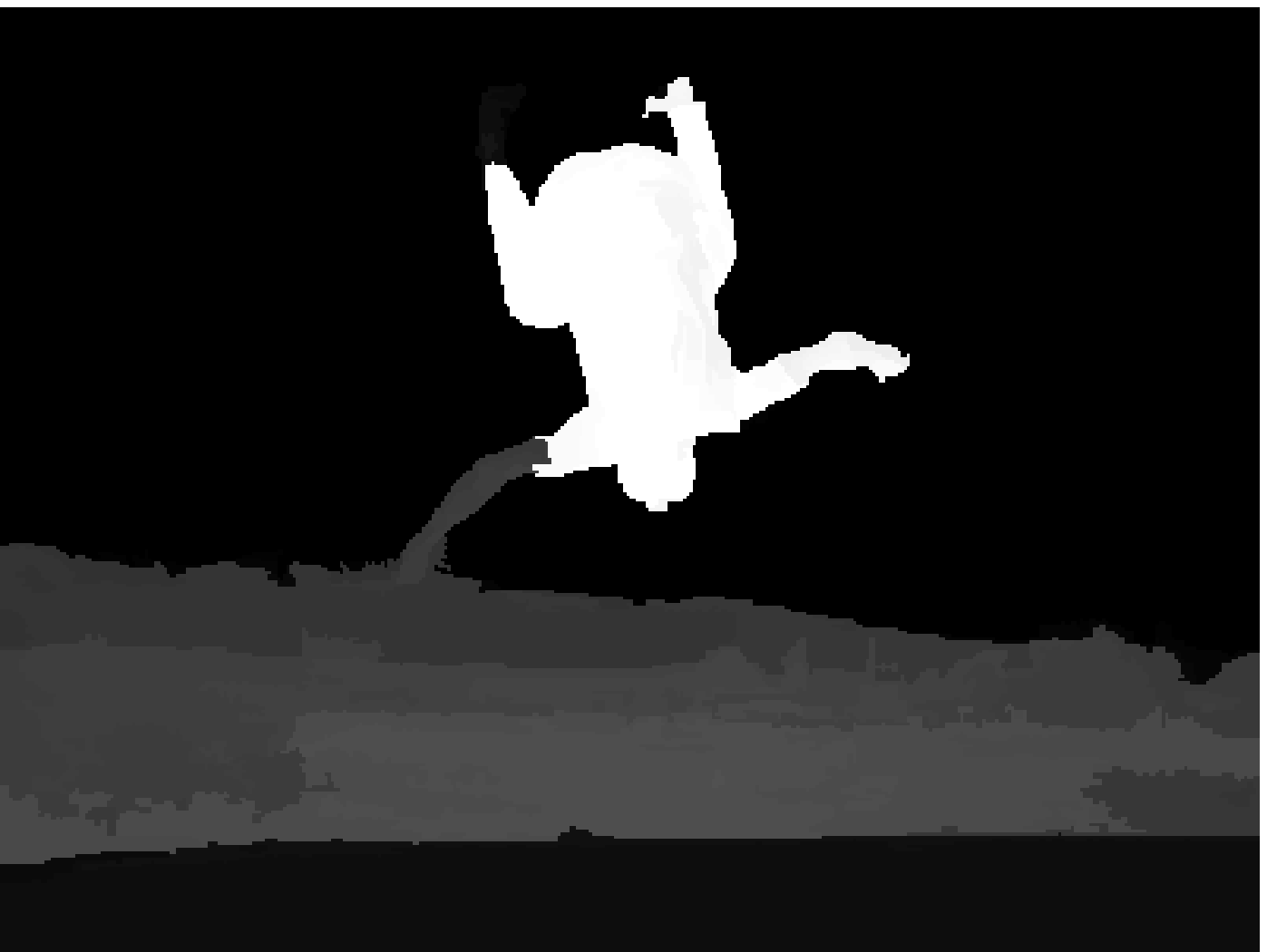} \\
ST&
\includegraphics[height=1.6cm]{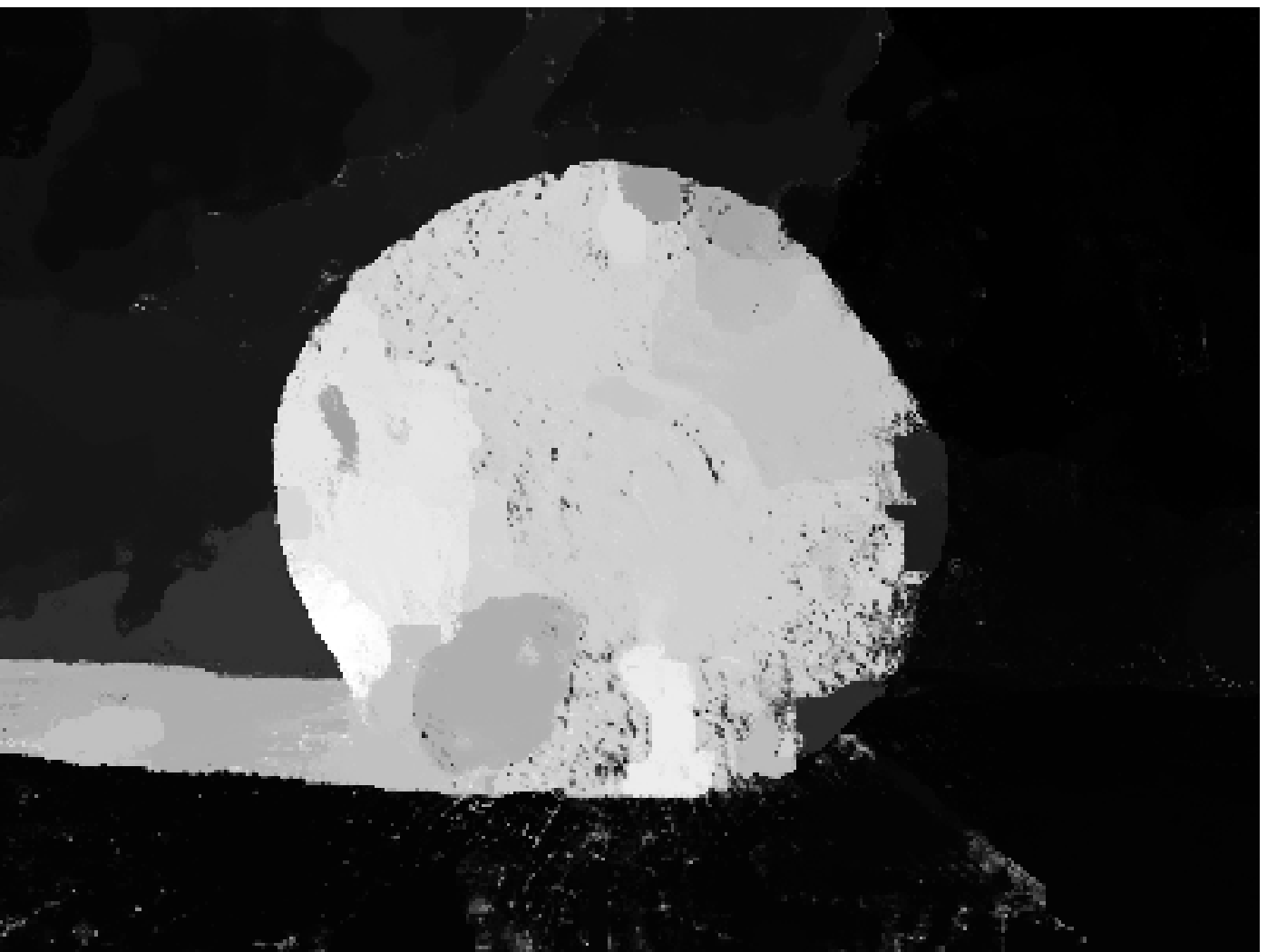} &
\includegraphics[height=1.6cm]{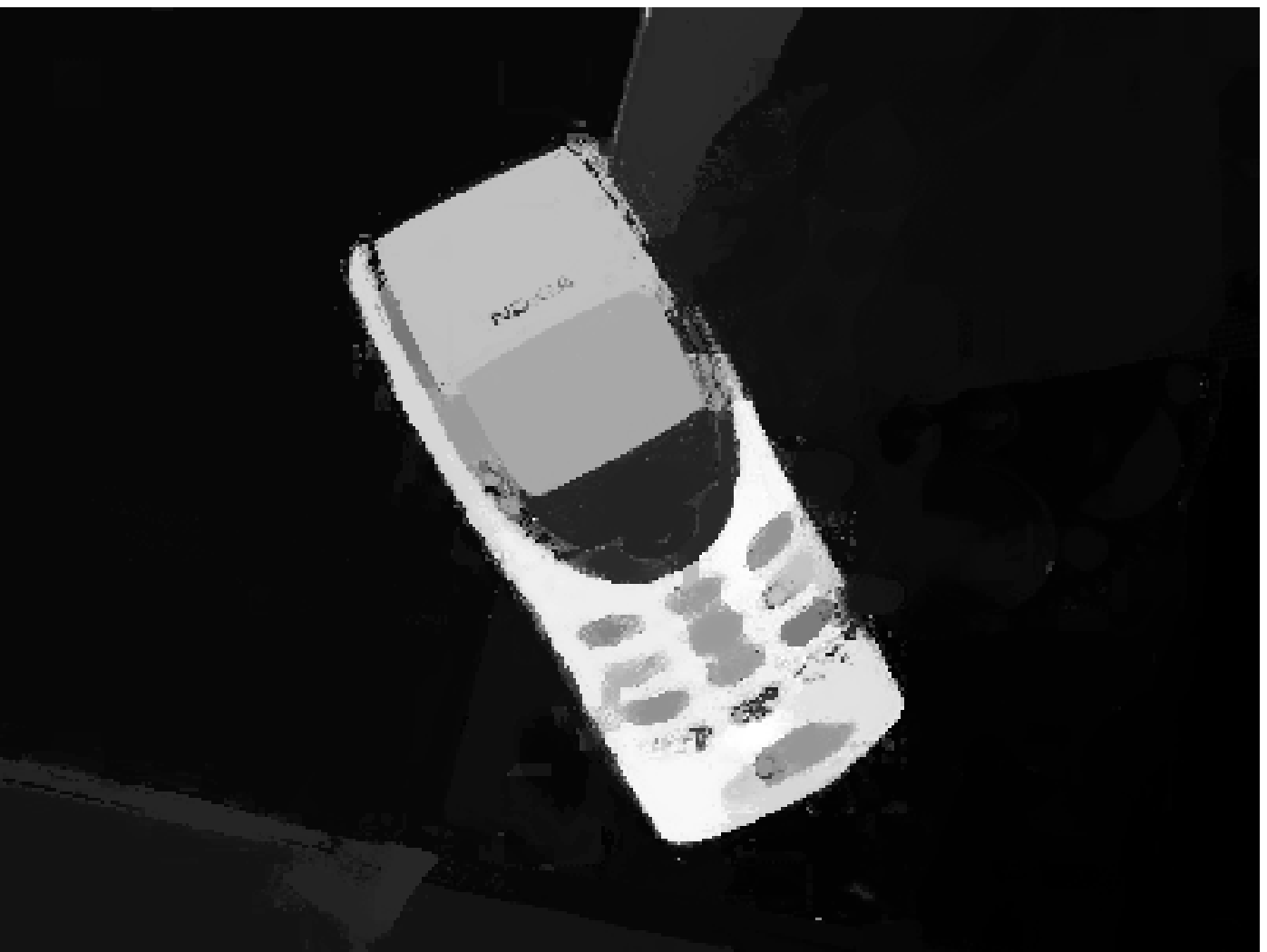} &
\includegraphics[height=1.6cm]{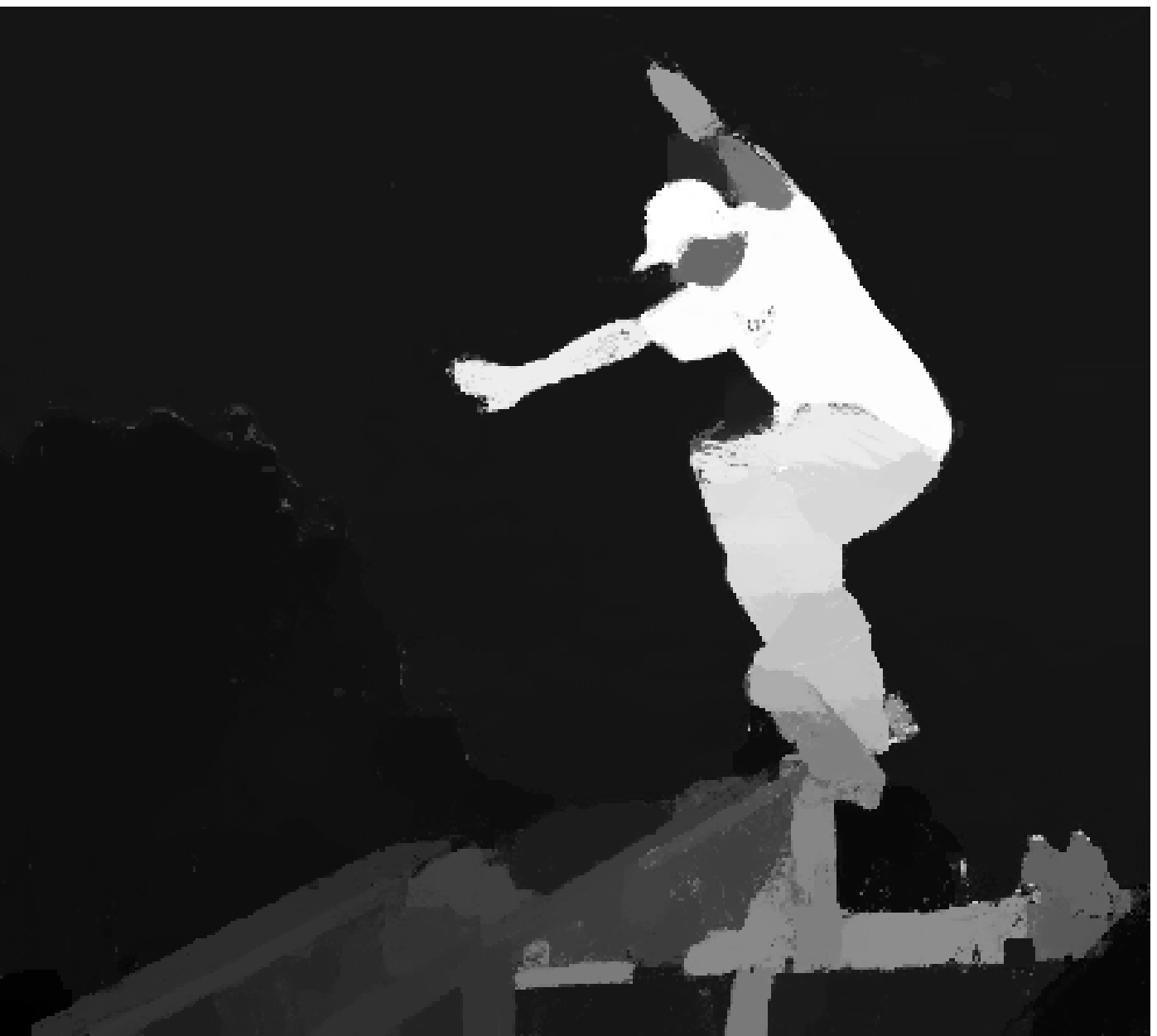} &
\includegraphics[height=1.6cm]{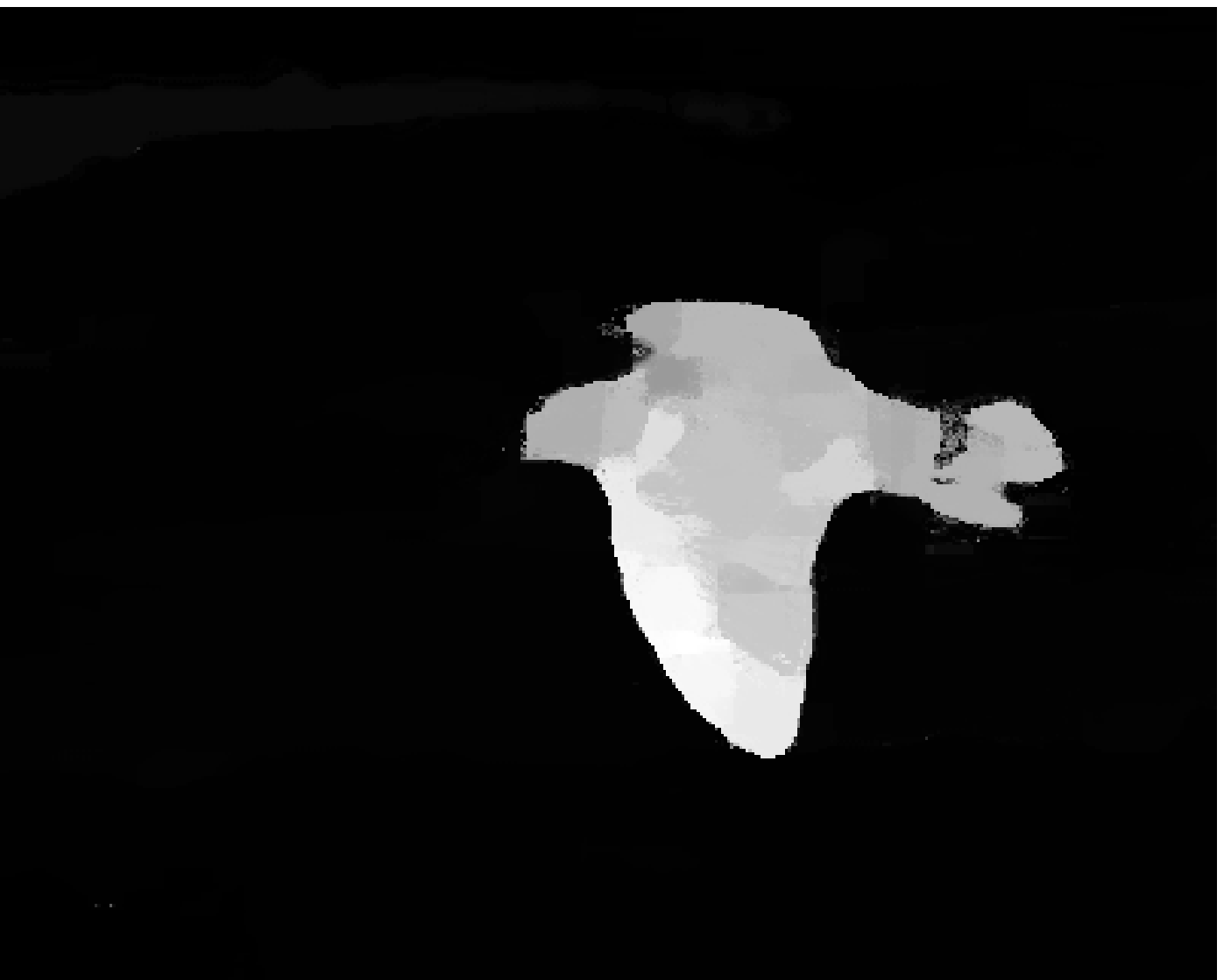} &
\includegraphics[height=1.6cm]{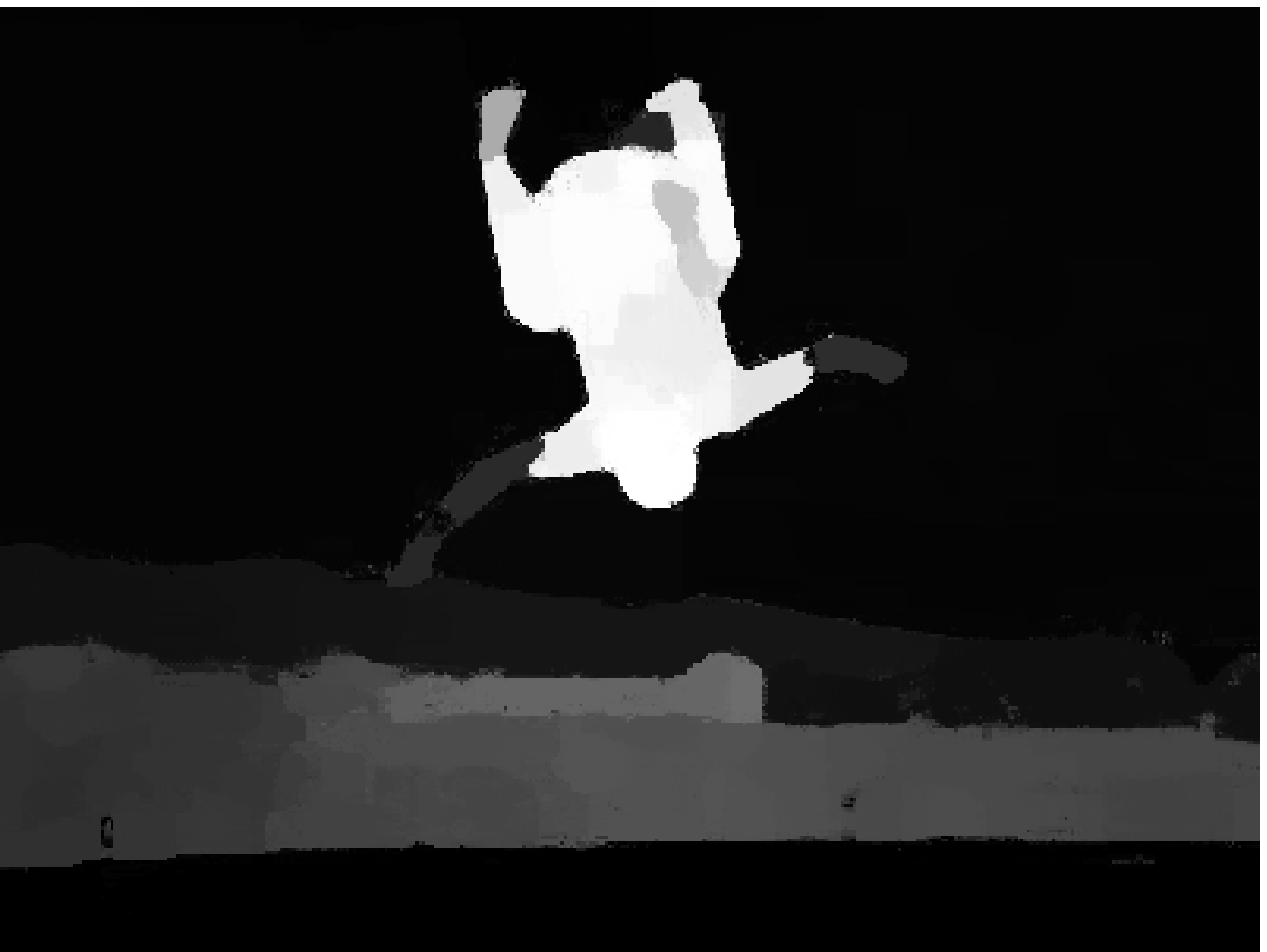} \\
HP&
\includegraphics[height=1.6cm]{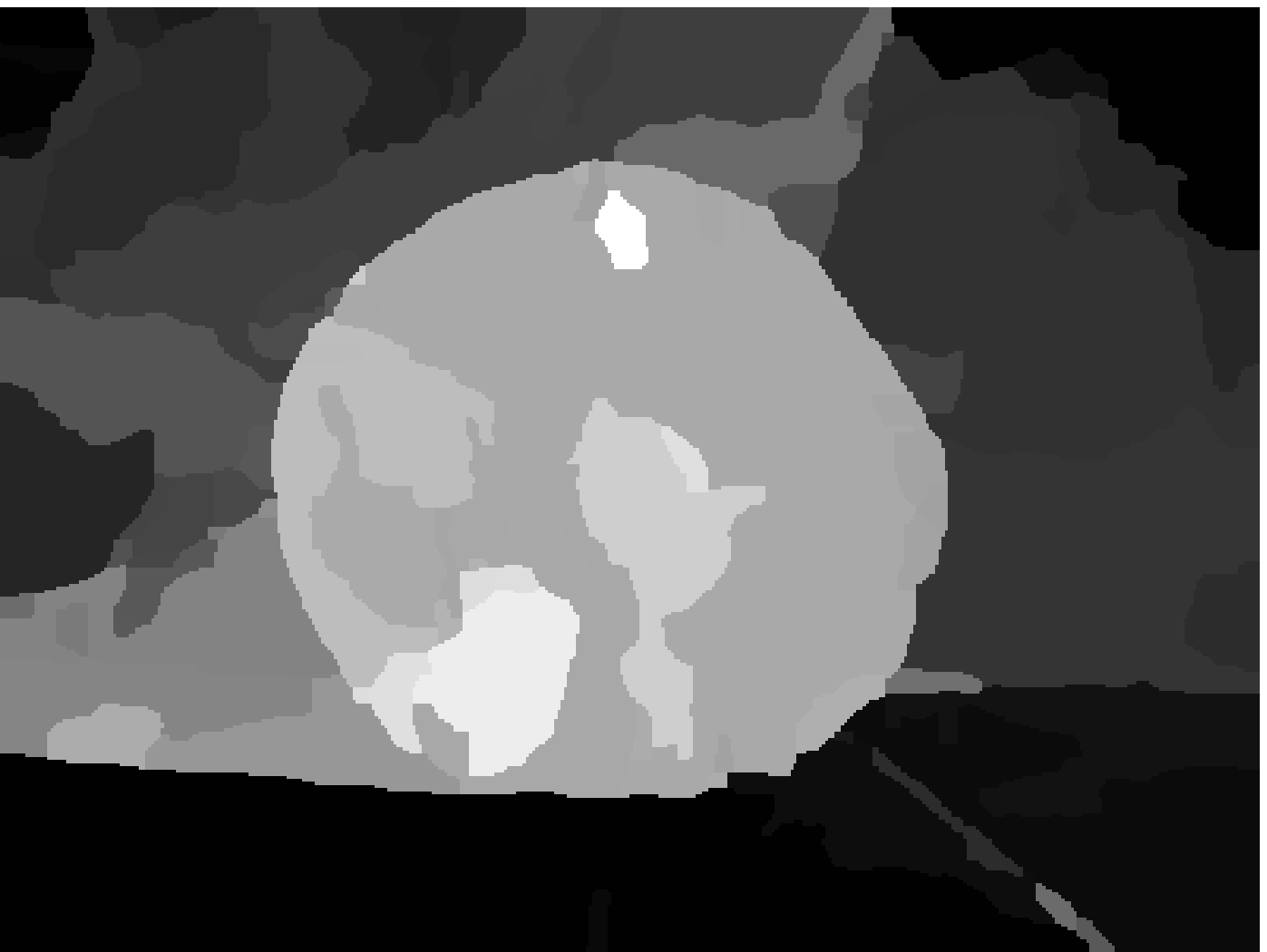} &
\includegraphics[height=1.6cm]{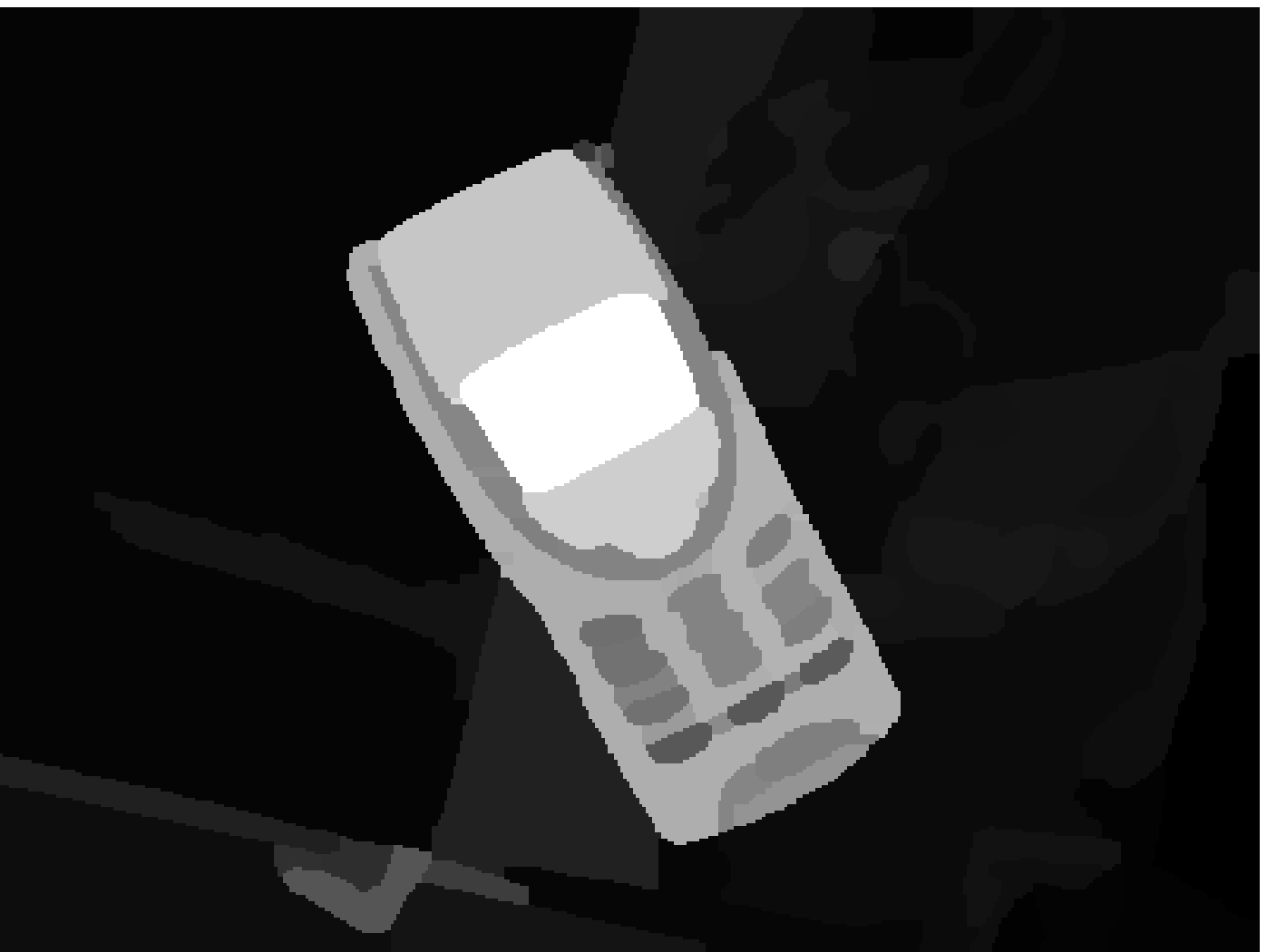} &
\includegraphics[height=1.6cm]{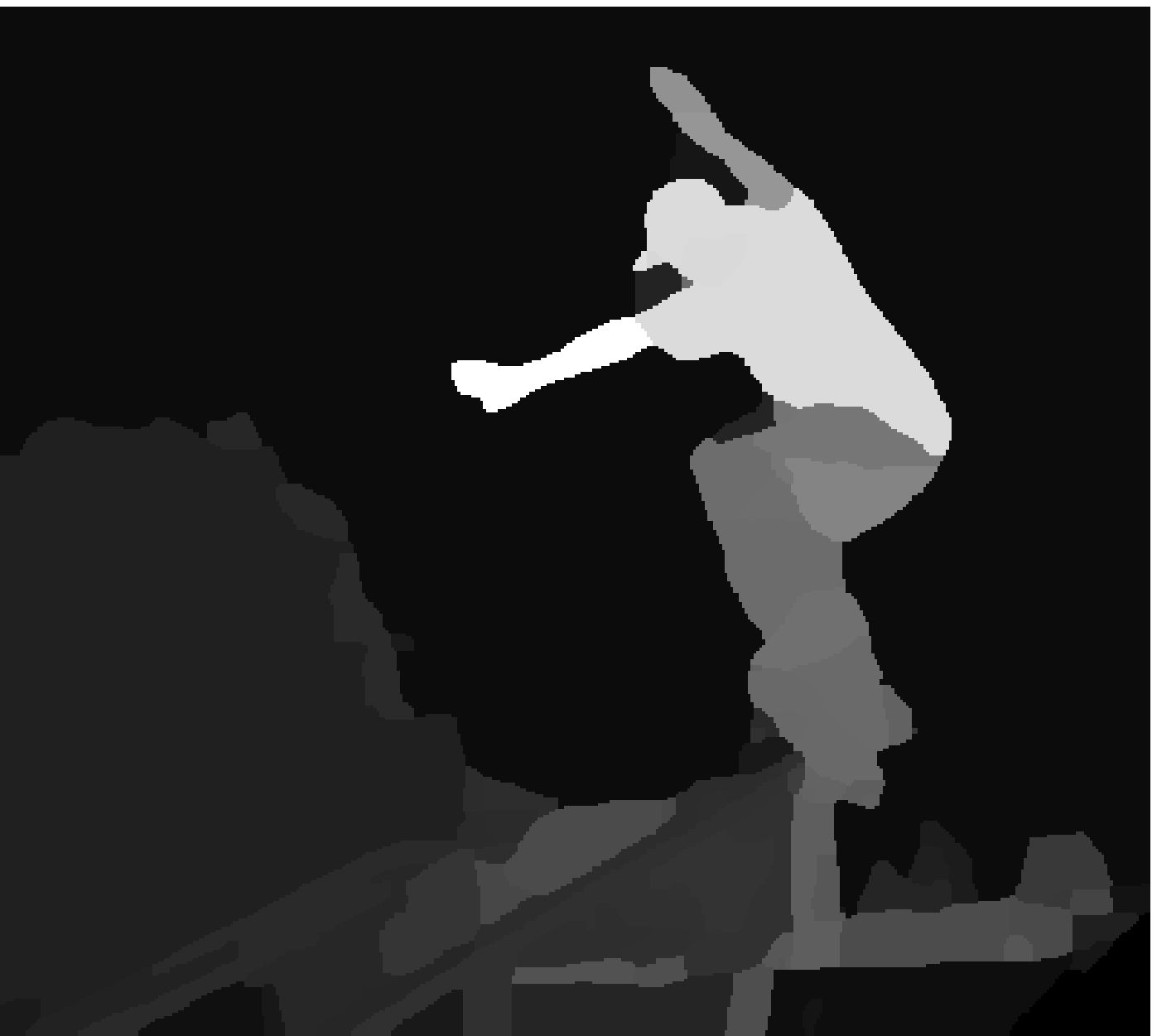} &
\includegraphics[height=1.6cm]{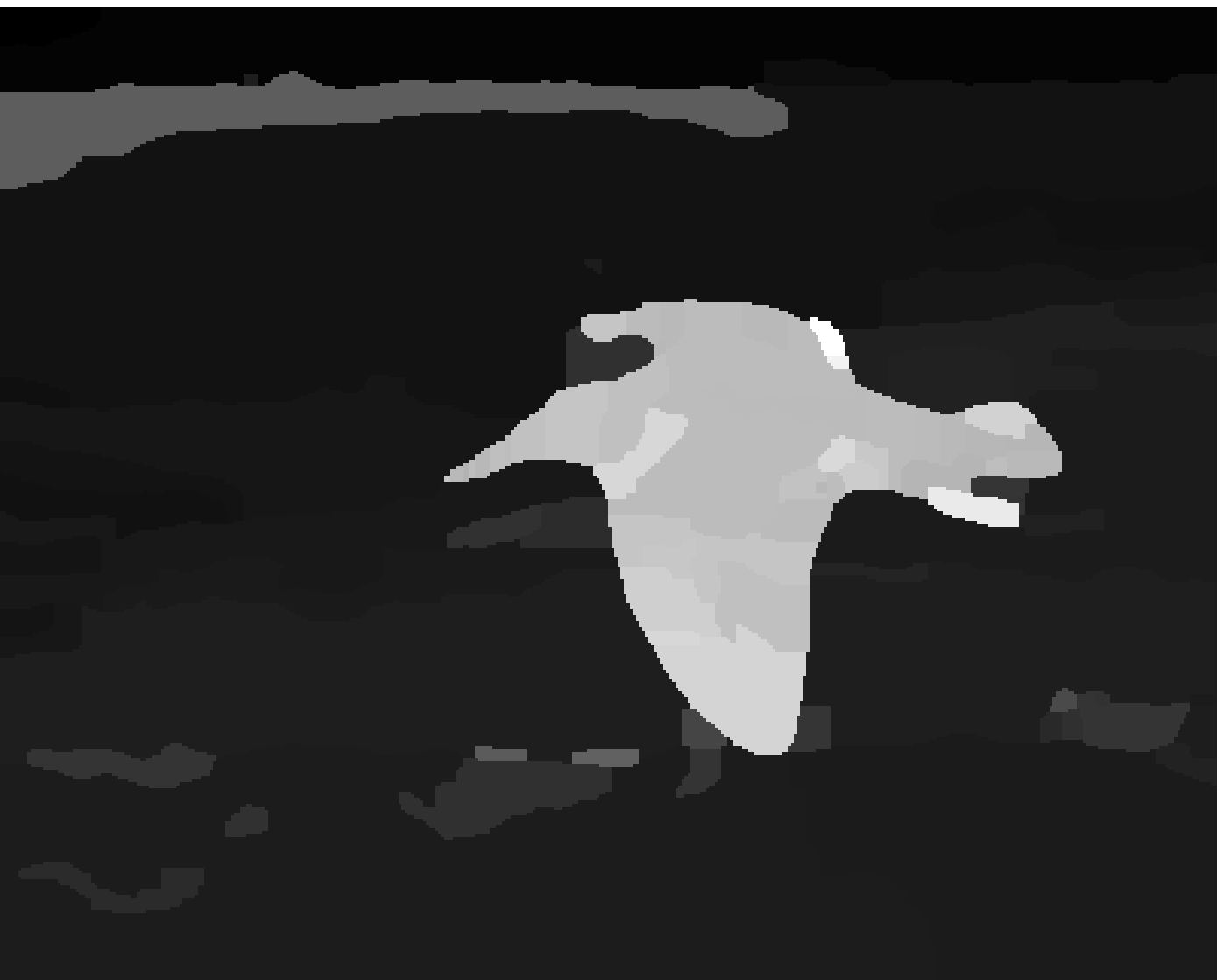} &
\includegraphics[height=1.6cm]{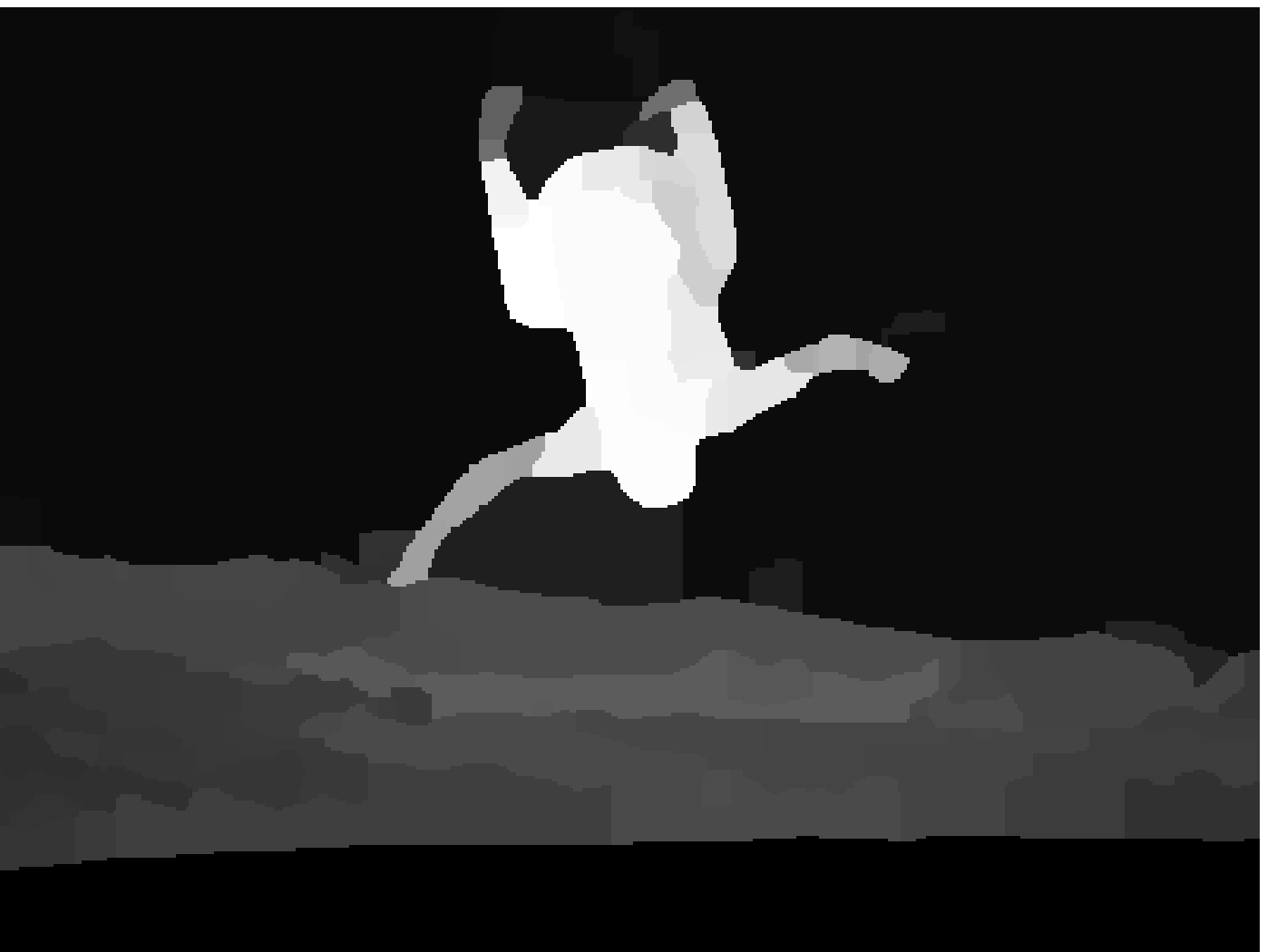} \\
SOH&
\includegraphics[height=1.6cm]{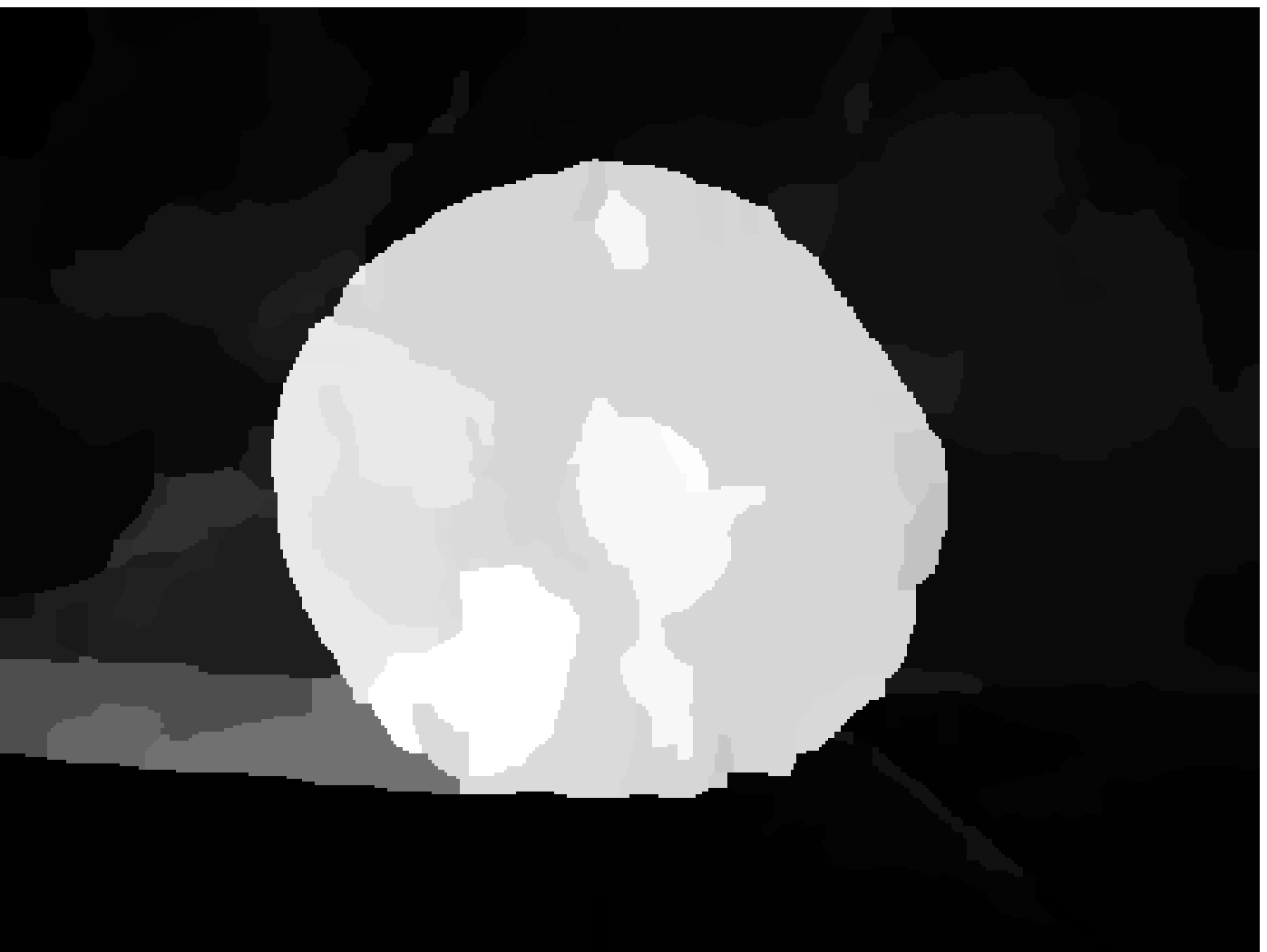} &
\includegraphics[height=1.6cm]{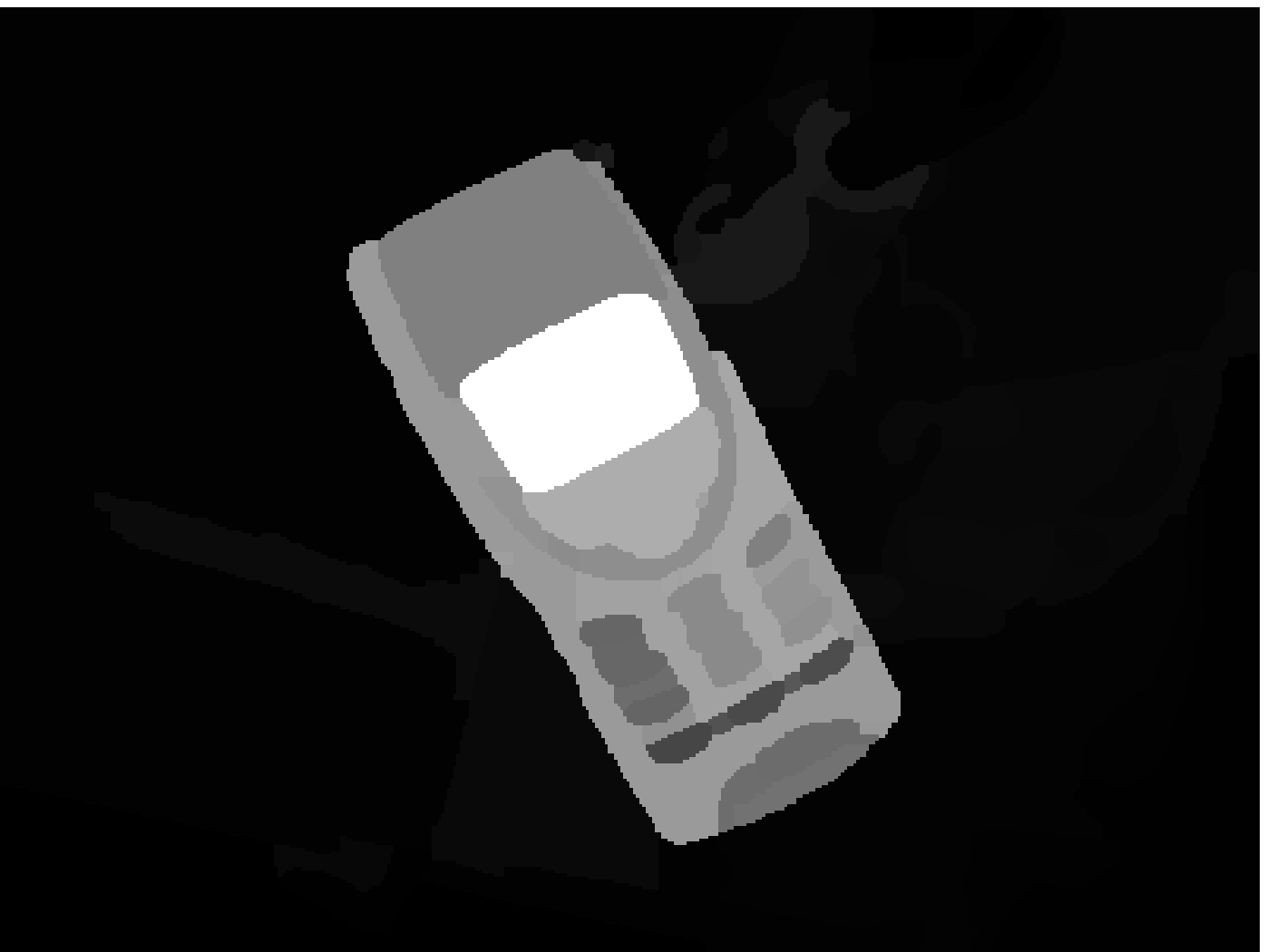} &
\includegraphics[height=1.6cm]{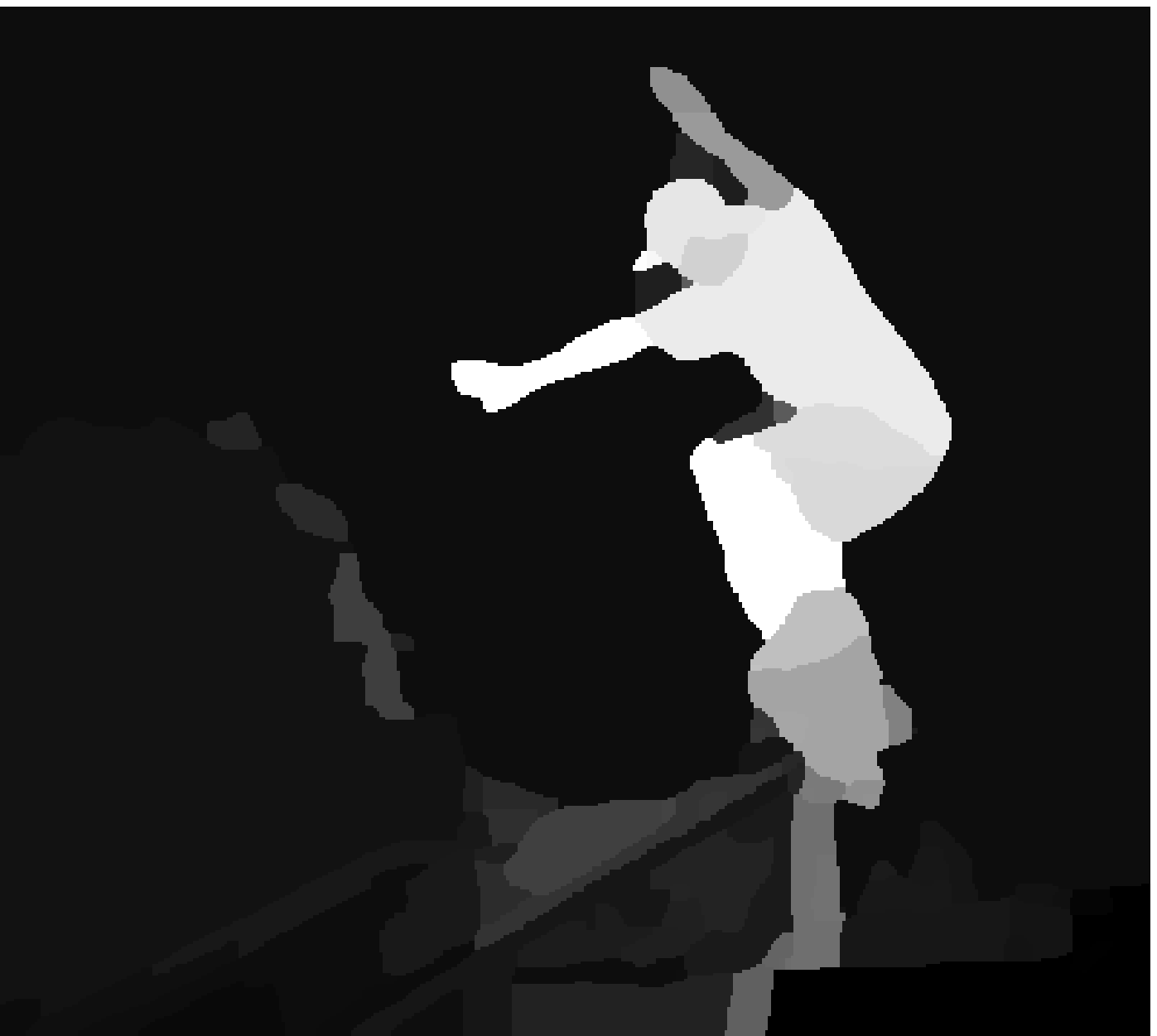} &
\includegraphics[height=1.6cm]{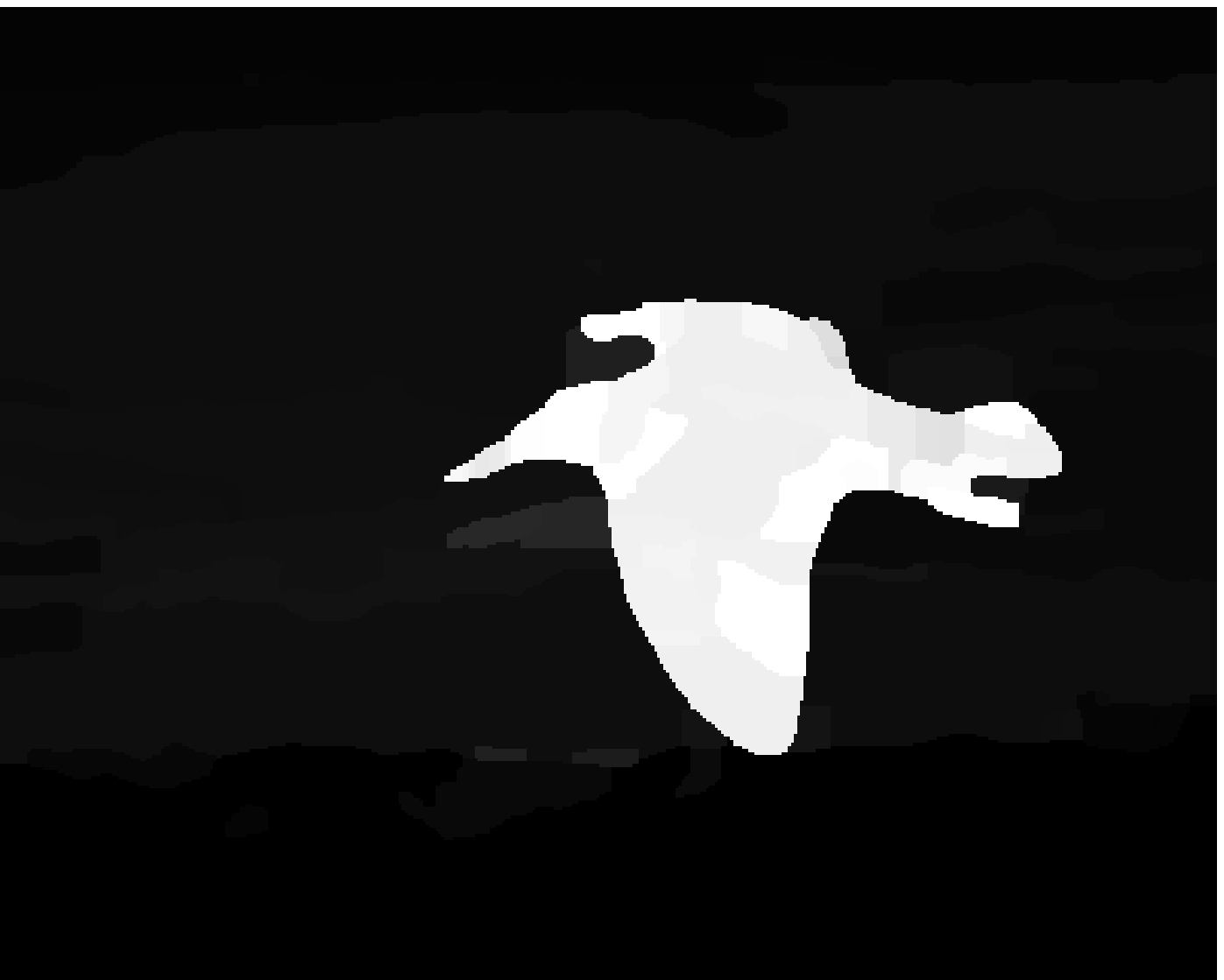} &
\includegraphics[height=1.6cm]{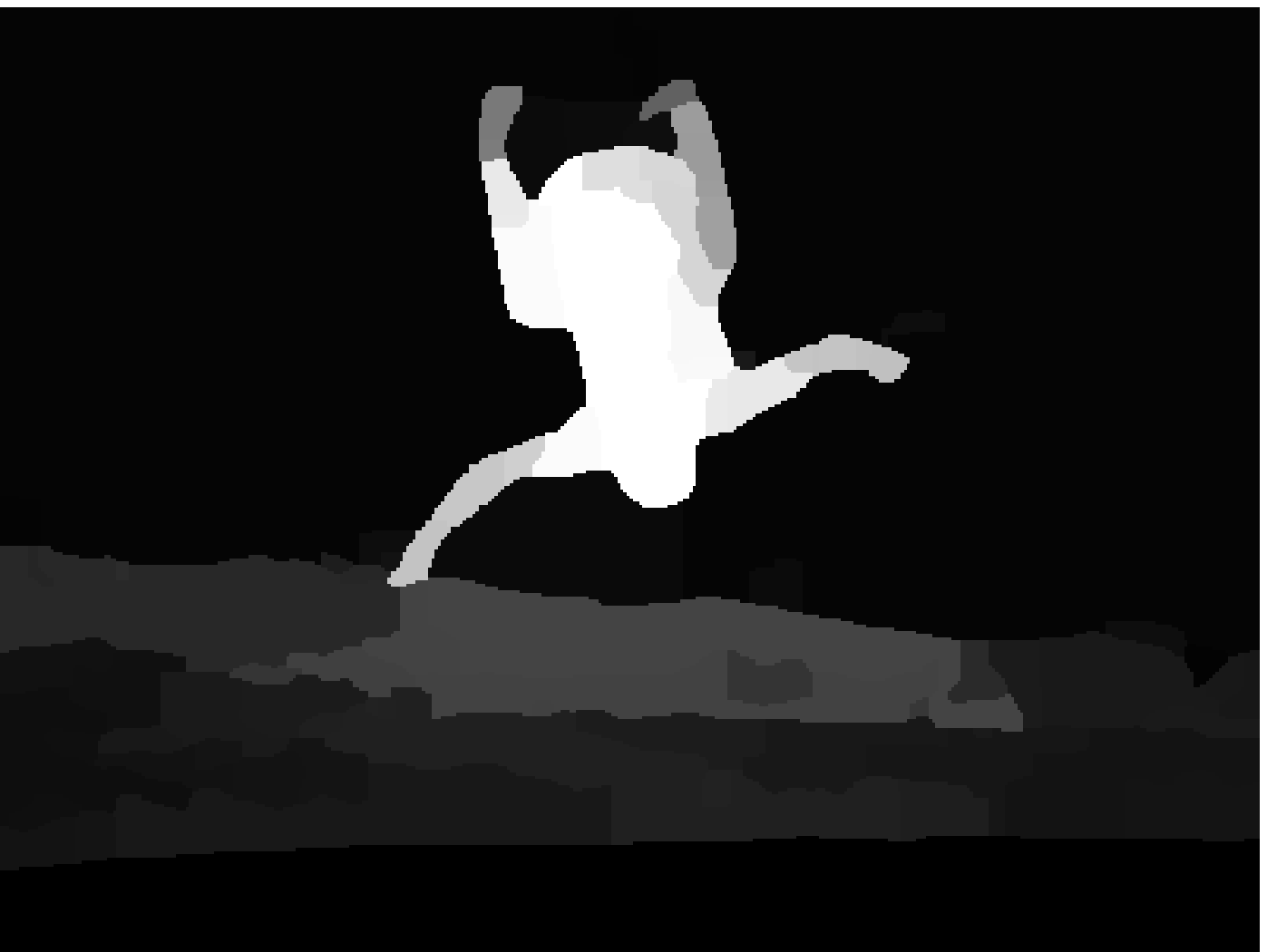} \\
GT&
\includegraphics[height=1.6cm]{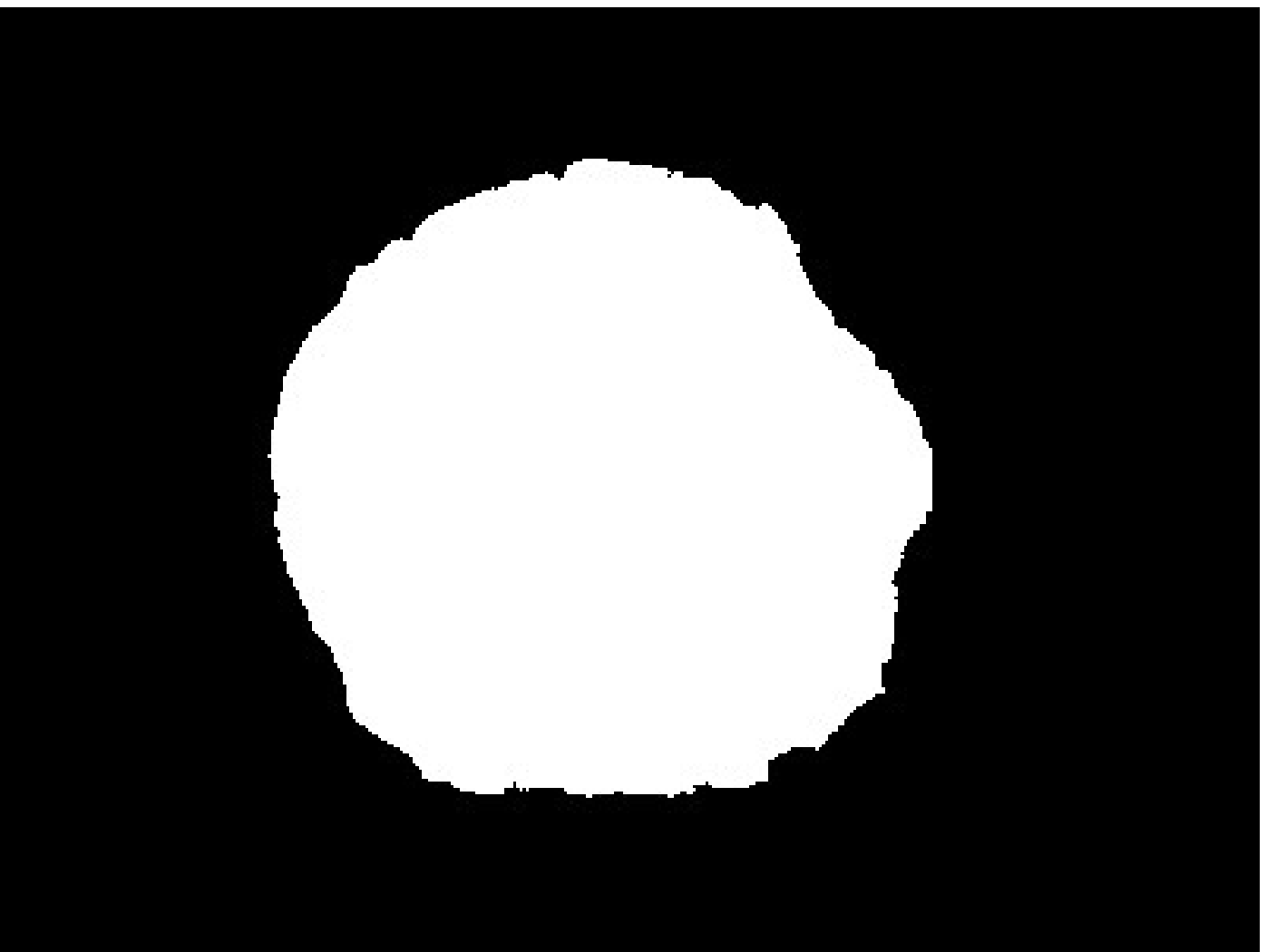} &
\includegraphics[height=1.6cm]{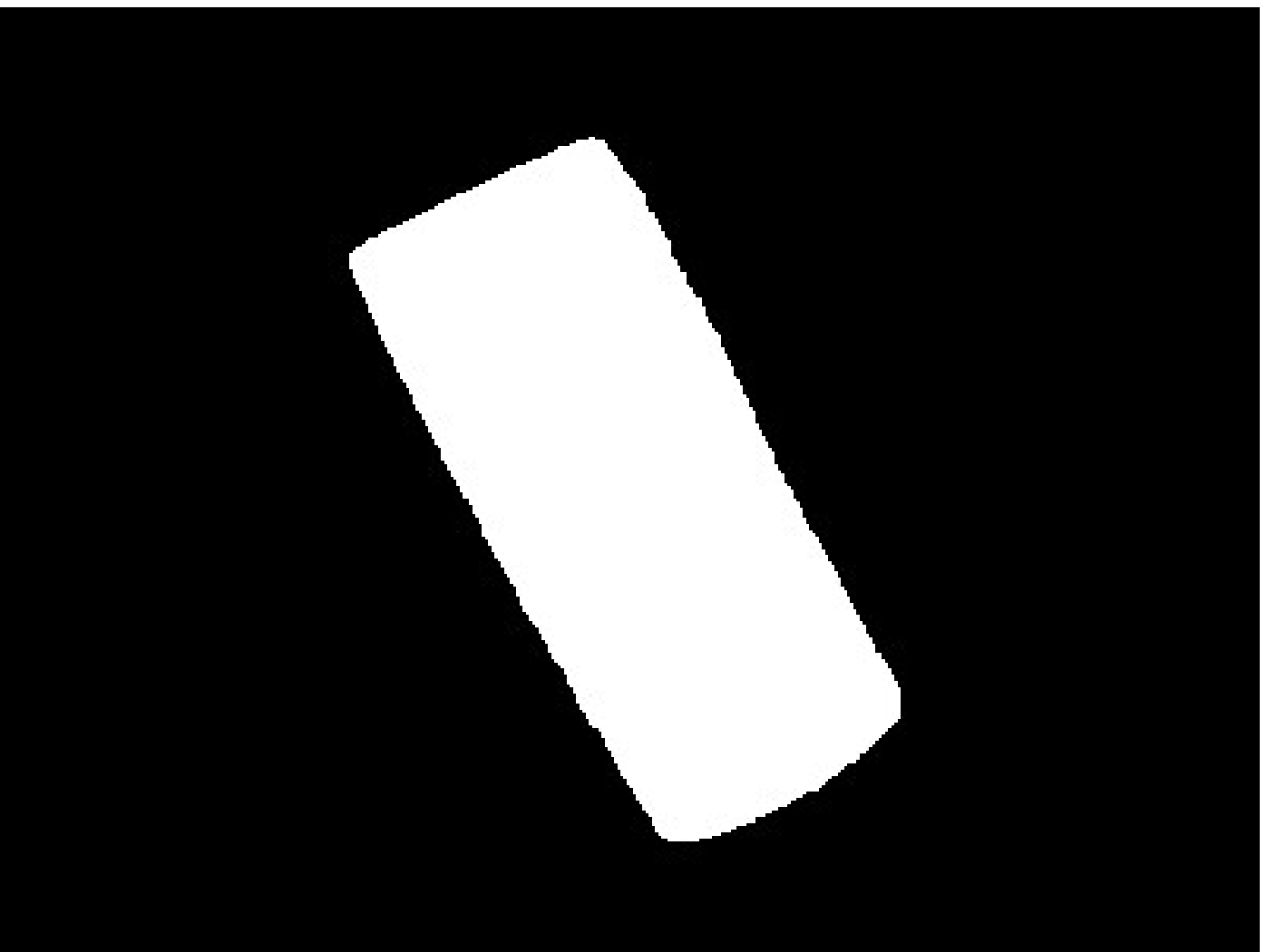} &
\includegraphics[height=1.6cm]{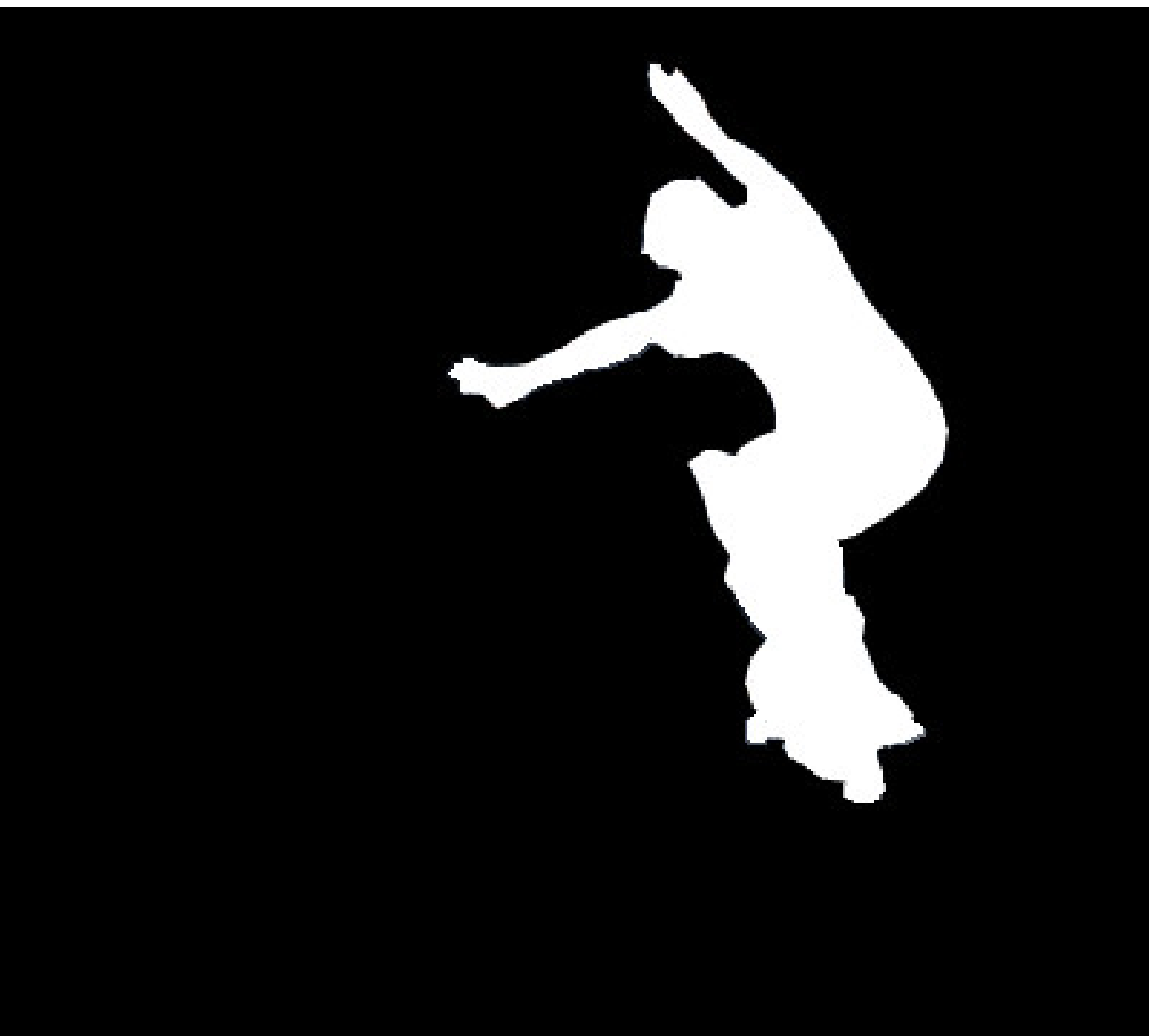} &
\includegraphics[height=1.6cm]{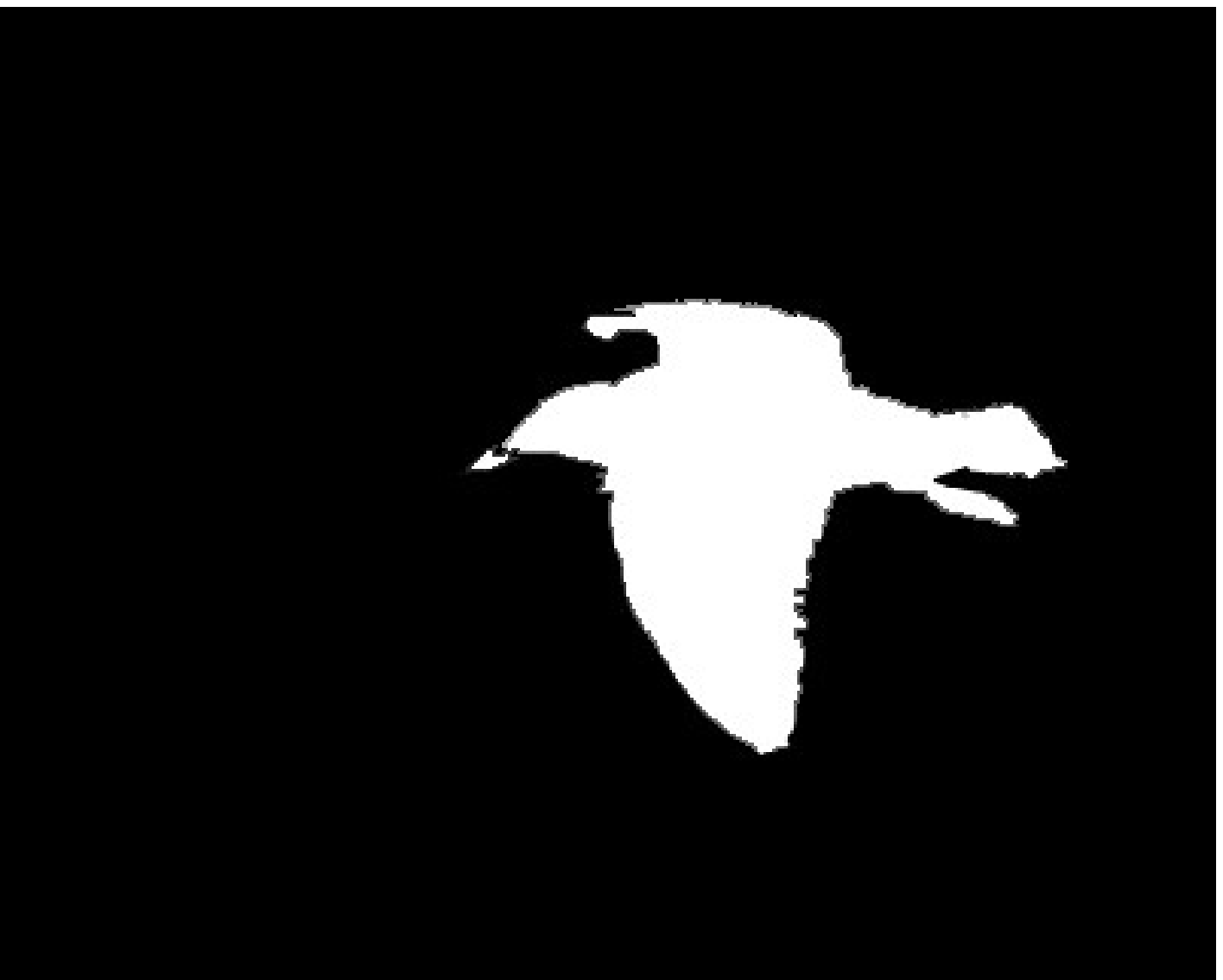} &
\includegraphics[height=1.6cm]{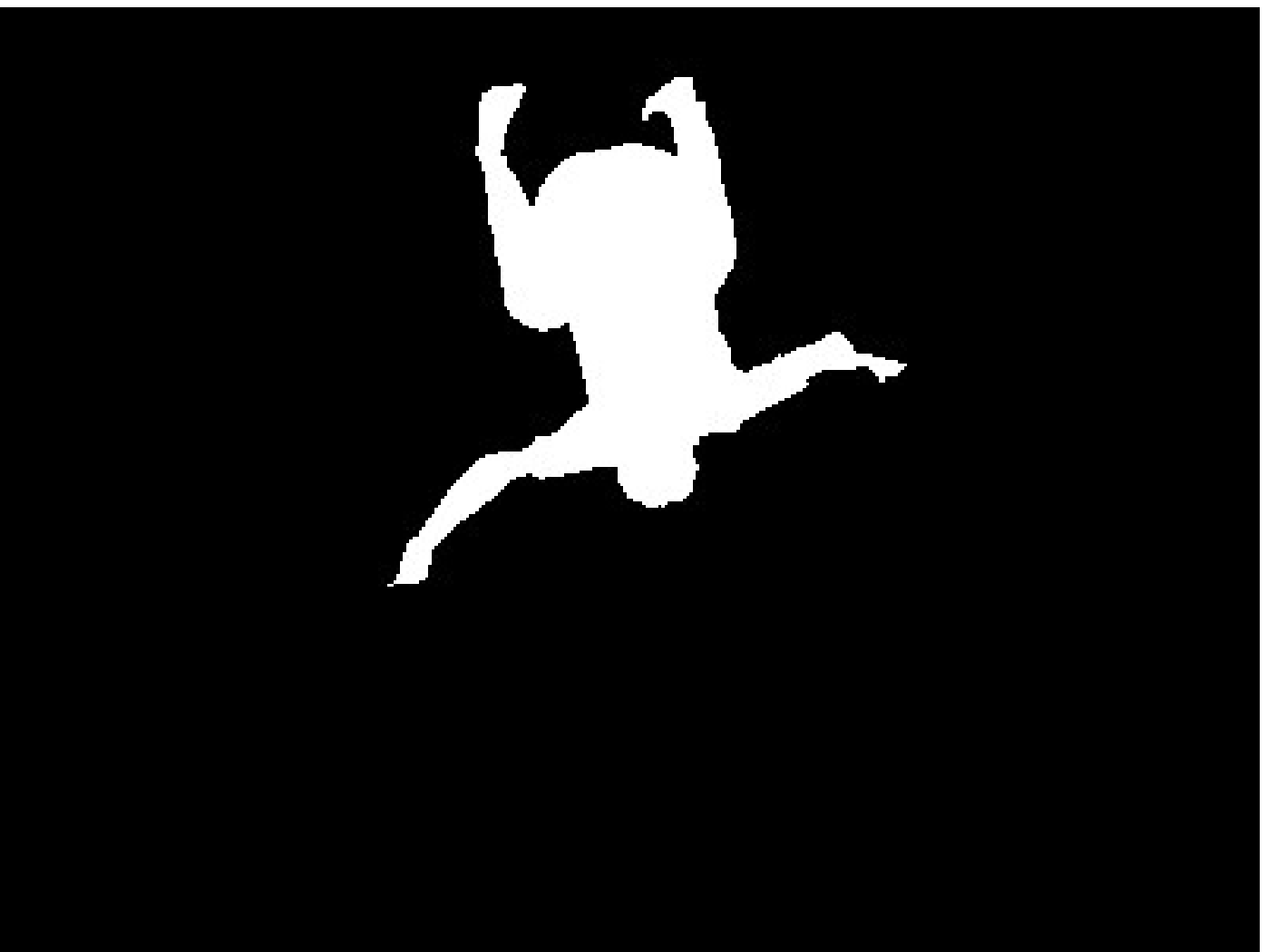} \\
& (a) & (b) & (c) & (d) & (e) \\
\end{tabular}
\caption{Experimental results on some ASD images. First row: IM original images; saliency maps generated using FT \cite{Achanta-09cvpr},  RC \cite{Cheng-11cvpr}, SF \cite{perazzi-cvpr12}, PCAS \cite{margolin-cvpr13}, HS \cite{Yan-cvpr13}, ST \cite{ZLiu-tip14}, our models HP and SOH,  and GT ground truth masks.} \label{fig:ASDimage_comp}
\end{figure}

For a visual comparison, saliency maps generated by the eight methods (pixel, partition and hierarchy-based) on the ASD dataset are shown in Figure~\ref{fig:ASDimage_comp}. It can be seen that FT, RC and SF detect some salient regions but fail to highlight complete objects, while PCAS detects mostly the outlines of objects. In turn, the four methods based on hierarchies capture both the outline and the inner parts of salient objects. On this simple set, all four methods perform well, though in some cases HS and HP fail to capture complete objects (such as Fig.\ref{fig:ASDimage_comp}(c)).

\begin{figure}[thbp]
\centering
\begin{tabular}{cc}
\includegraphics[height=4.5cm]{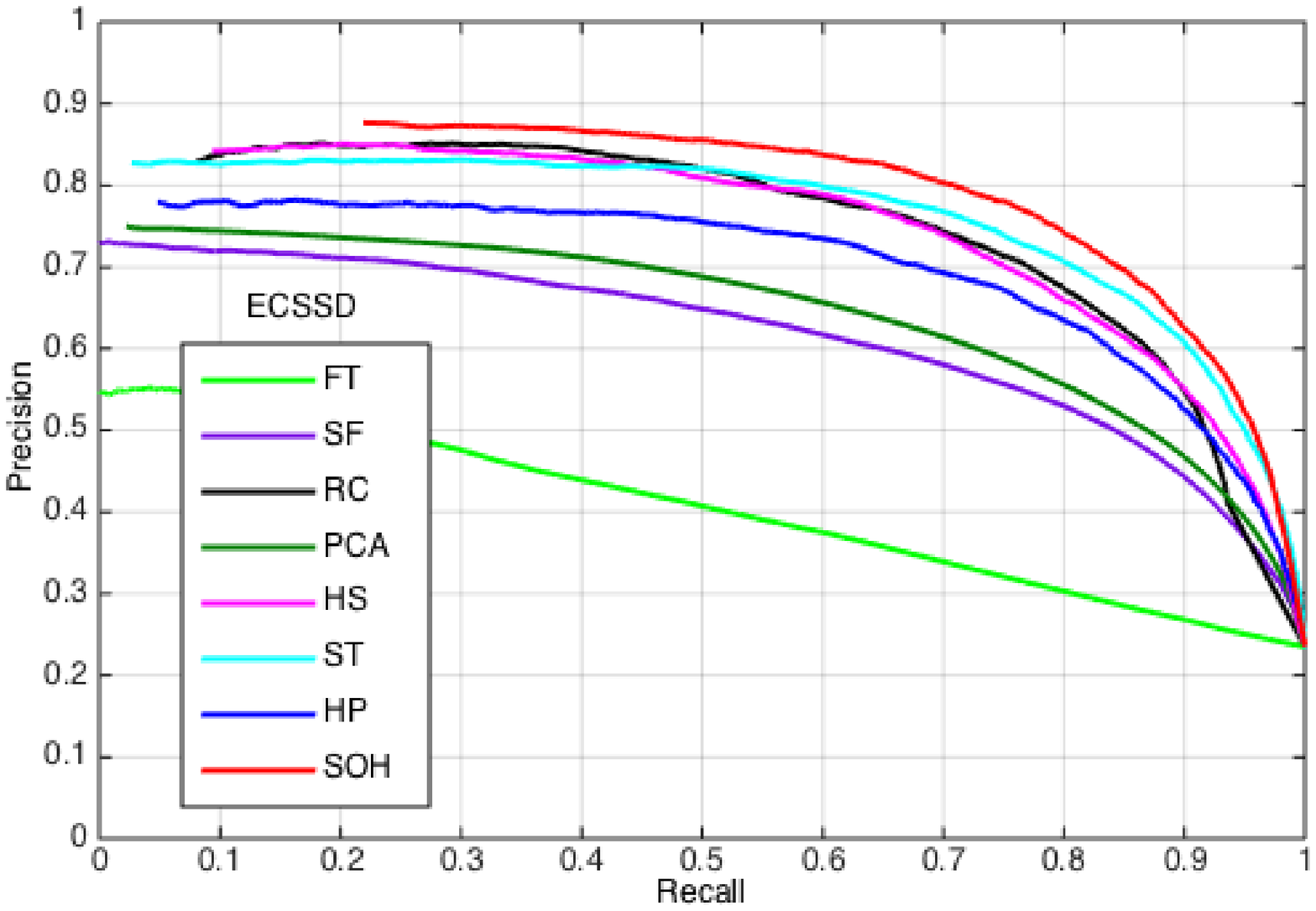} &\
\raisebox{.2\height}{\includegraphics[height=3.5cm]{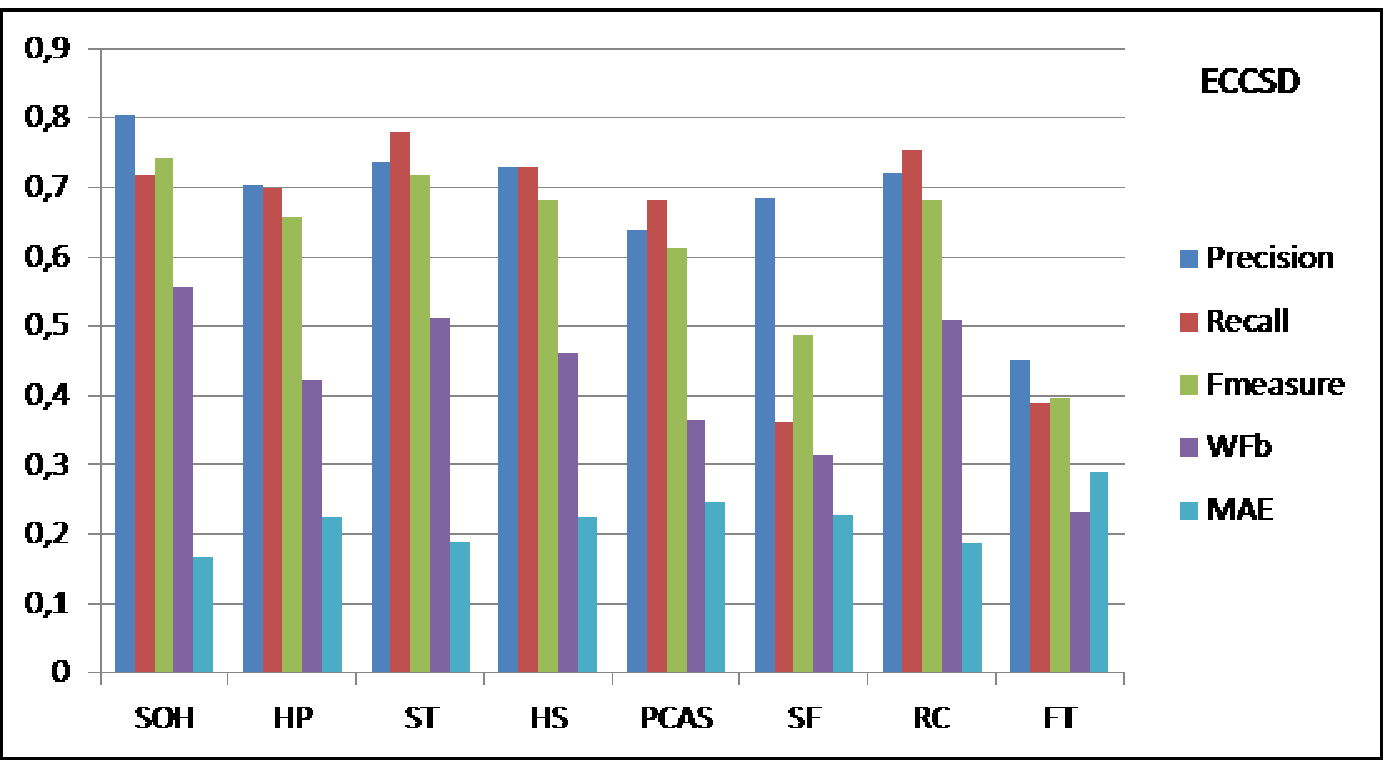}} \\
(a) & (b)\\
\end{tabular}
\caption{Precision-recall curves (left) and precision, recall, $F_{\beta}$-measure, $Wf_{\beta}$ and Mae values (right) for ECSSD (our methods SOH and HP, ST \cite{ZLiu-tip14}, HS \cite{Yan-cvpr13}, FT \cite{Achanta-09cvpr}, RC \cite{Cheng-11cvpr}, SF \cite{perazzi-cvpr12}, PCAS \cite{margolin-cvpr13})}
\label{fig:prfm_ecssd}
\end{figure}

Precision-recall curves and metrics for adaptive-thresholded maps for MSRA and ECSSD are given in Figures~\ref{fig:prfm_msra} and \ref{fig:prfm_ecssd}.

For MSRA our direct model SOH and ST achieve the same (best) performance for high recall values ($>0.70$), RC gives the highest precision for low recall ($<0.70$) , while HP outperforms HS for high recall values ($>0.75$).  For ECSSD, SOH is the best performing method, both in terms of P-R curves and all the other metrics, followed by ST, while HS and RC have similar performance. For this set, HP is the fifth method, followed by PCAS, SF and FT.

\begin{figure}[thbp]
\centering
\begin{tabular}{cccccc}
IM&
\includegraphics[height=1.6cm]{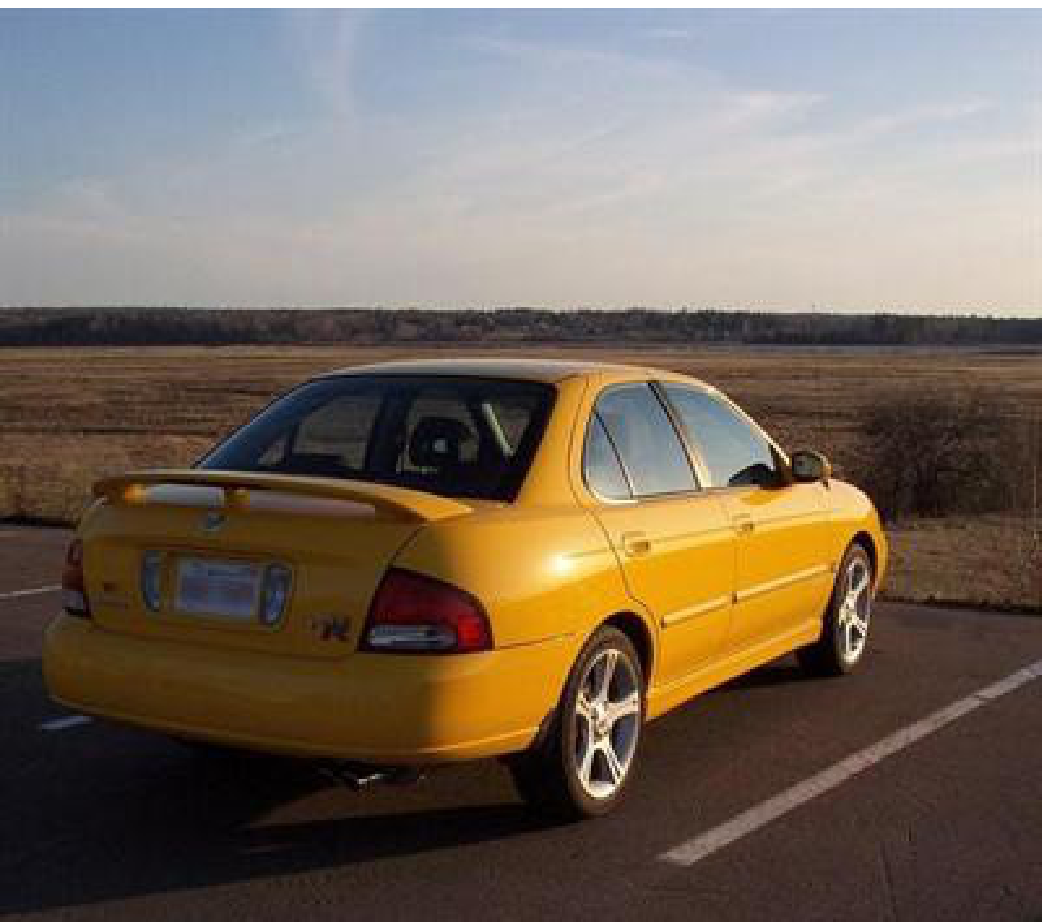} &
\includegraphics[height=1.6cm]{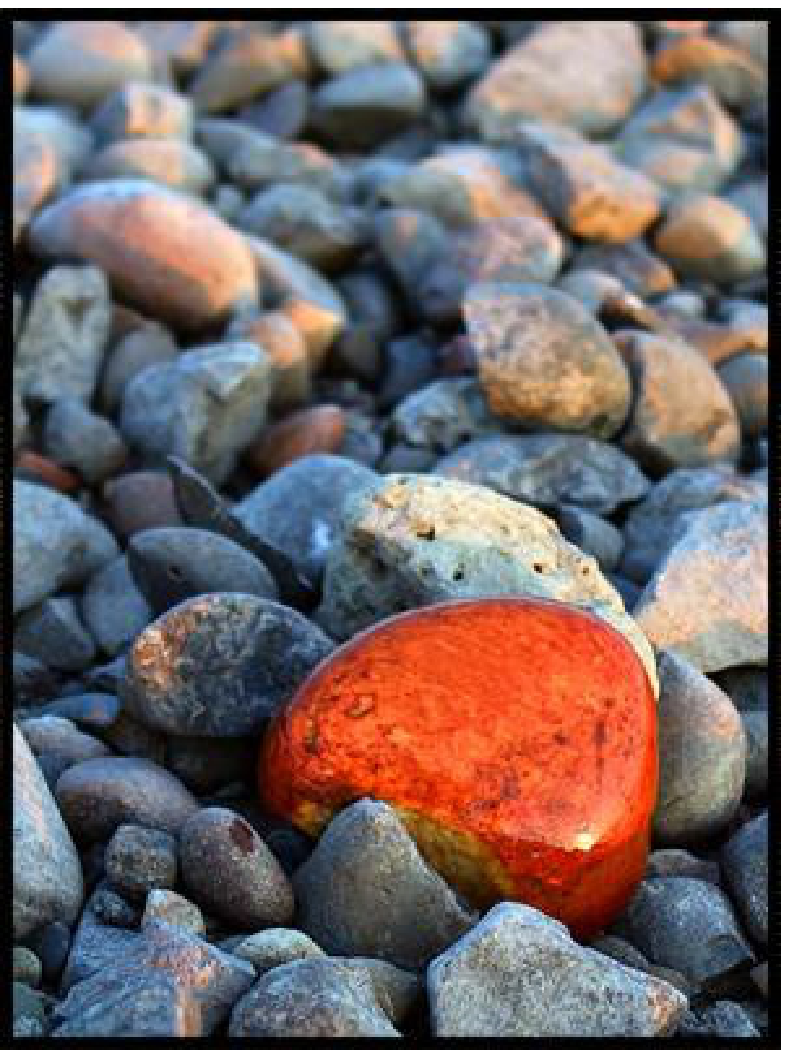} &
\includegraphics[height=1.6cm]{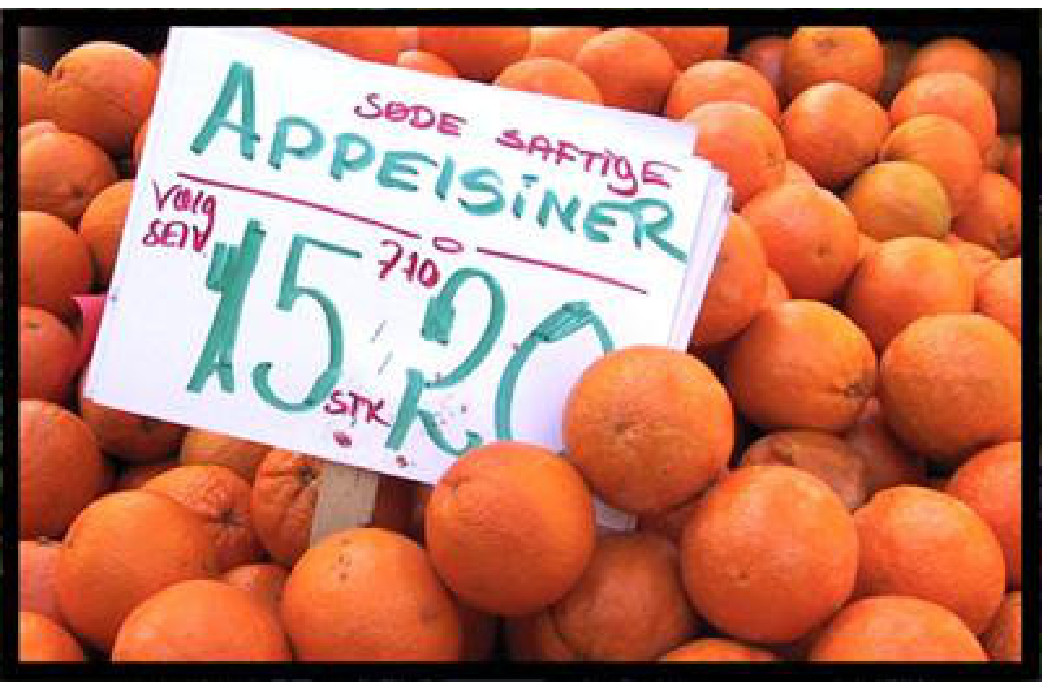} &
\includegraphics[height=1.6cm]{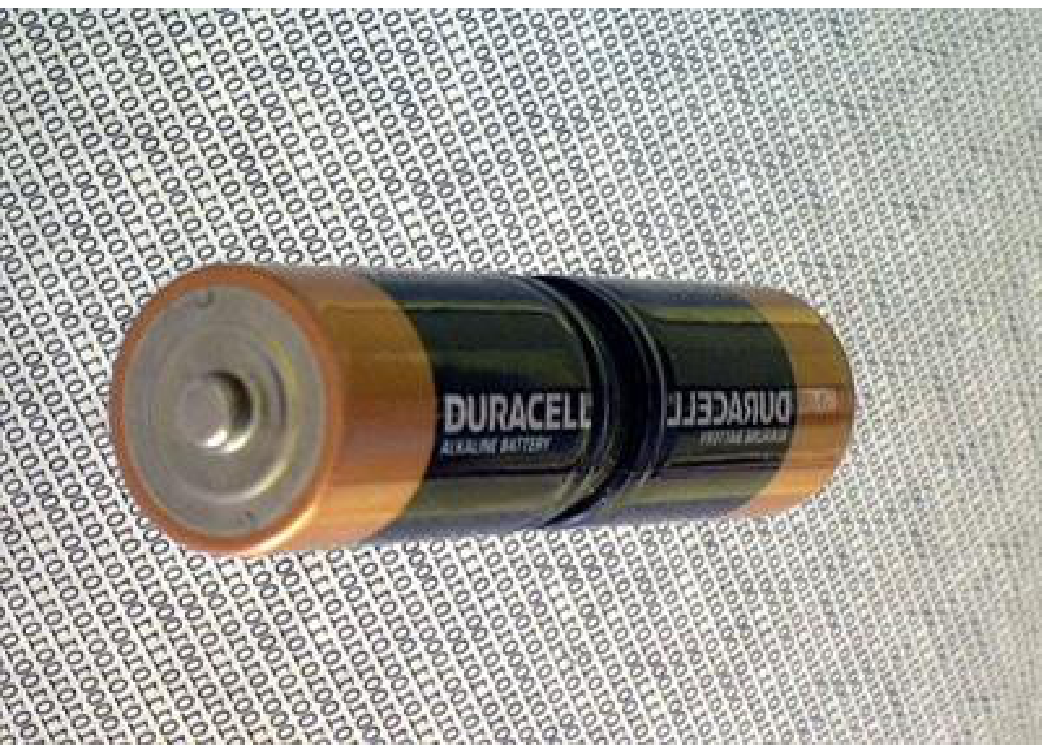} &
\includegraphics[height=1.6cm]{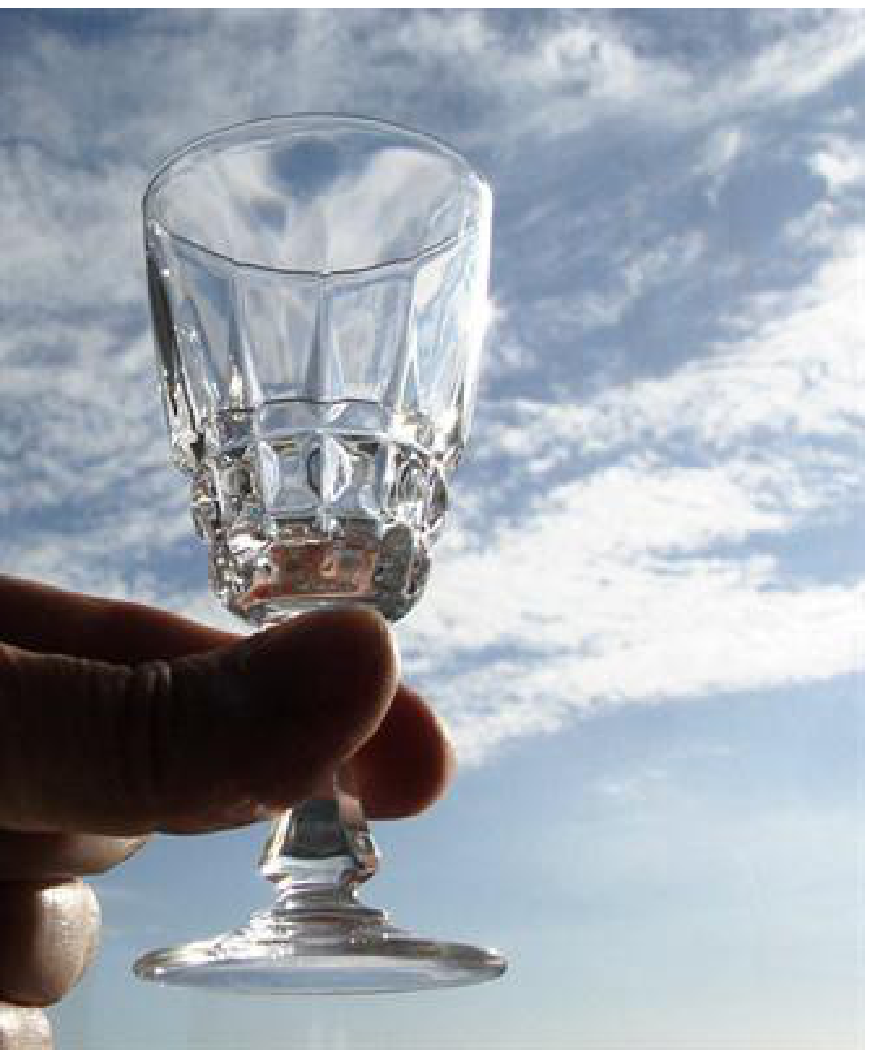} \\
FT&
\includegraphics[height=1.6cm]{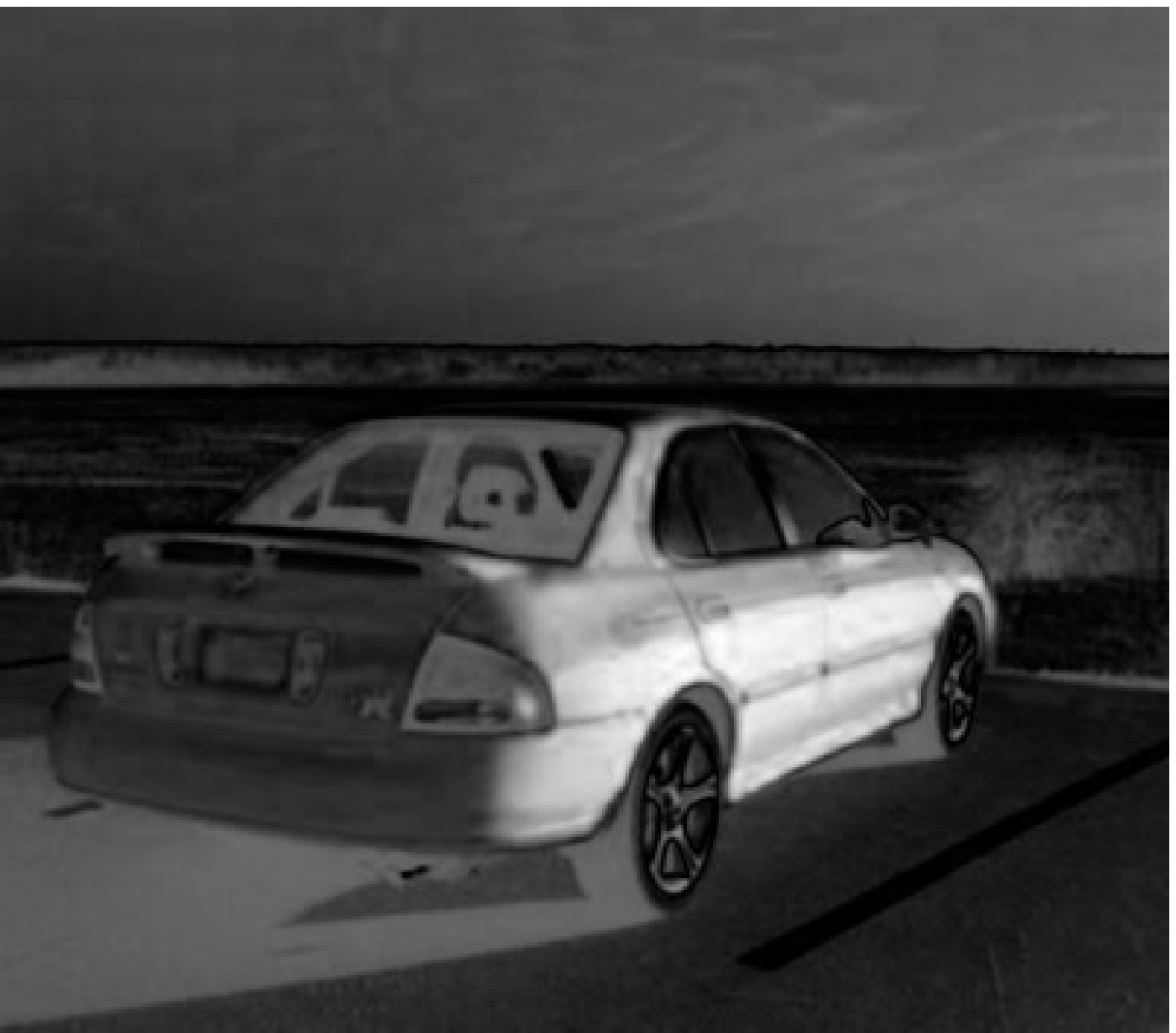} &
\includegraphics[height=1.6cm]{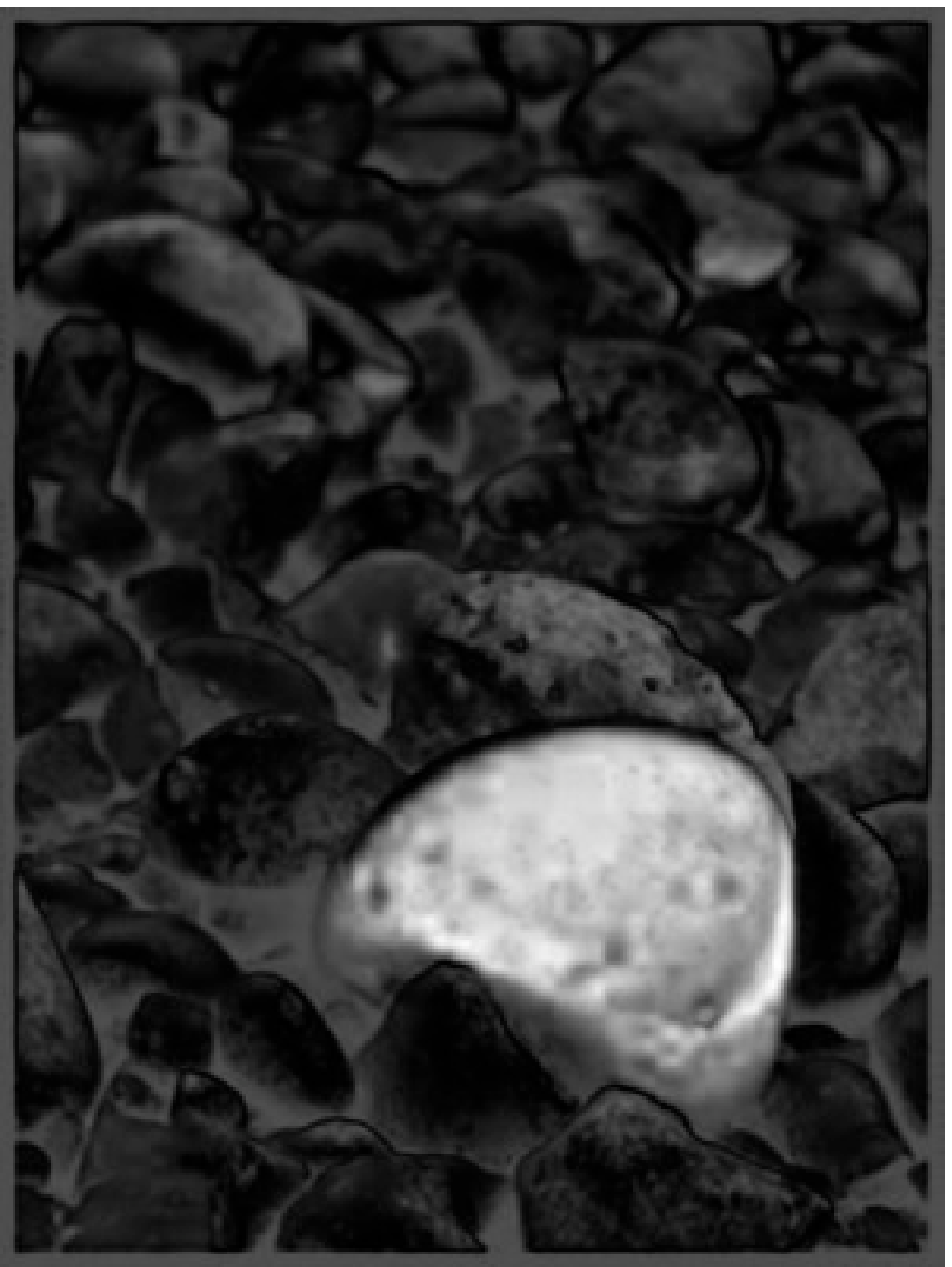} &
\includegraphics[height=1.6cm]{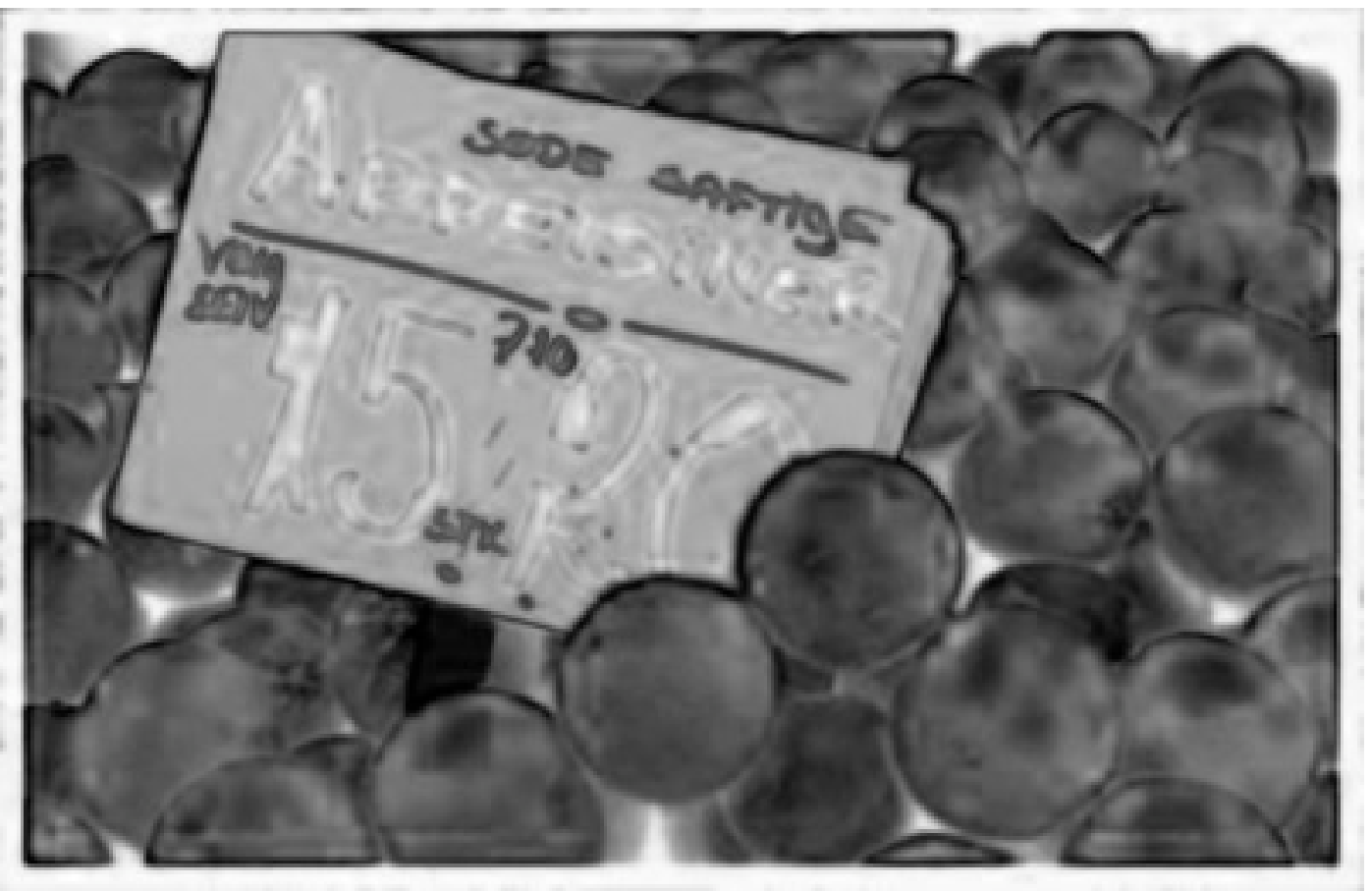} &
\includegraphics[height=1.6cm]{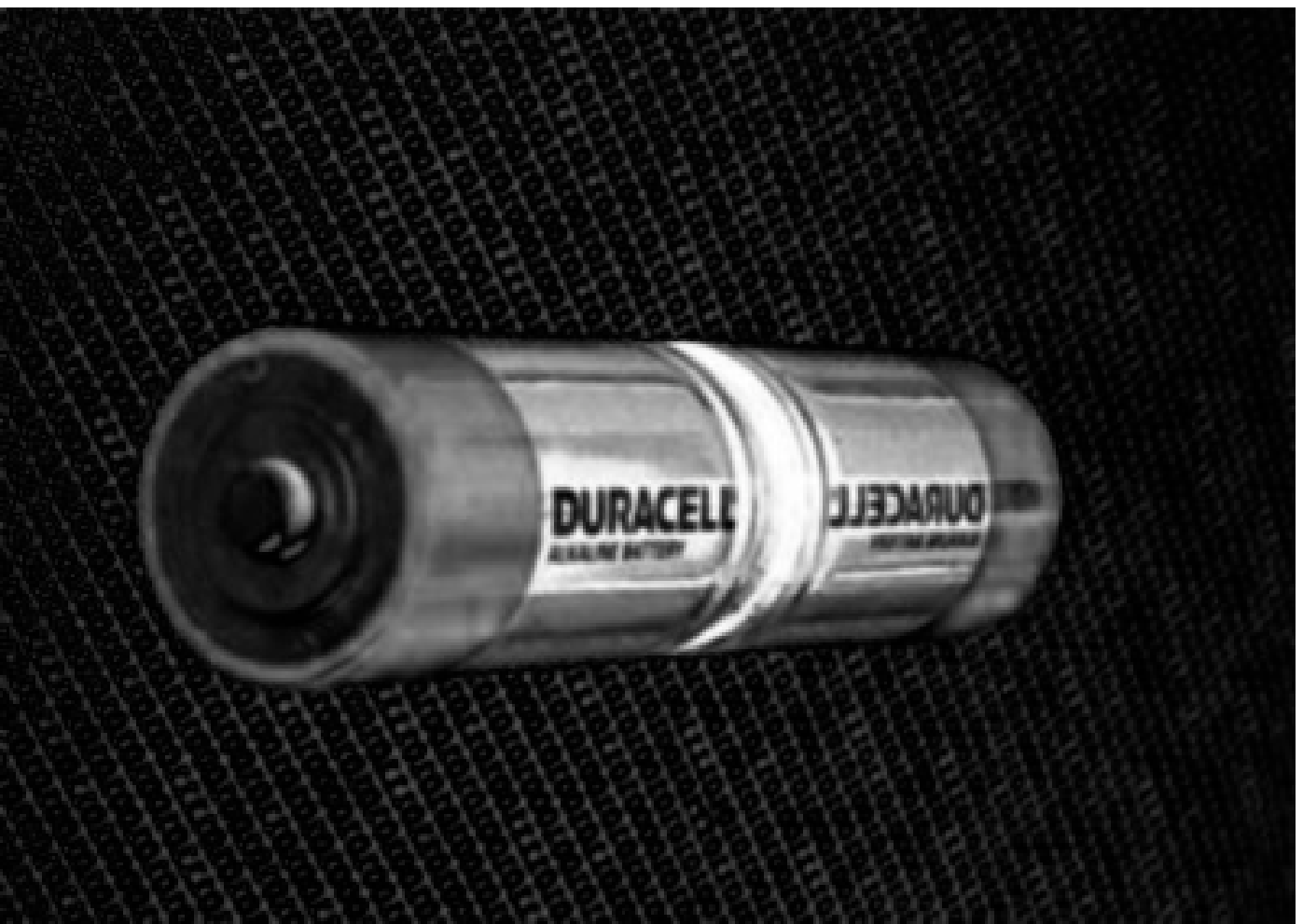} &
\includegraphics[height=1.6cm]{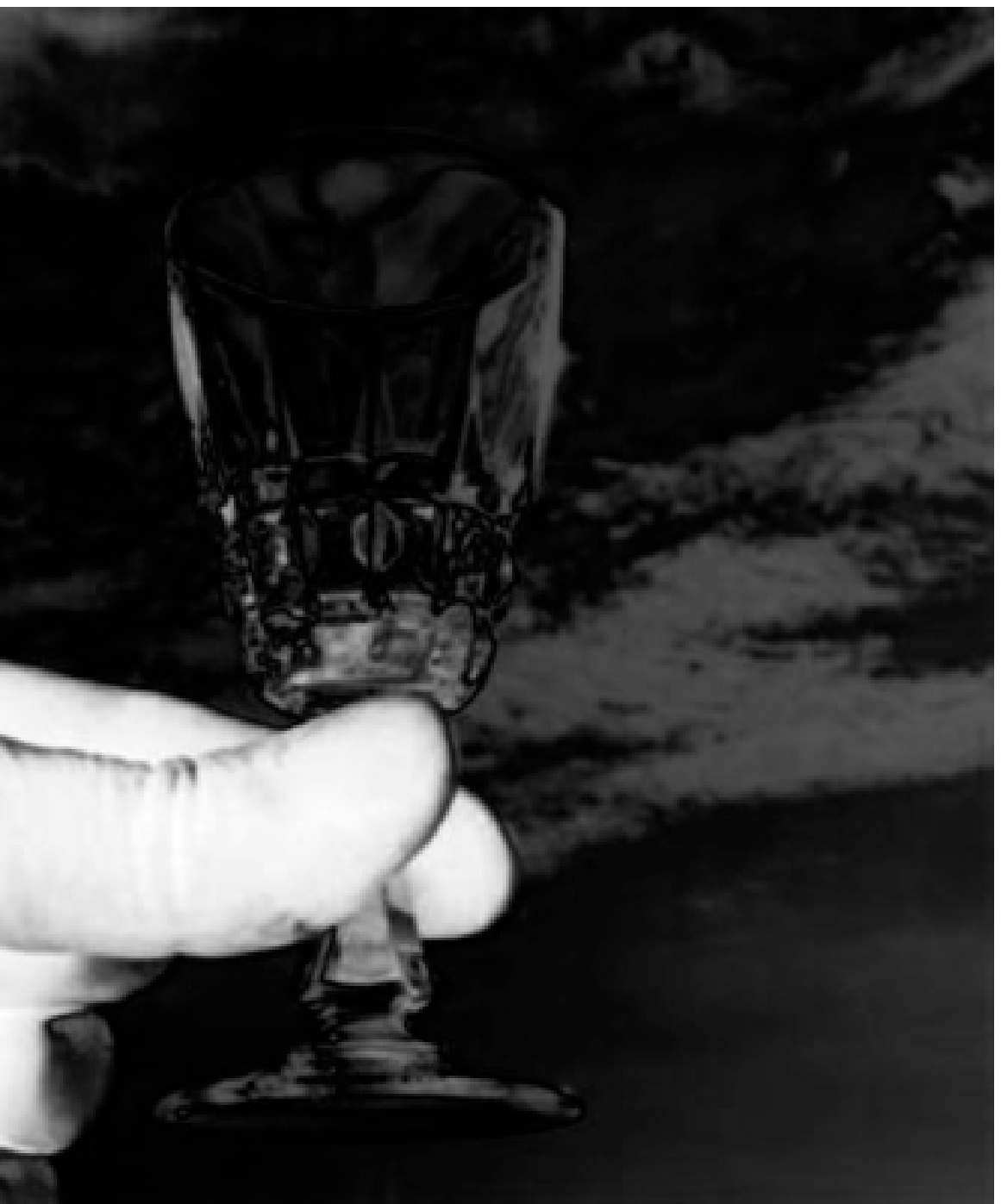} \\
RC&
\includegraphics[height=1.6cm]{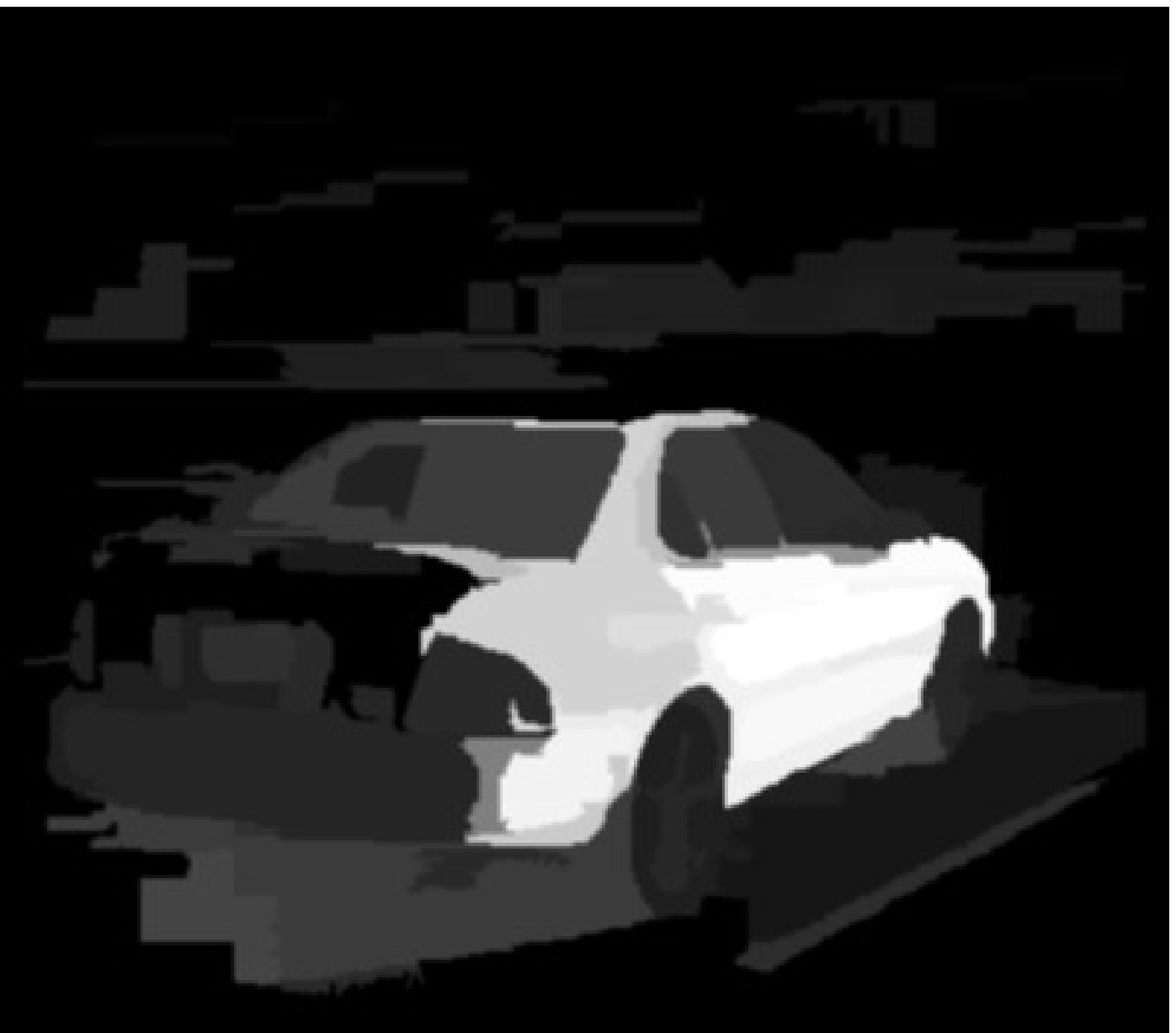} &
\includegraphics[height=1.6cm]{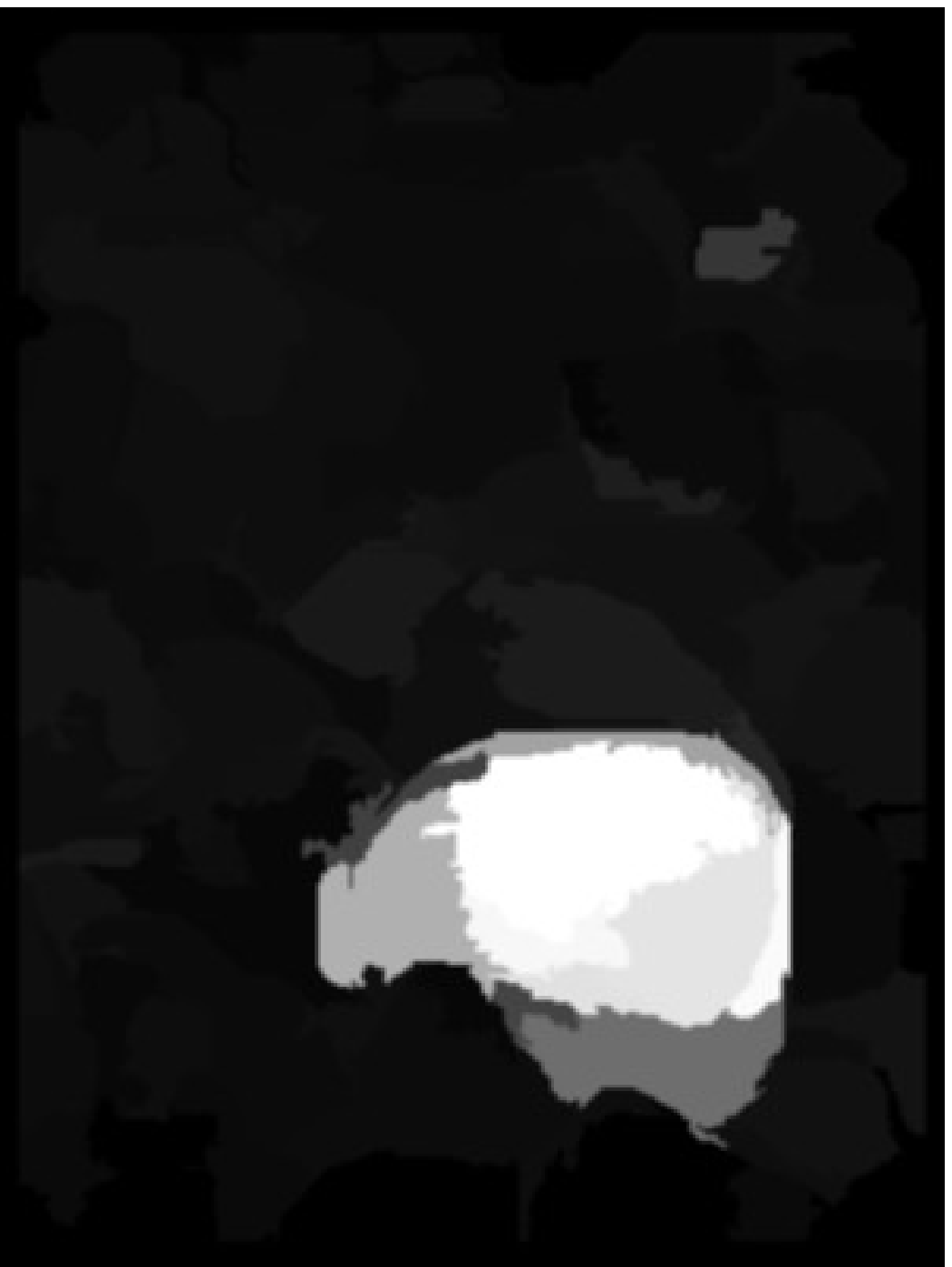} &
\includegraphics[height=1.6cm]{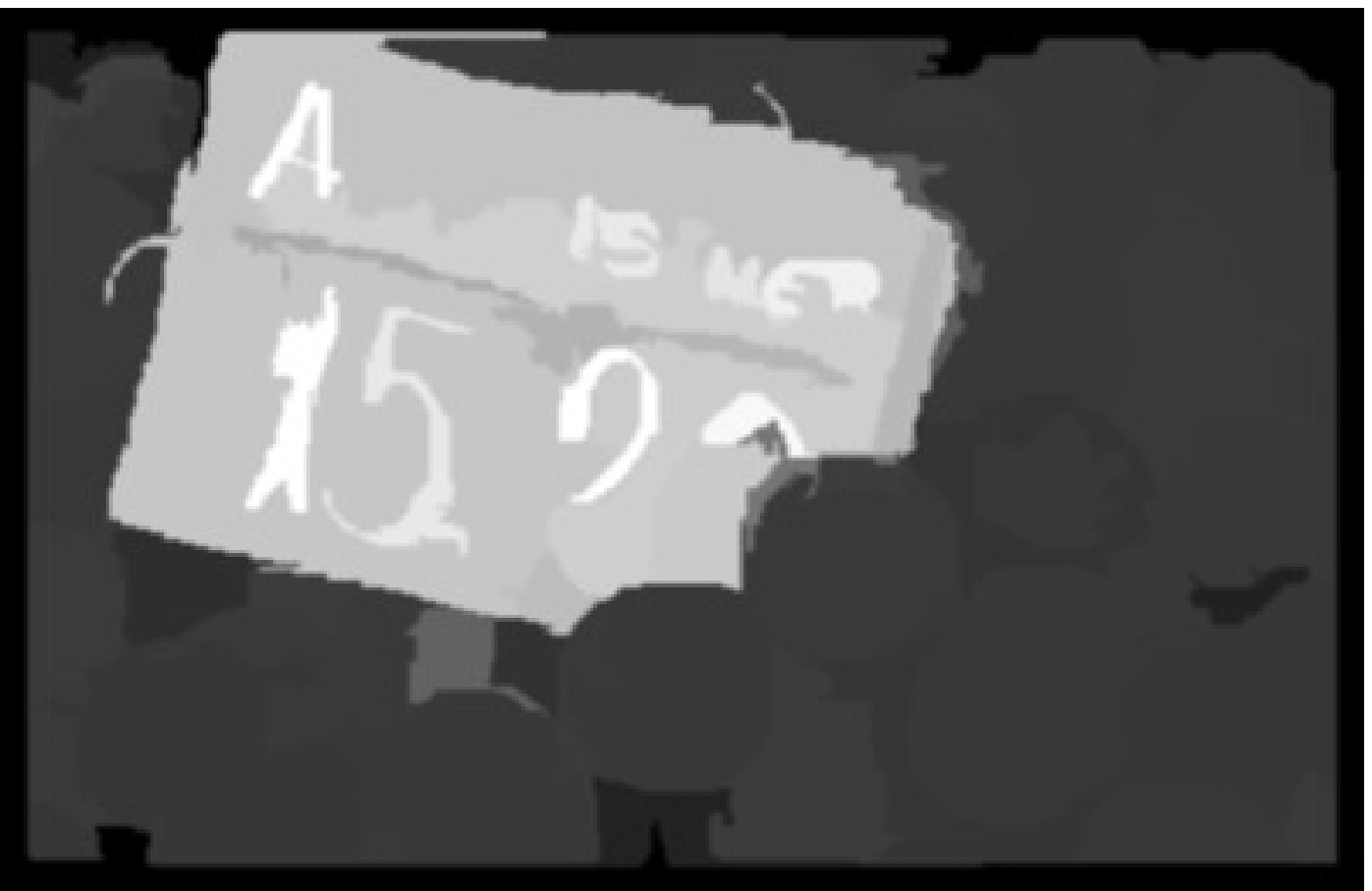} &
\includegraphics[height=1.6cm]{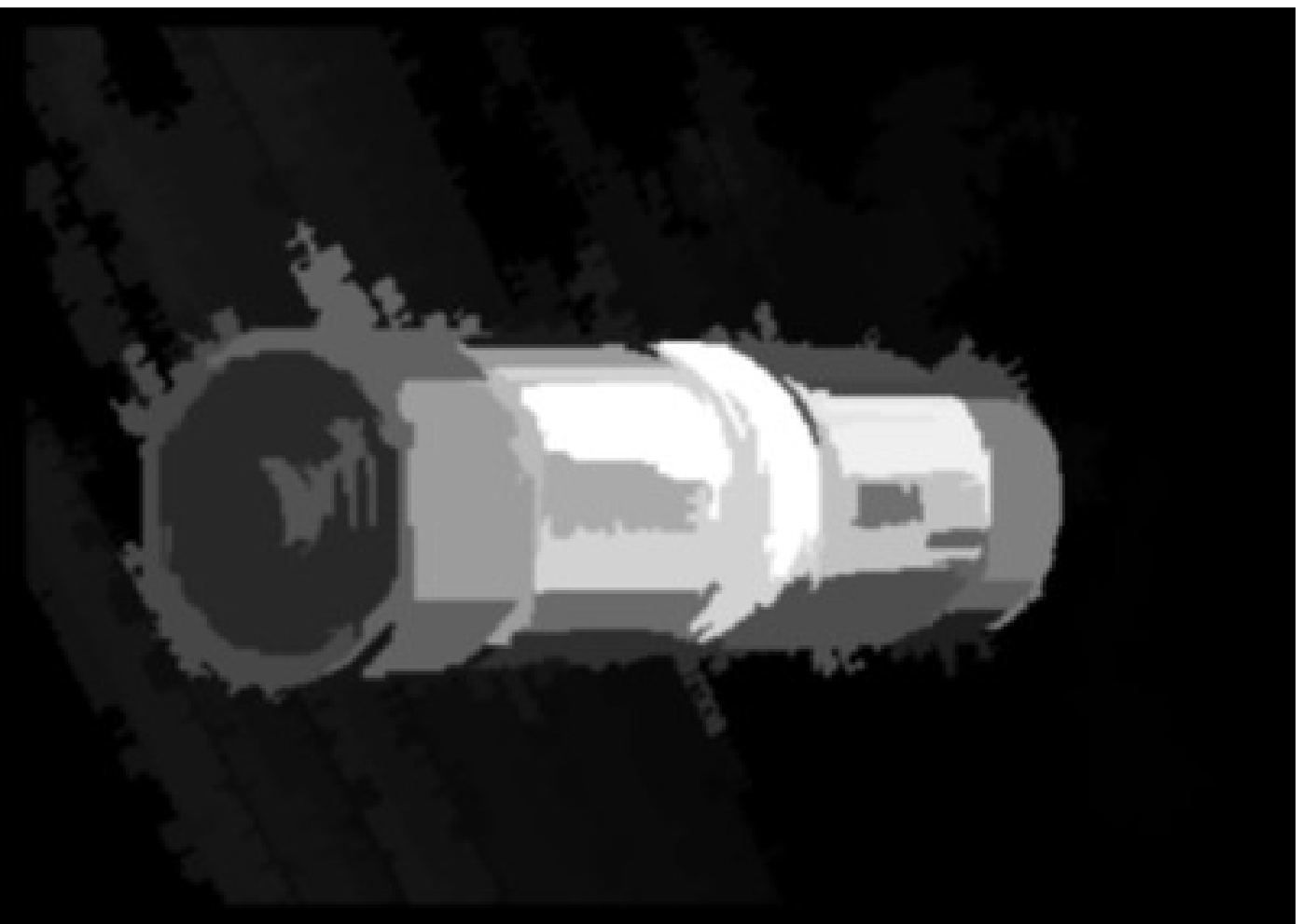} &
\includegraphics[height=1.6cm]{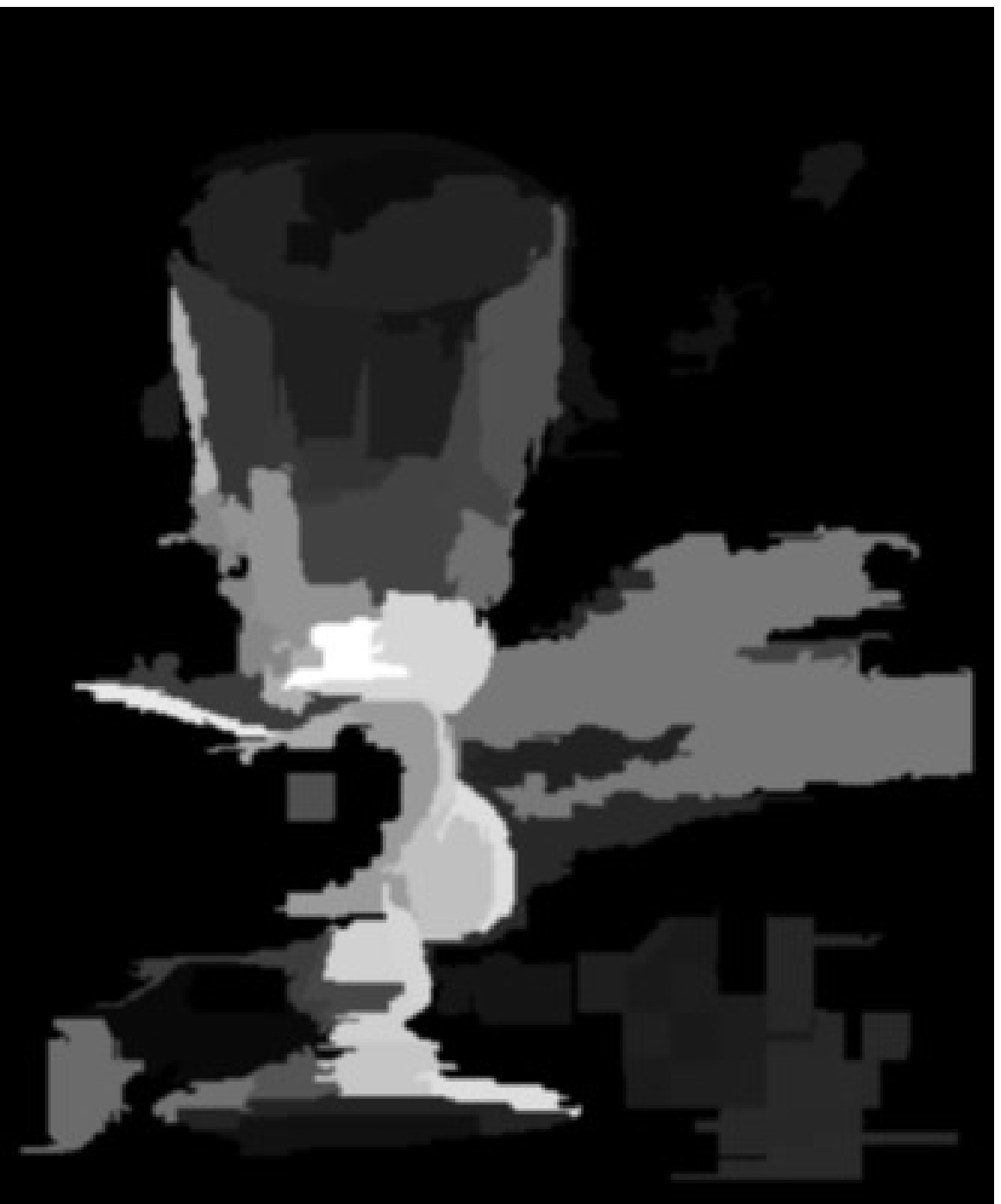} \\
SF&
\includegraphics[height=1.6cm]{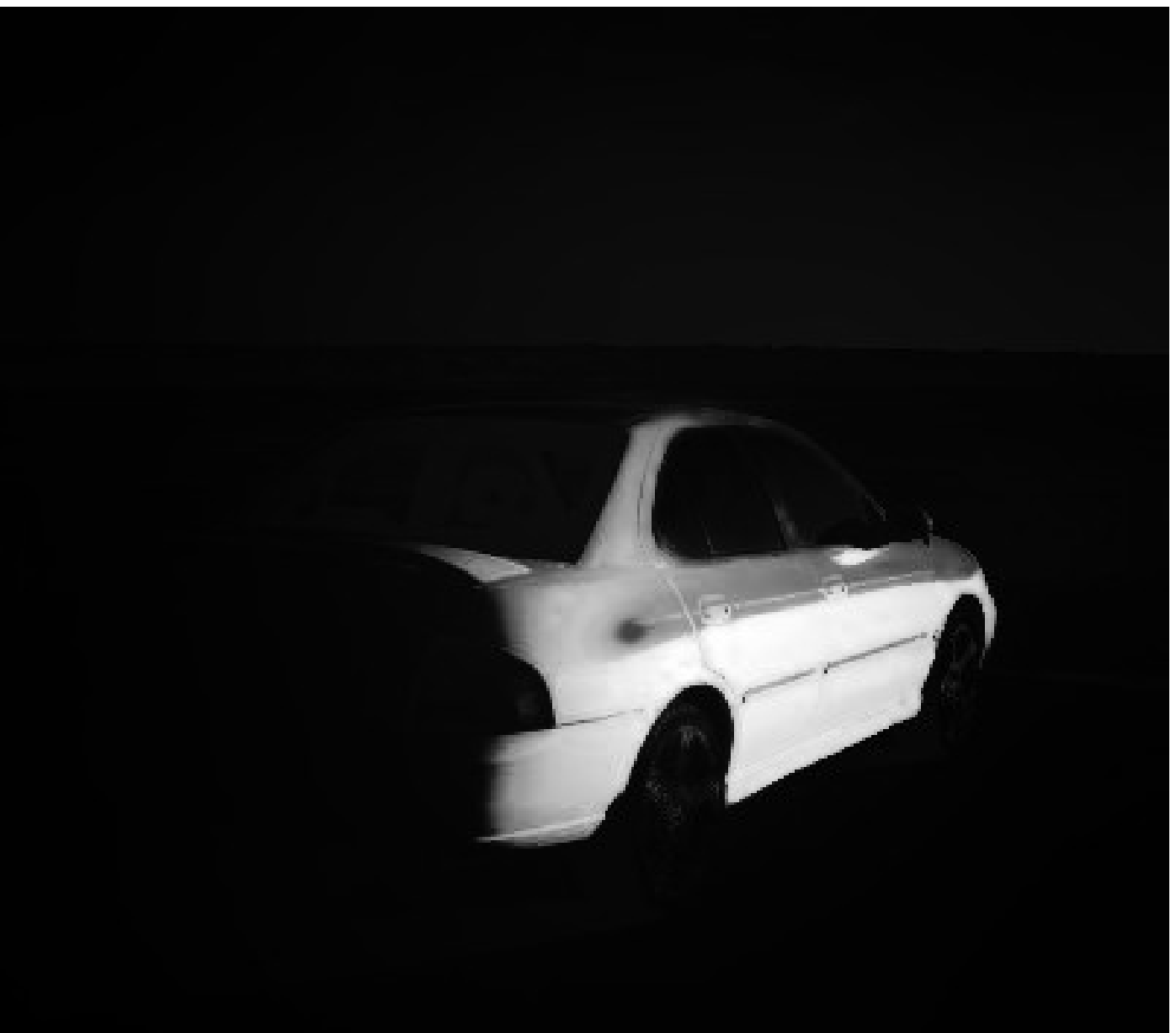} &
\includegraphics[height=1.6cm]{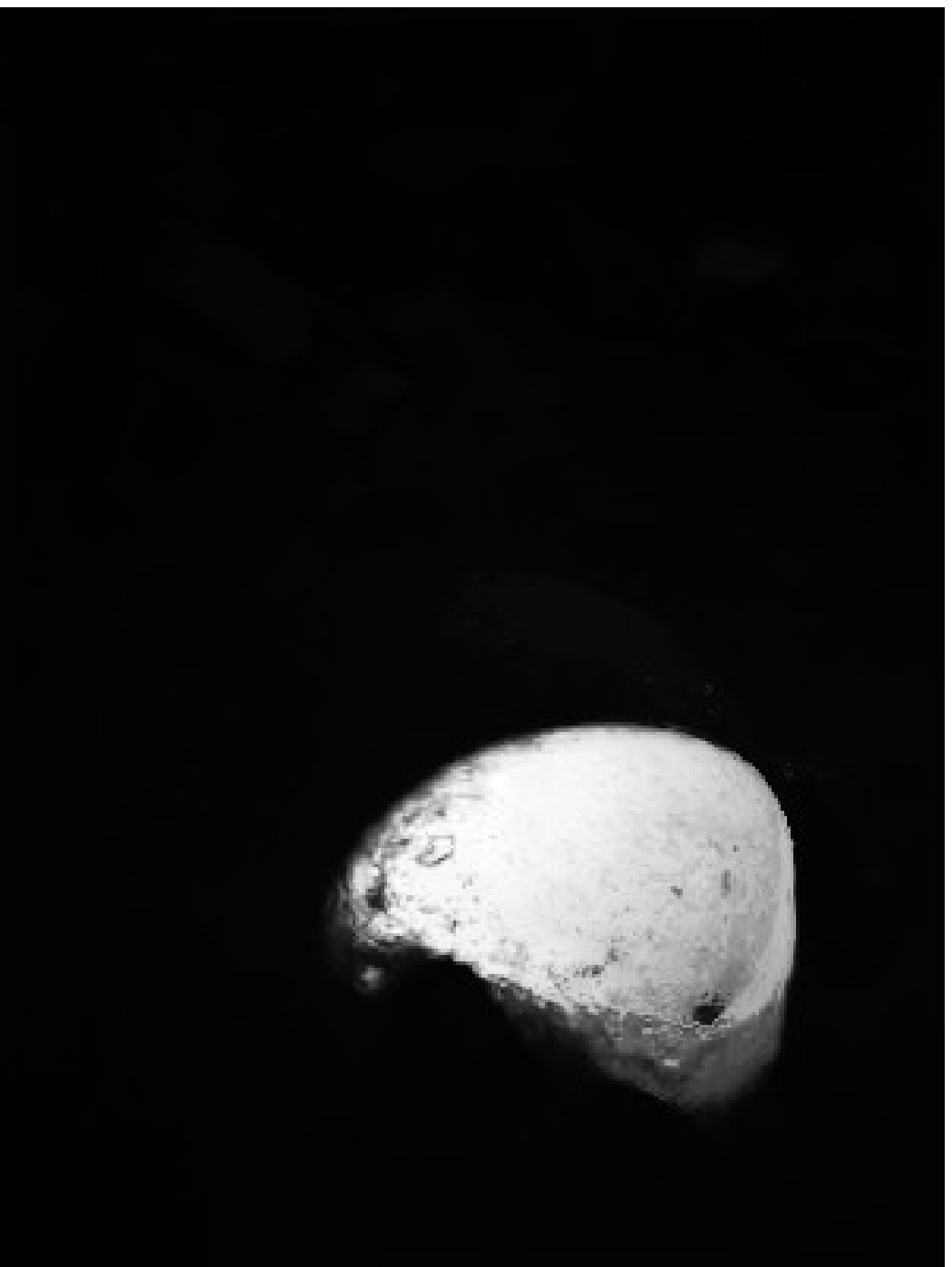} &
\includegraphics[height=1.6cm]{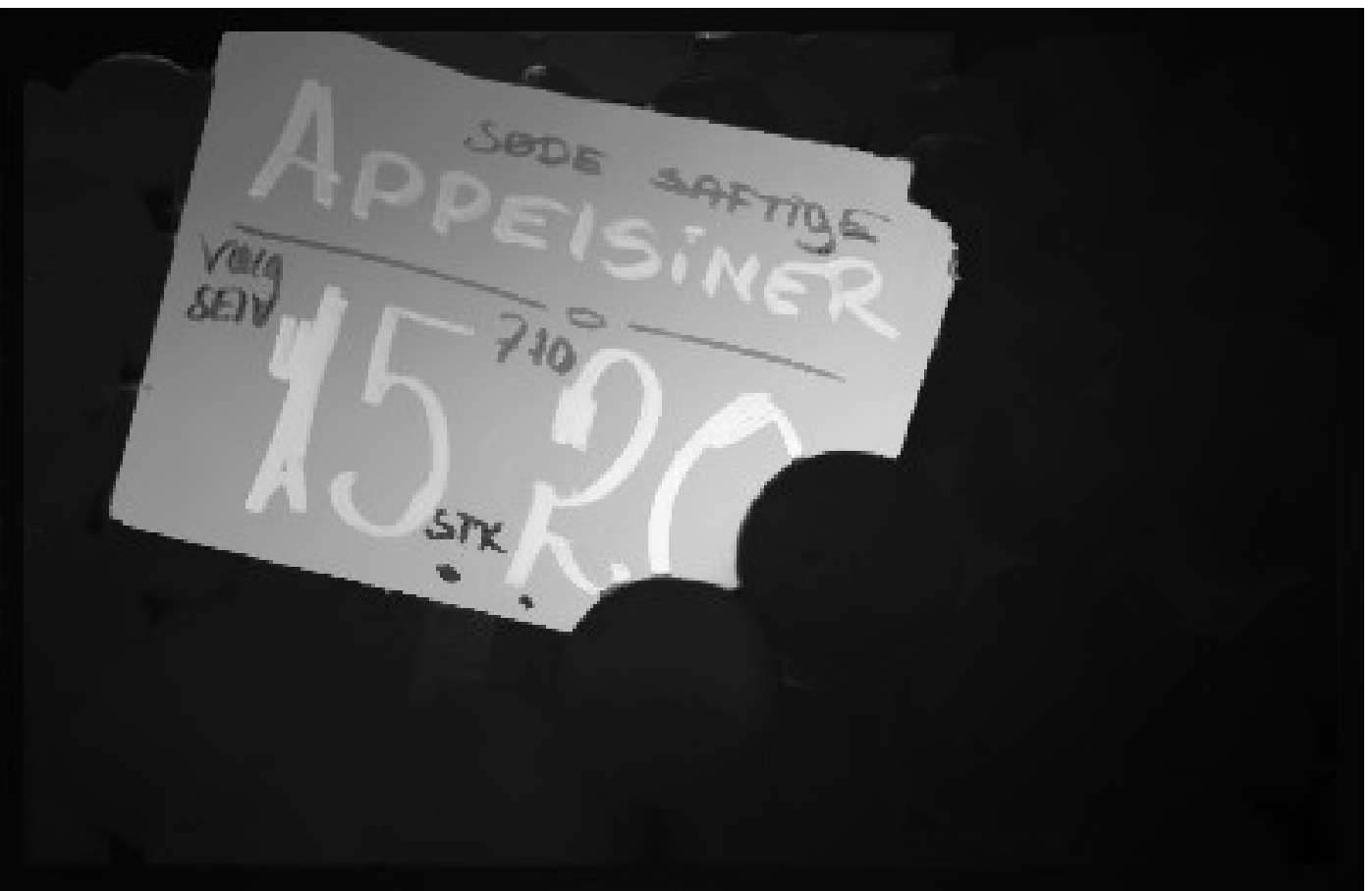} &
\includegraphics[height=1.6cm]{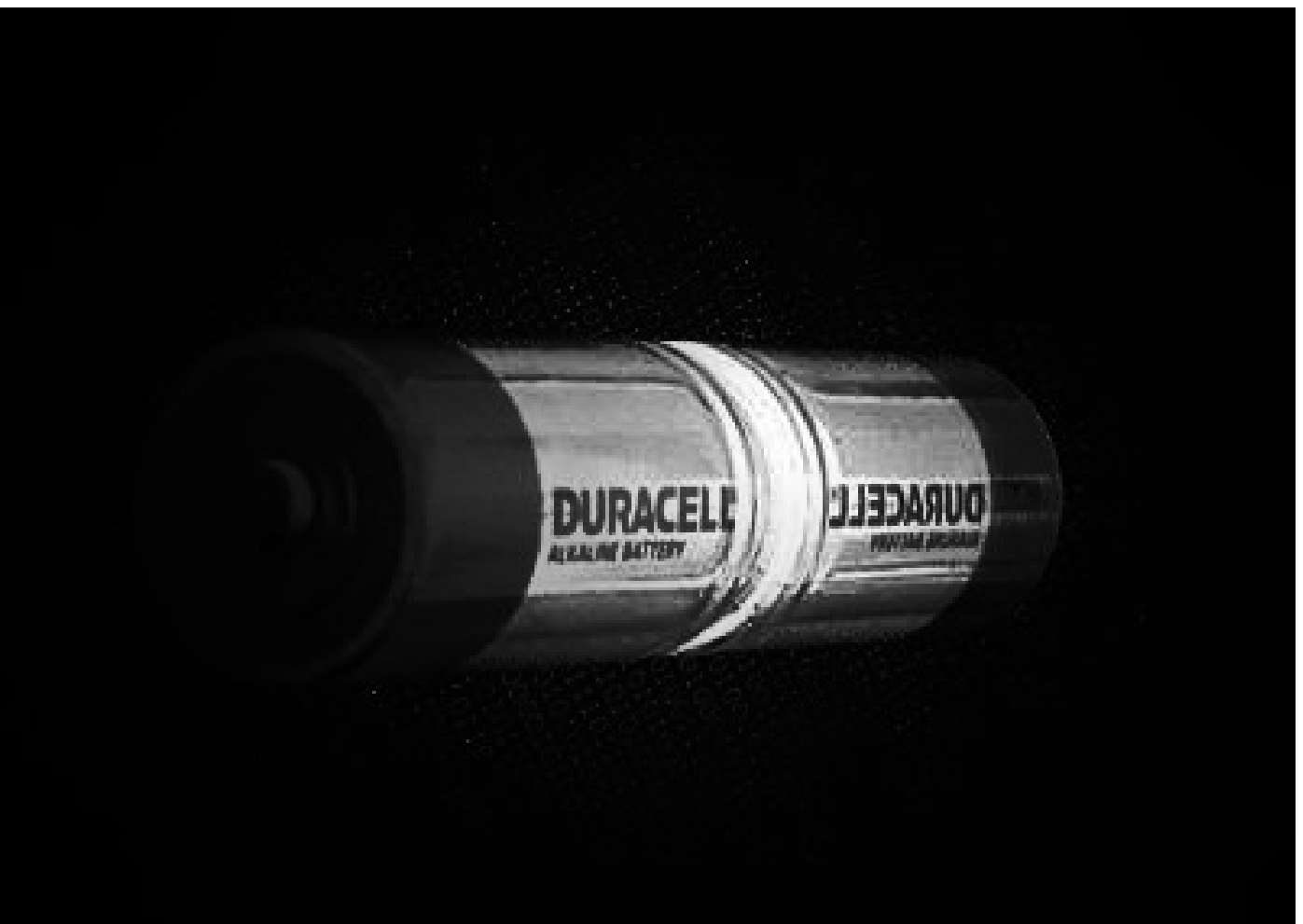} &
\includegraphics[height=1.6cm]{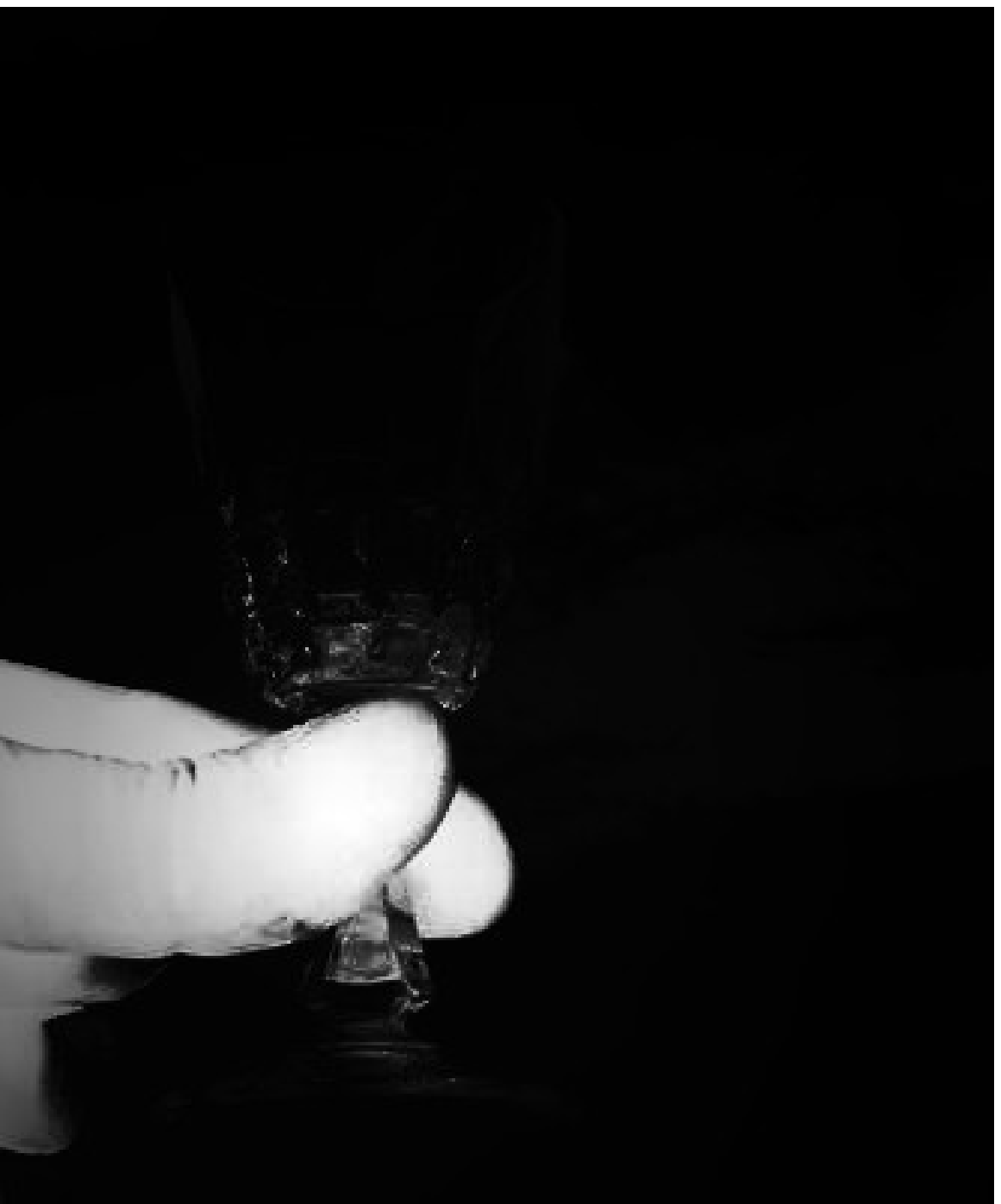} \\
PCAS&
\includegraphics[height=1.6cm]{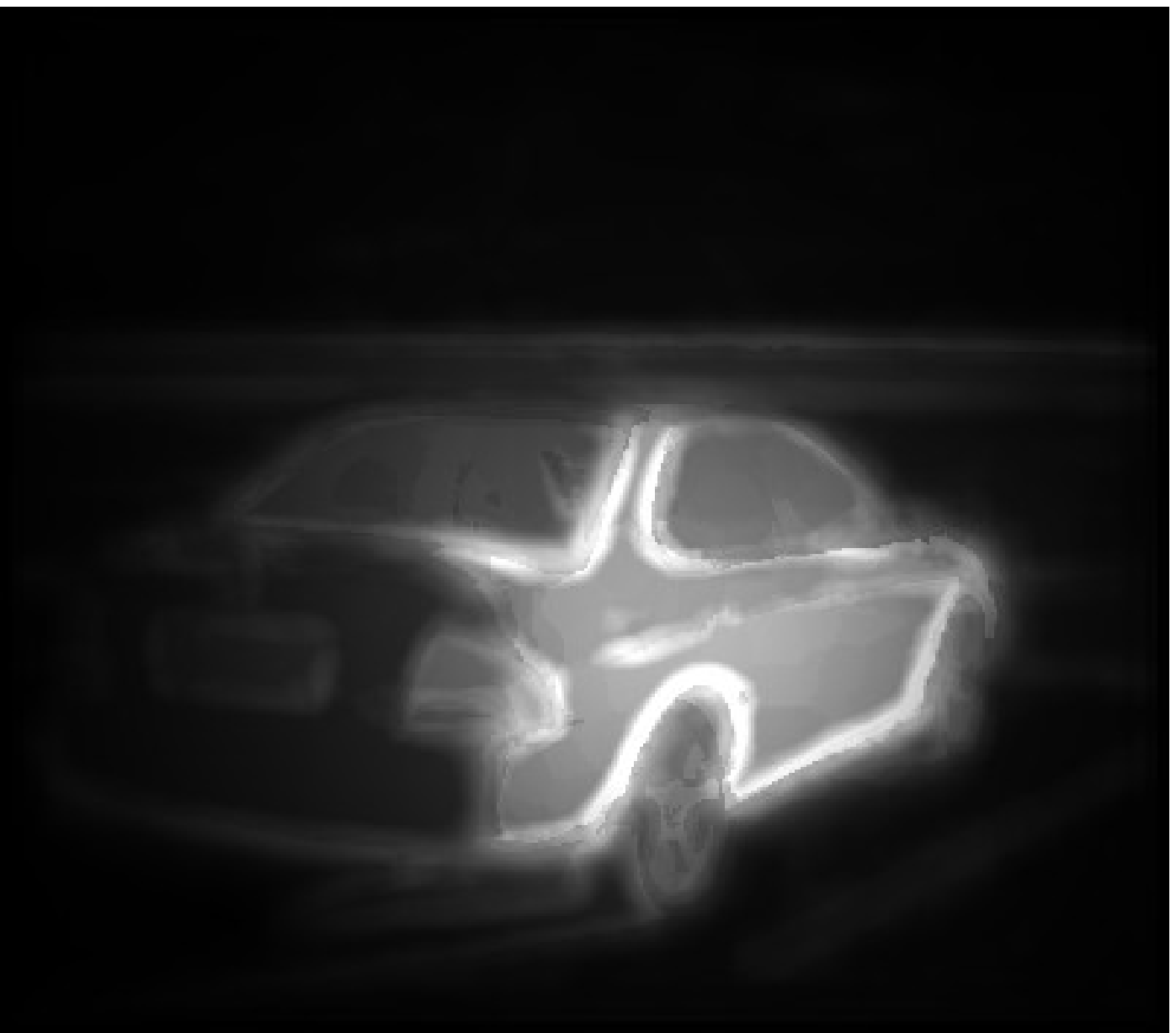} &
\includegraphics[height=1.6cm]{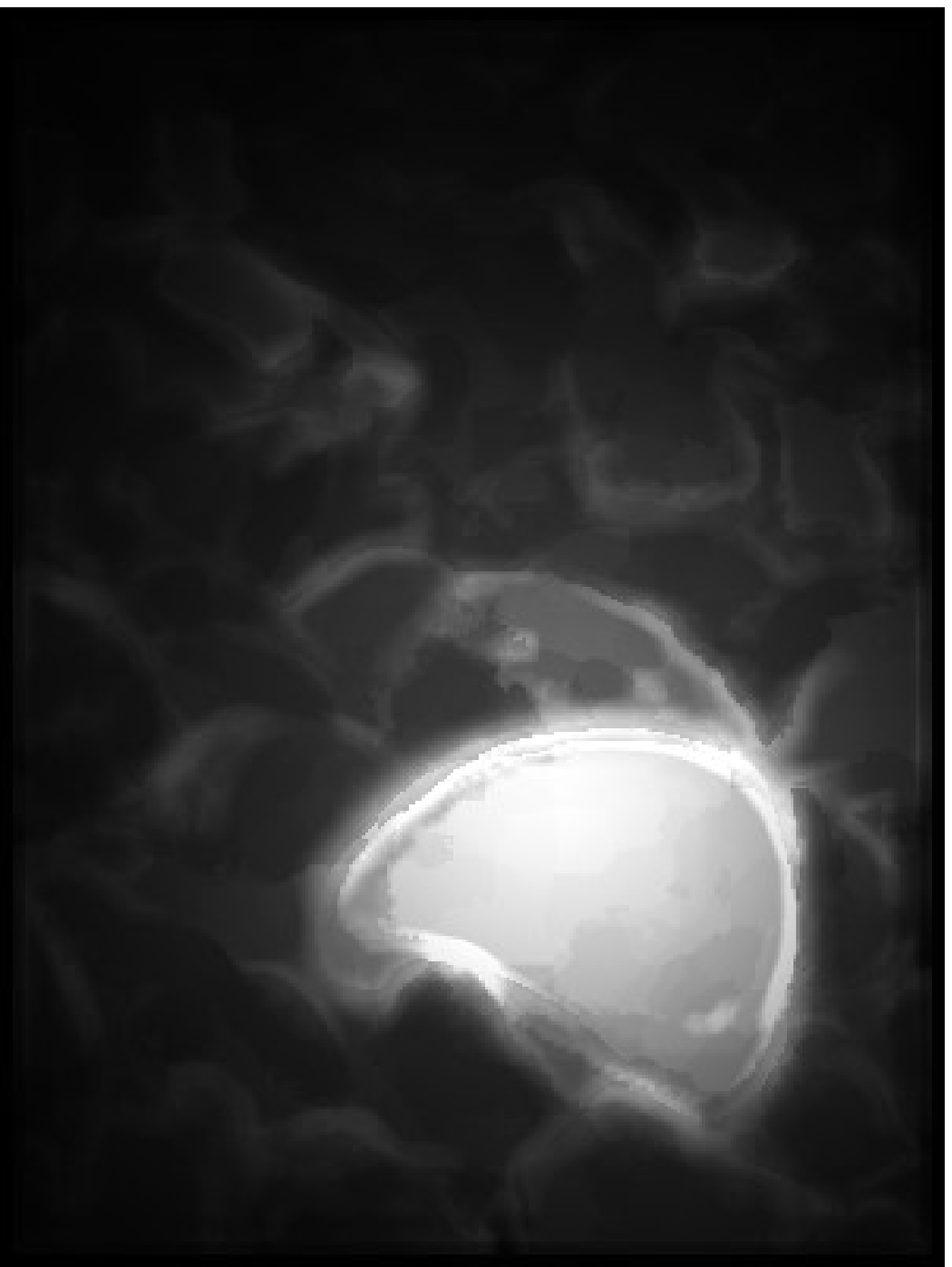} &
\includegraphics[height=1.6cm]{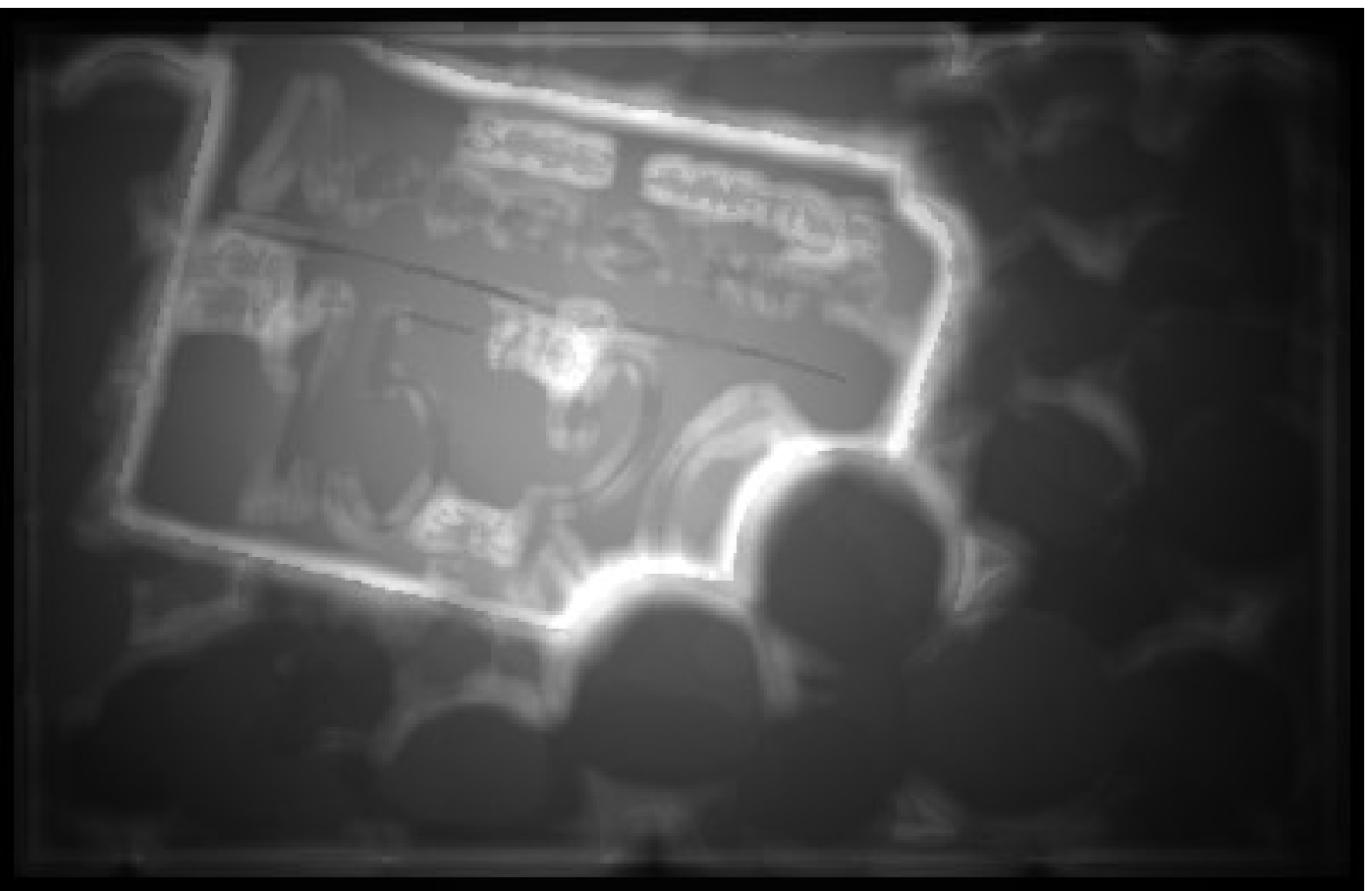} &
\includegraphics[height=1.6cm]{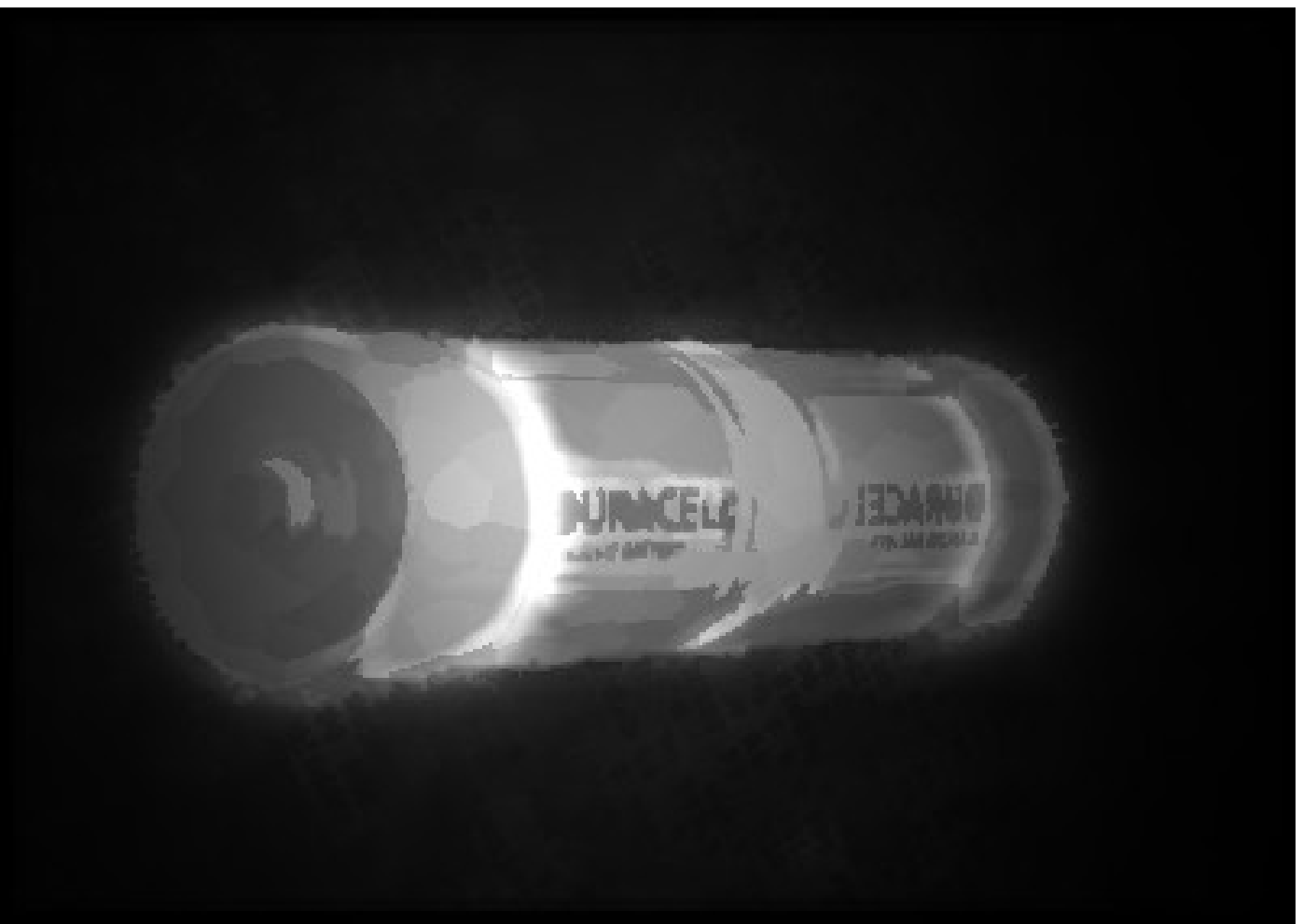} &
\includegraphics[height=1.6cm]{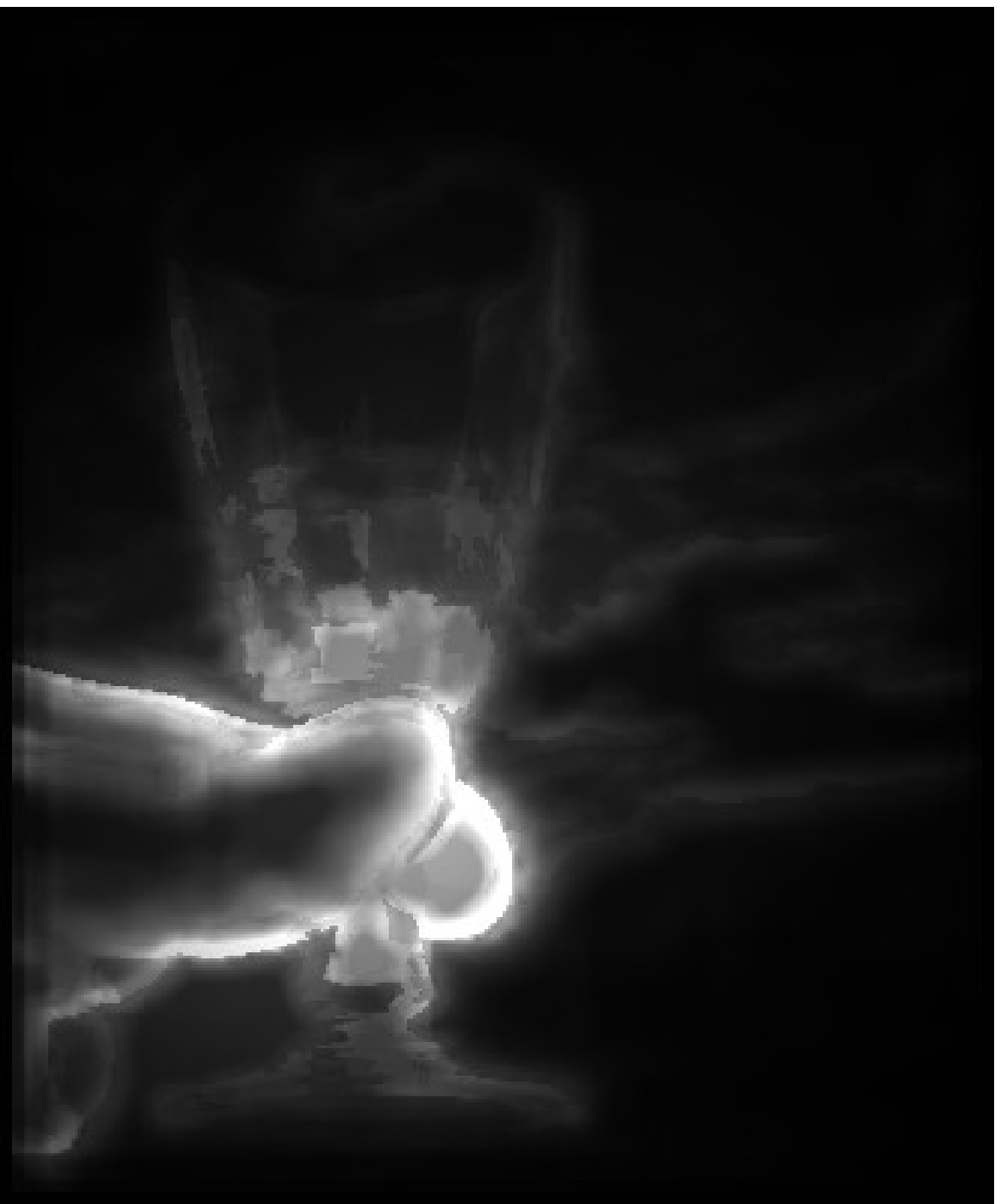} \\
HS&
\includegraphics[height=1.6cm]{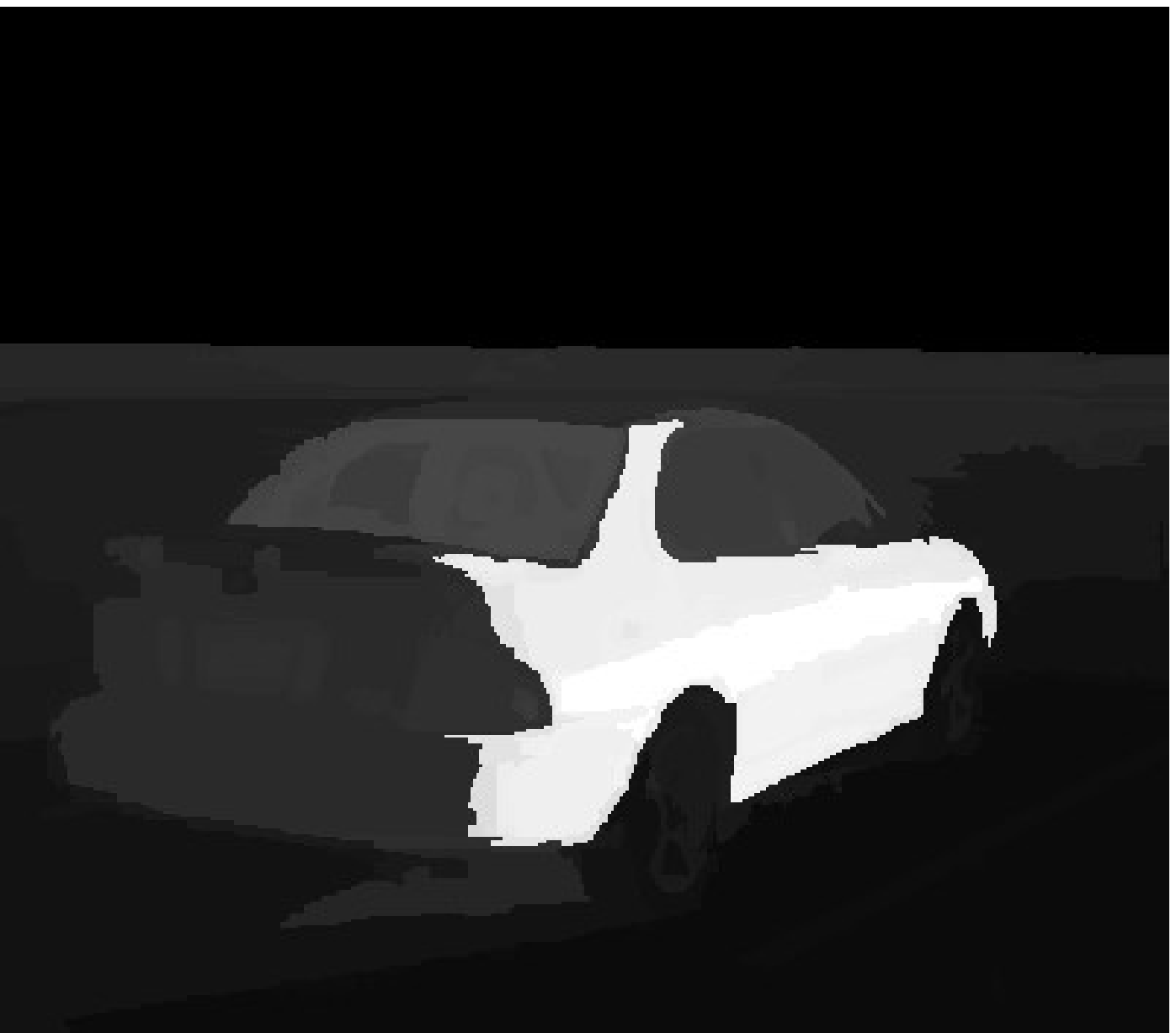} &
\includegraphics[height=1.6cm]{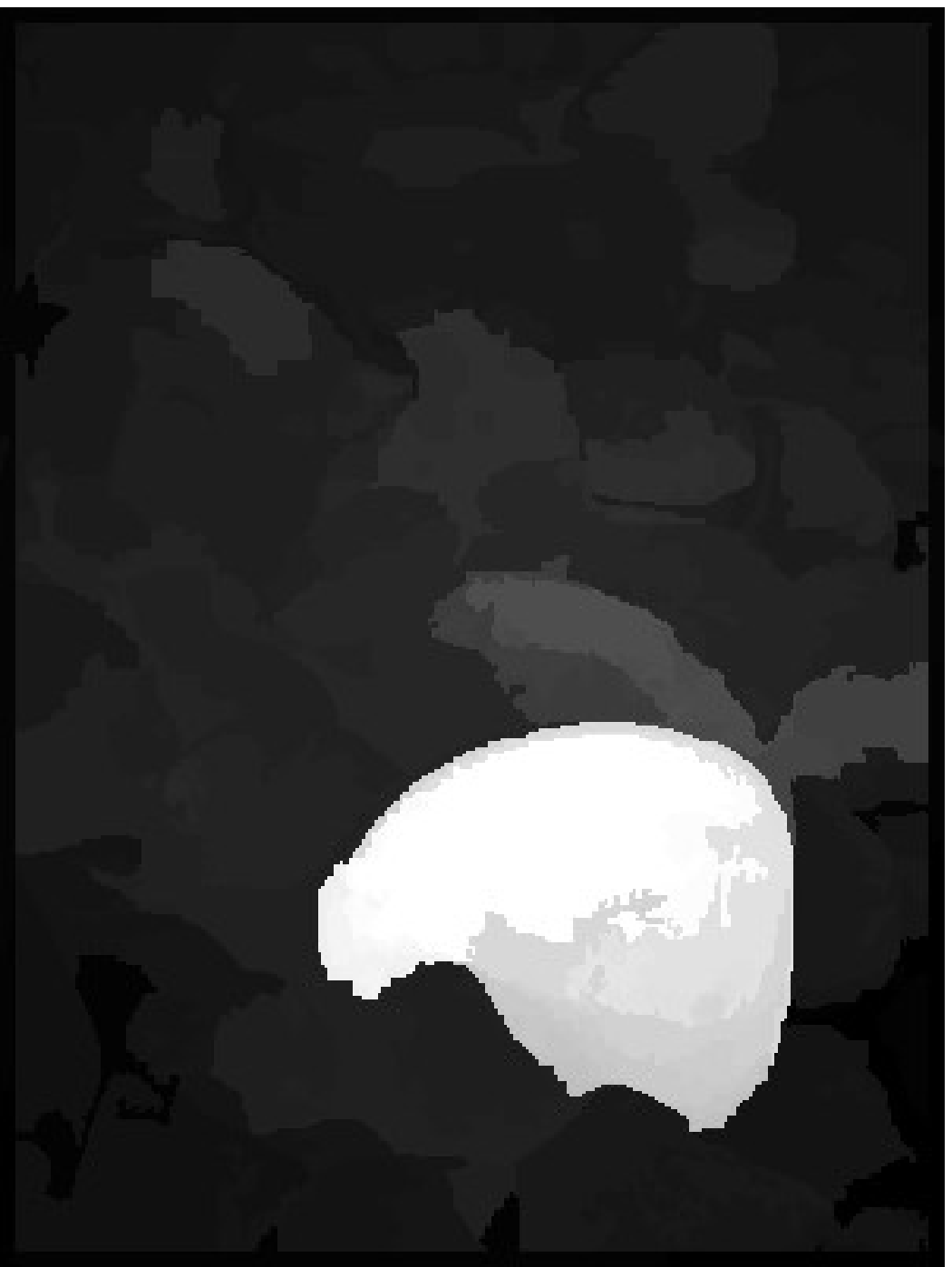} &
\includegraphics[height=1.6cm]{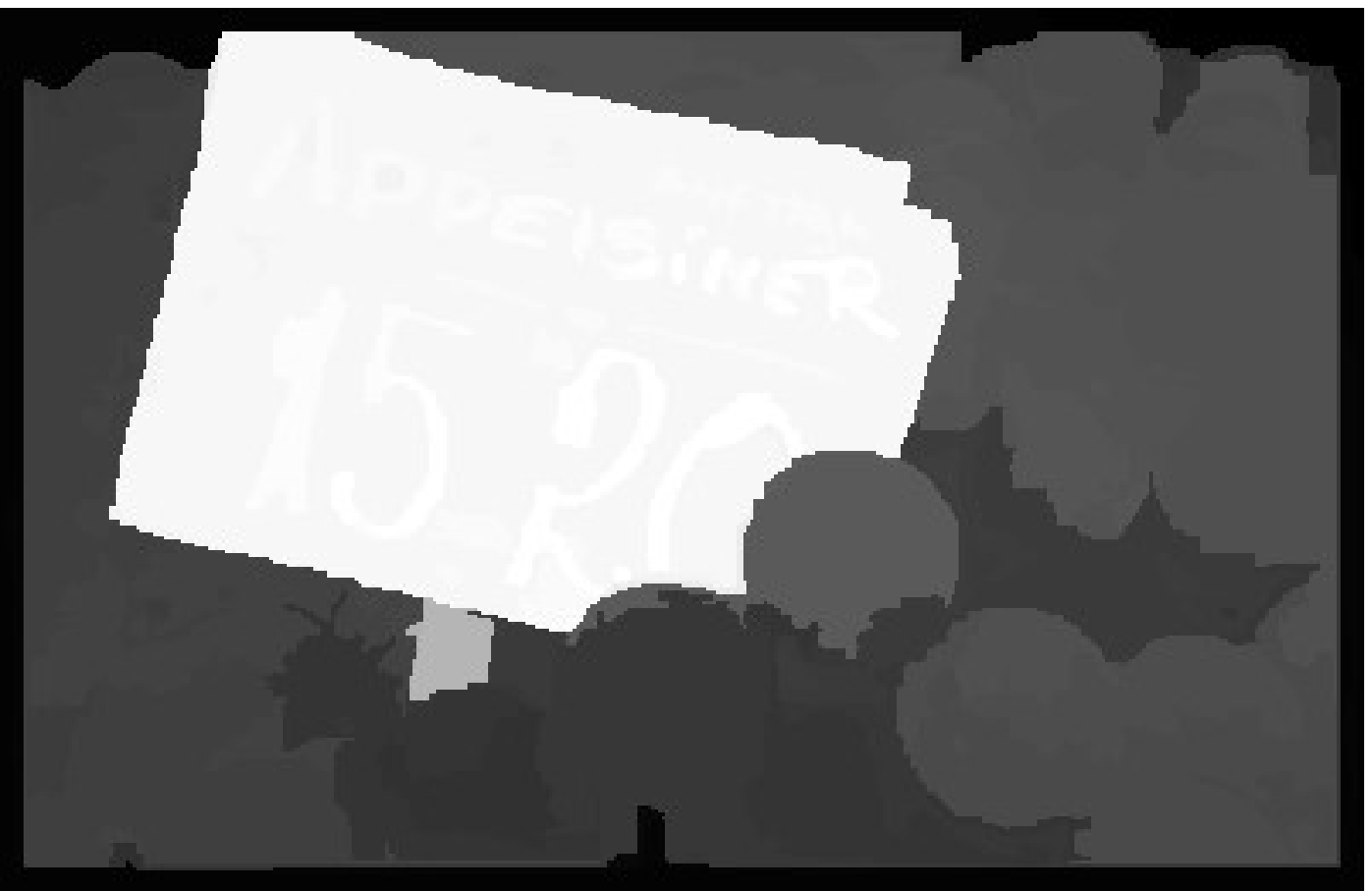} &
\includegraphics[height=1.6cm]{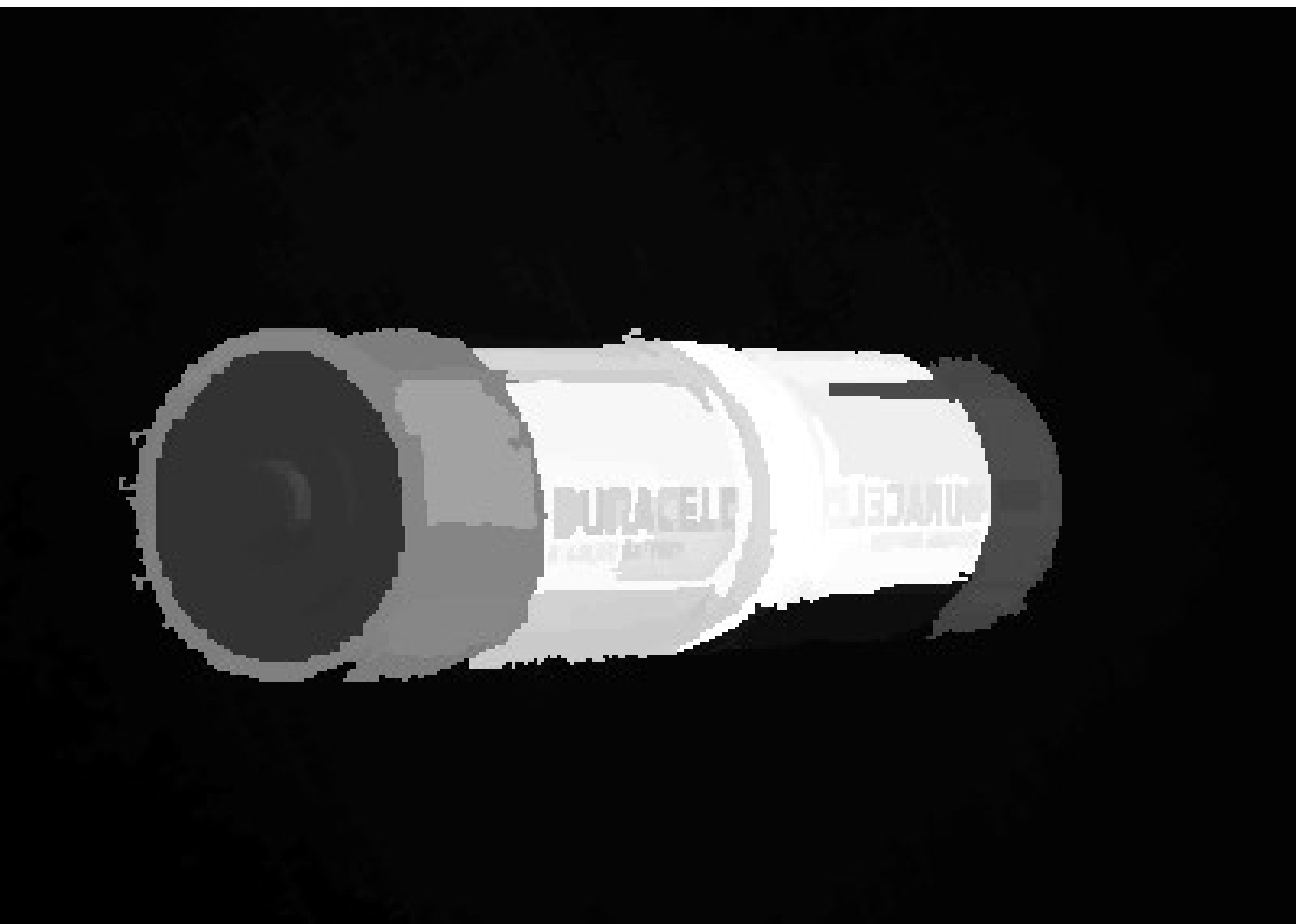} &
\includegraphics[height=1.6cm]{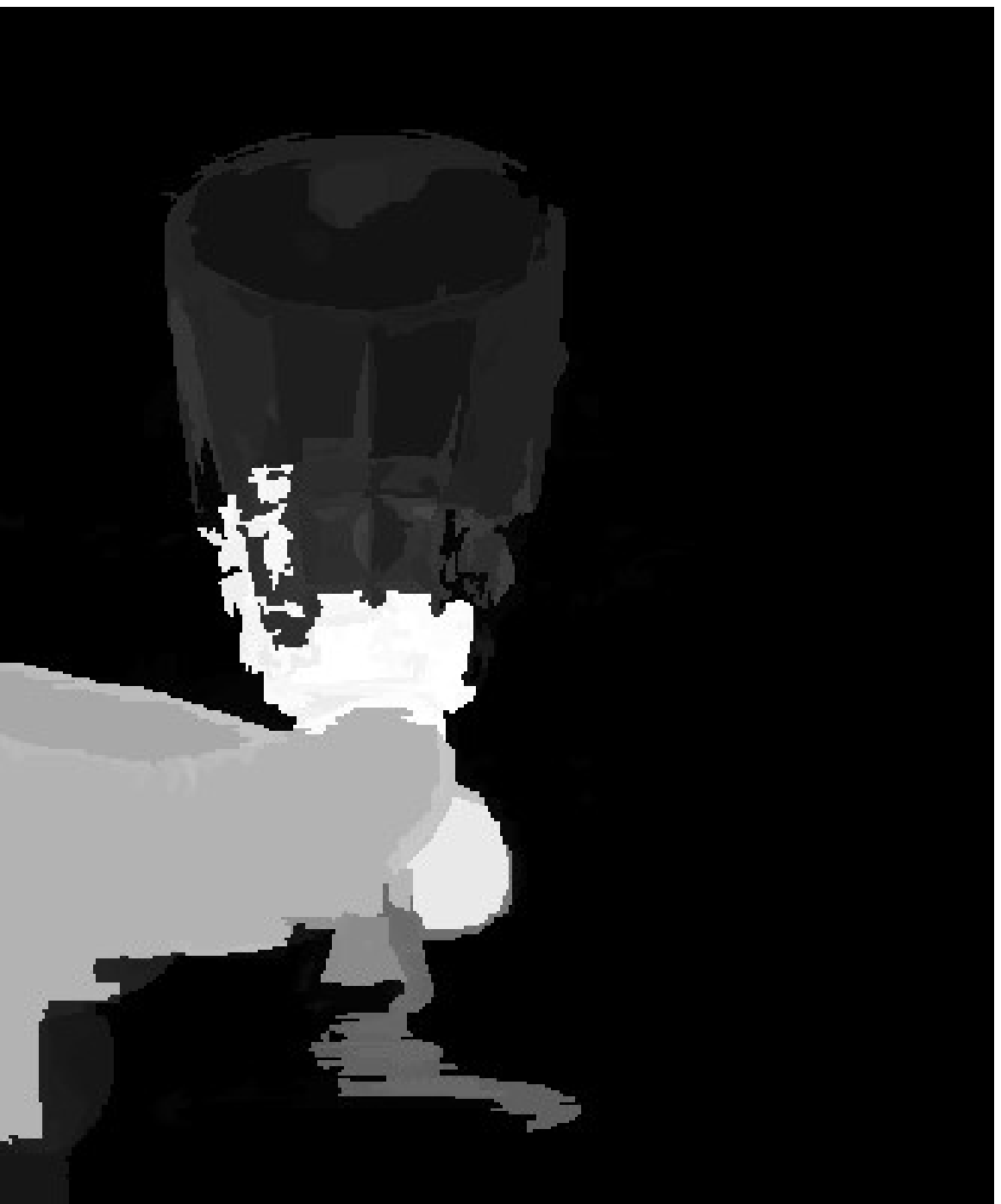} \\
ST&
\includegraphics[height=1.6cm]{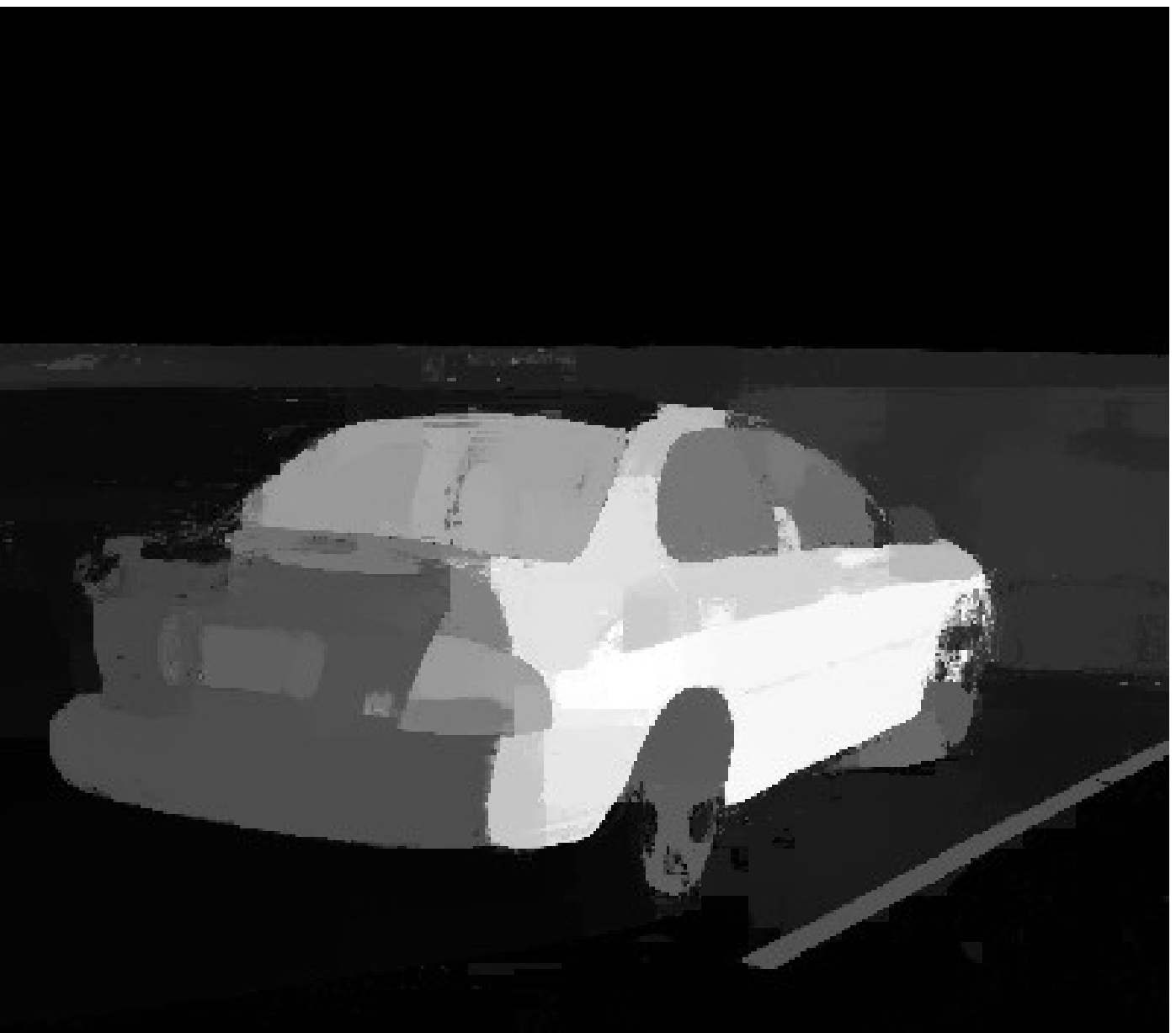} &
\includegraphics[height=1.6cm]{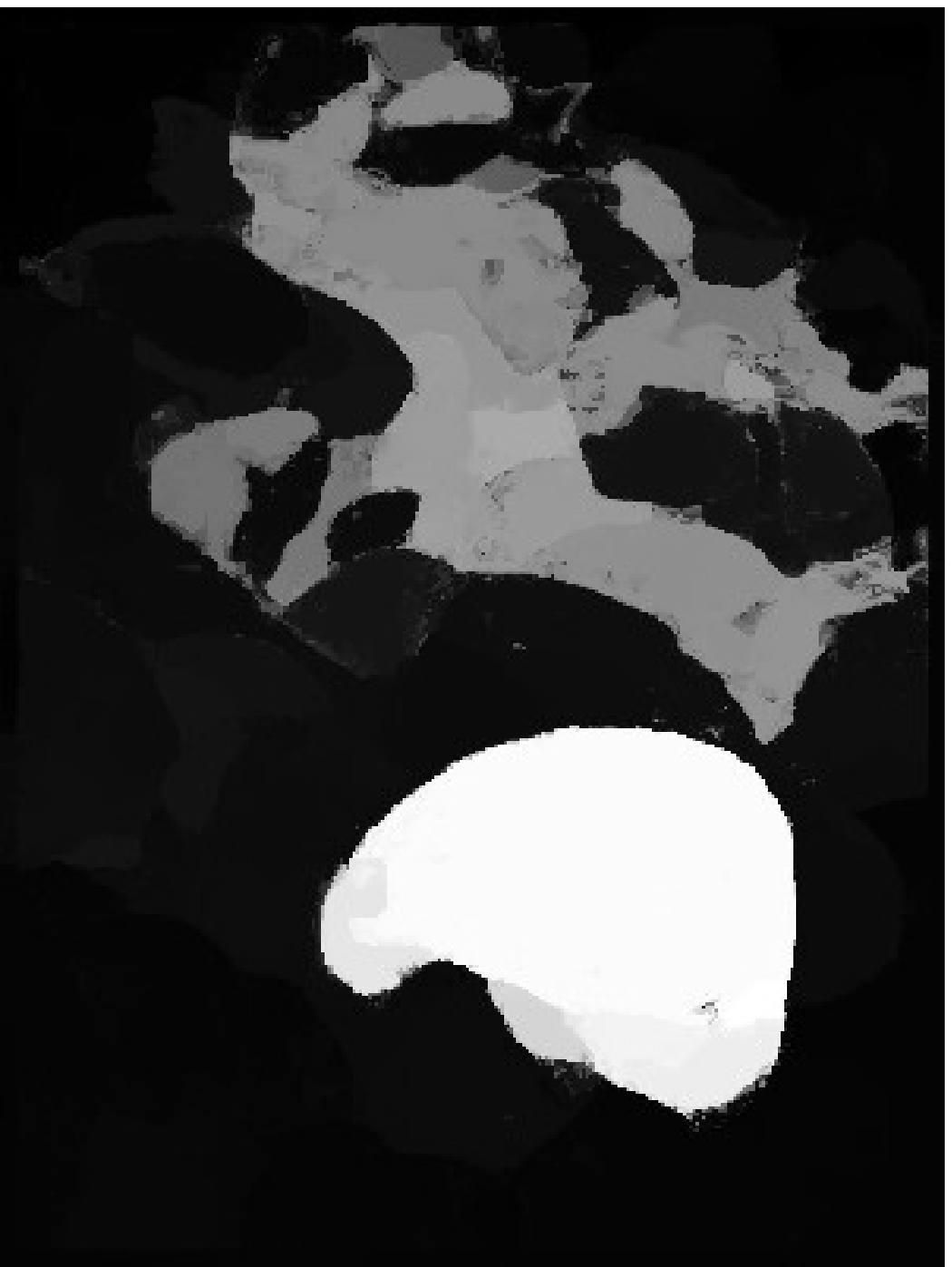} &
\includegraphics[height=1.6cm]{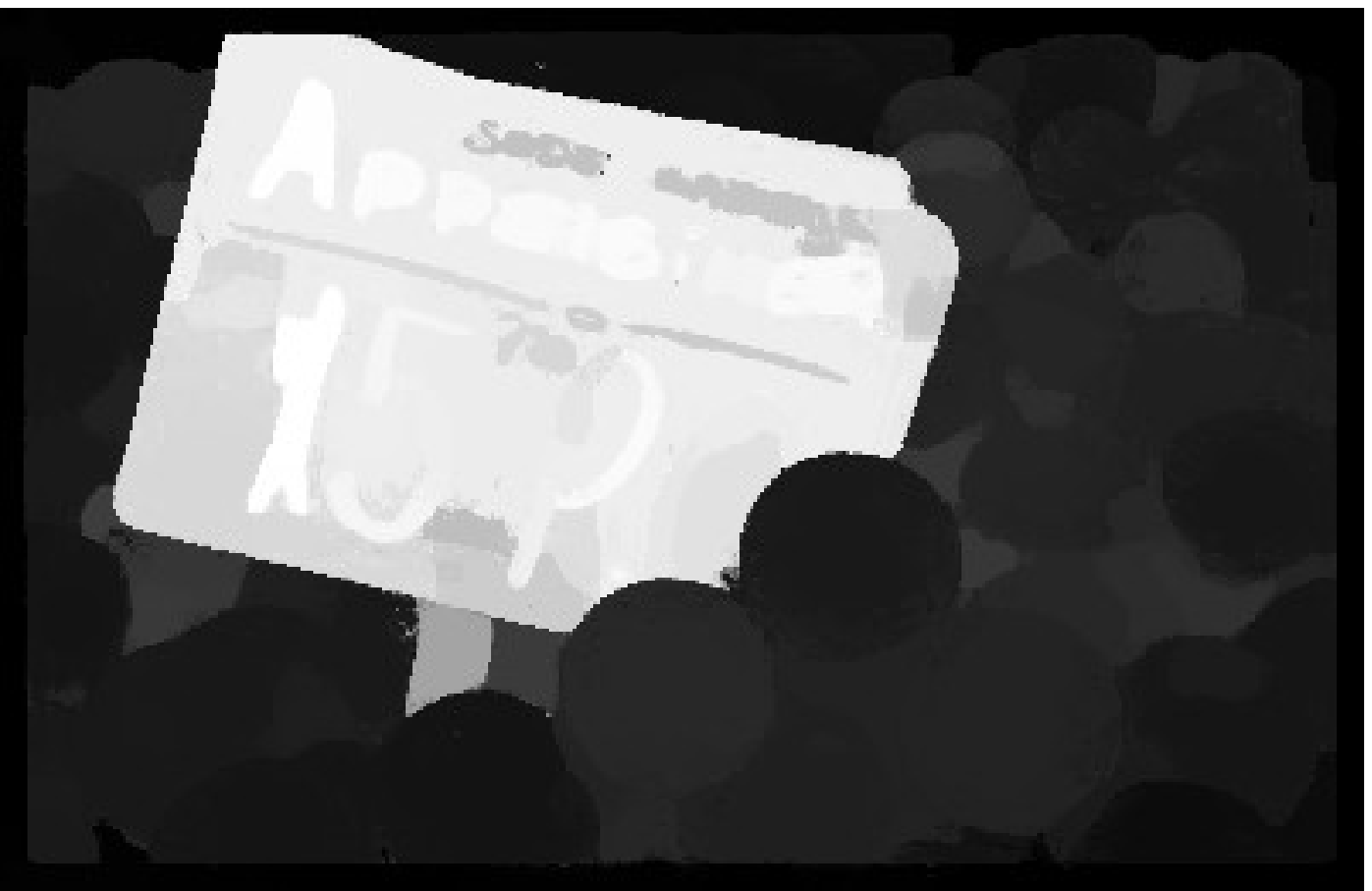} &
\includegraphics[height=1.6cm]{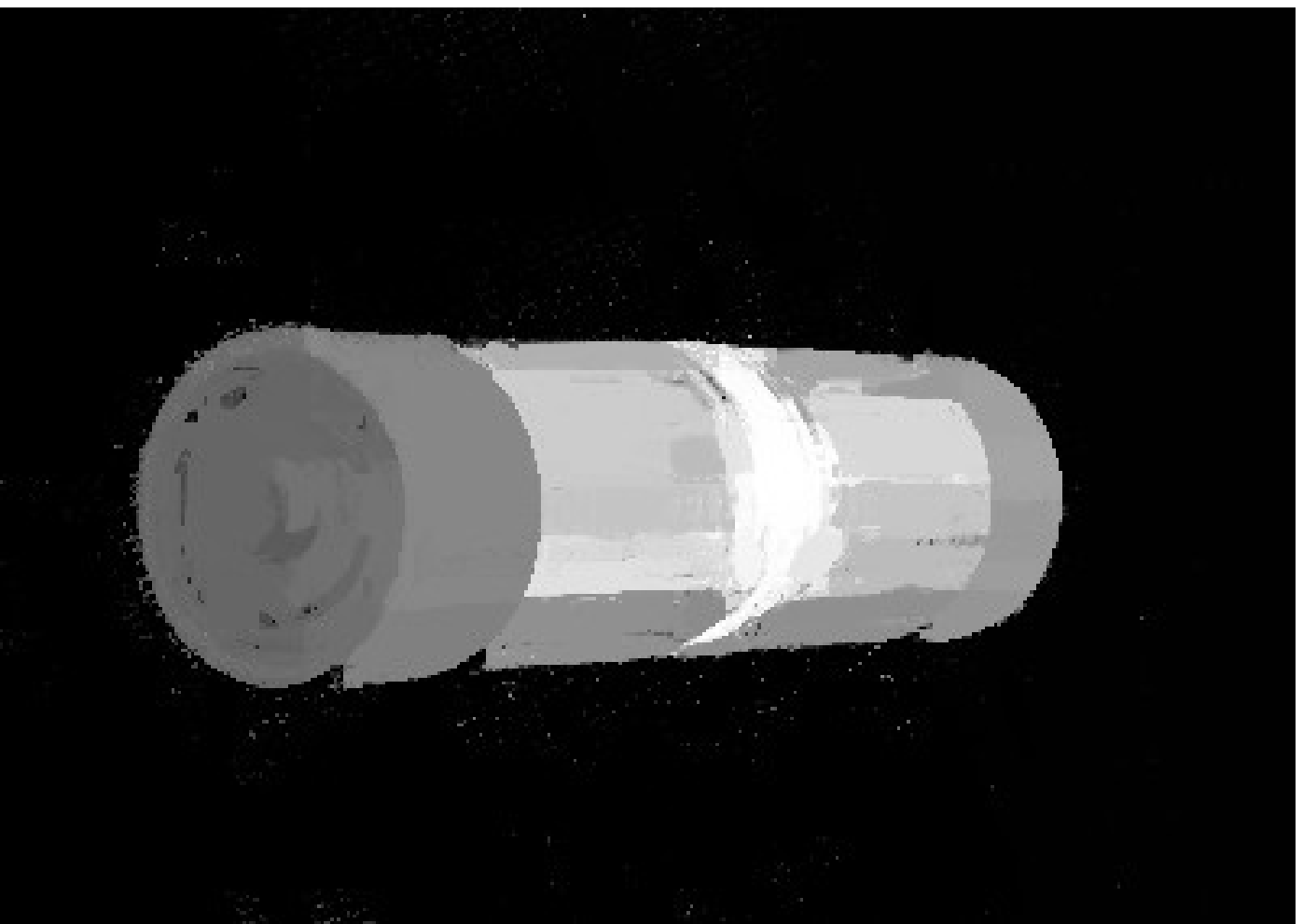} &
\includegraphics[height=1.6cm]{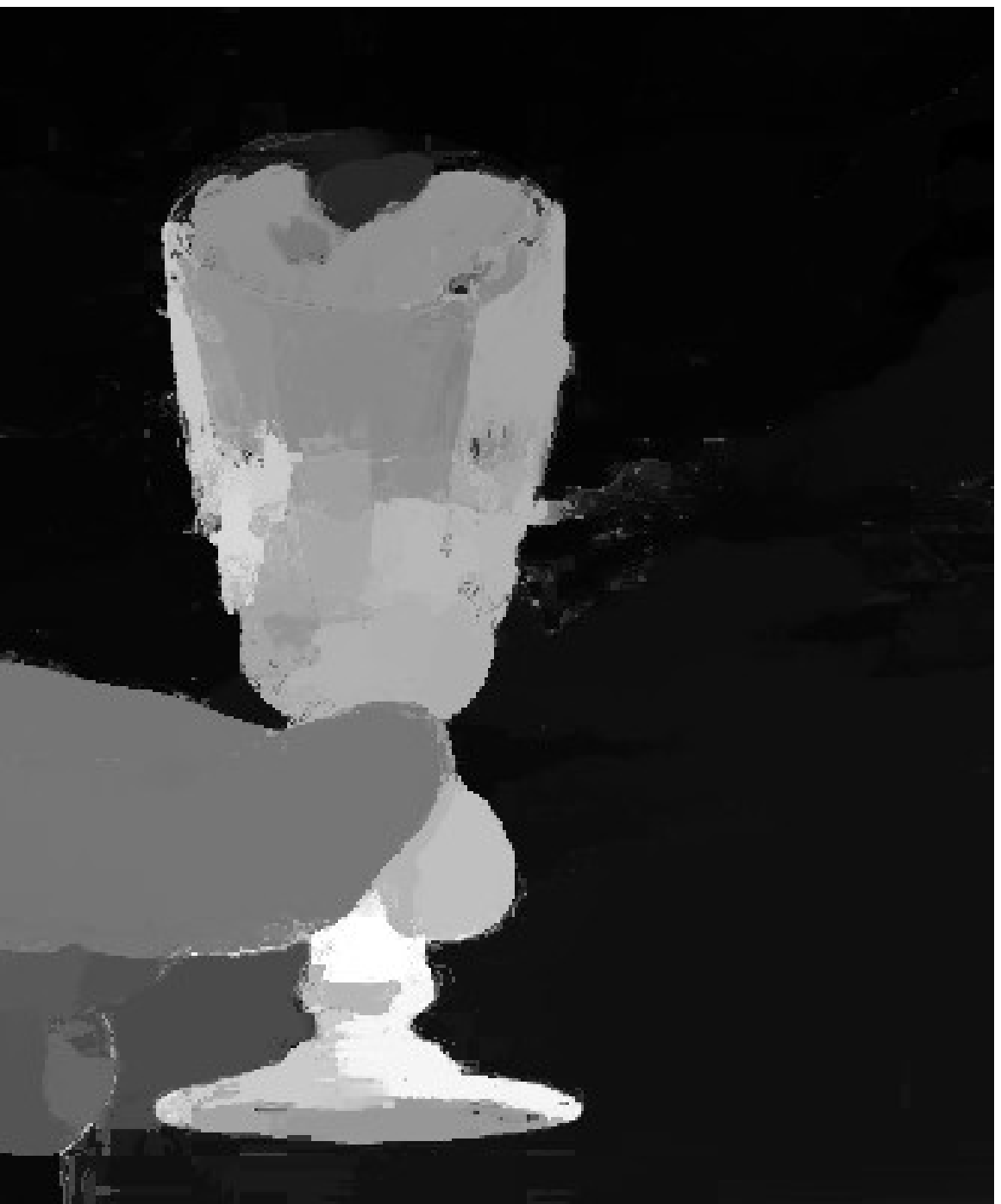} \\
HP&
\includegraphics[height=1.6cm]{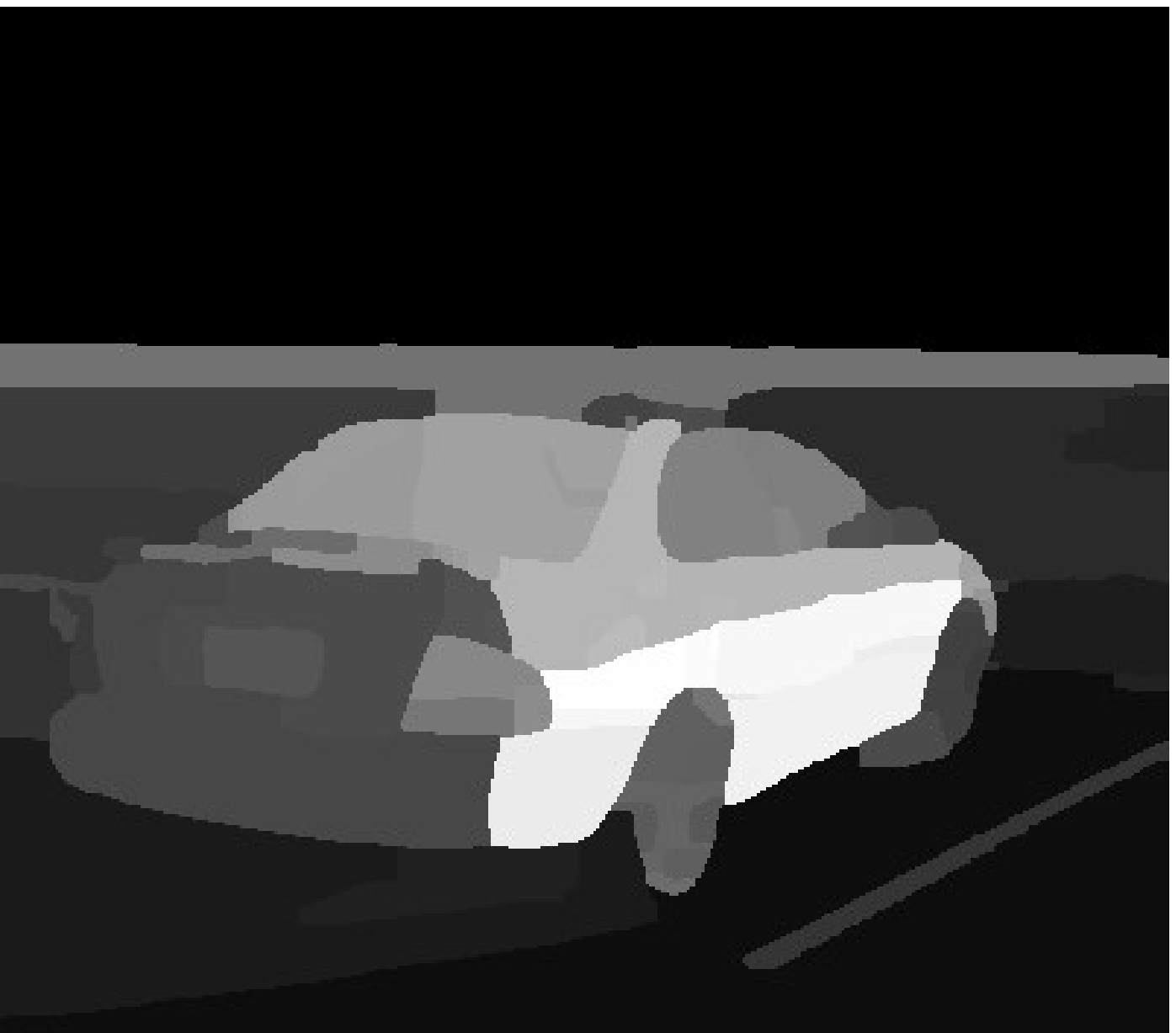} &
\includegraphics[height=1.6cm]{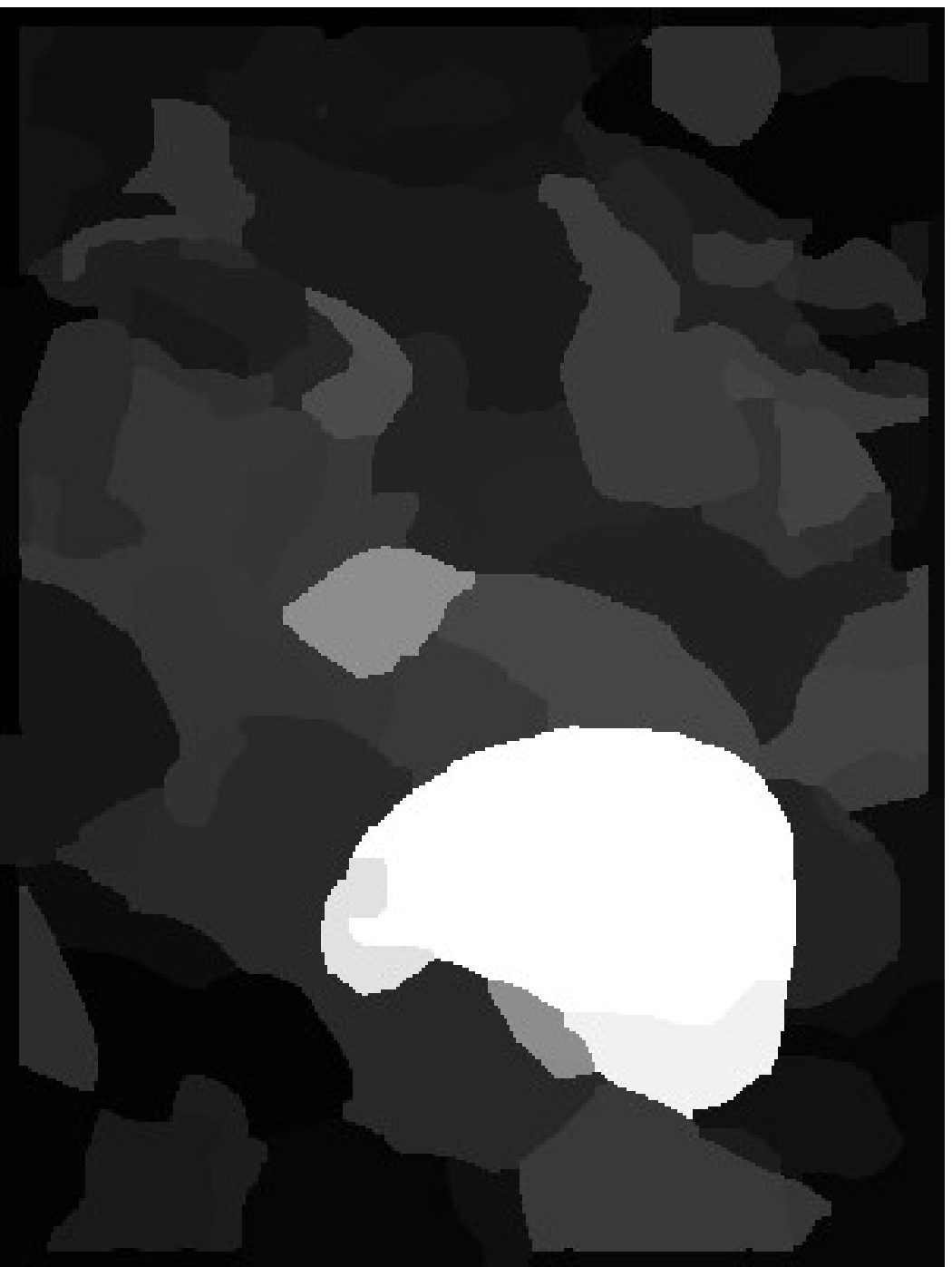} &
\includegraphics[height=1.6cm]{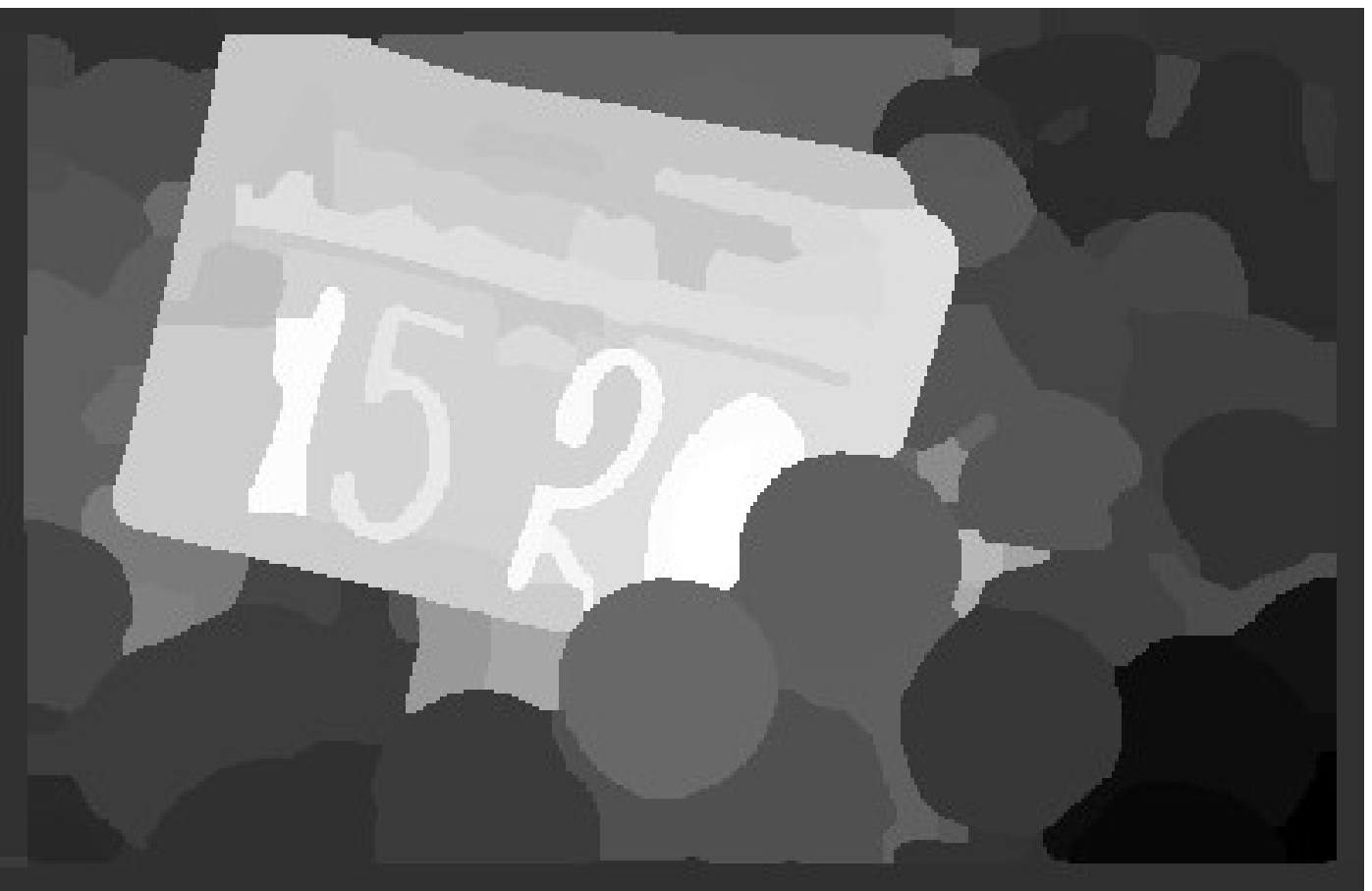} &
\includegraphics[height=1.6cm]{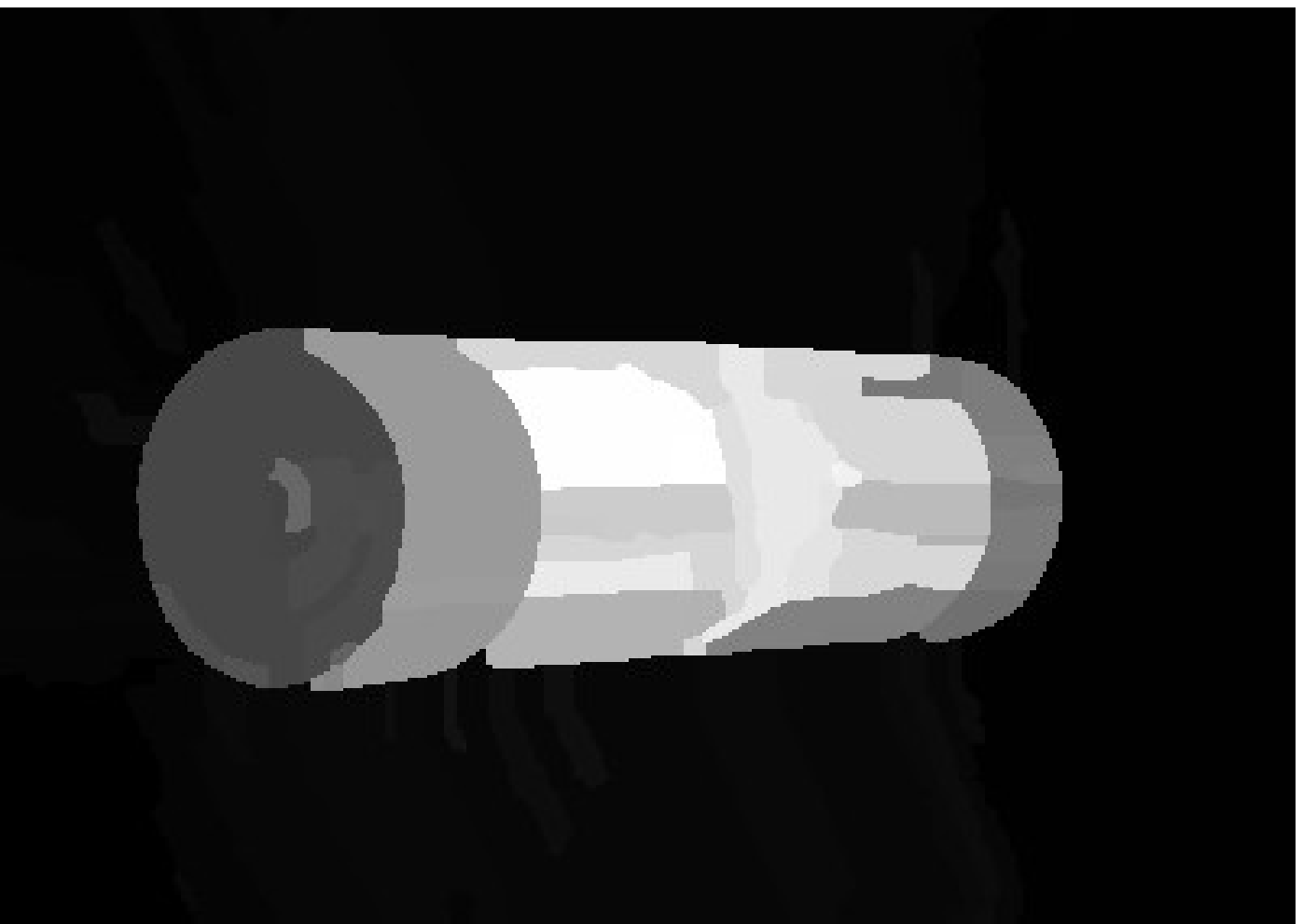} &
\includegraphics[height=1.6cm]{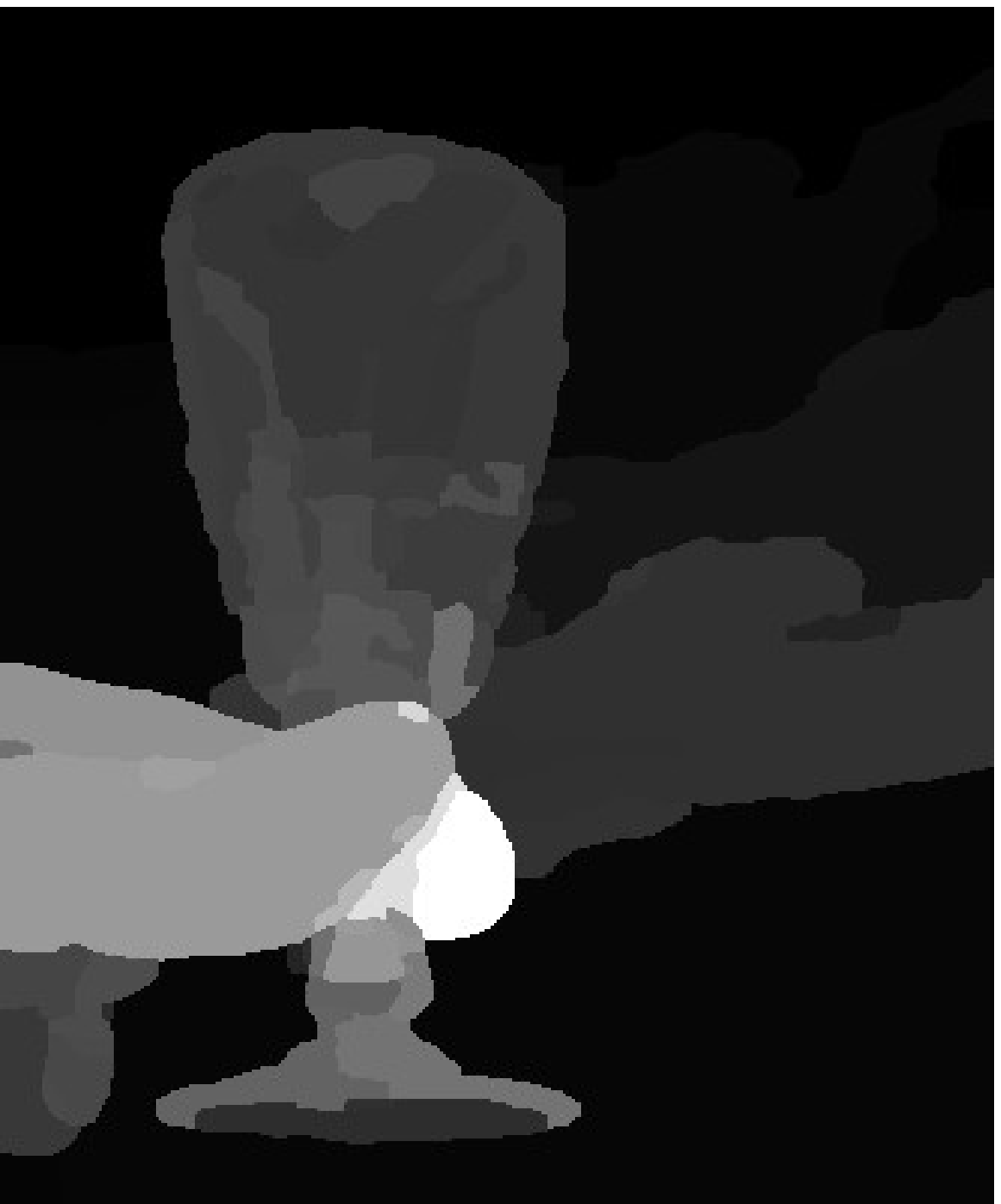} \\
SOH&
\includegraphics[height=1.6cm]{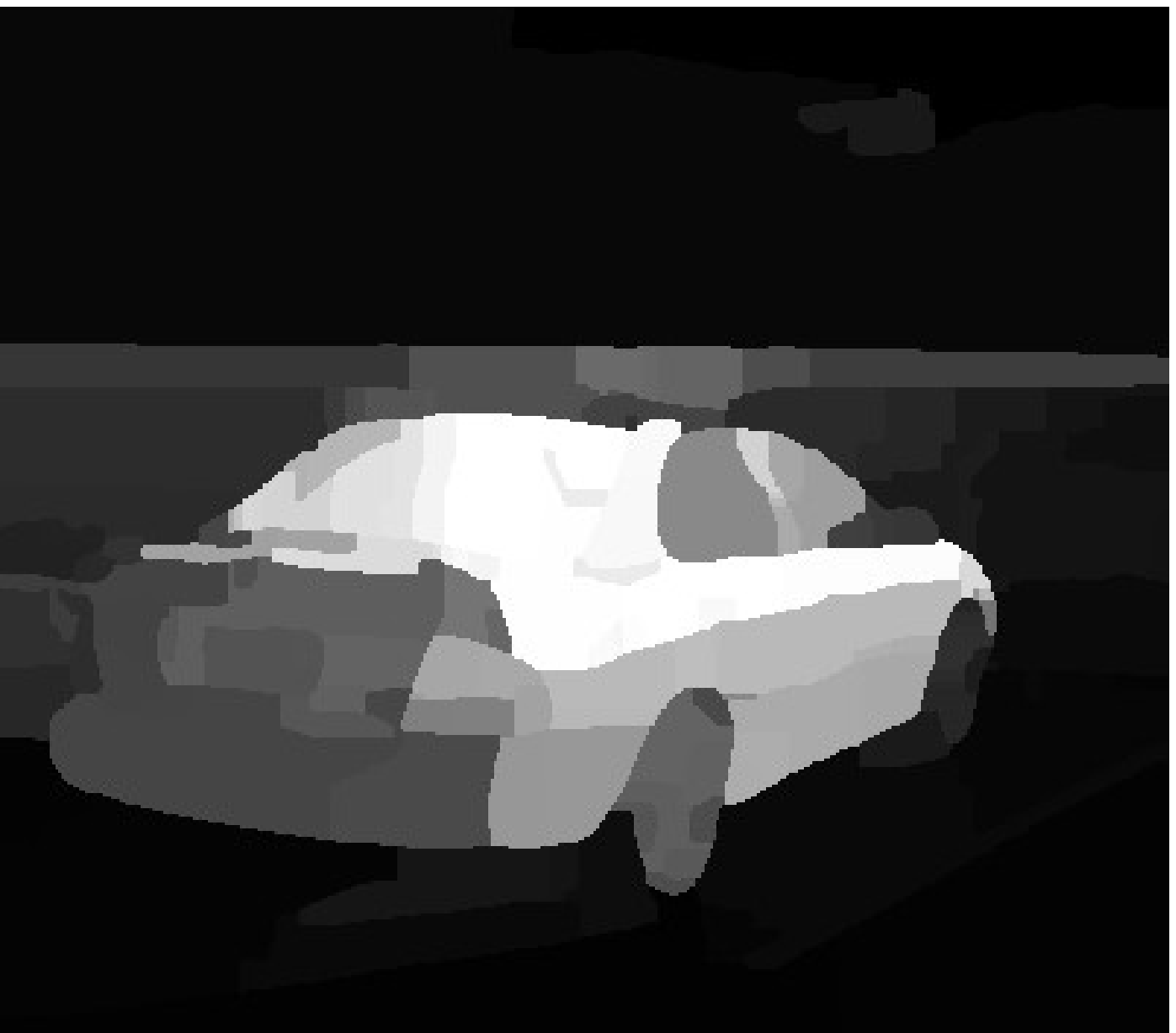} &
\includegraphics[height=1.6cm]{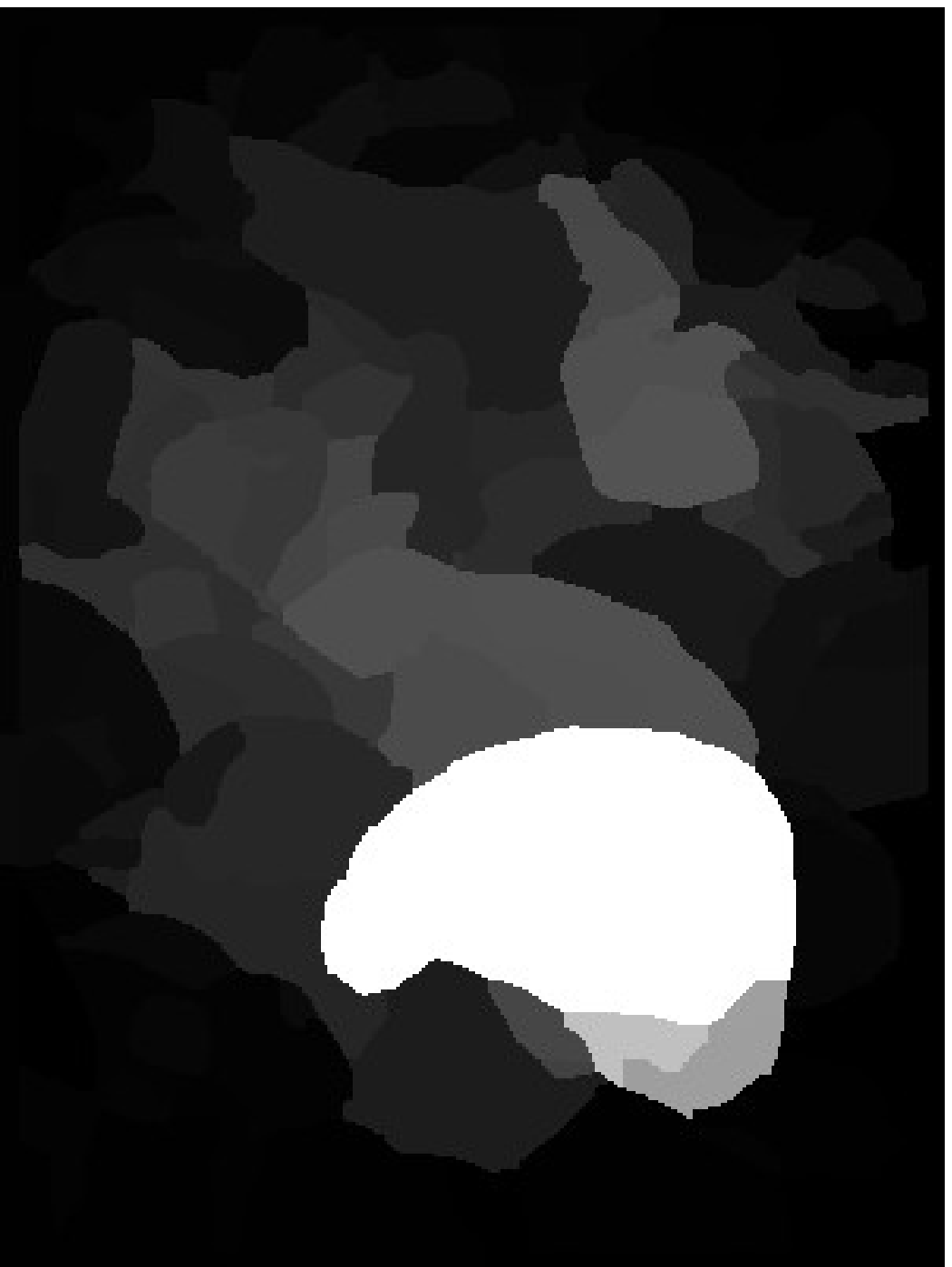} &
\includegraphics[height=1.6cm]{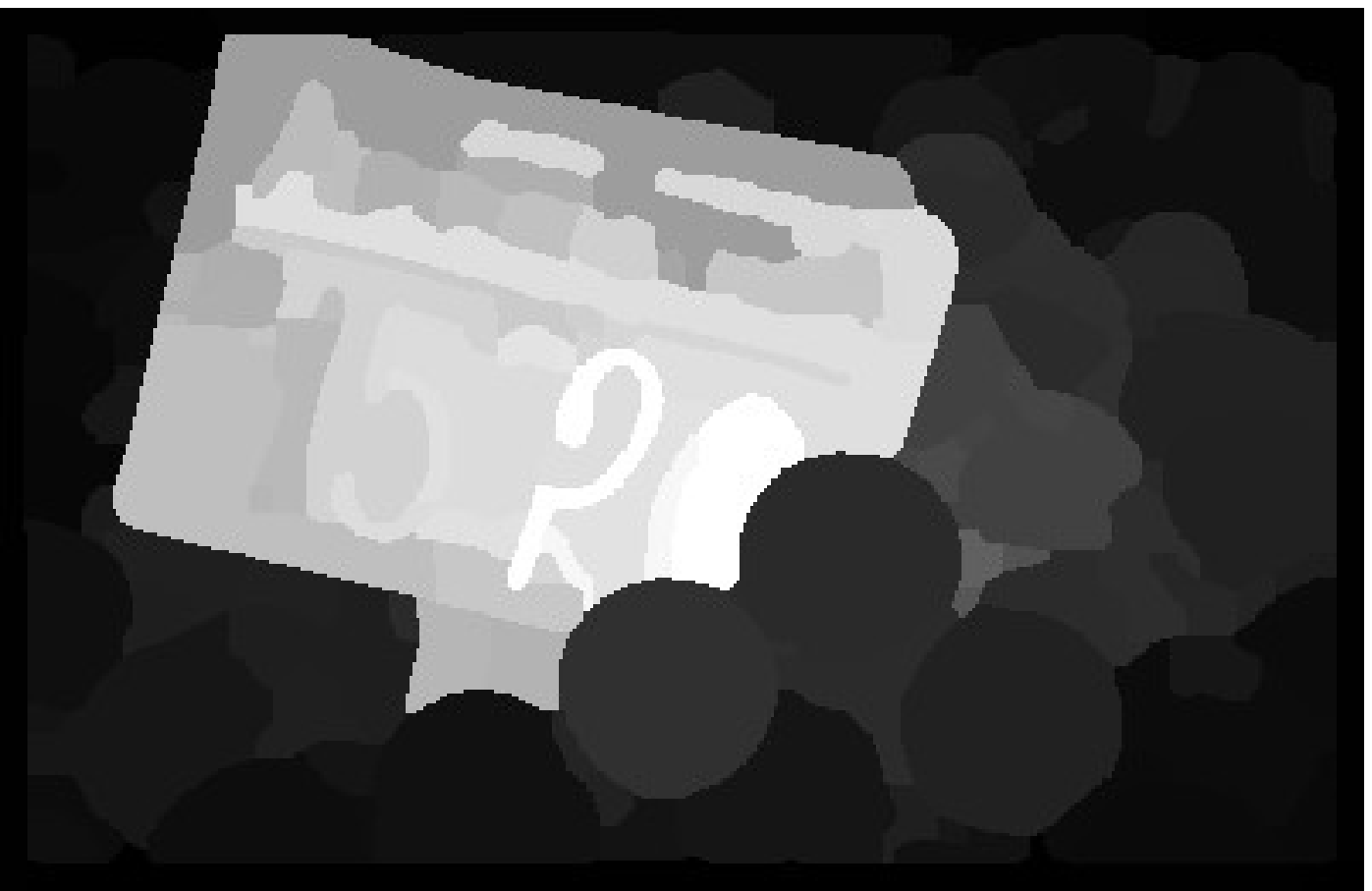} &
\includegraphics[height=1.6cm]{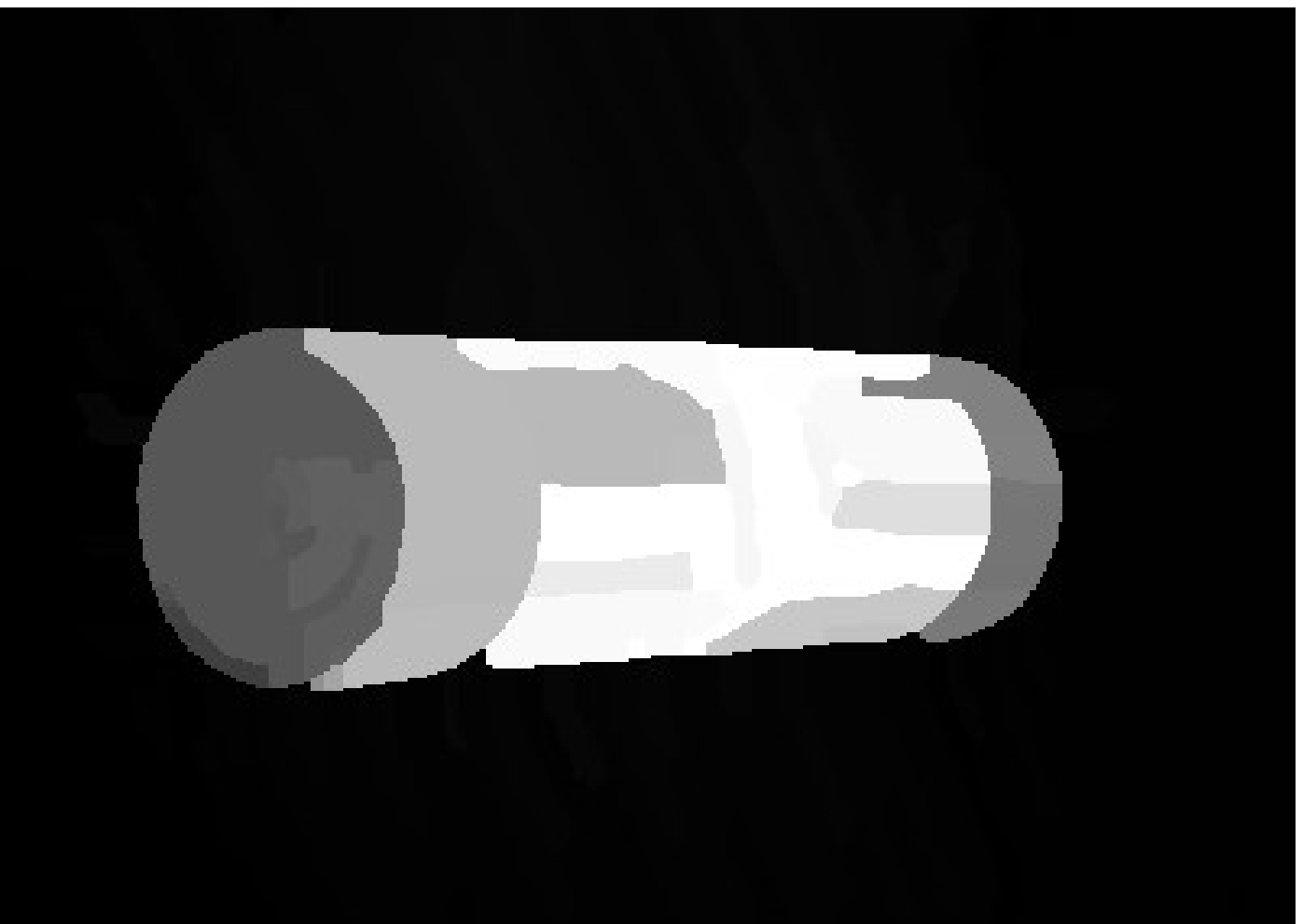} &
\includegraphics[height=1.6cm]{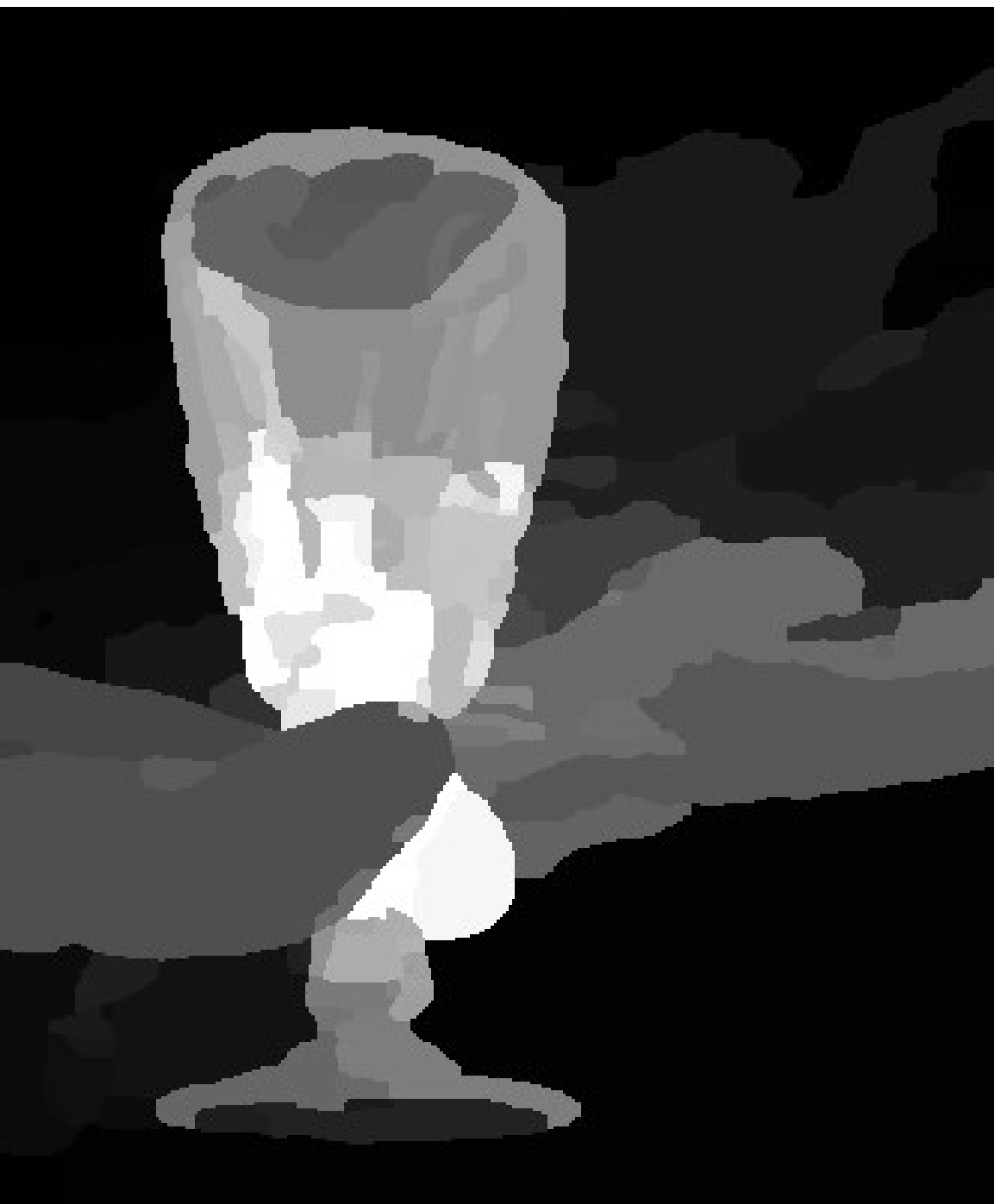} \\
GT&
\includegraphics[height=1.6cm]{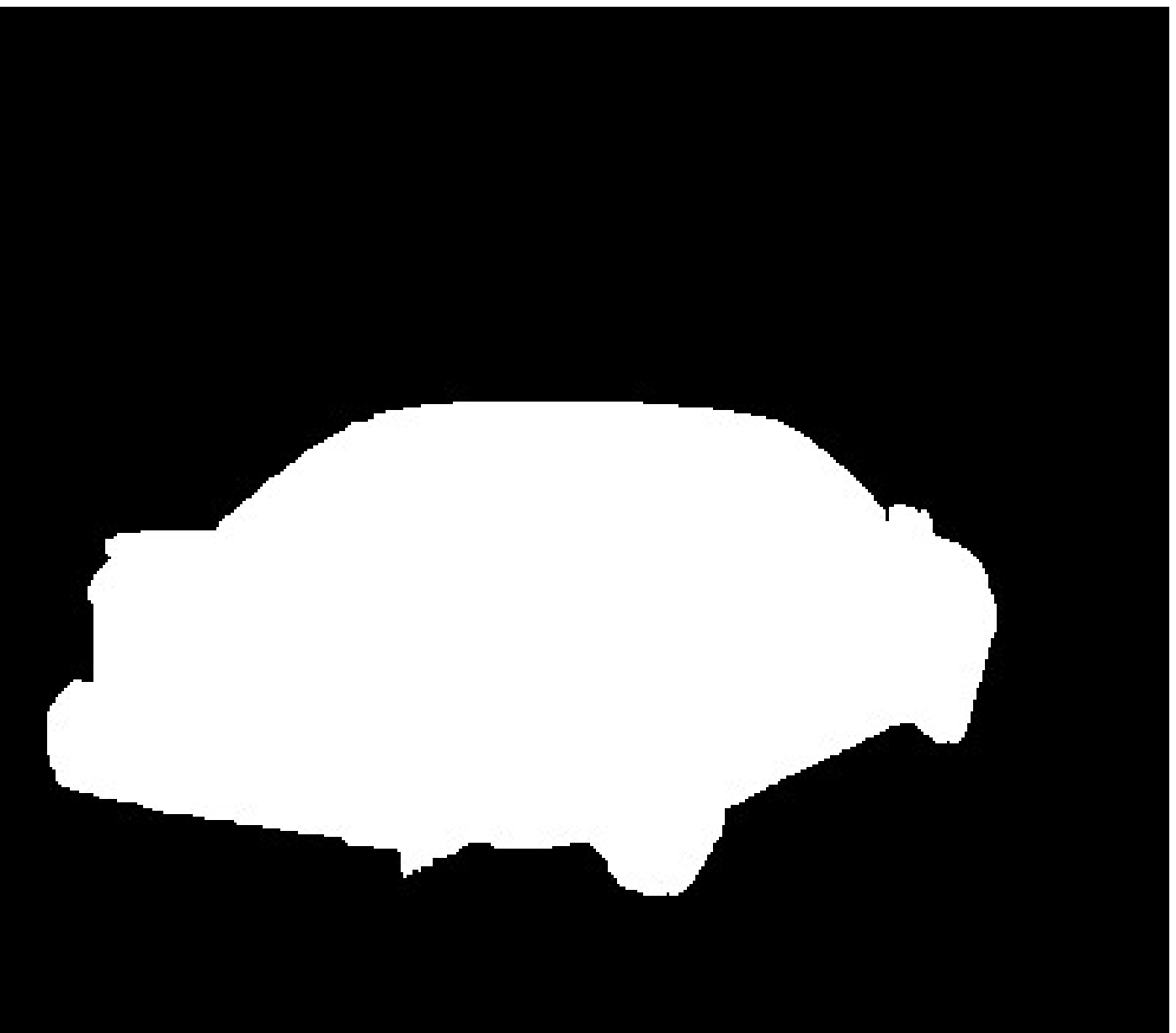} &
\includegraphics[height=1.6cm]{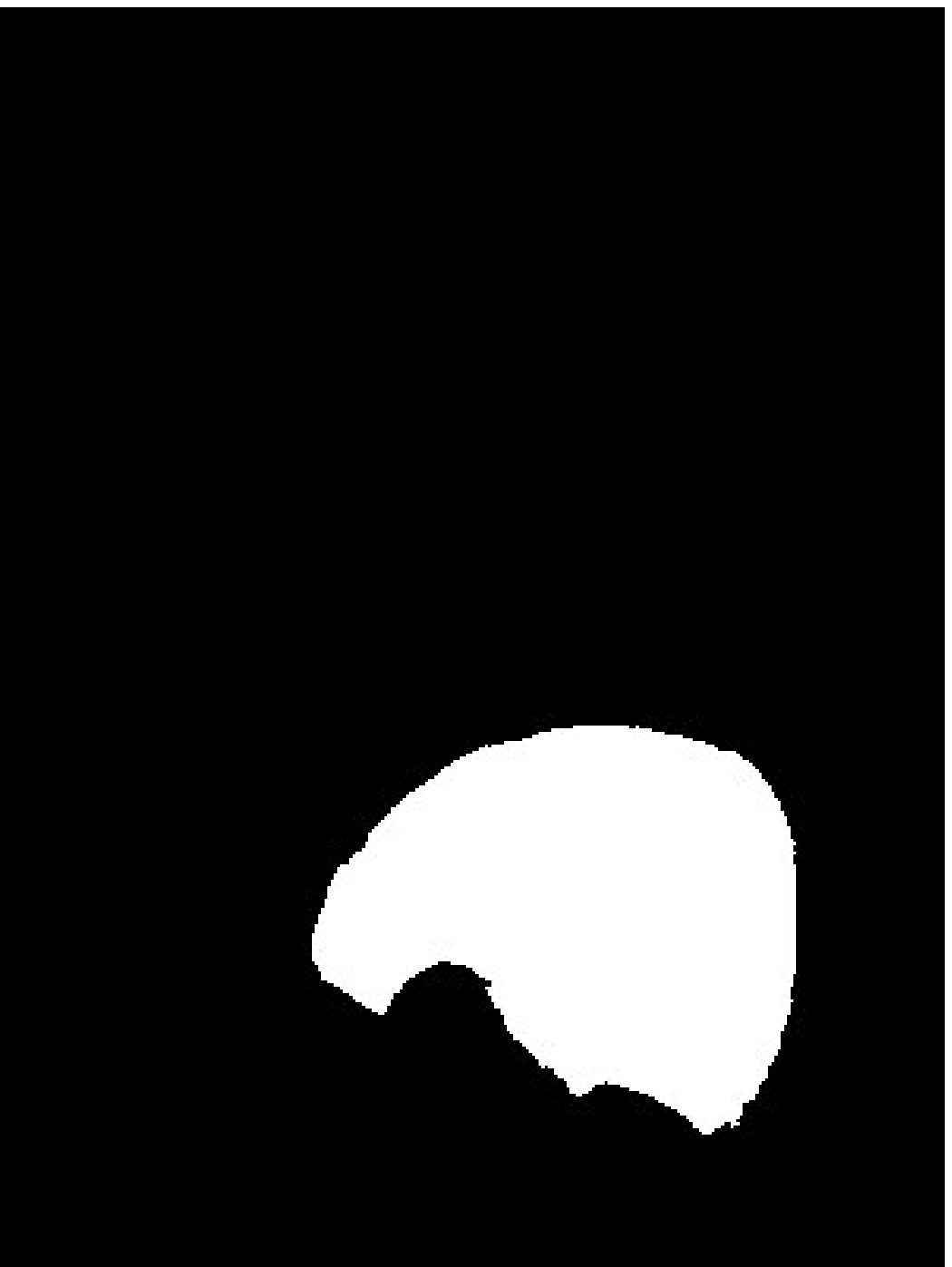} &
\includegraphics[height=1.6cm]{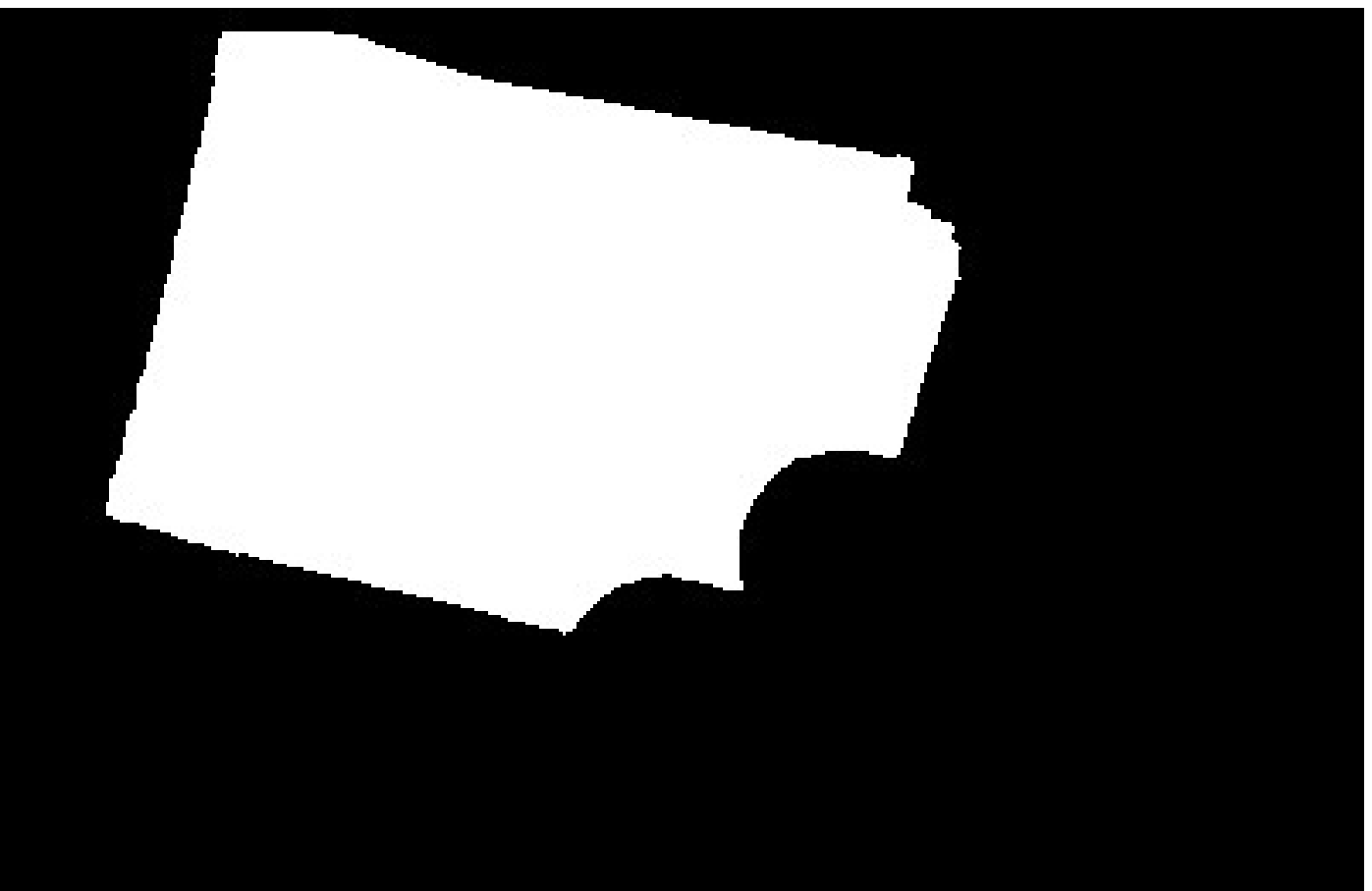} &
\includegraphics[height=1.6cm]{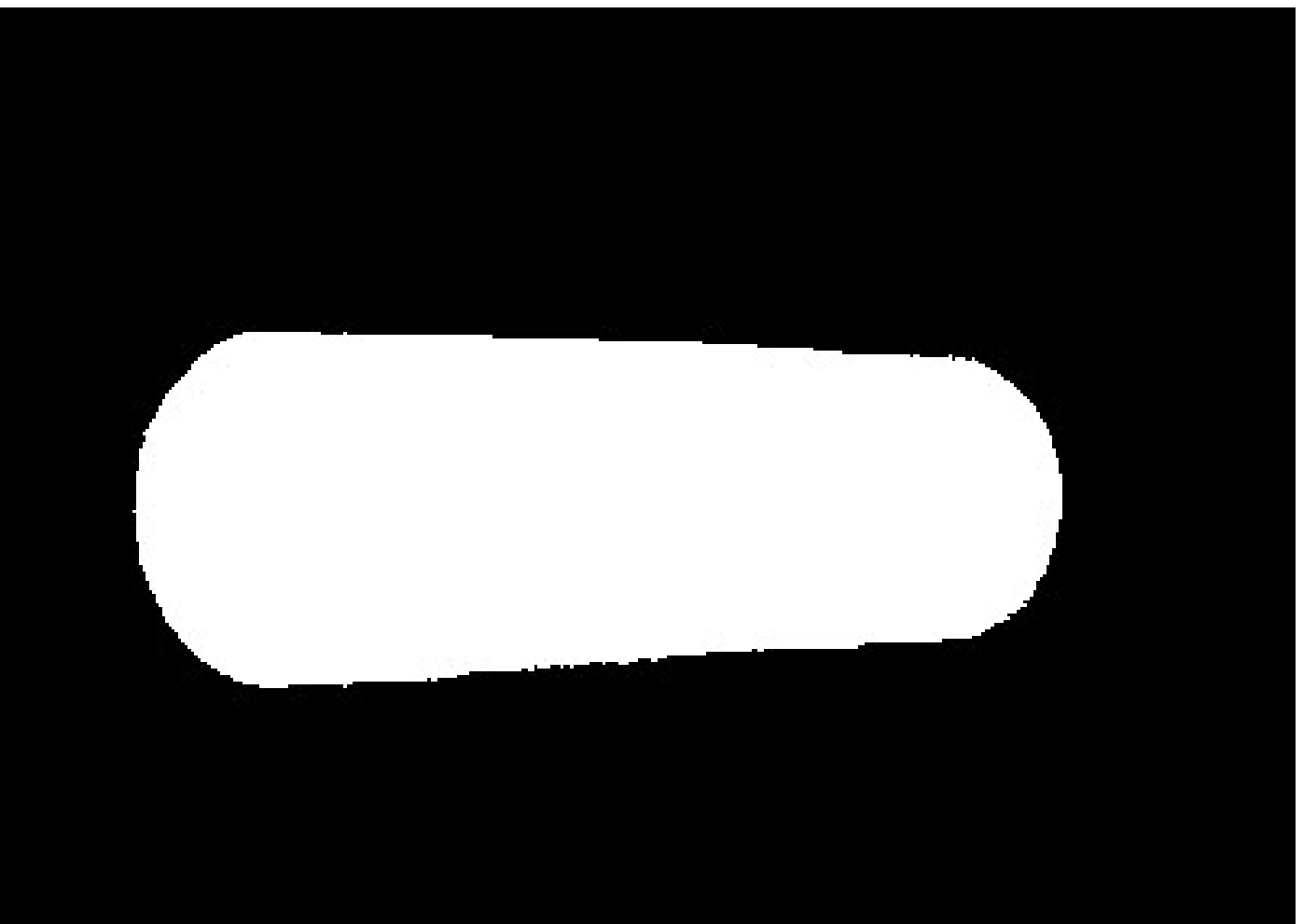} &
\includegraphics[height=1.6cm]{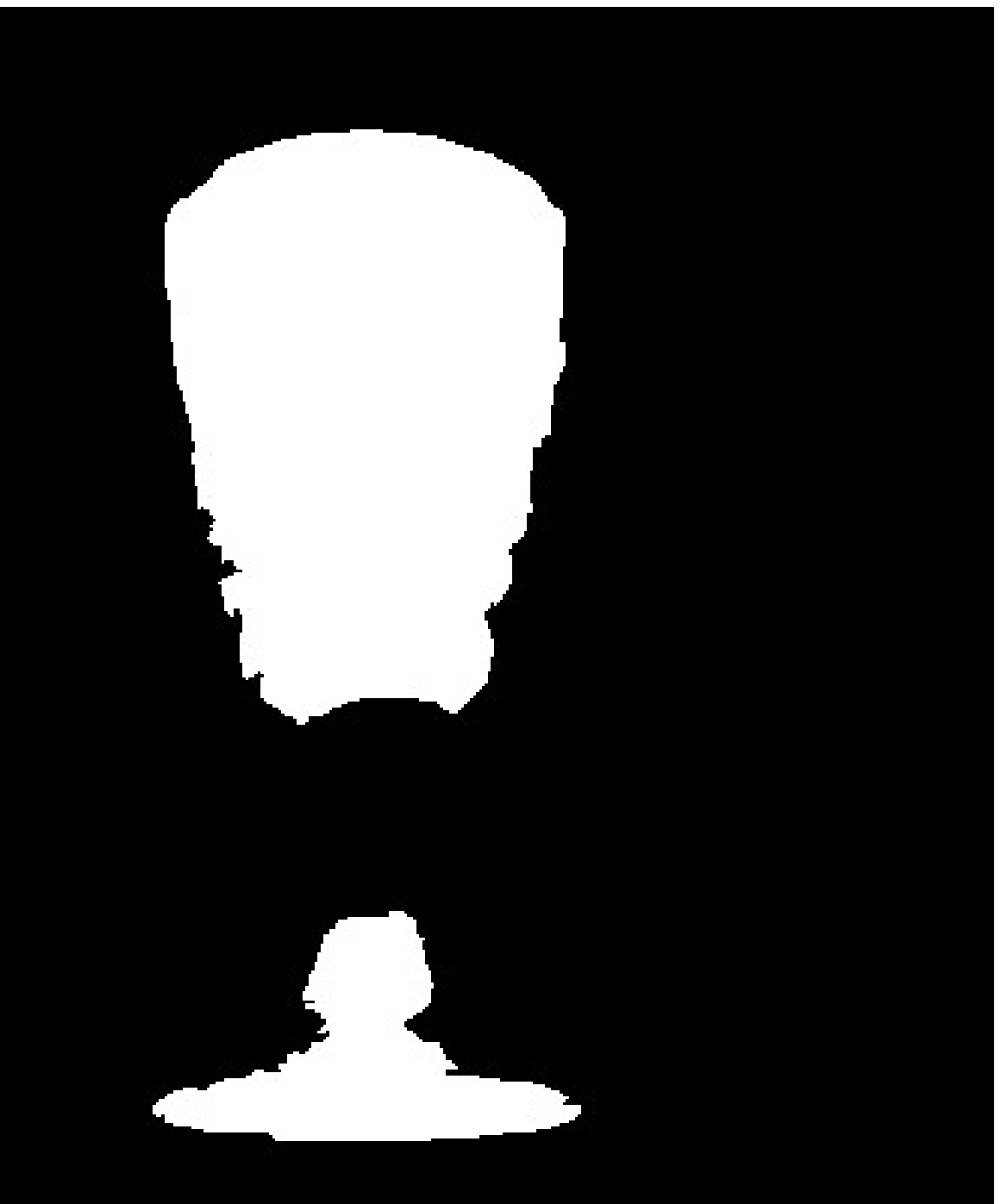} \\
& (a) & (b) & (c) & (d) & (e) \\
\end{tabular}
\caption{Experimental results on some MSRA images. First row: IM original images; saliency maps generated using FT \cite{Achanta-09cvpr},  RC \cite{Cheng-11cvpr}, SF \cite{perazzi-cvpr12}, PCAS \cite{margolin-cvpr13}, HS \cite{Yan-cvpr13}, ST \cite{ZLiu-tip14}, our models HP and SOH,  and GT ground truth masks.} \label{fig:MSRAimage_comp}
\end{figure}

\begin{figure}[htbp]
\centering
\begin{tabular}{cccccc}

IM&
\includegraphics[height=1.6cm]{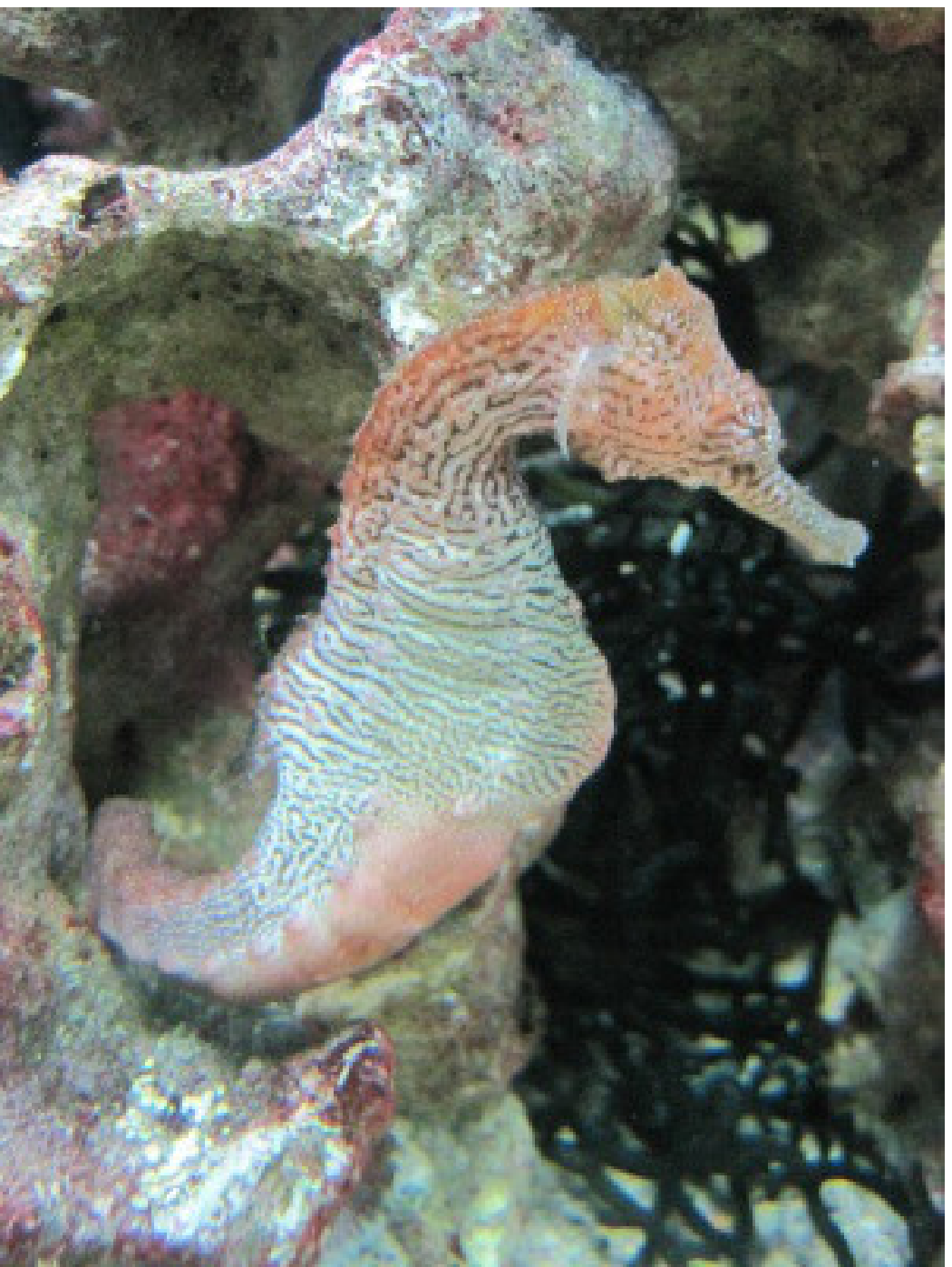} &
\includegraphics[height=1.6cm]{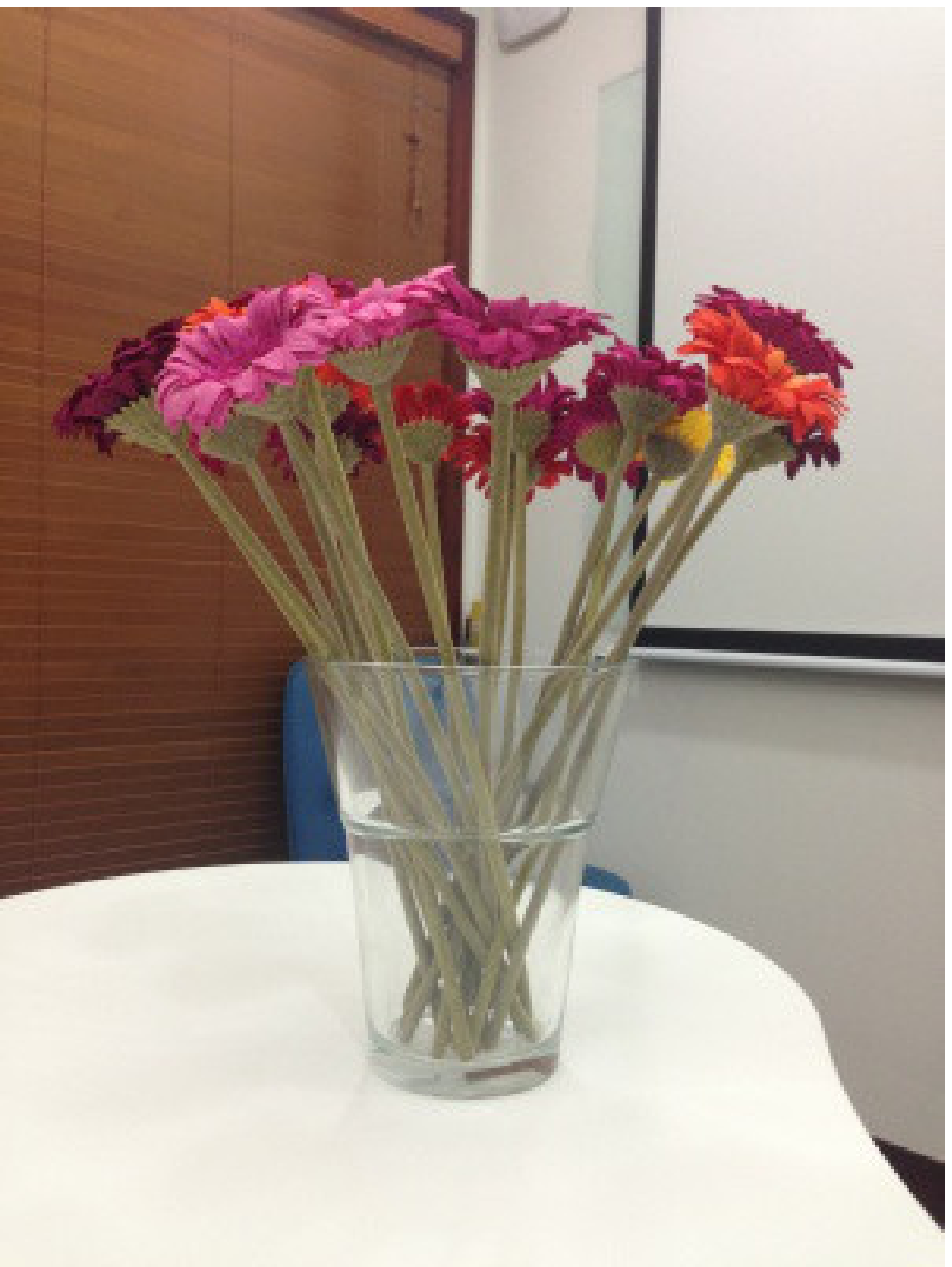} &
\includegraphics[height=1.6cm]{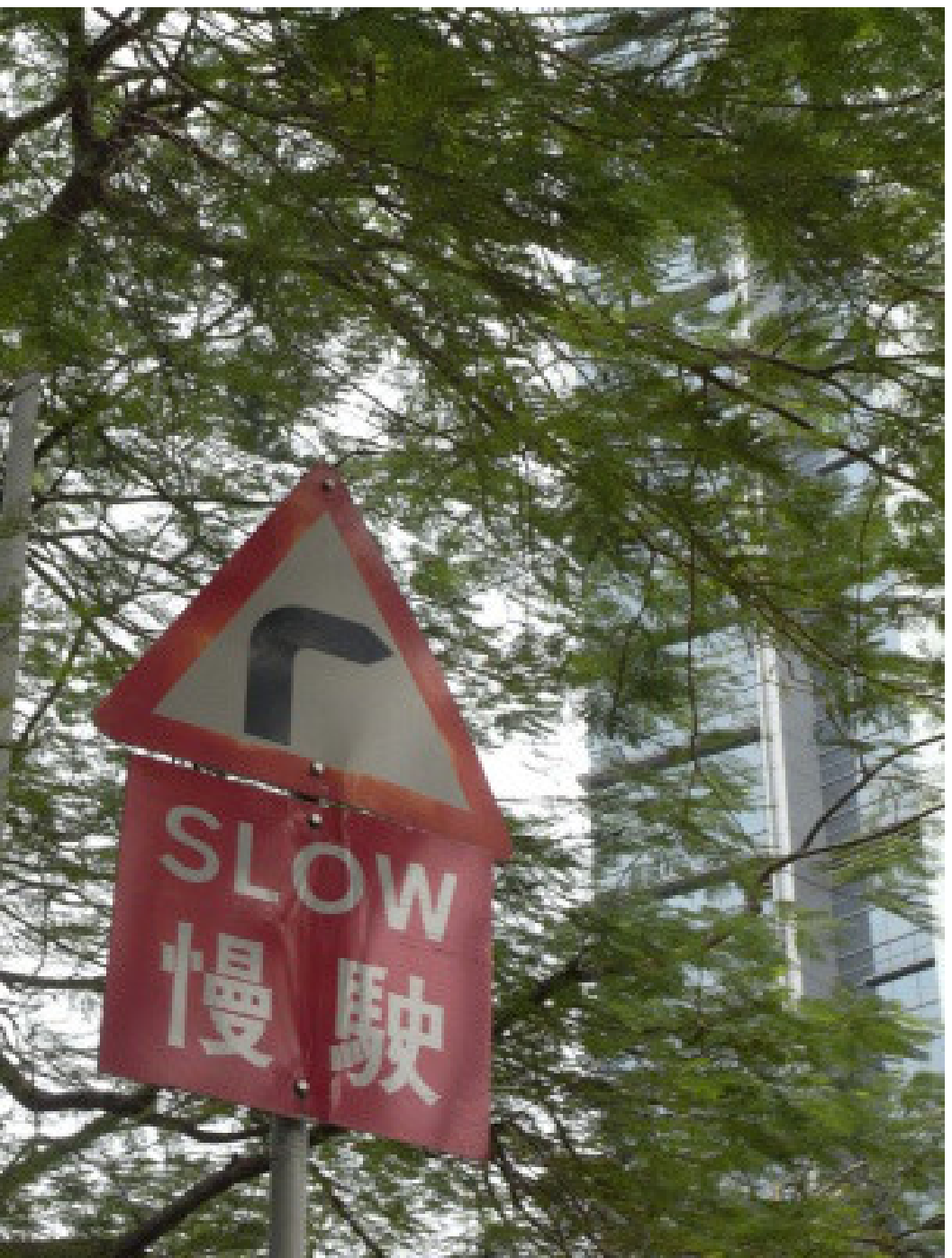} &
\includegraphics[height=1.6cm]{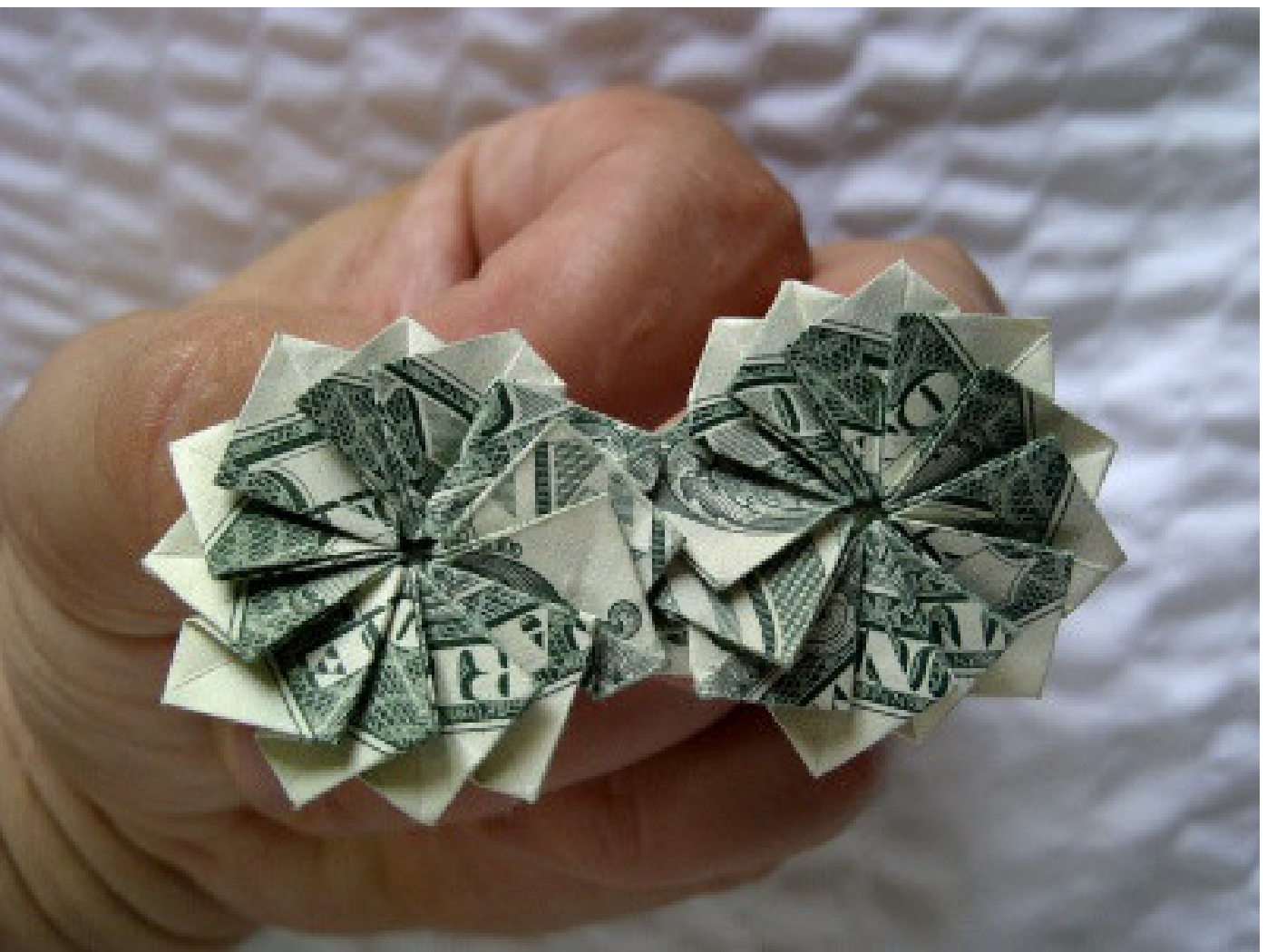} &
\includegraphics[height=1.6cm]{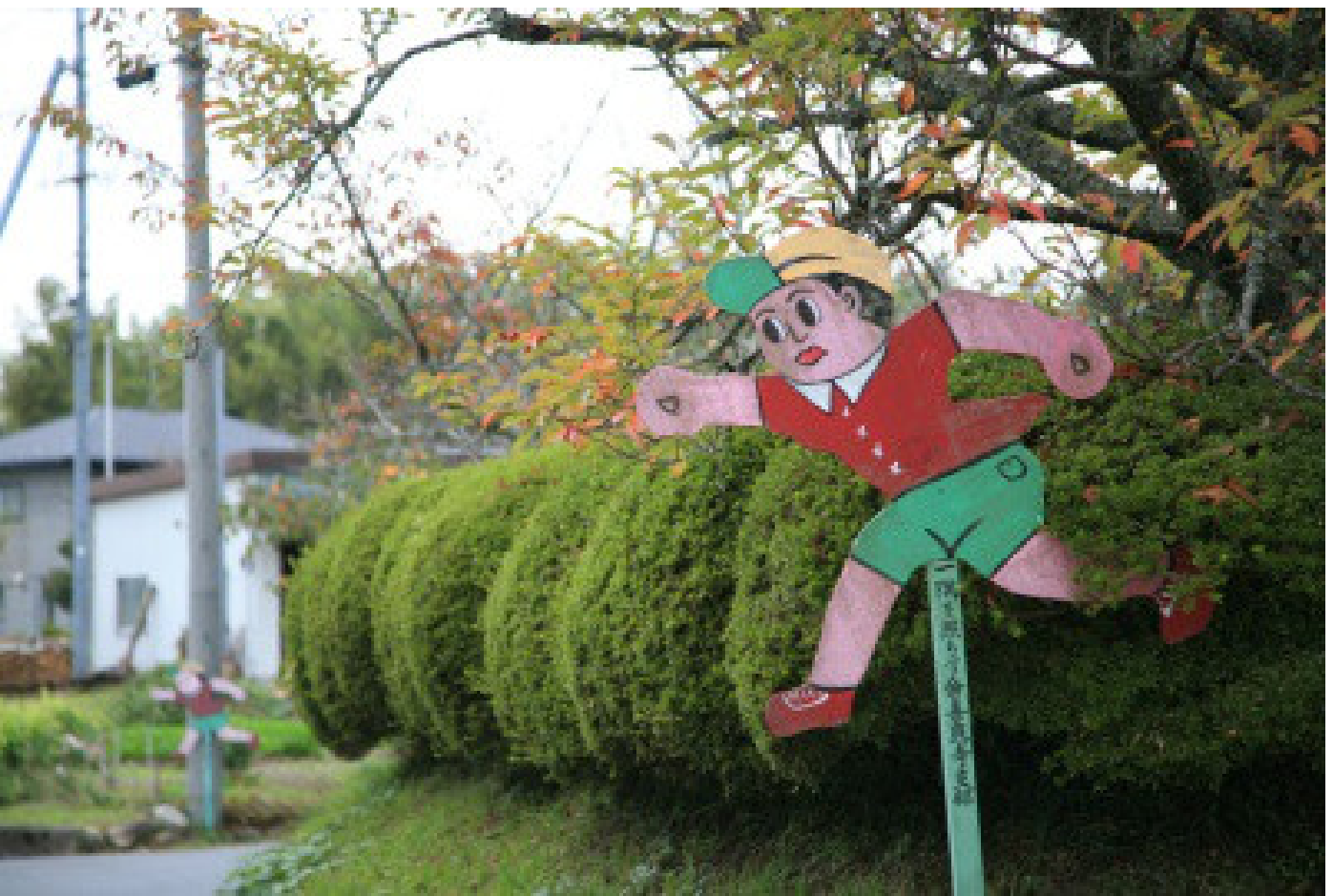} \\

FT&
\includegraphics[height=1.6cm]{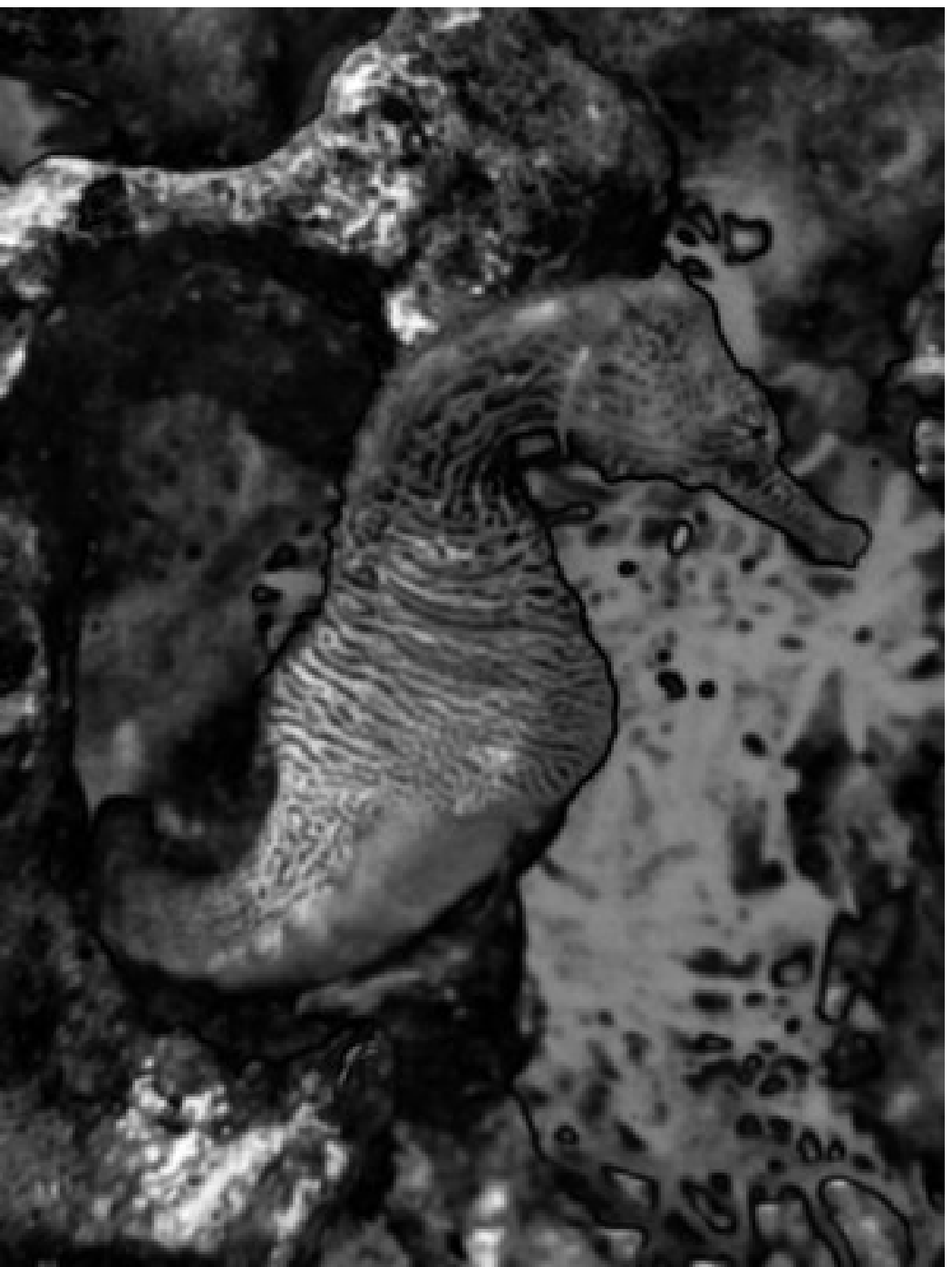} &
\includegraphics[height=1.6cm]{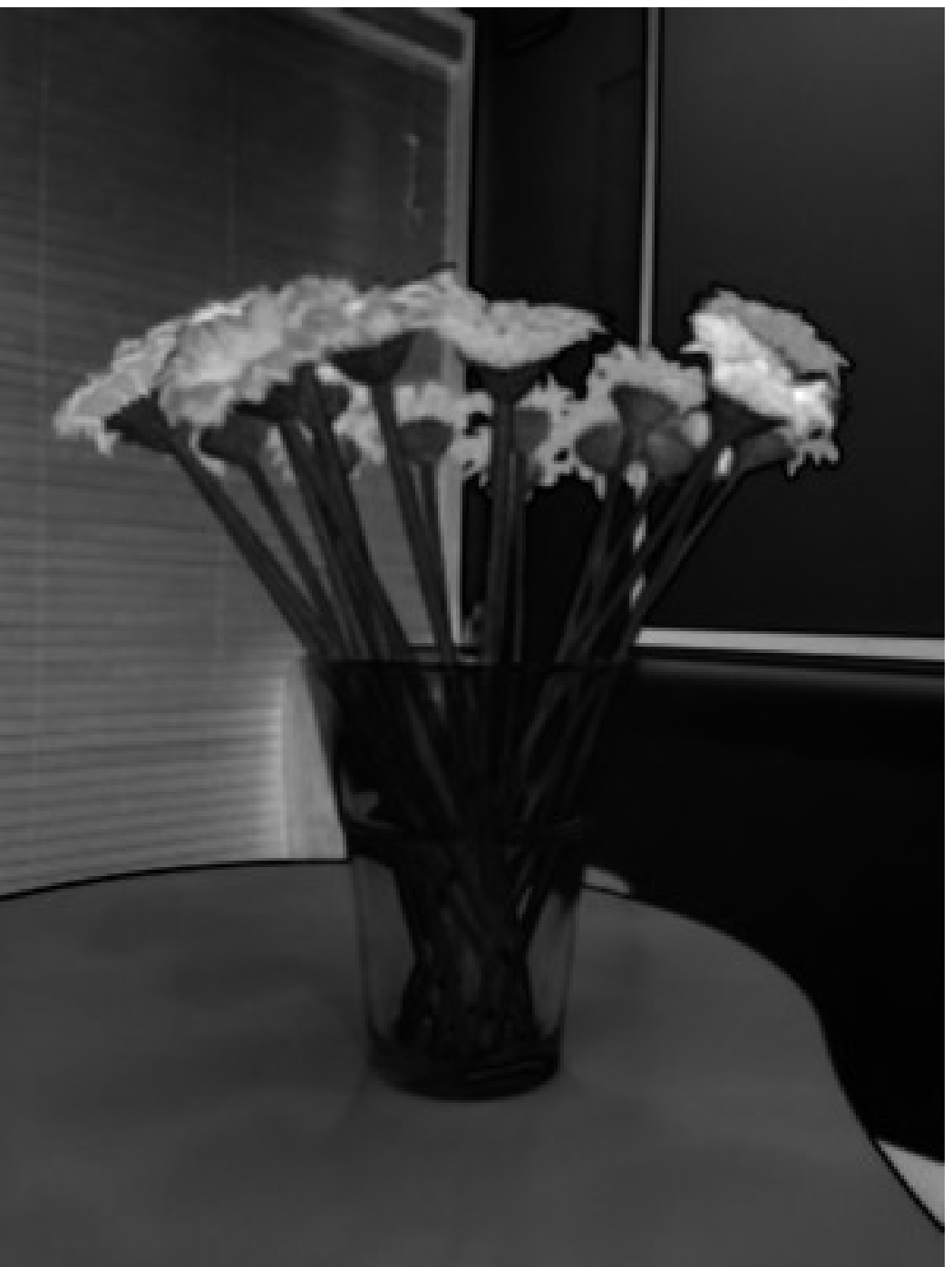} &
\includegraphics[height=1.6cm]{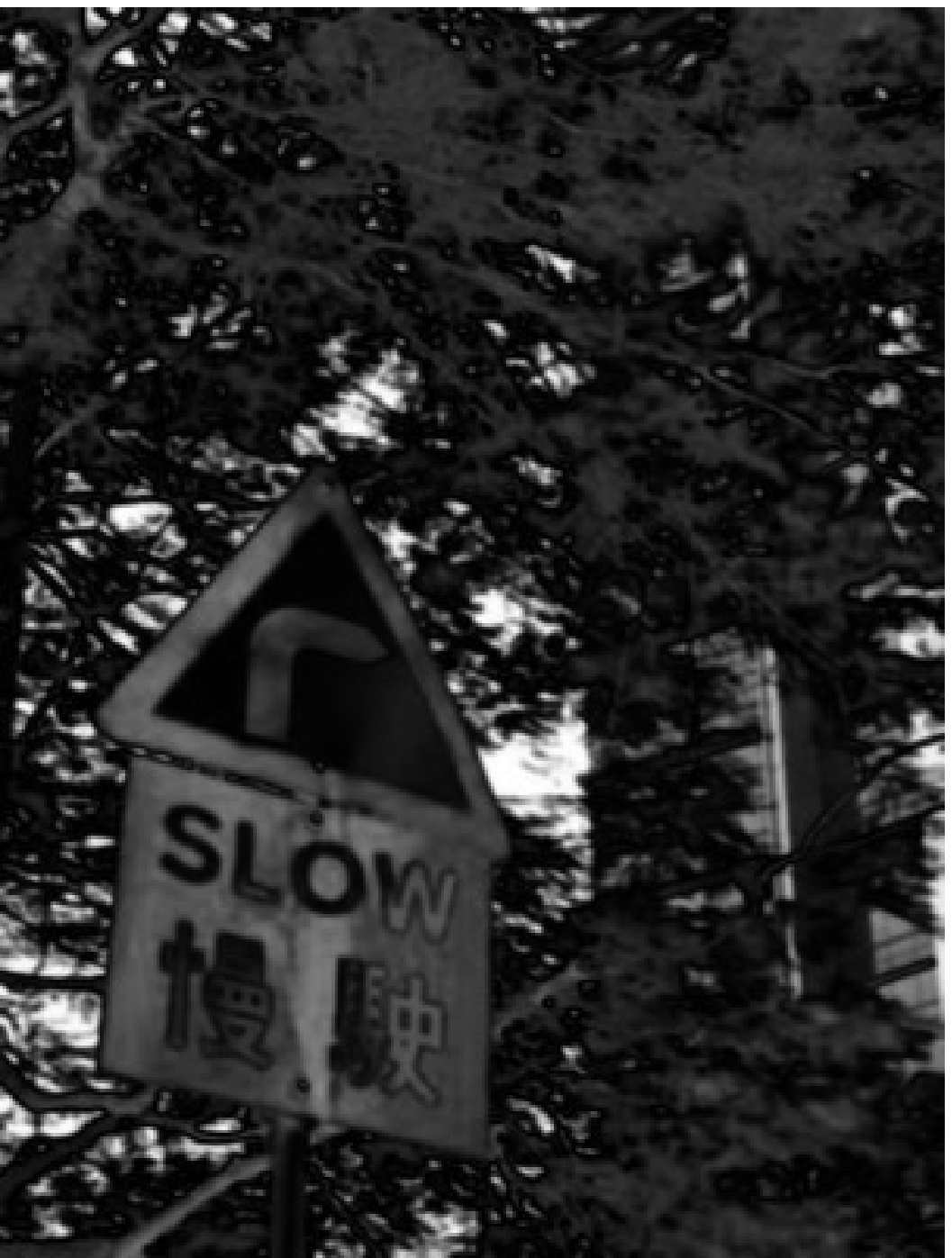} &
\includegraphics[height=1.6cm]{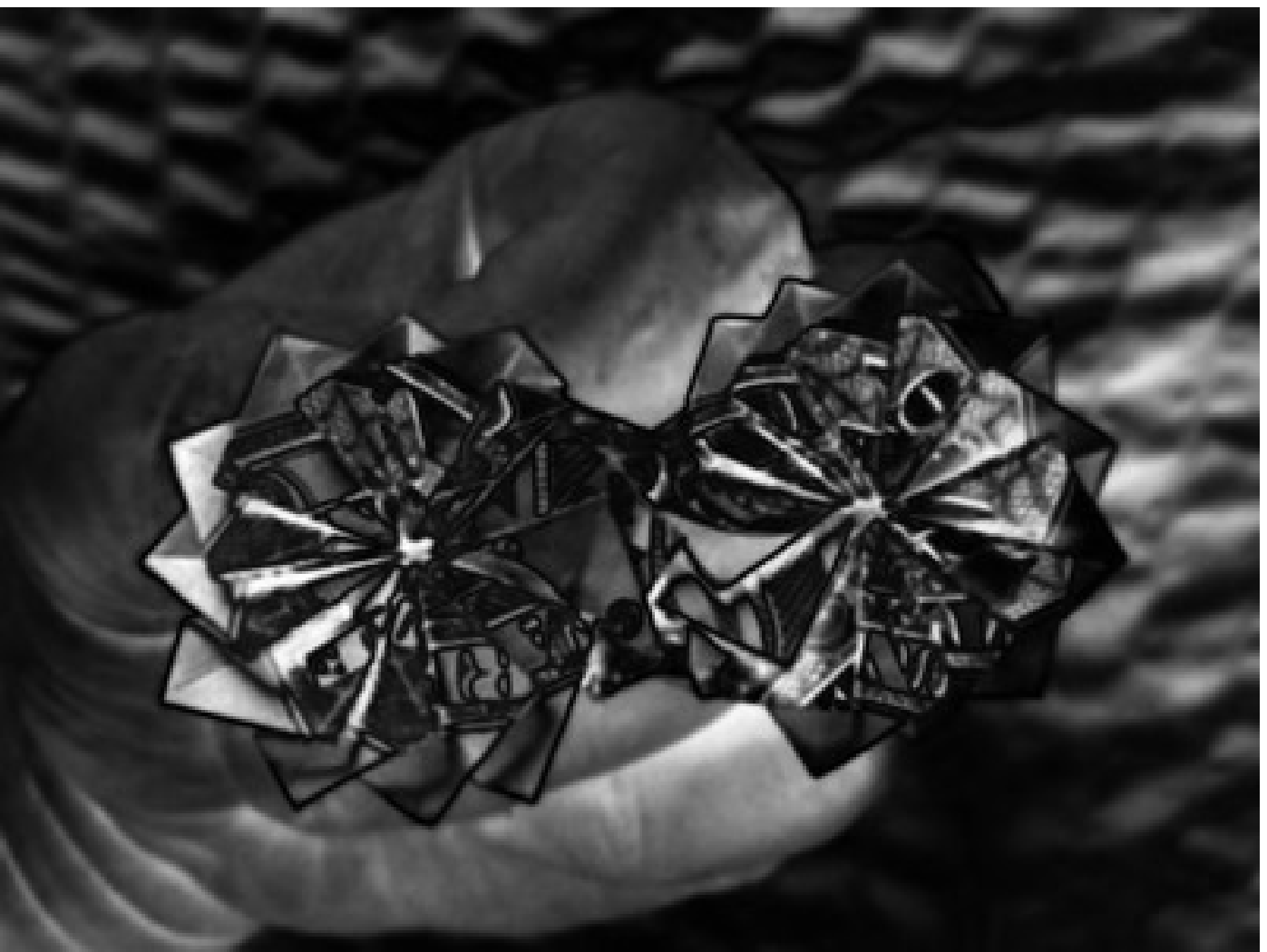} &
\includegraphics[height=1.6cm]{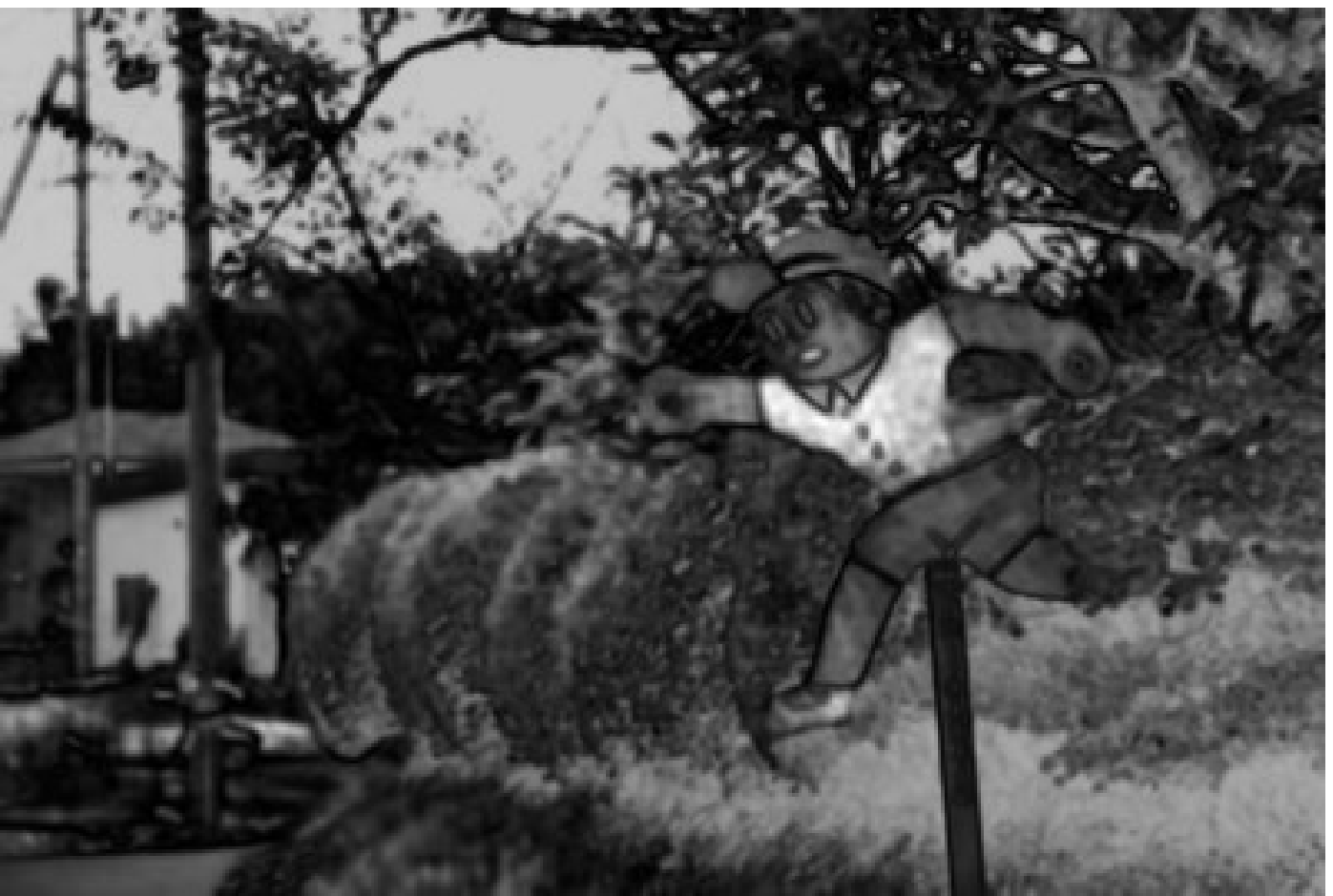} \\

RC&
\includegraphics[height=1.6cm]{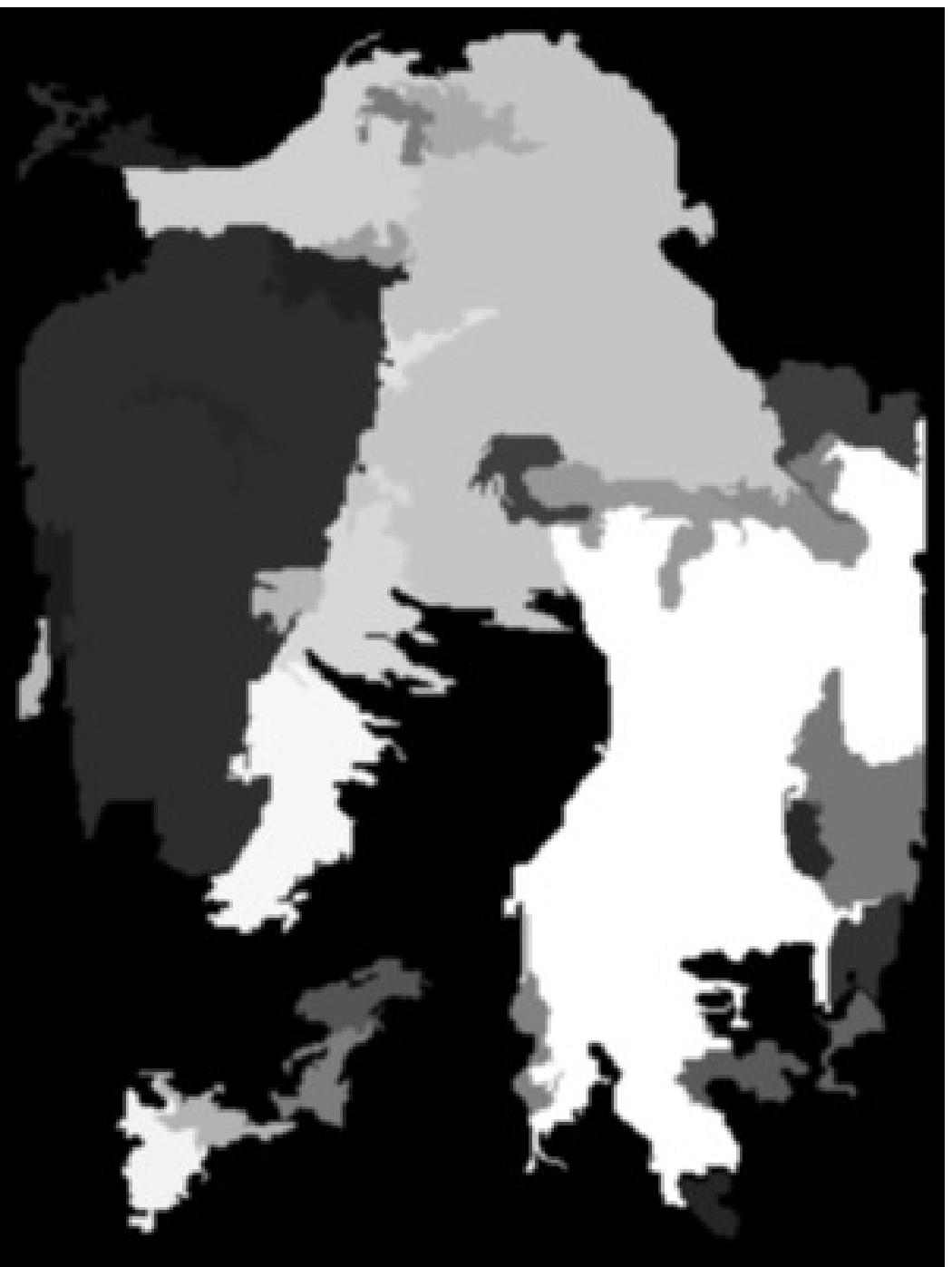} &
\includegraphics[height=1.6cm]{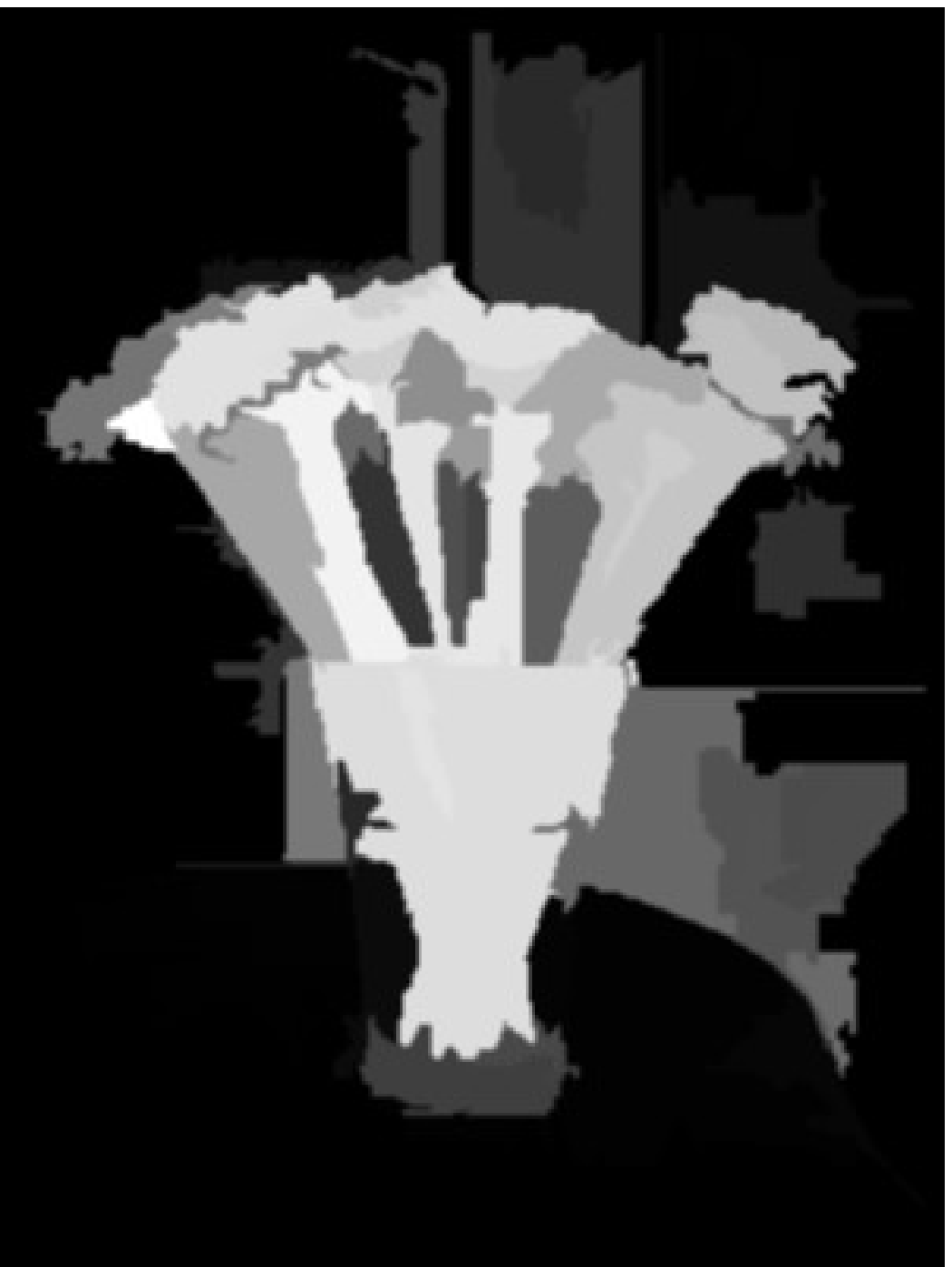} &
\includegraphics[height=1.6cm]{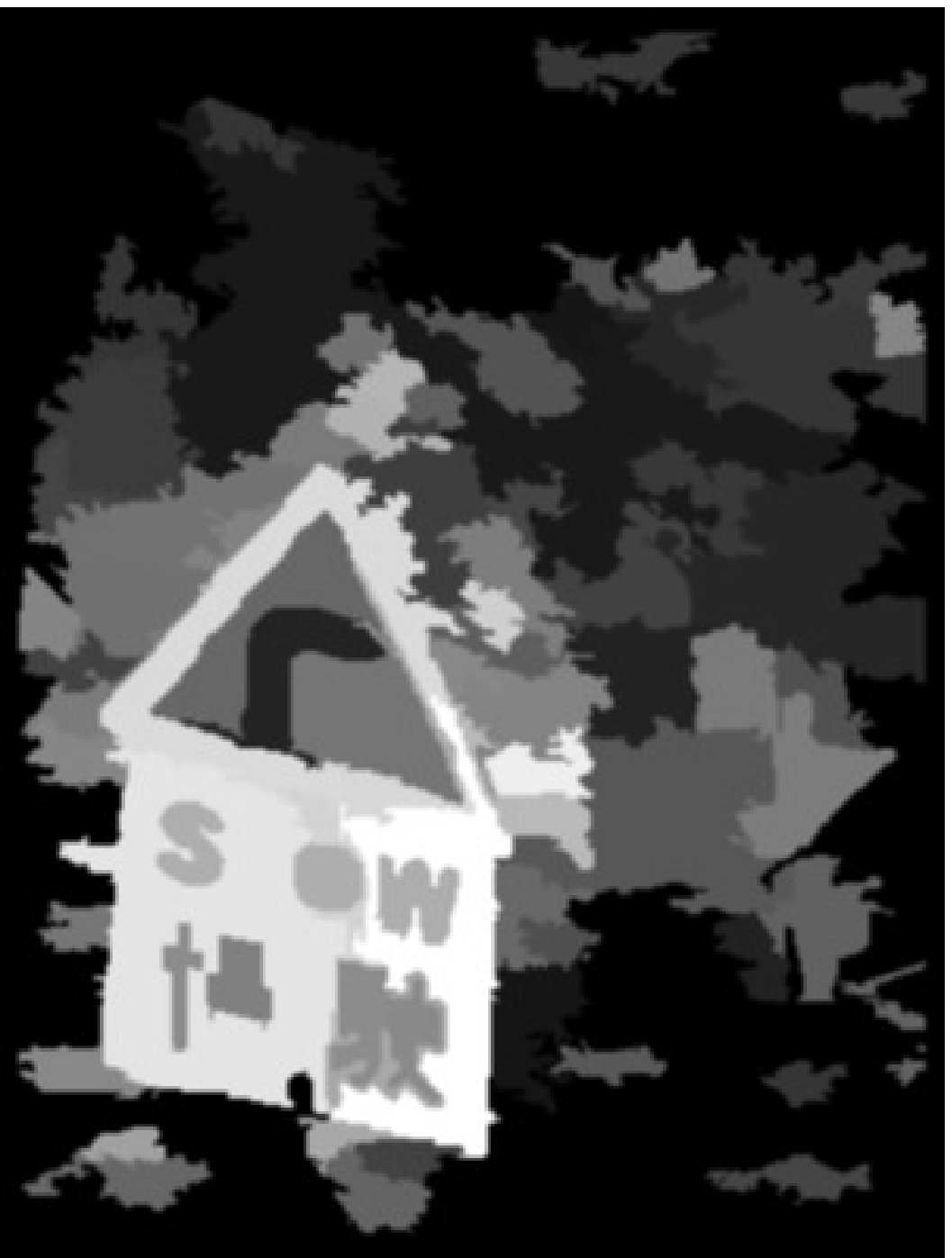} &
\includegraphics[height=1.6cm]{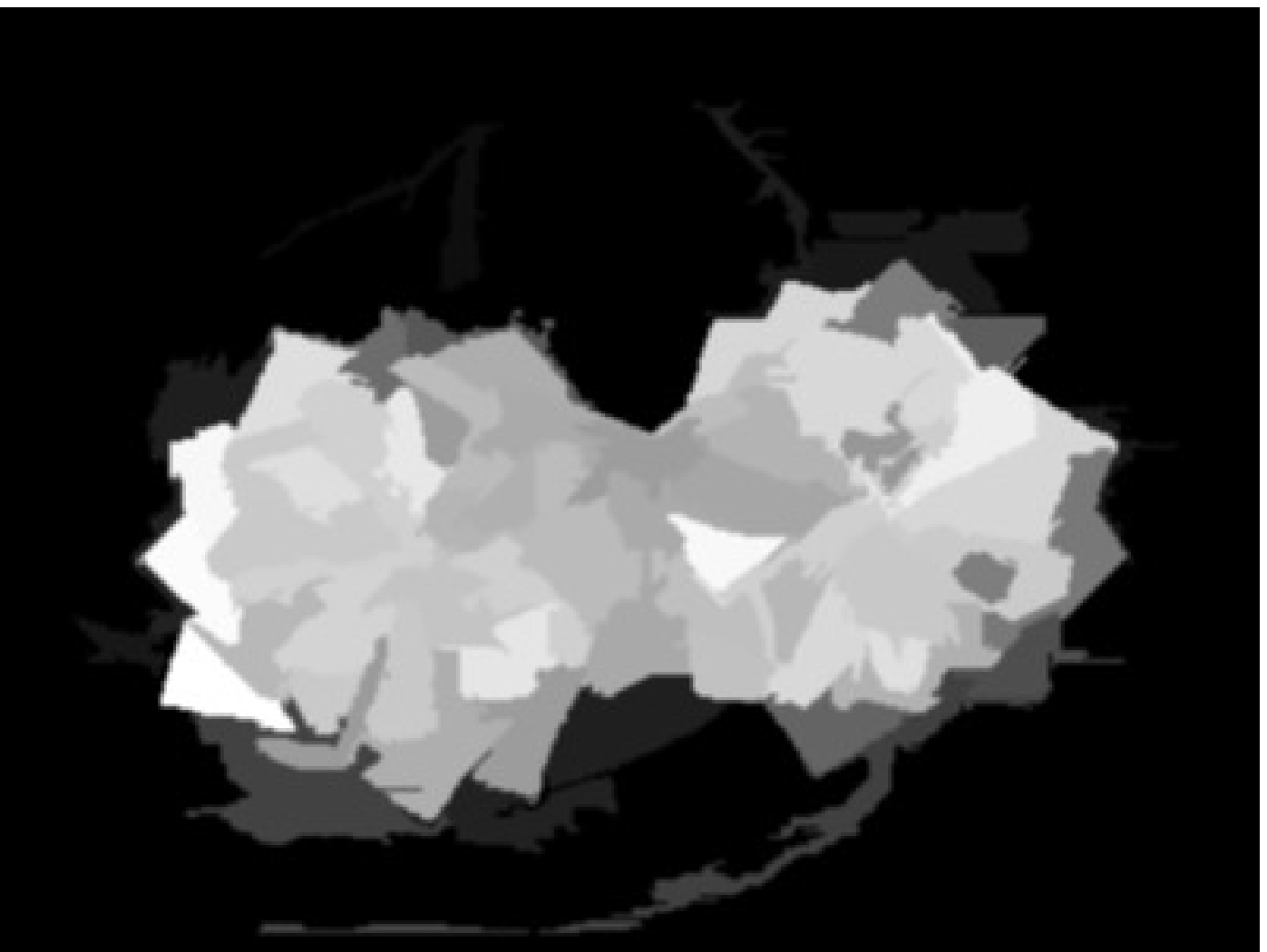} &
\includegraphics[height=1.6cm]{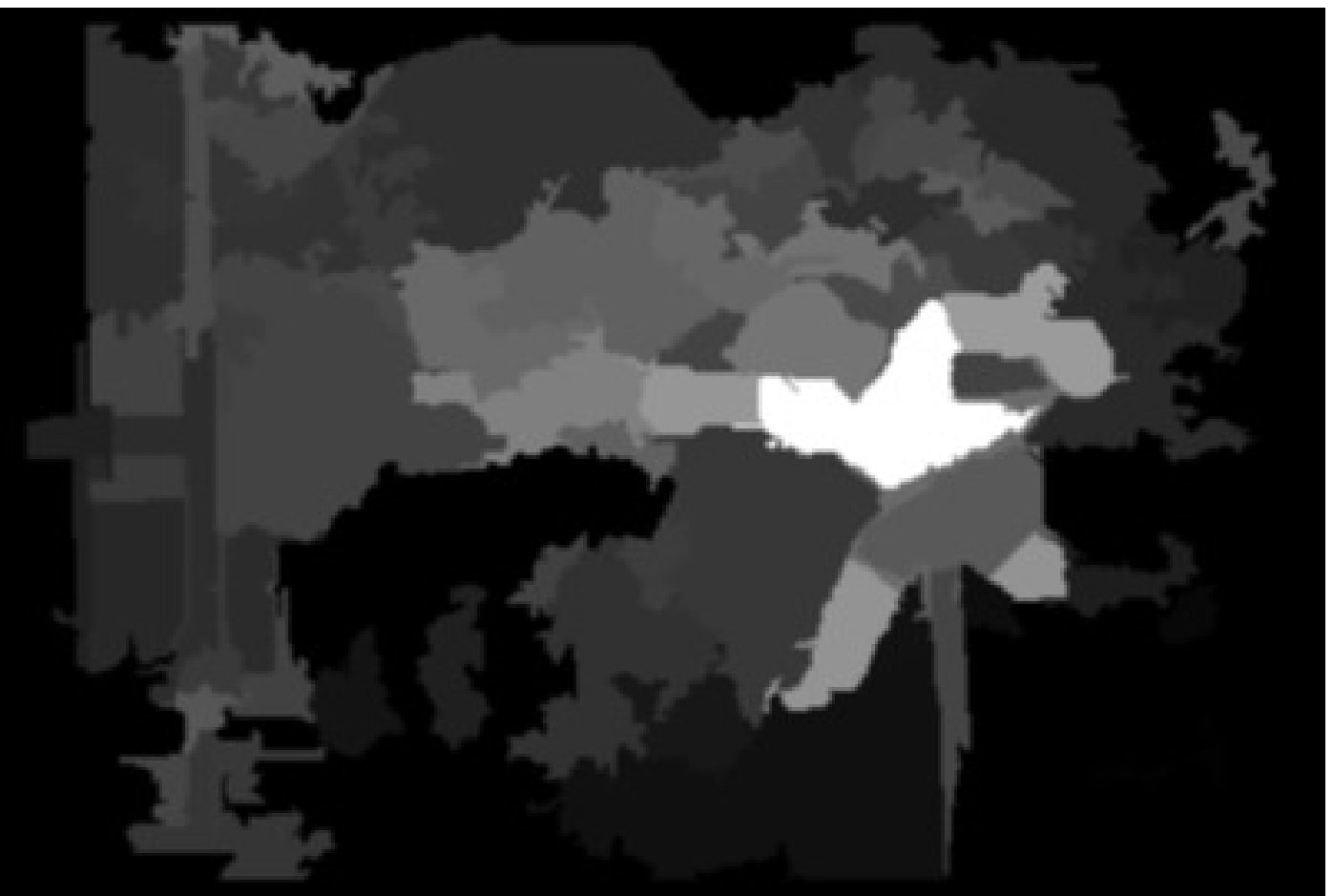} \\

SF&
\includegraphics[height=1.6cm]{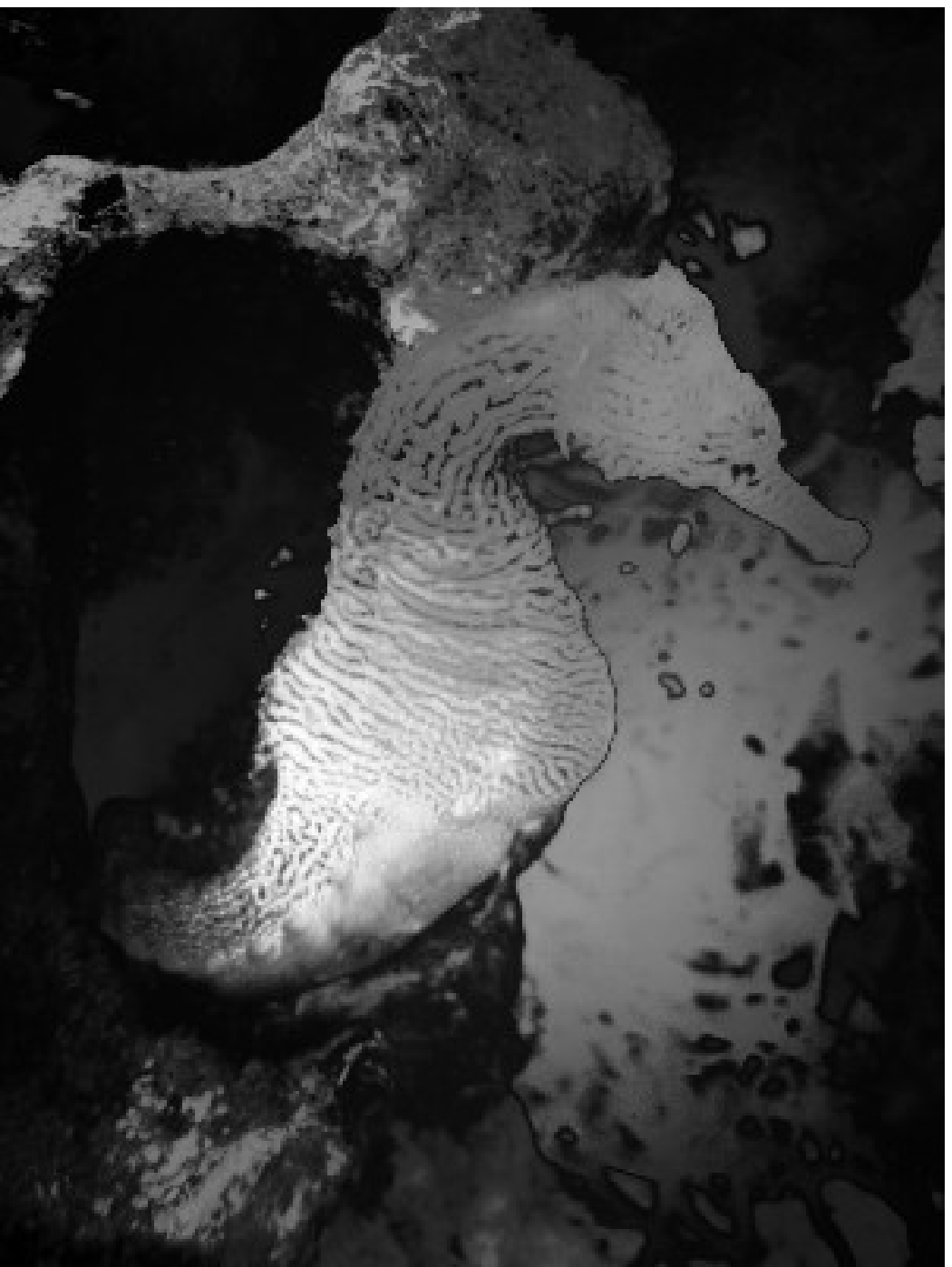} &
\includegraphics[height=1.6cm]{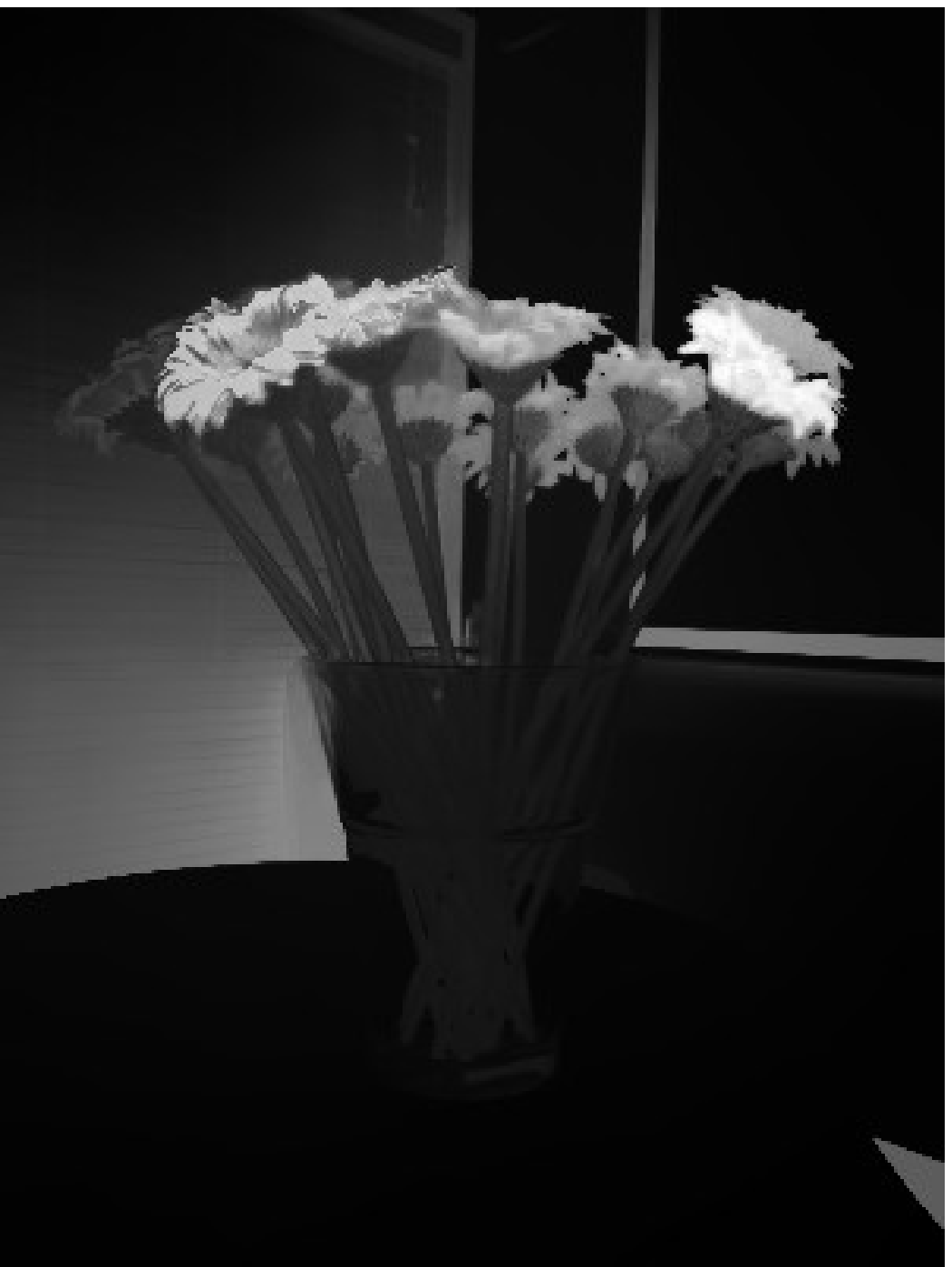} &
\includegraphics[height=1.6cm]{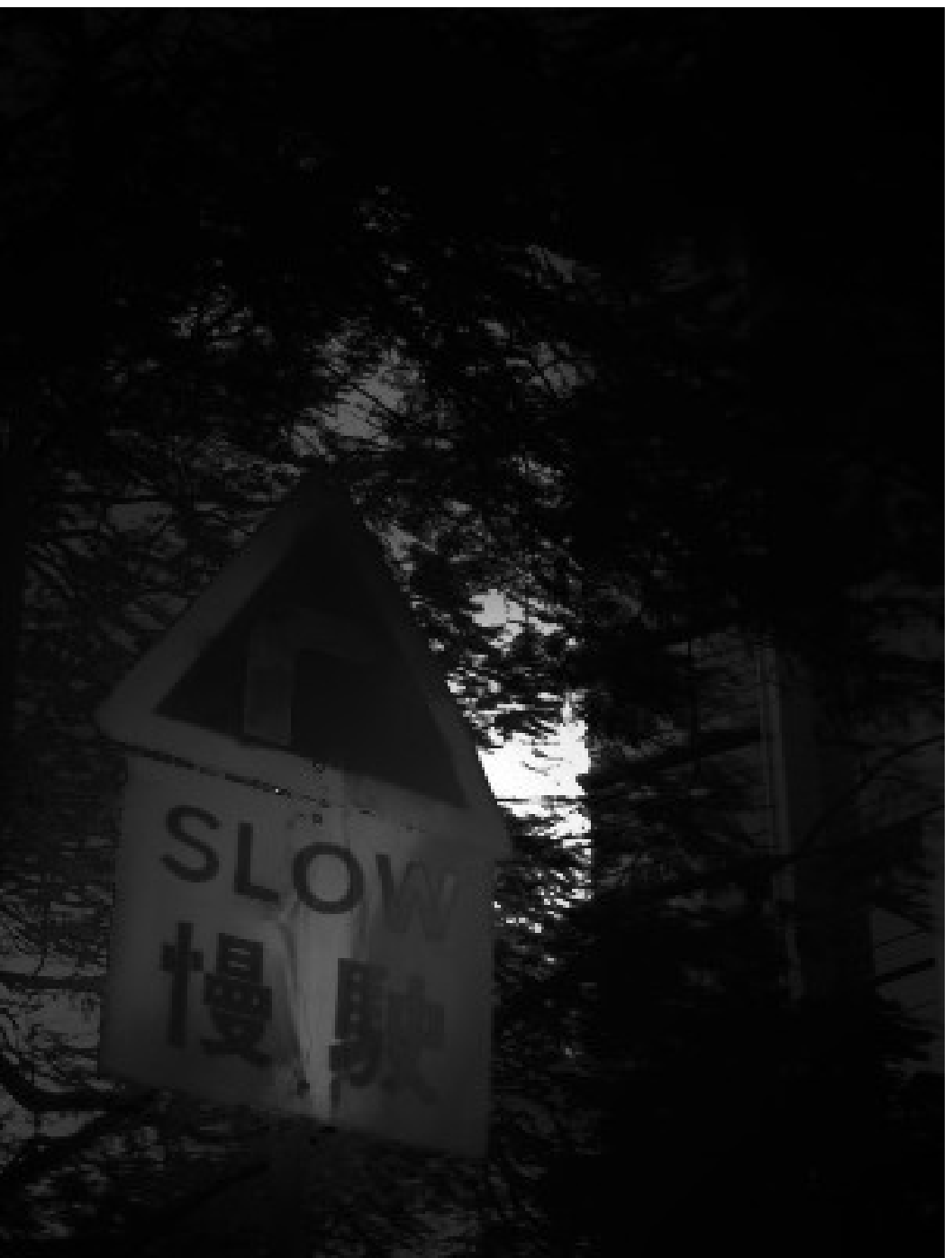} &
\includegraphics[height=1.6cm]{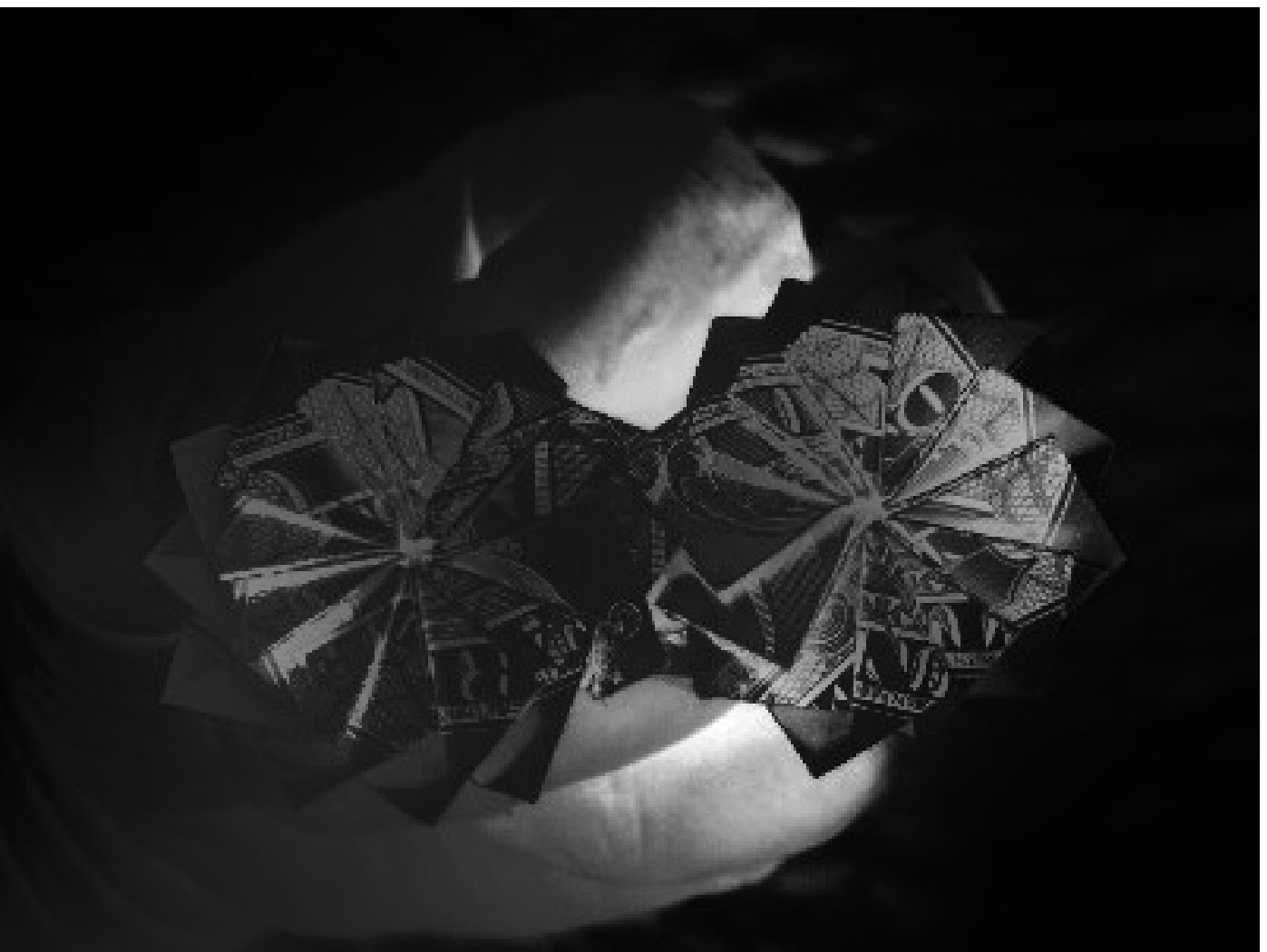} &
\includegraphics[height=1.6cm]{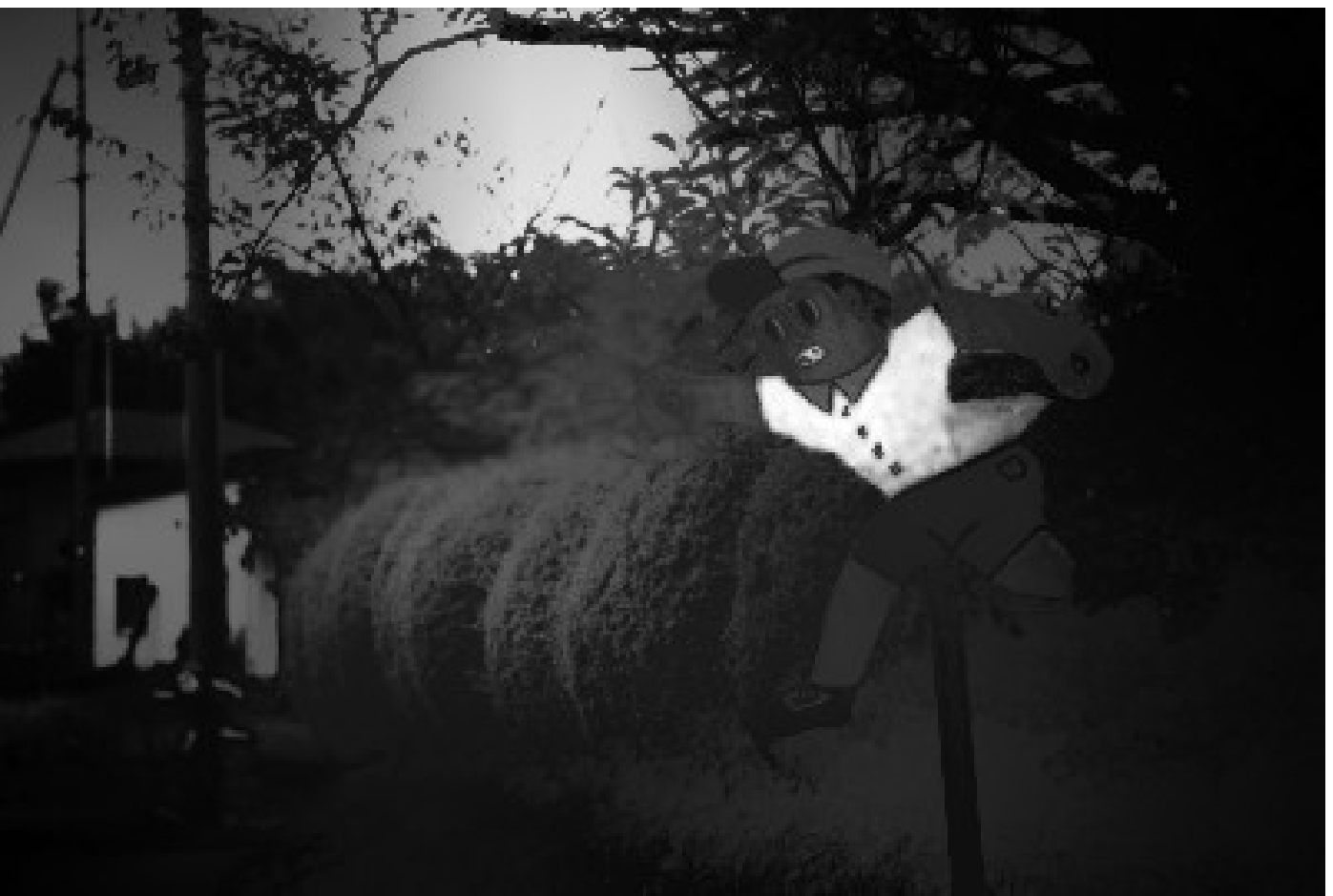} \\

PCAS&
\includegraphics[height=1.6cm]{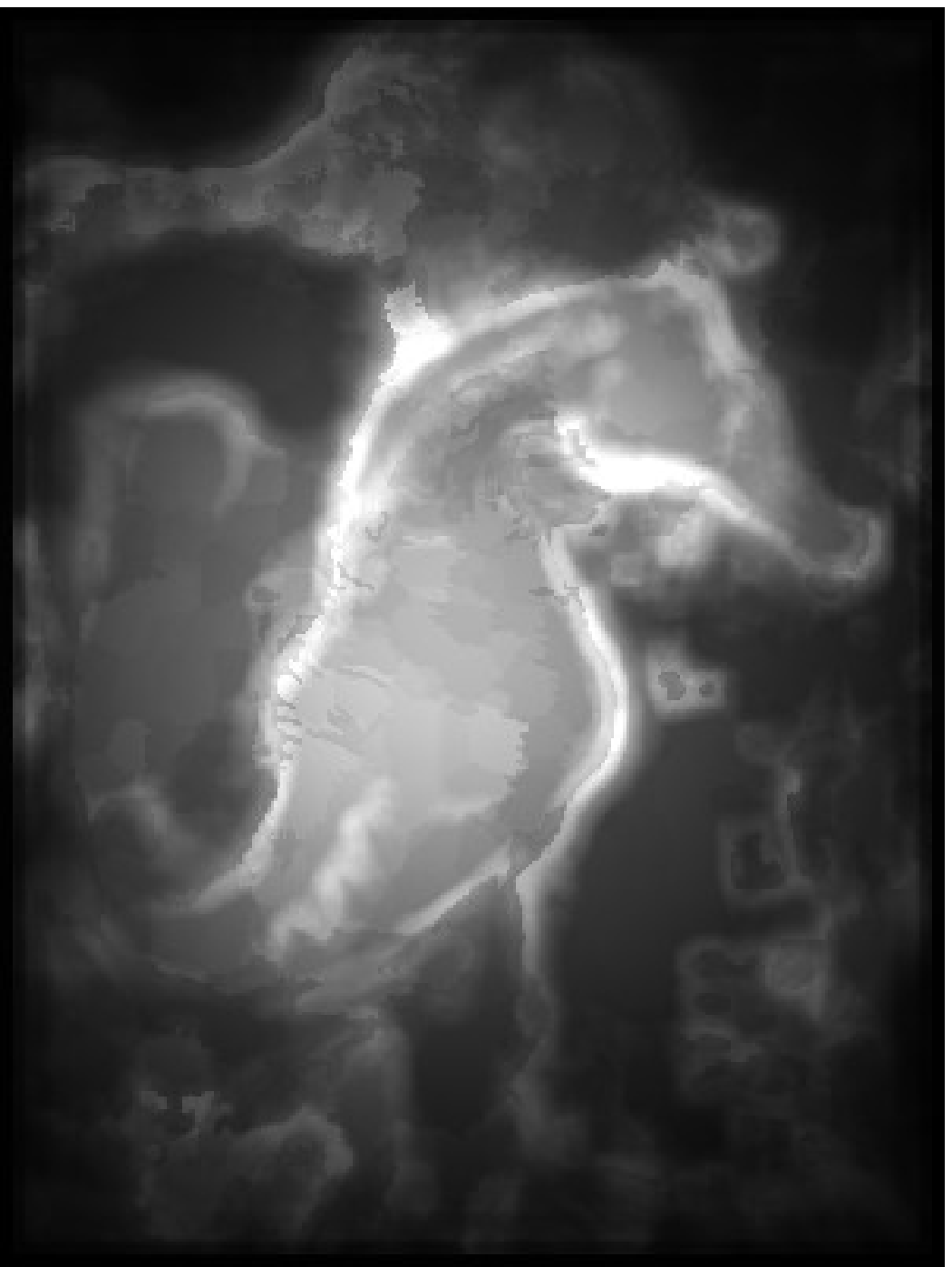} &
\includegraphics[height=1.6cm]{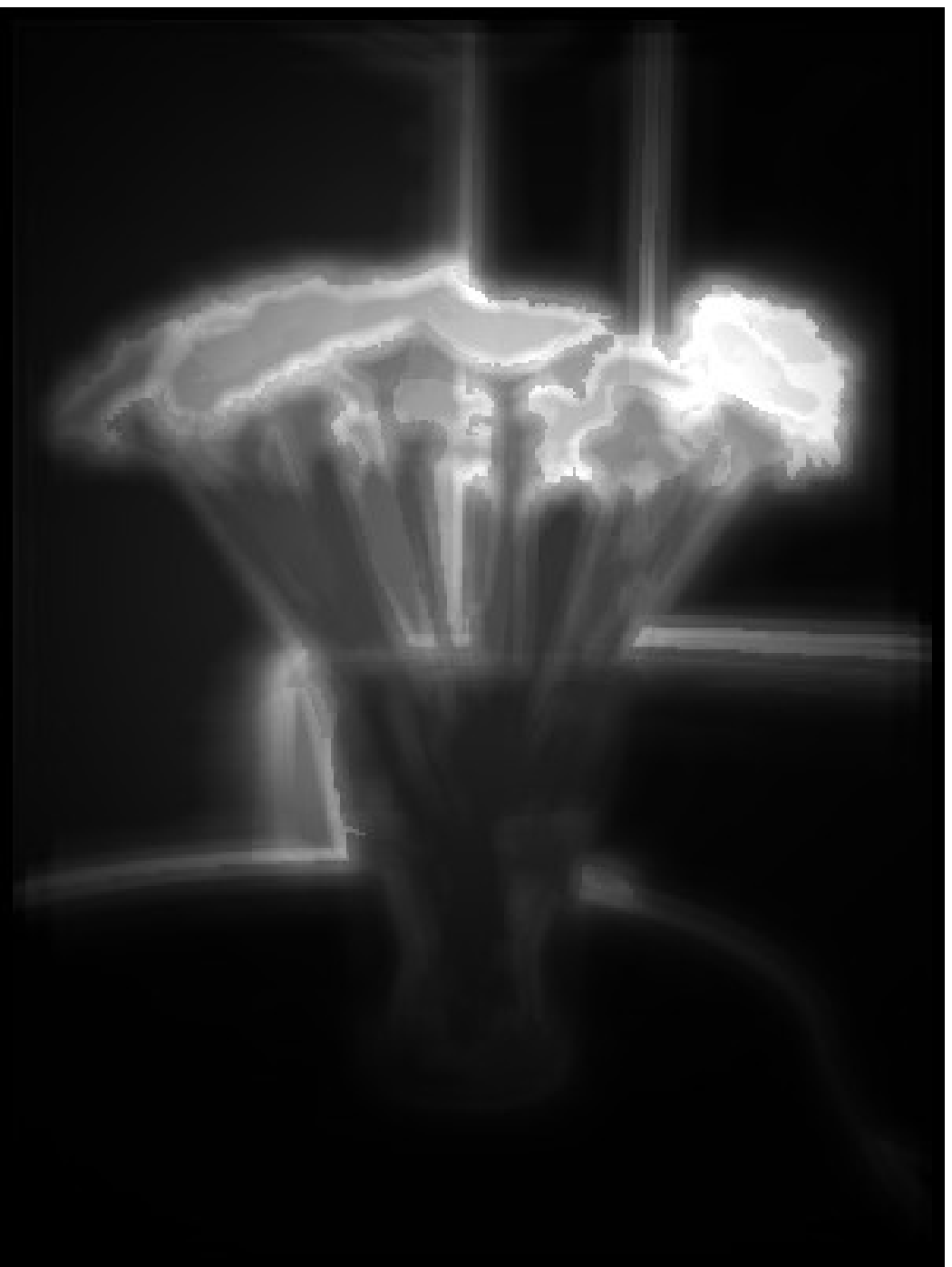} &
\includegraphics[height=1.6cm]{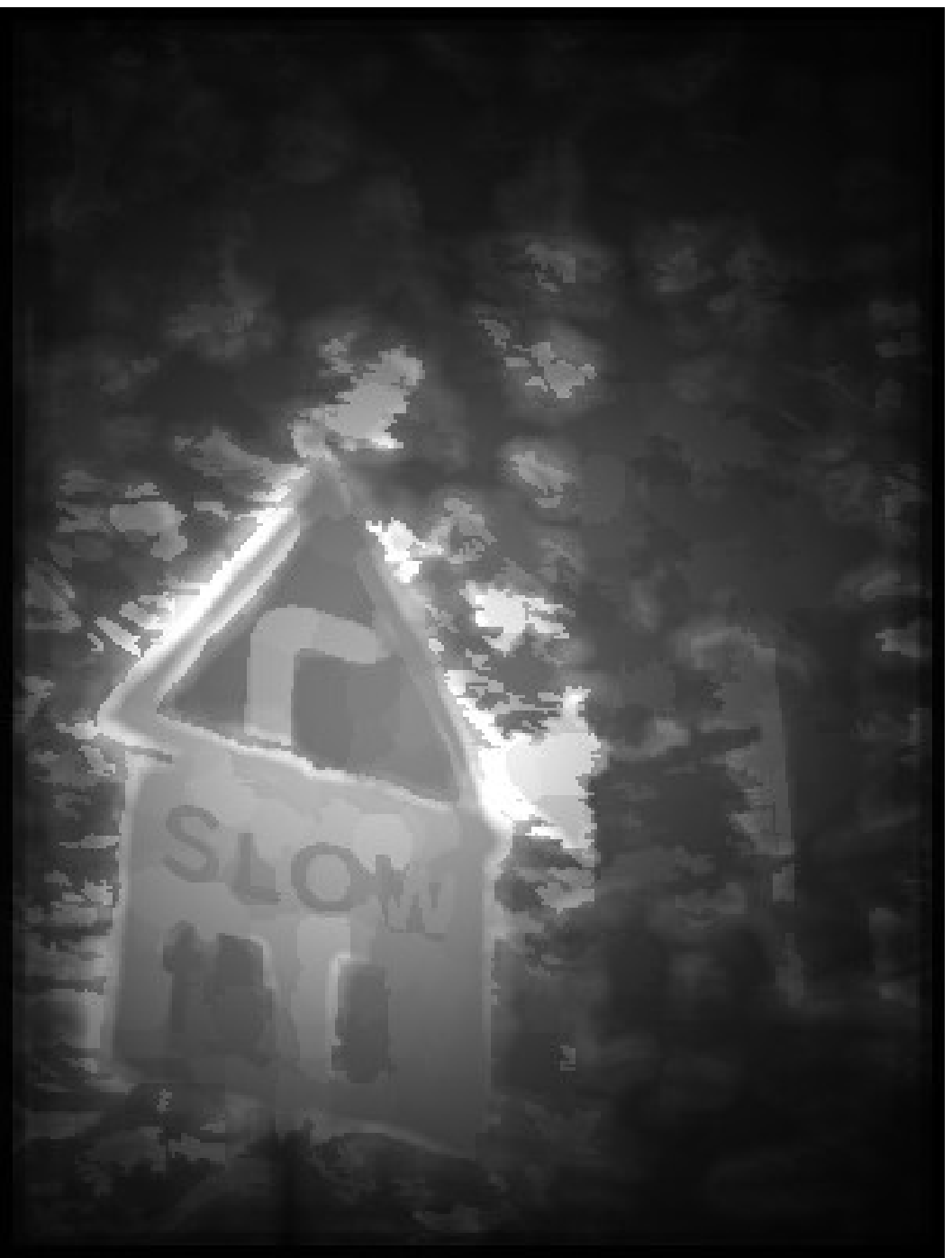} &
\includegraphics[height=1.6cm]{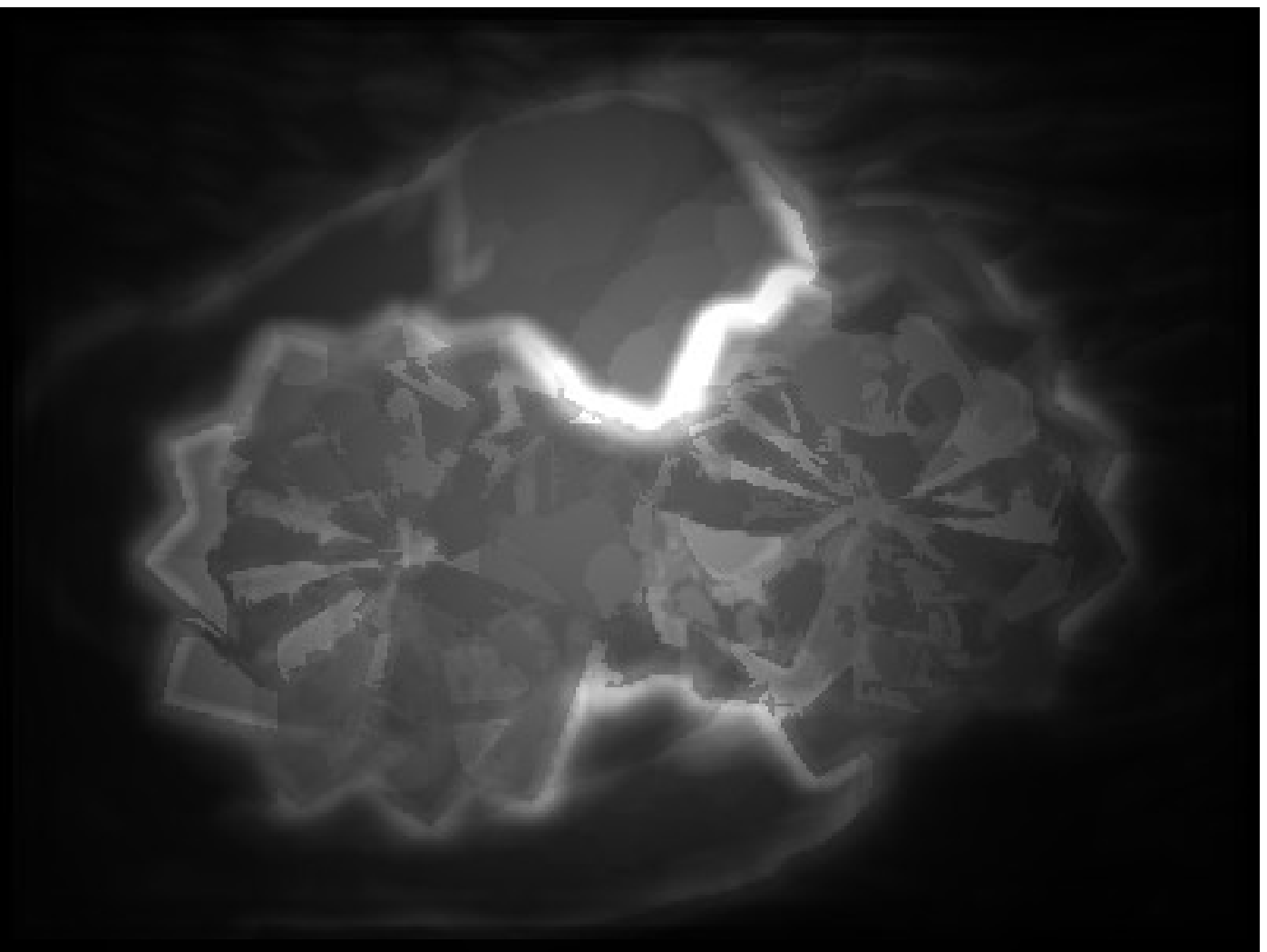} &
\includegraphics[height=1.6cm]{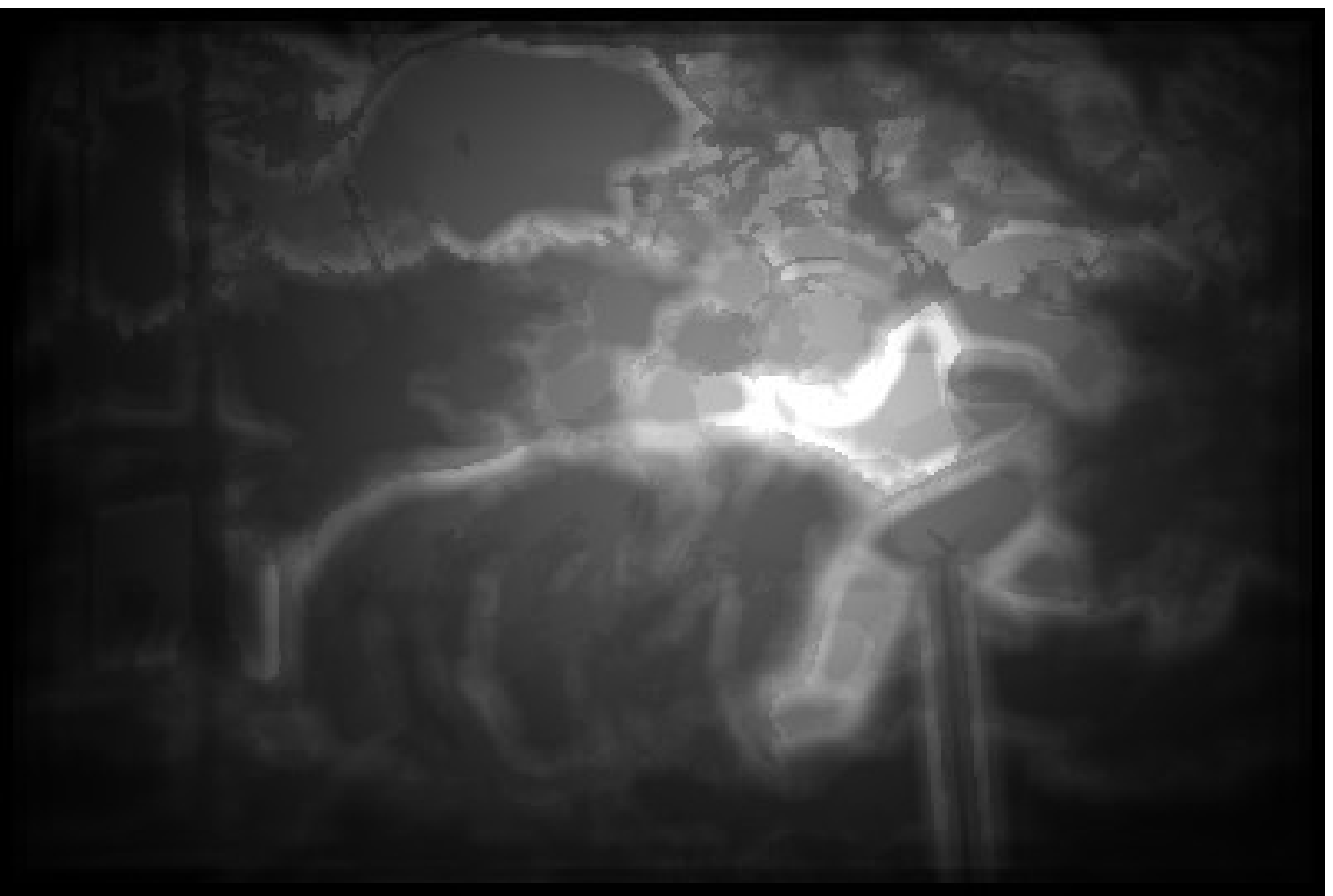} \\

HS&
\includegraphics[height=1.6cm]{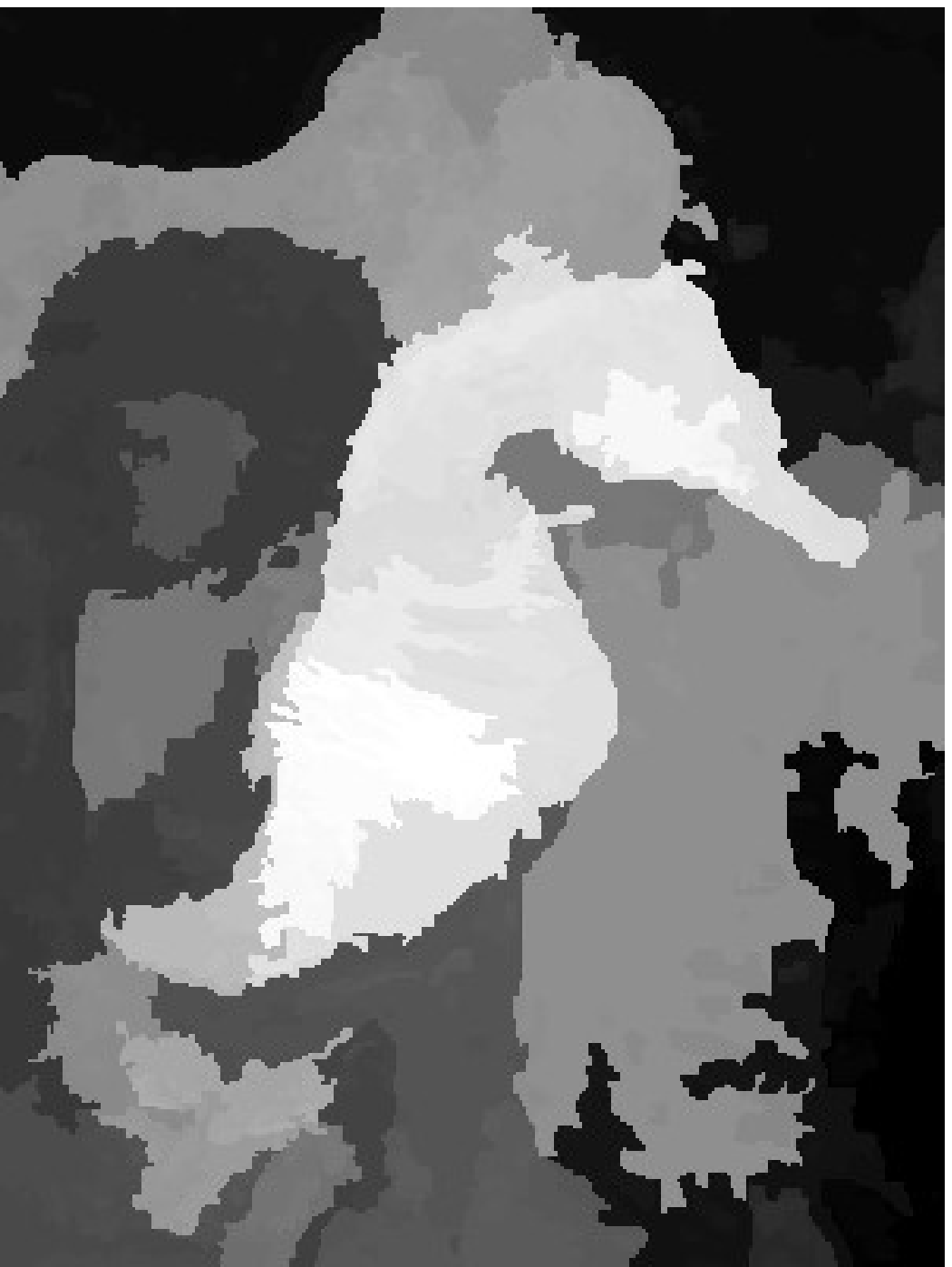} &
\includegraphics[height=1.6cm]{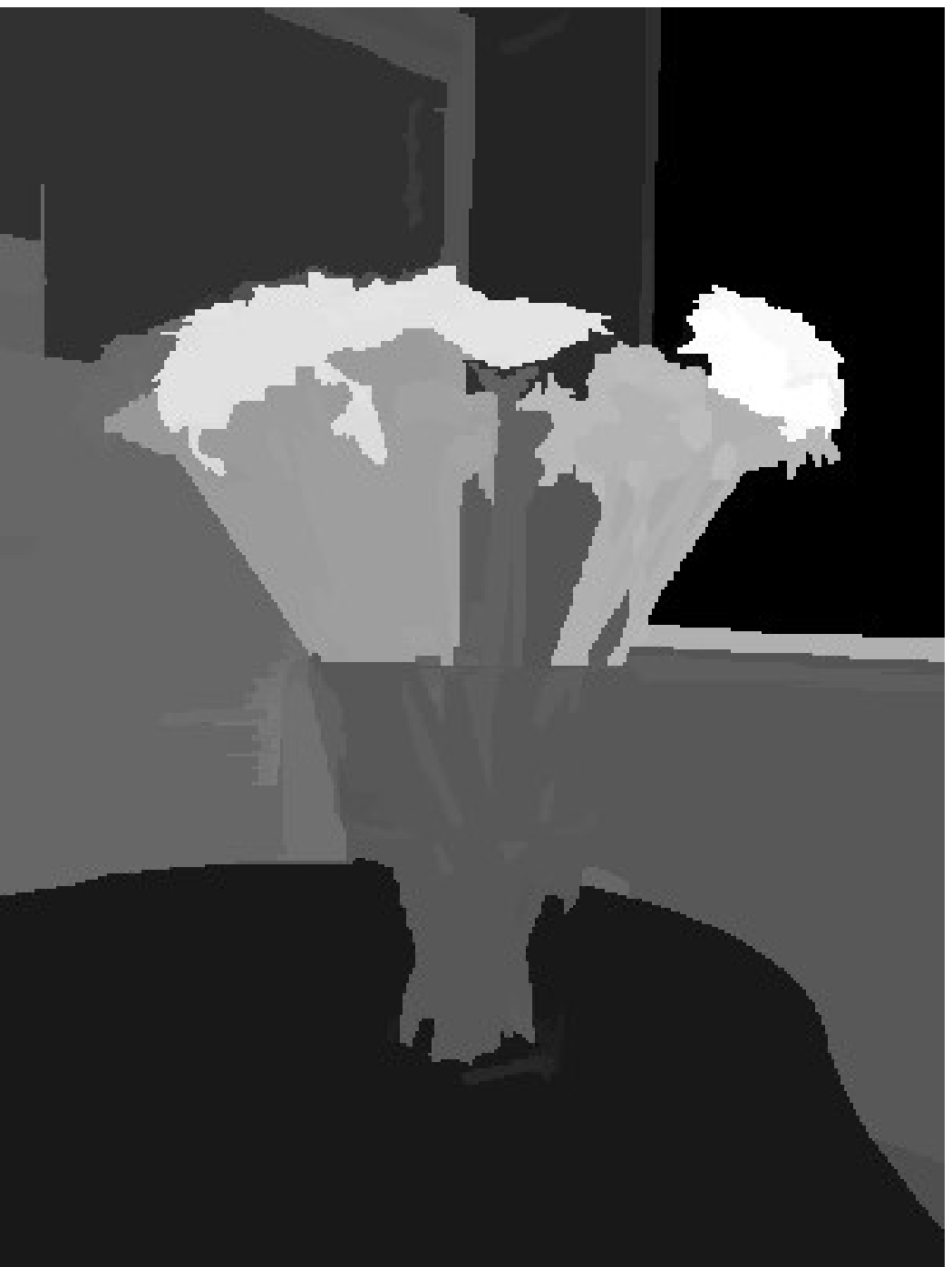} &
\includegraphics[height=1.6cm]{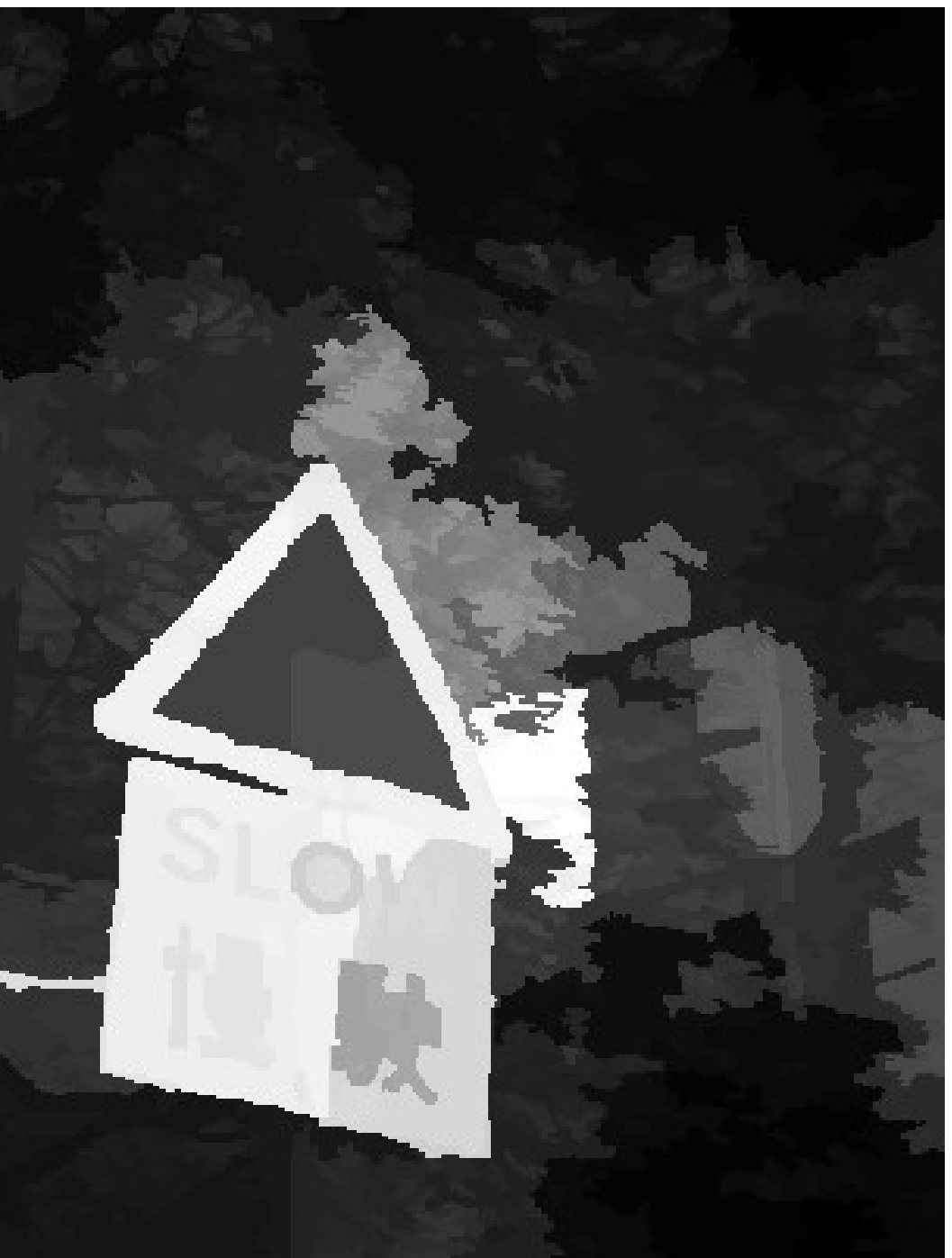} &
\includegraphics[height=1.6cm]{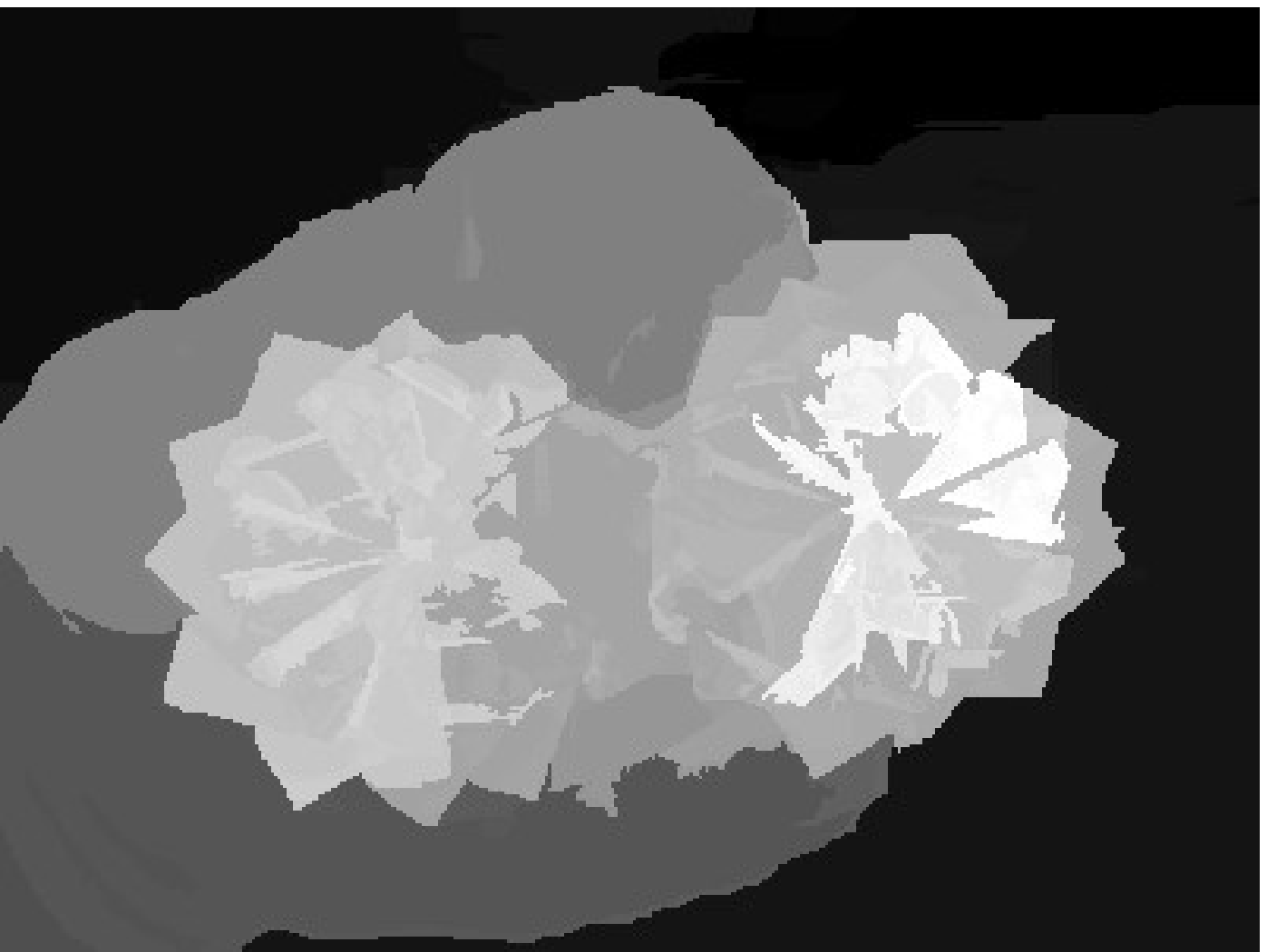} &
\includegraphics[height=1.6cm]{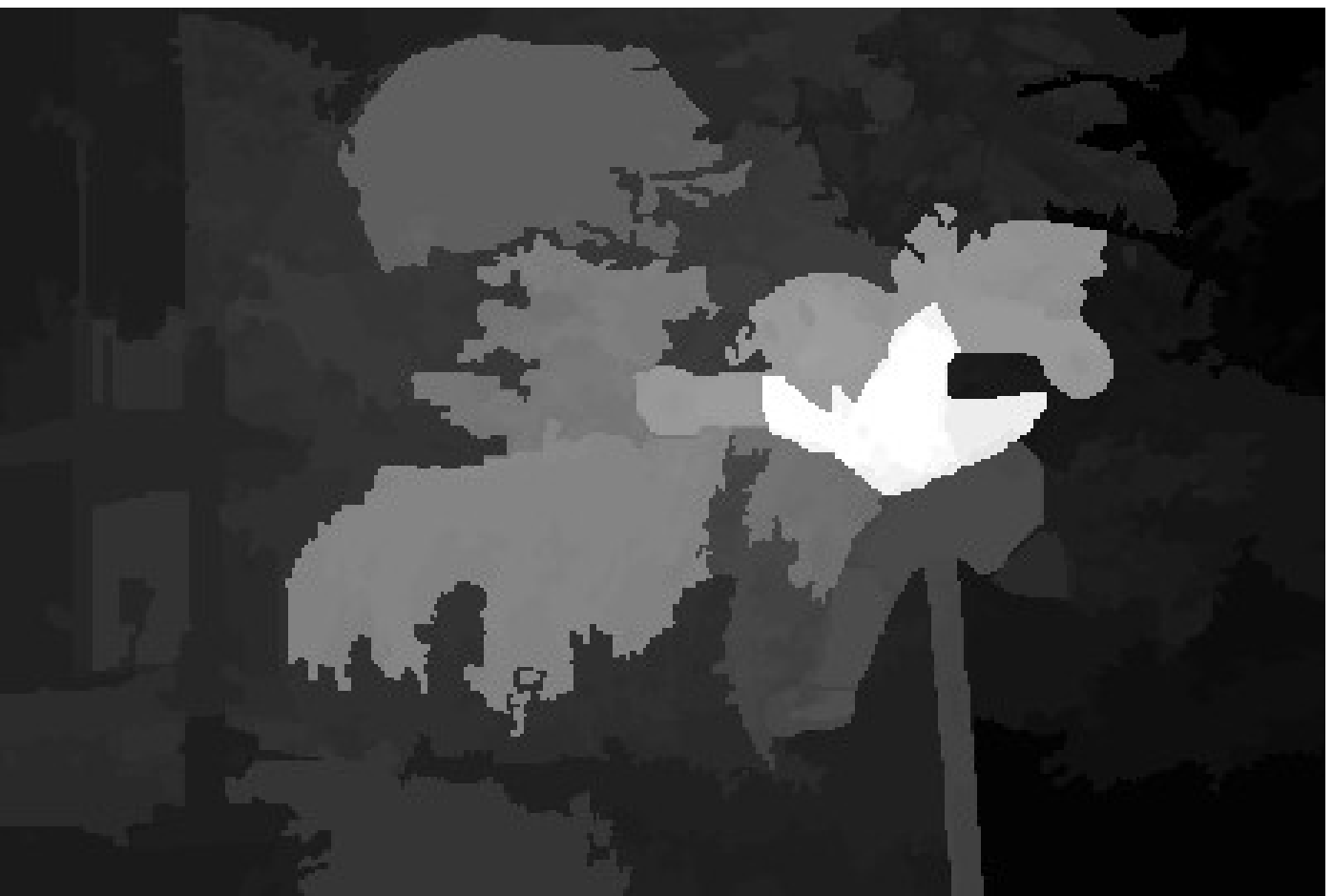} \\

ST&
\includegraphics[height=1.6cm]{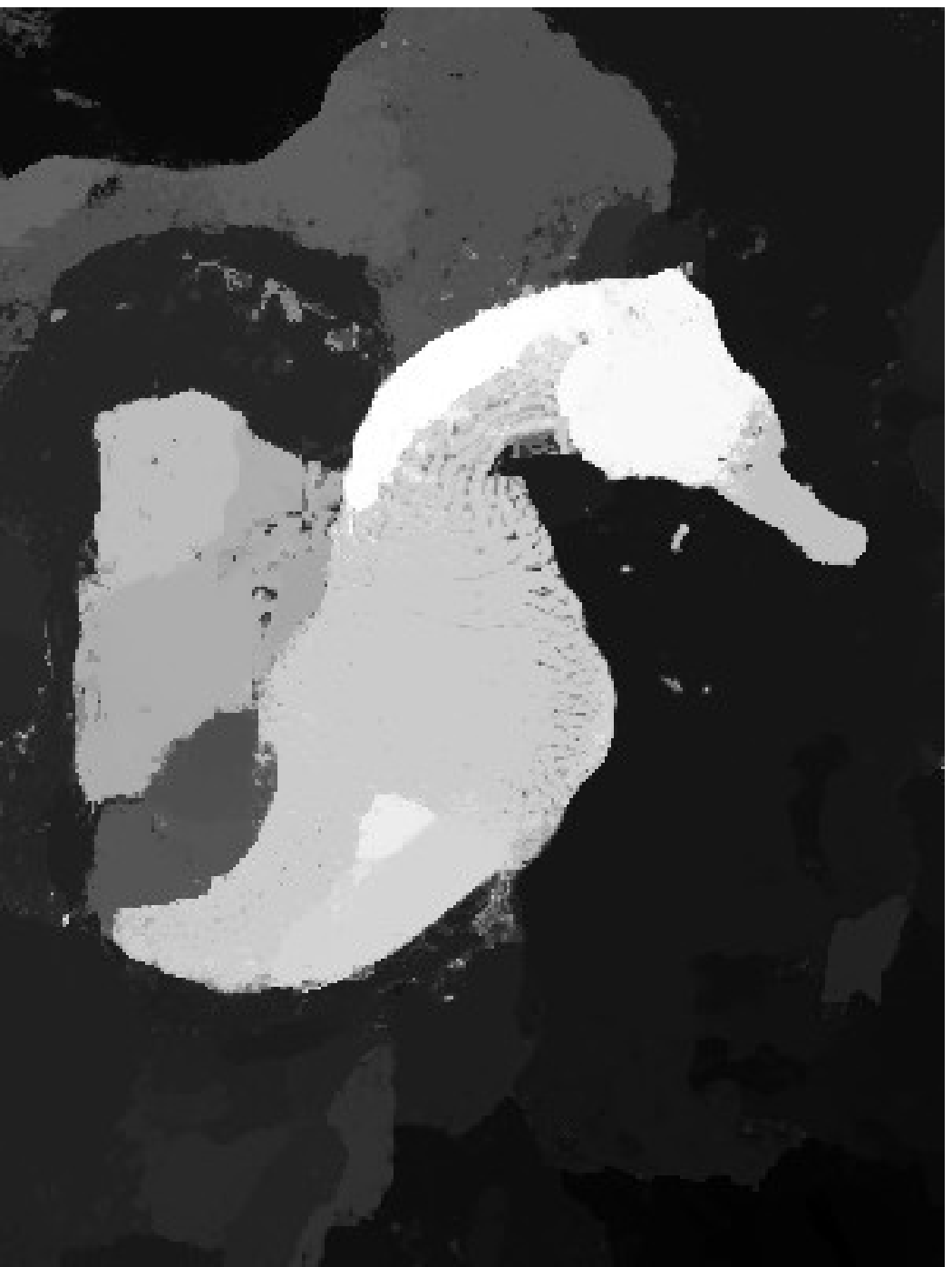} &
\includegraphics[height=1.6cm]{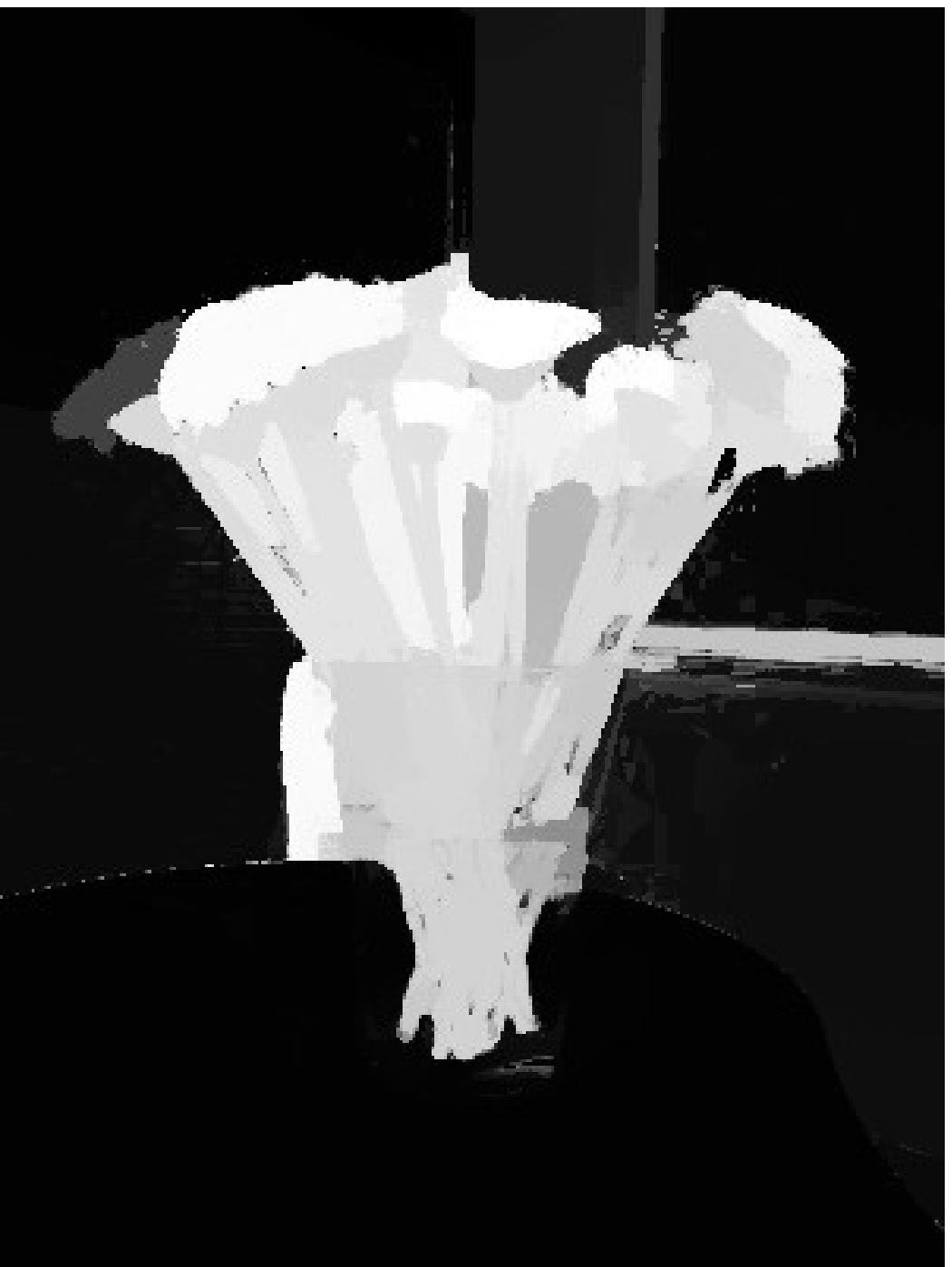} &
\includegraphics[height=1.6cm]{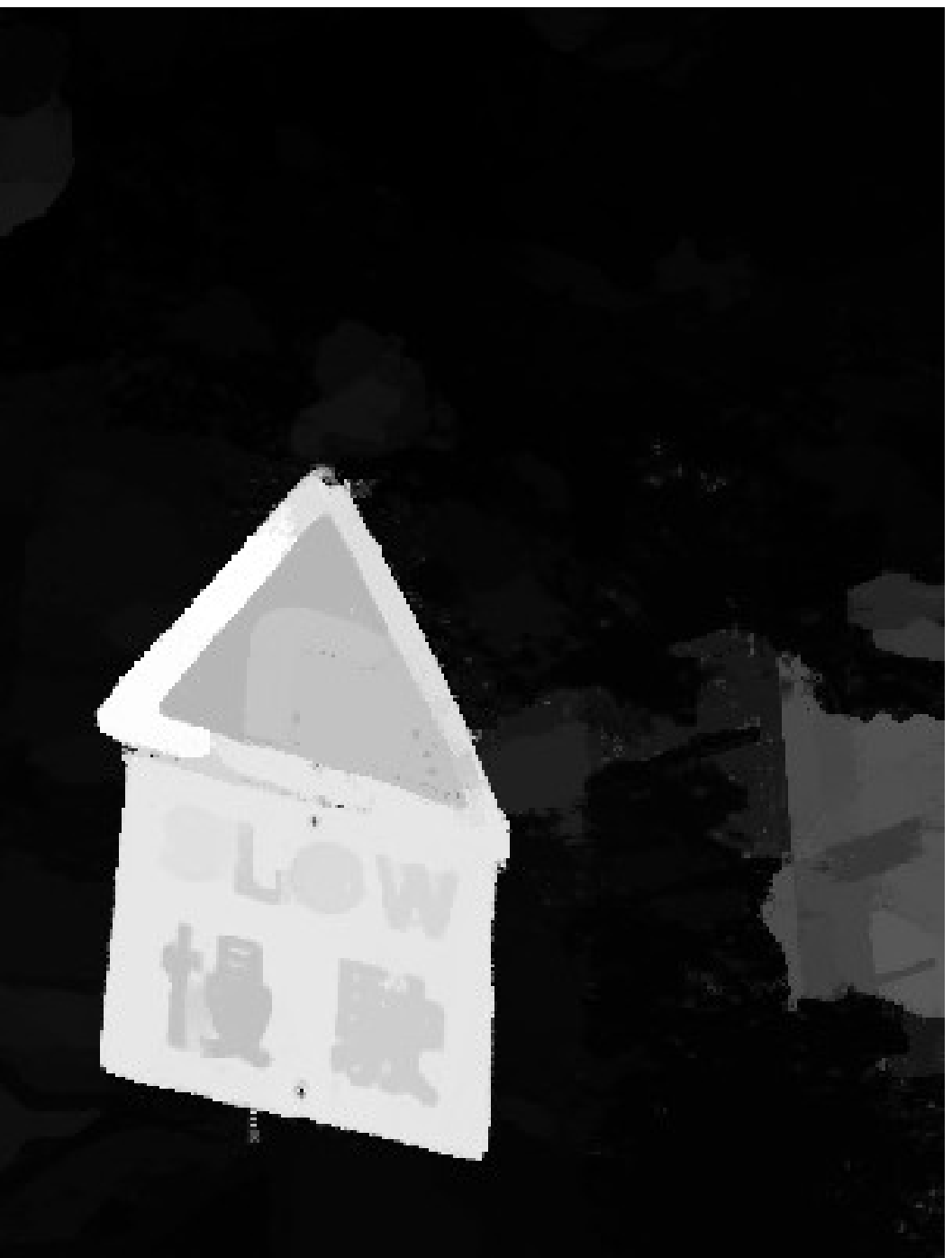} &
\includegraphics[height=1.6cm]{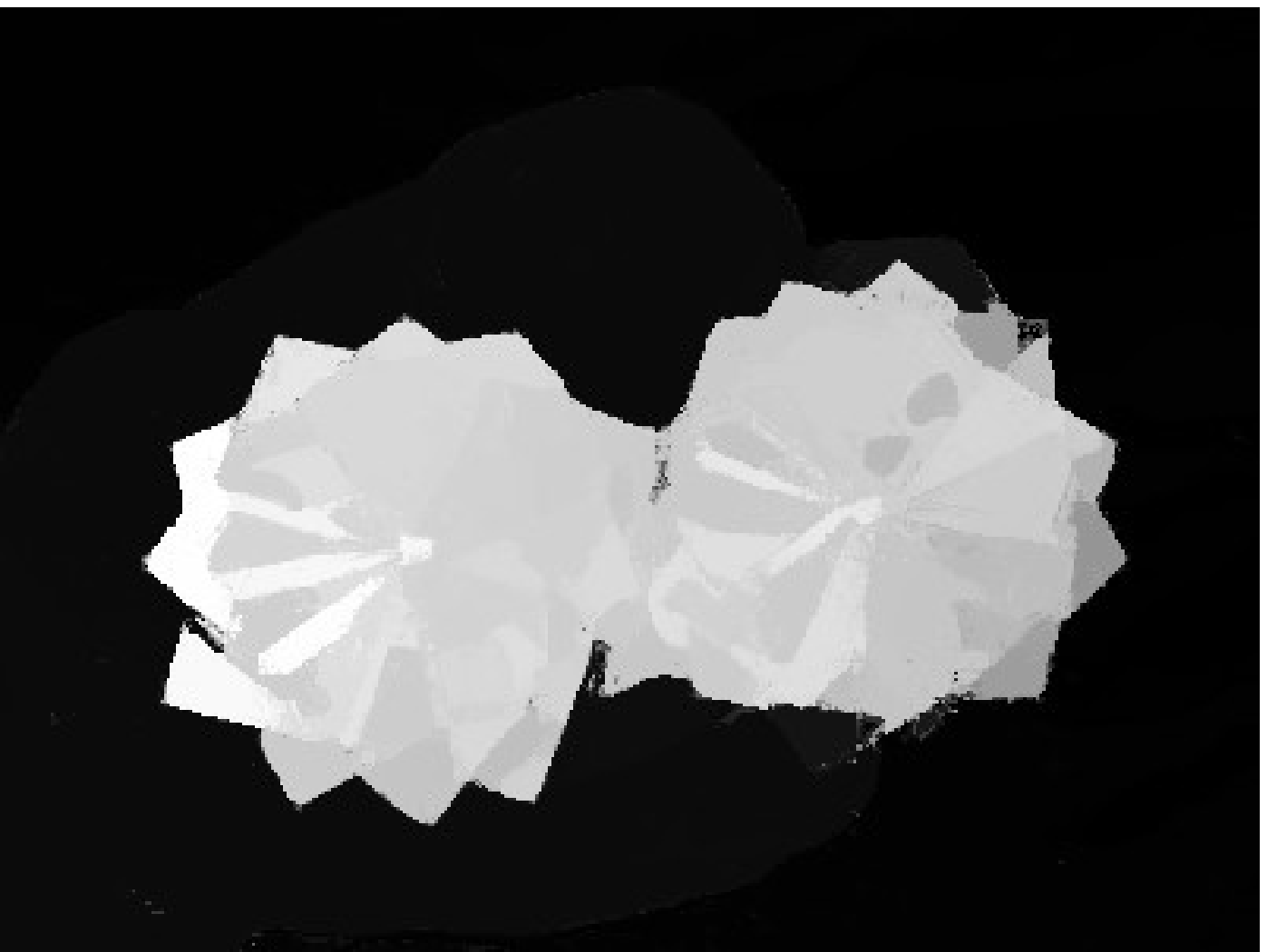} &
\includegraphics[height=1.6cm]{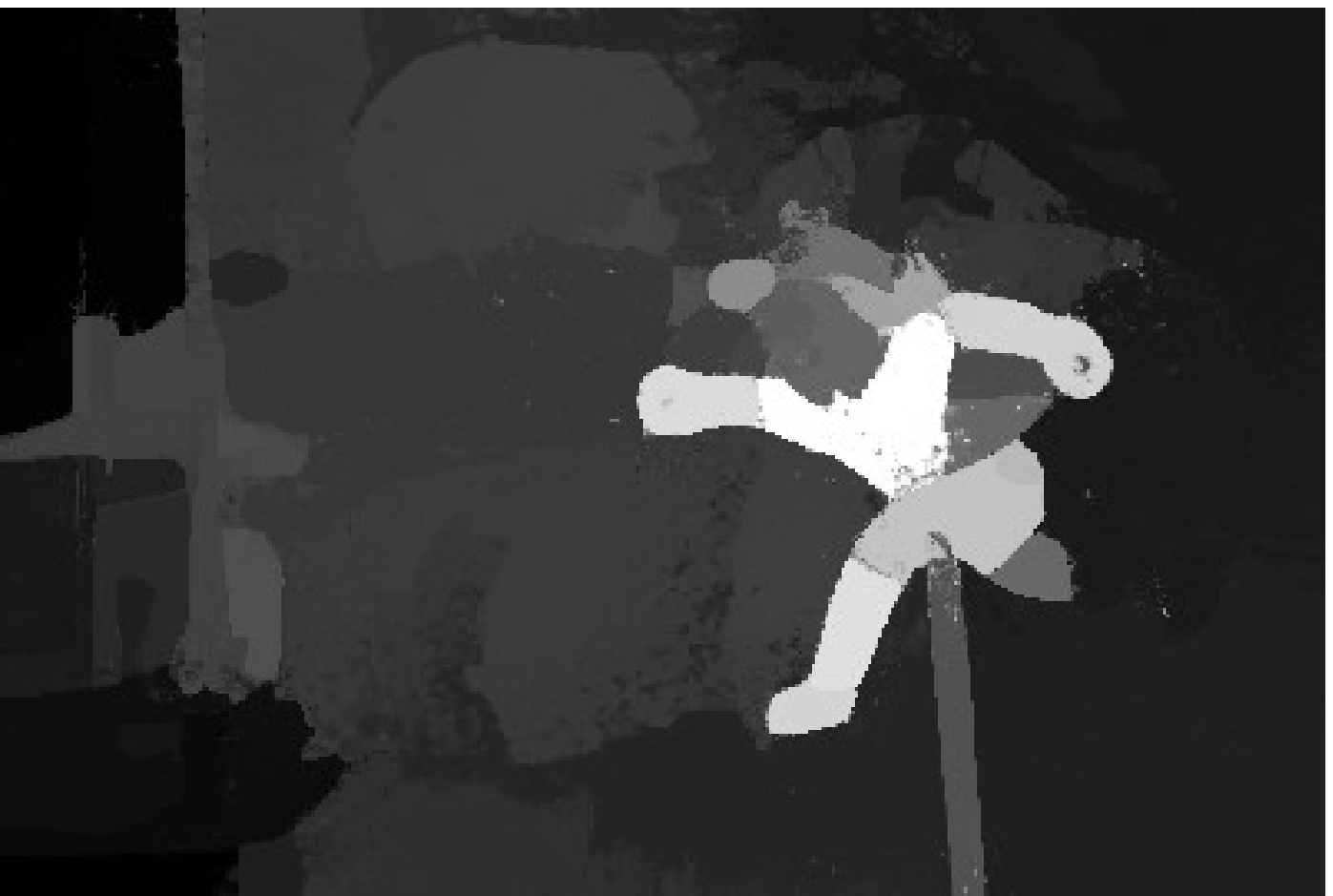} \\

HP&
\includegraphics[height=1.6cm]{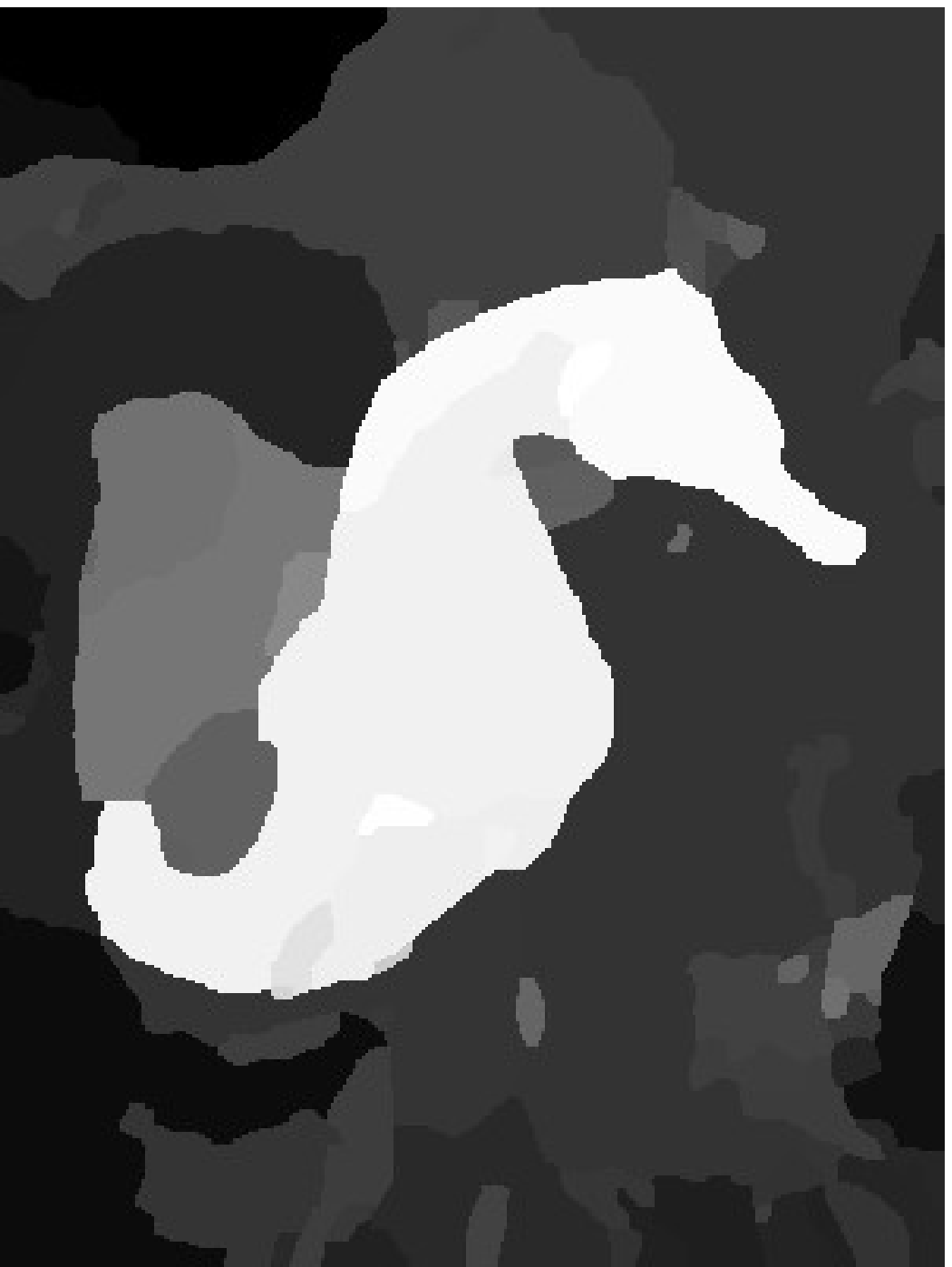} &
\includegraphics[height=1.6cm]{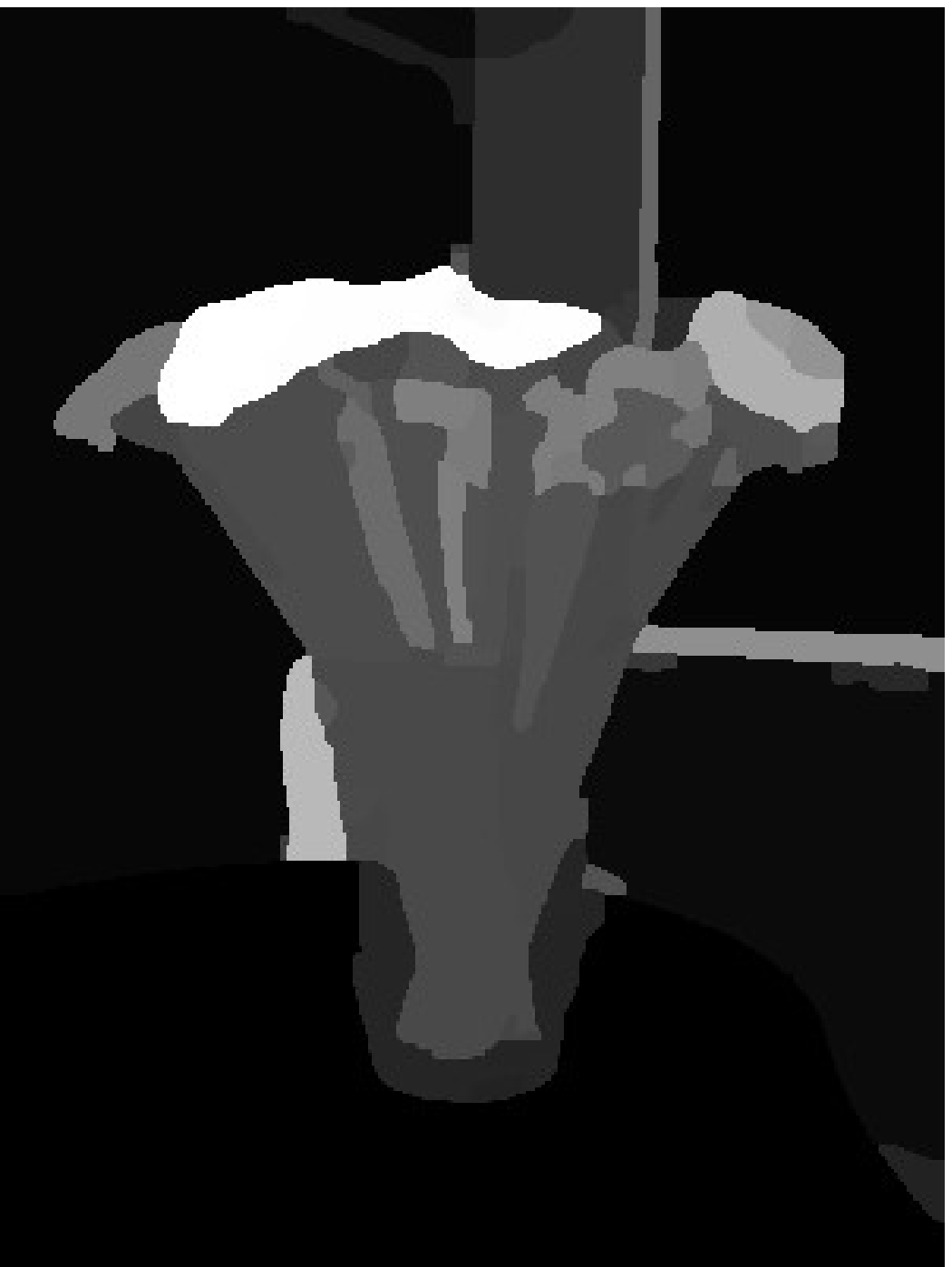} &
\includegraphics[height=1.6cm]{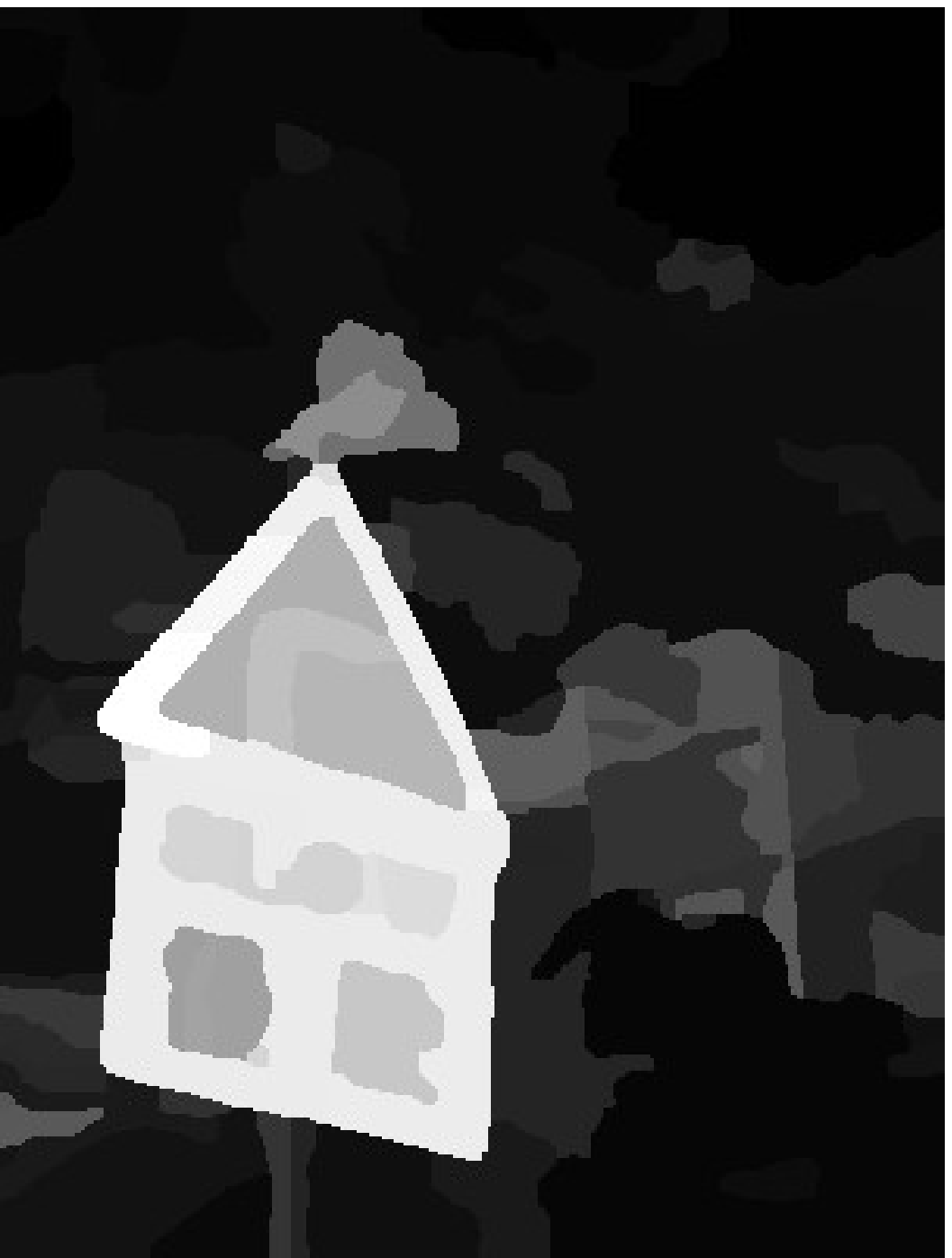} &
\includegraphics[height=1.6cm]{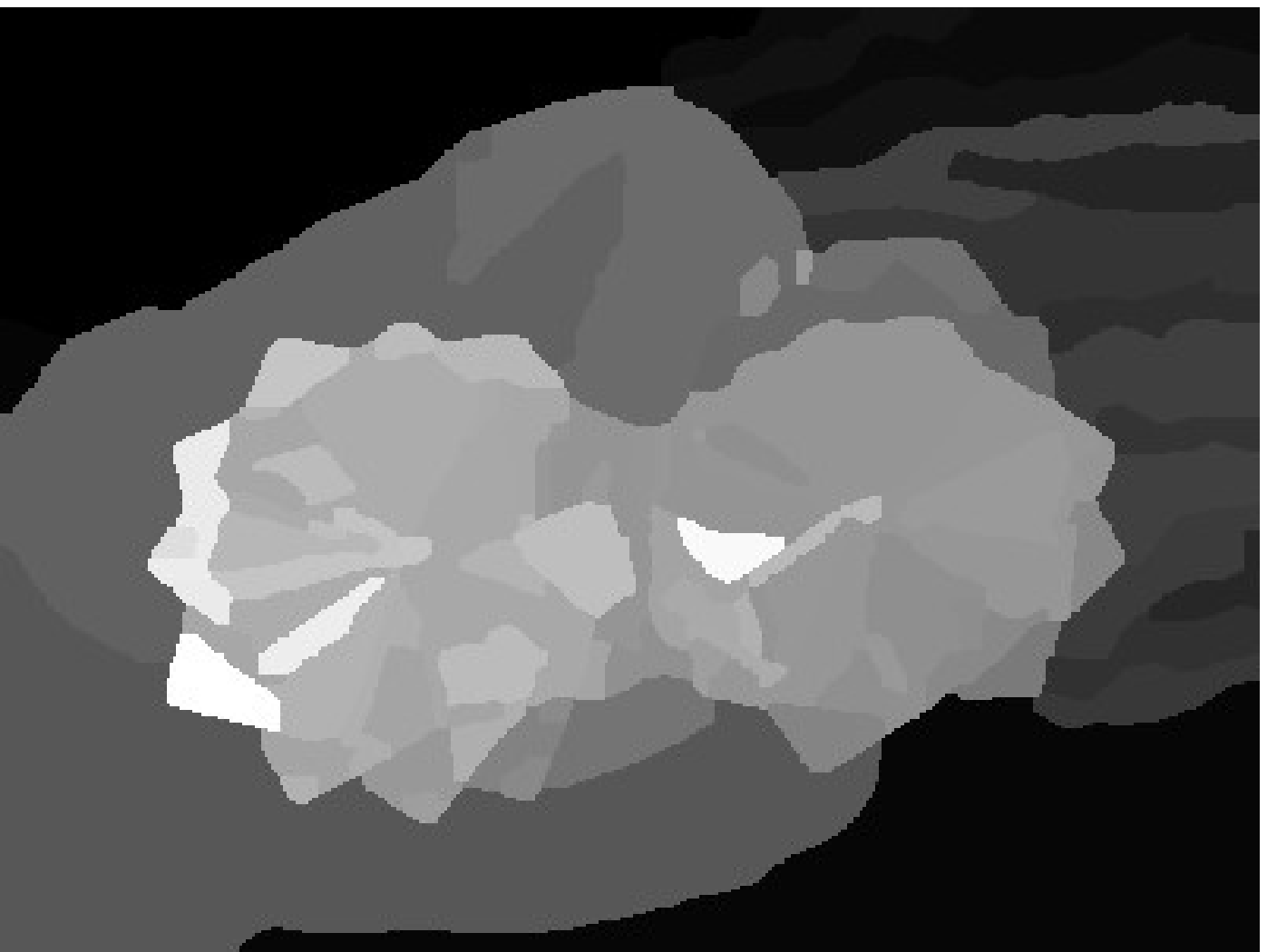} &
\includegraphics[height=1.6cm]{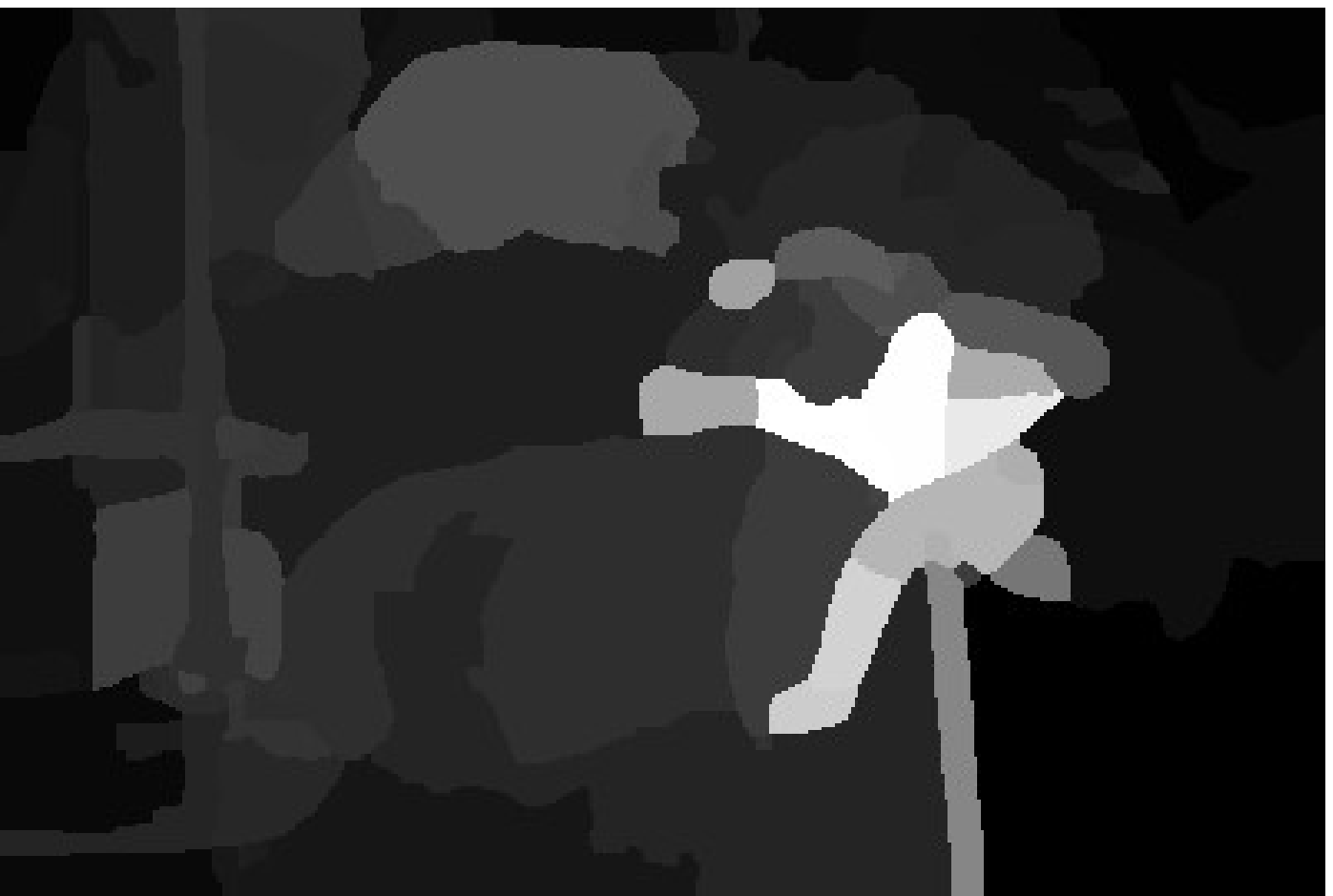} \\

SOH&
\includegraphics[height=1.6cm]{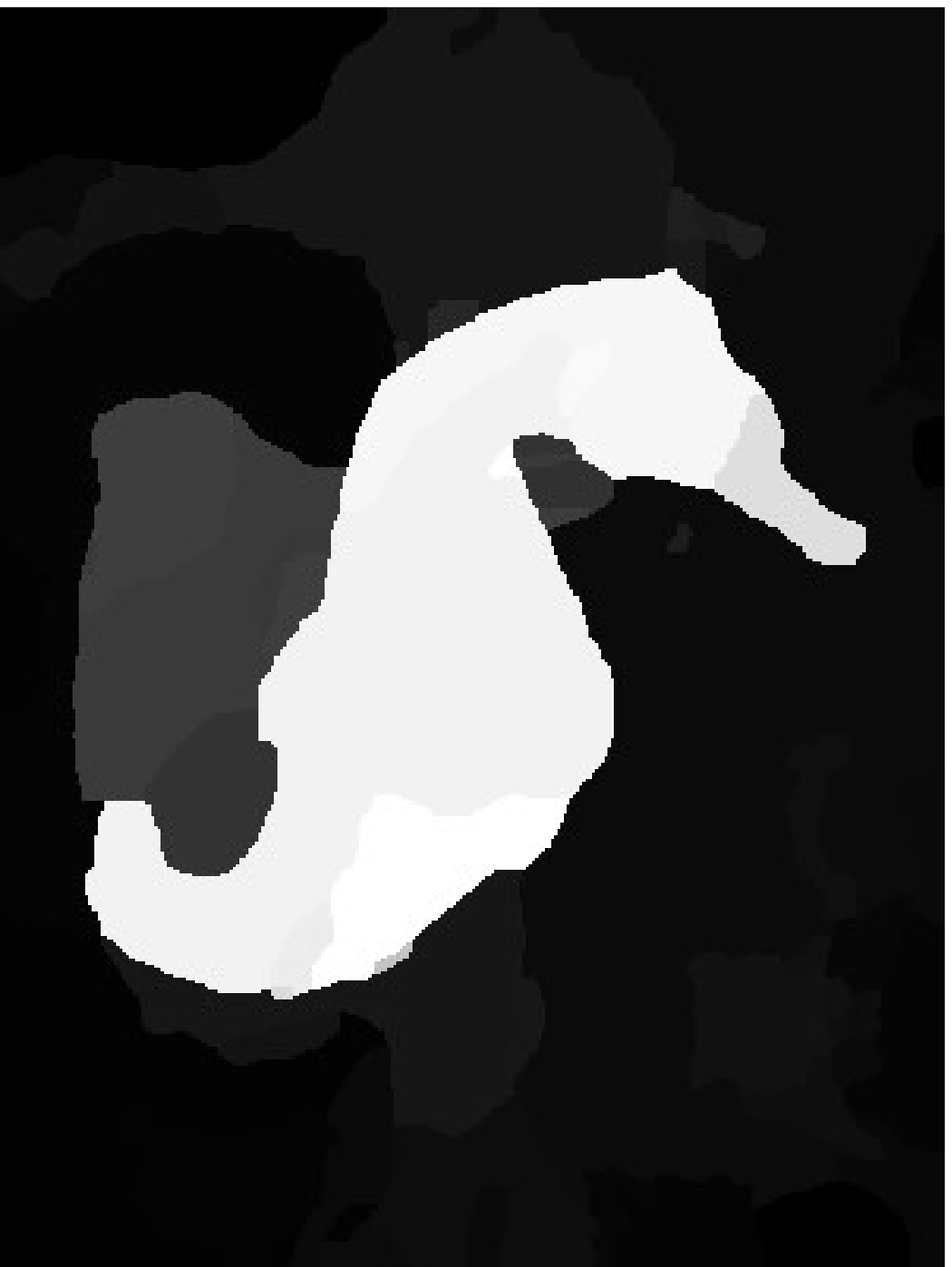} &
\includegraphics[height=1.6cm]{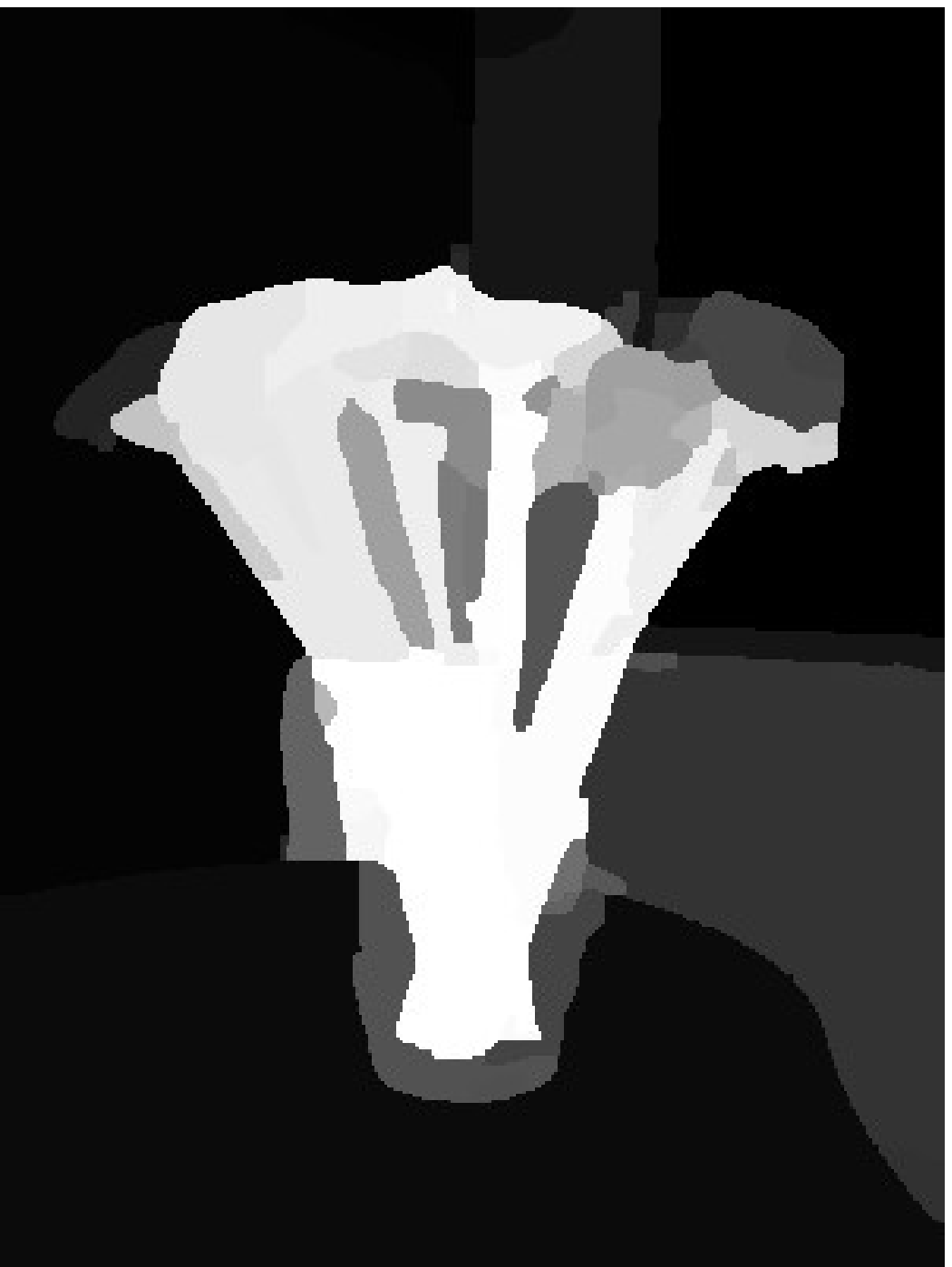} &
\includegraphics[height=1.6cm]{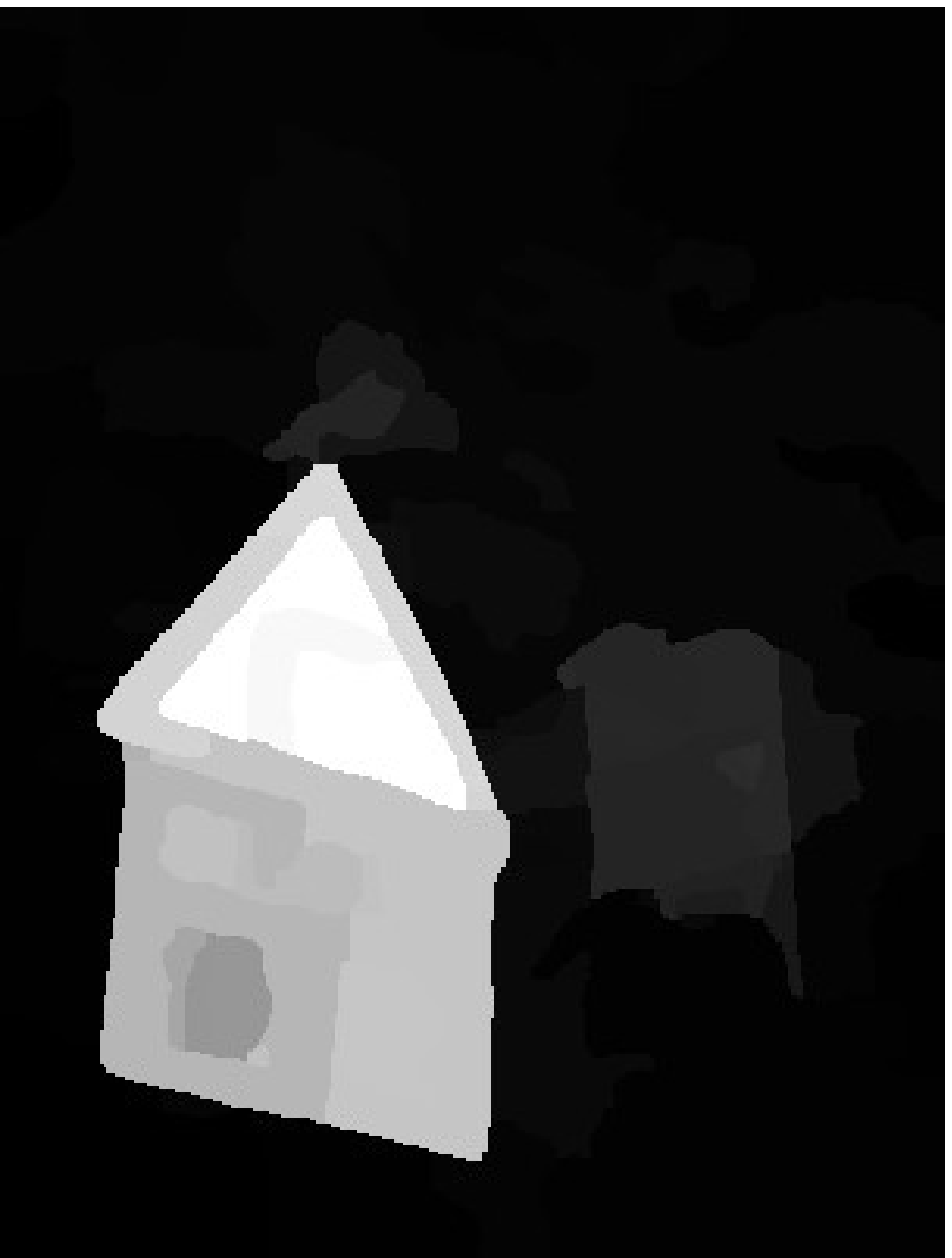} &
\includegraphics[height=1.6cm]{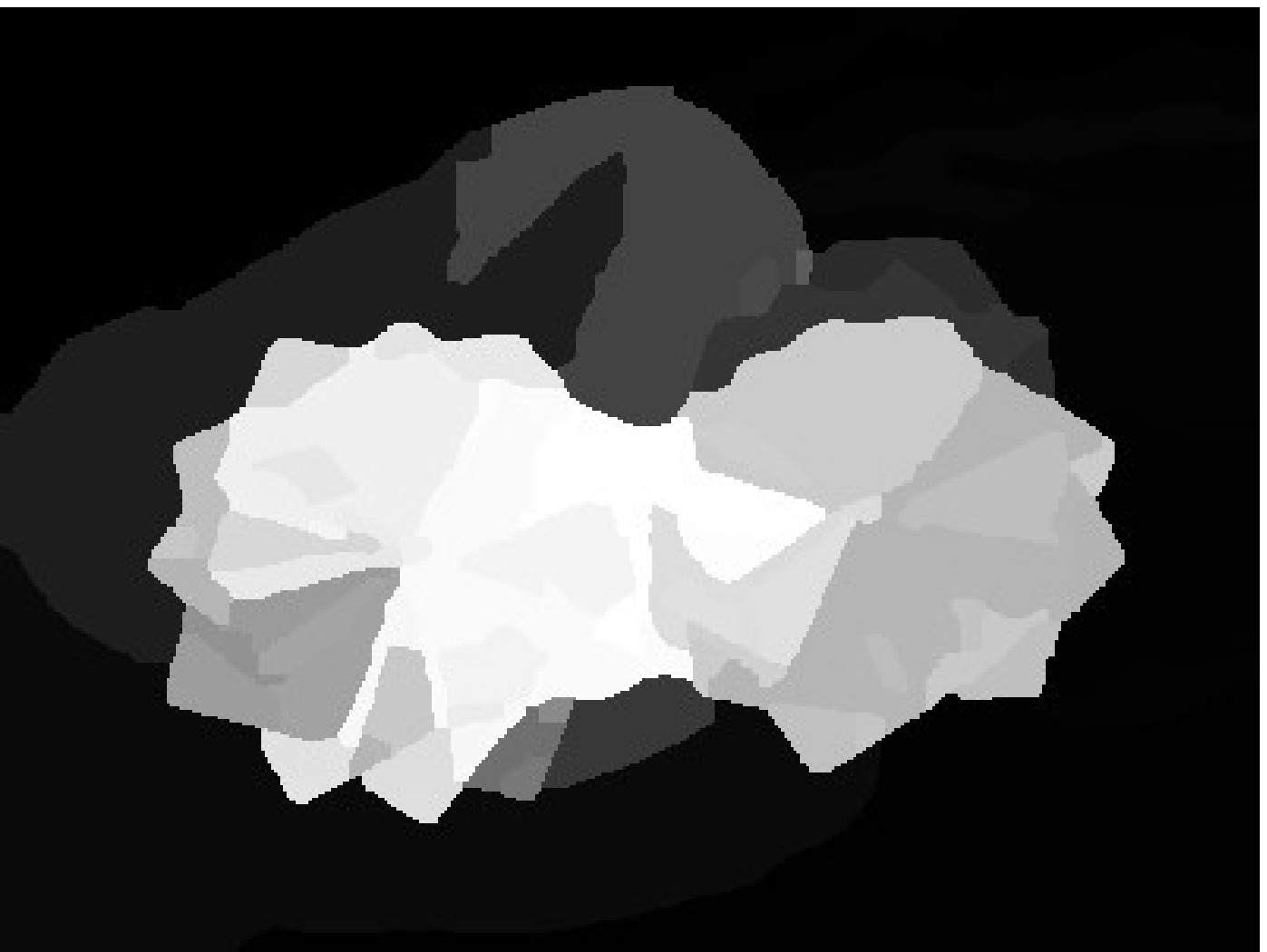} &
\includegraphics[height=1.6cm]{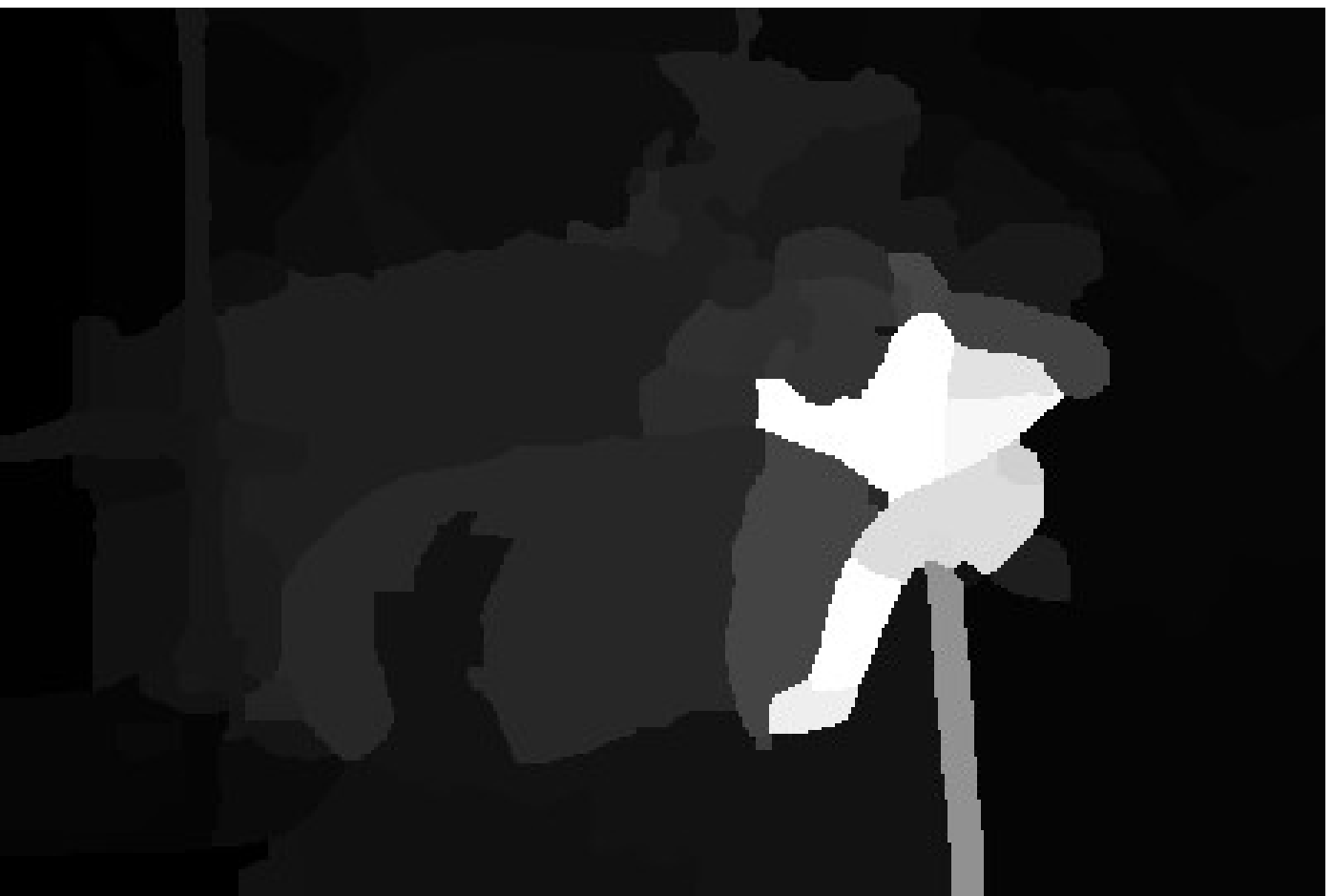} \\

GT&
\includegraphics[height=1.6cm]{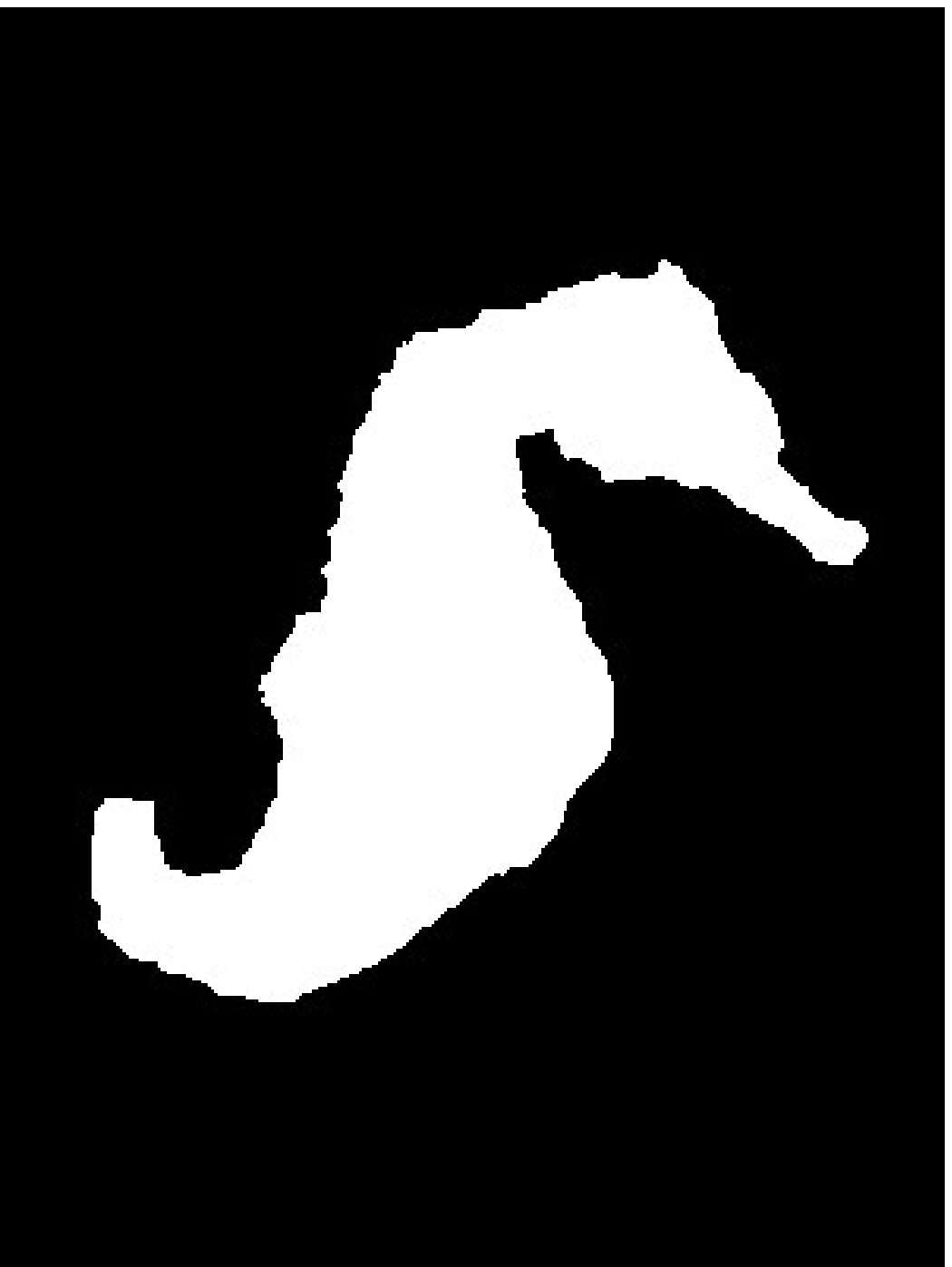} &
\includegraphics[height=1.6cm]{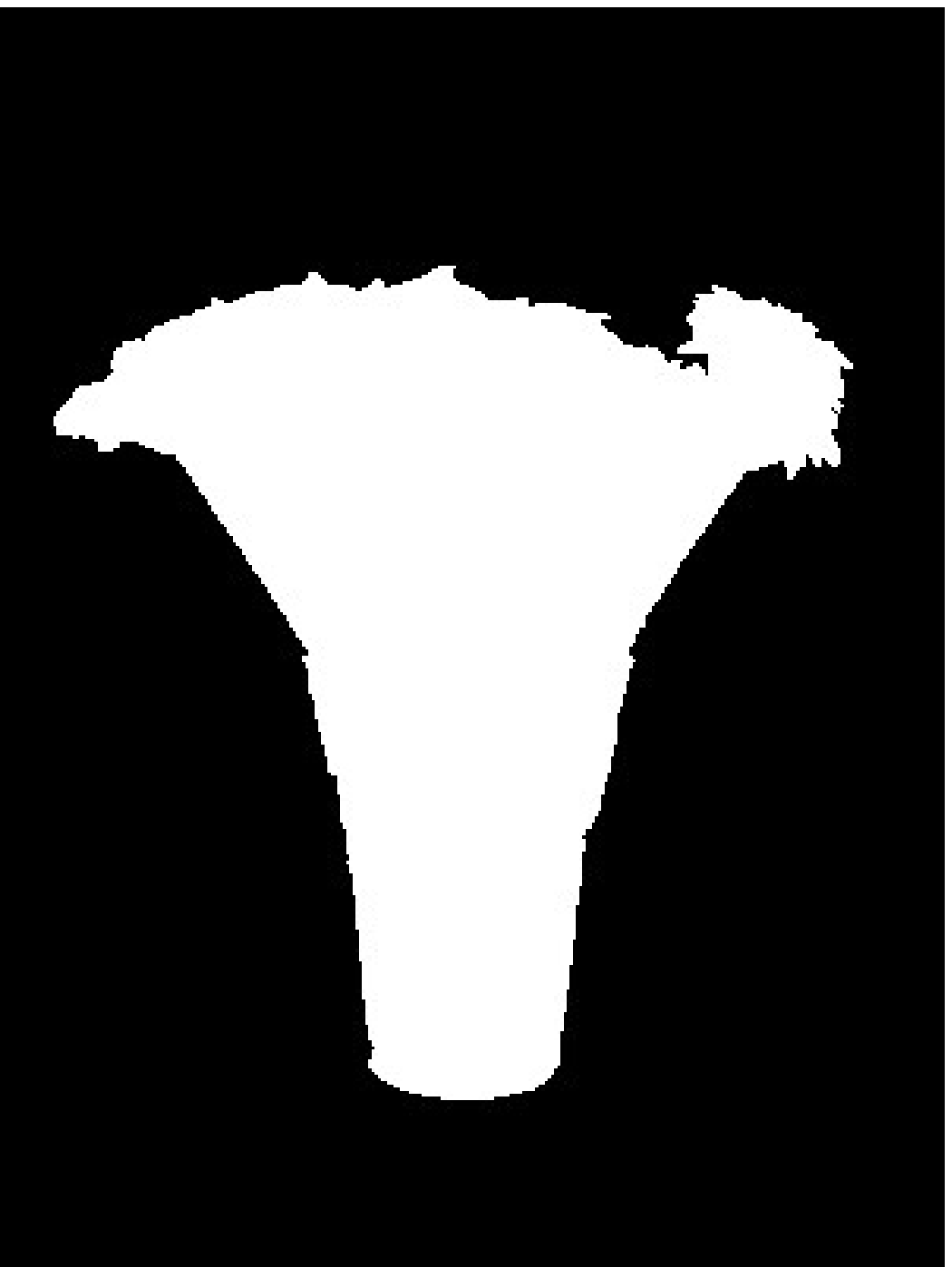} &
\includegraphics[height=1.6cm]{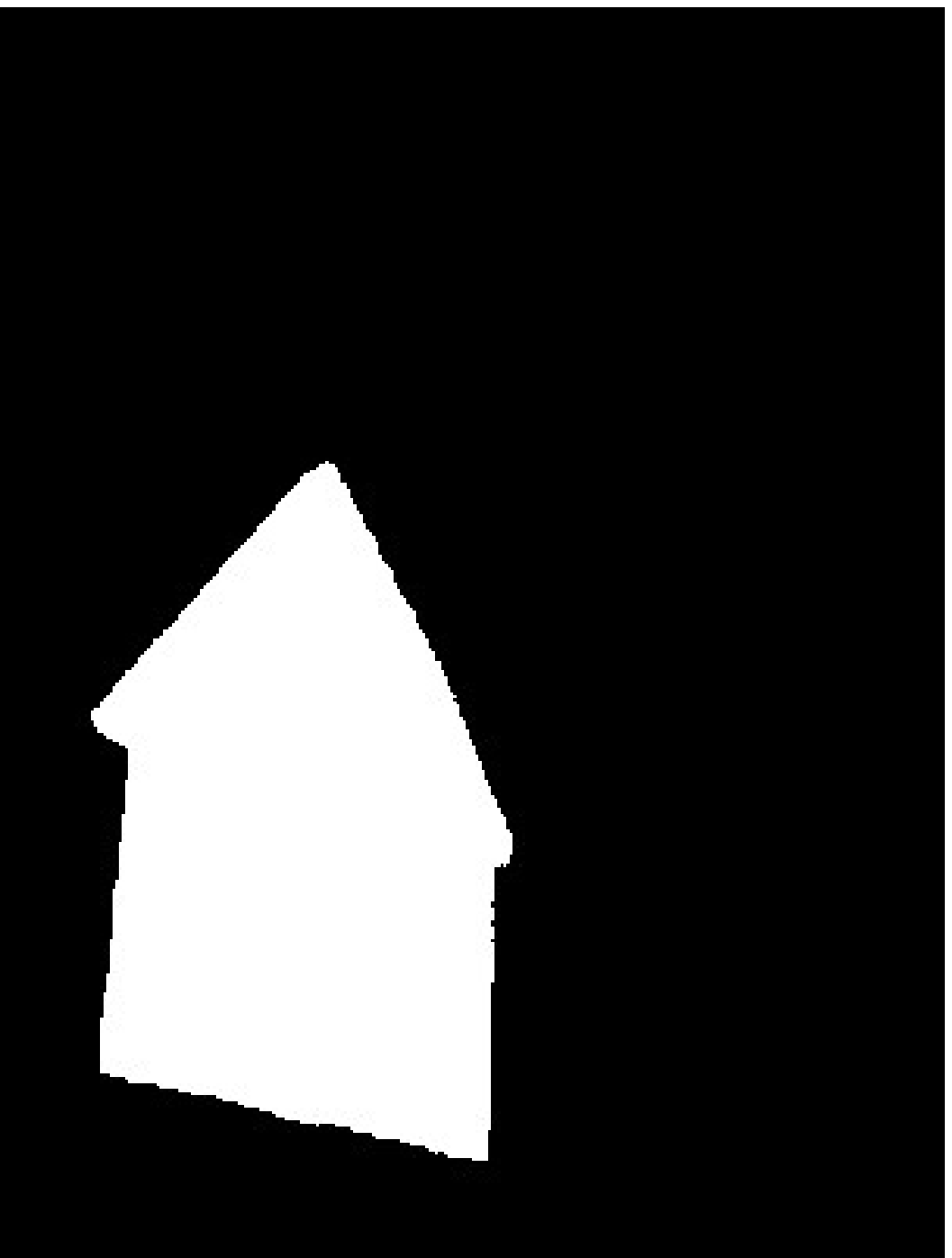} &
\includegraphics[height=1.6cm]{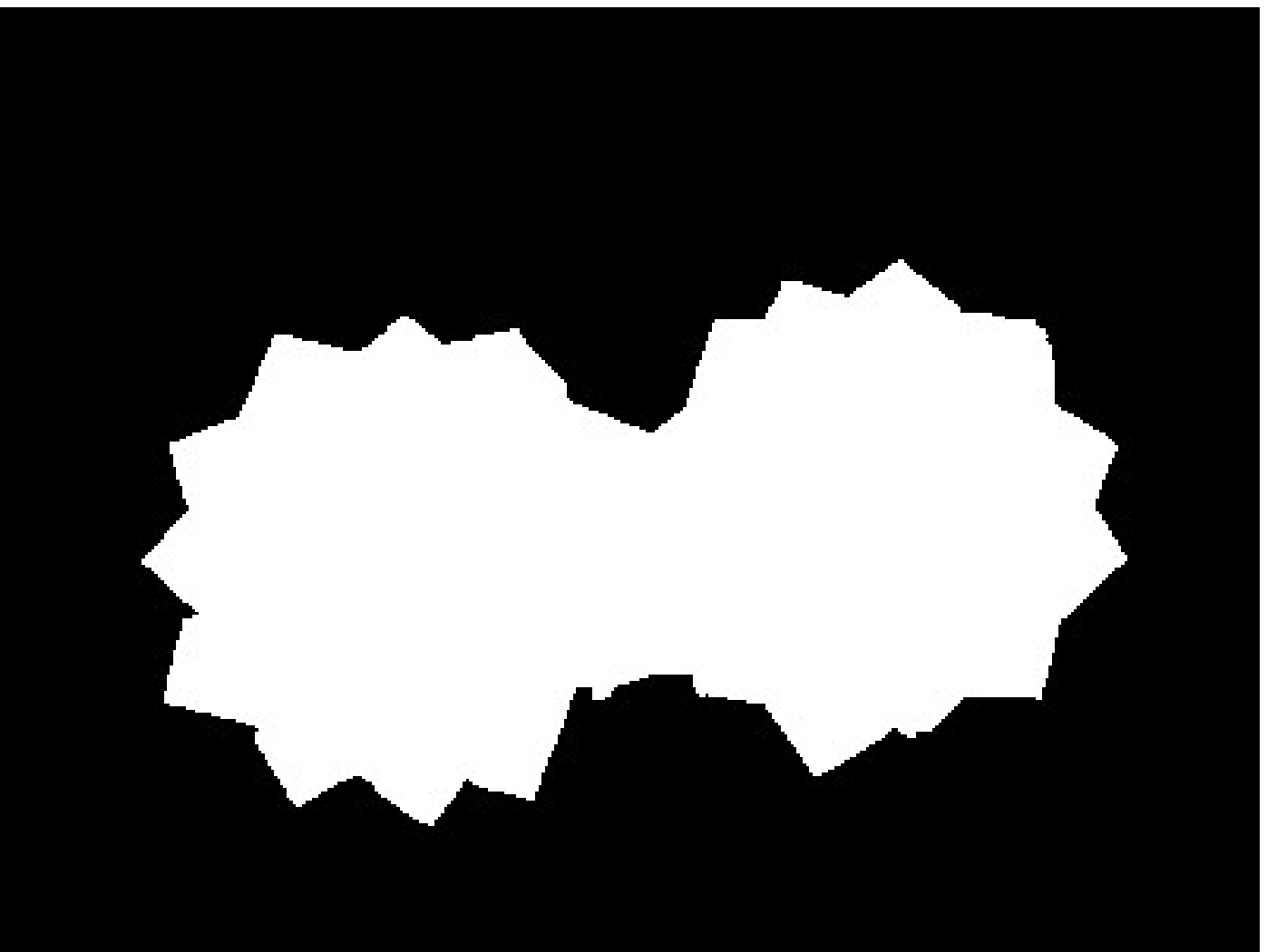} &
\includegraphics[height=1.6cm]{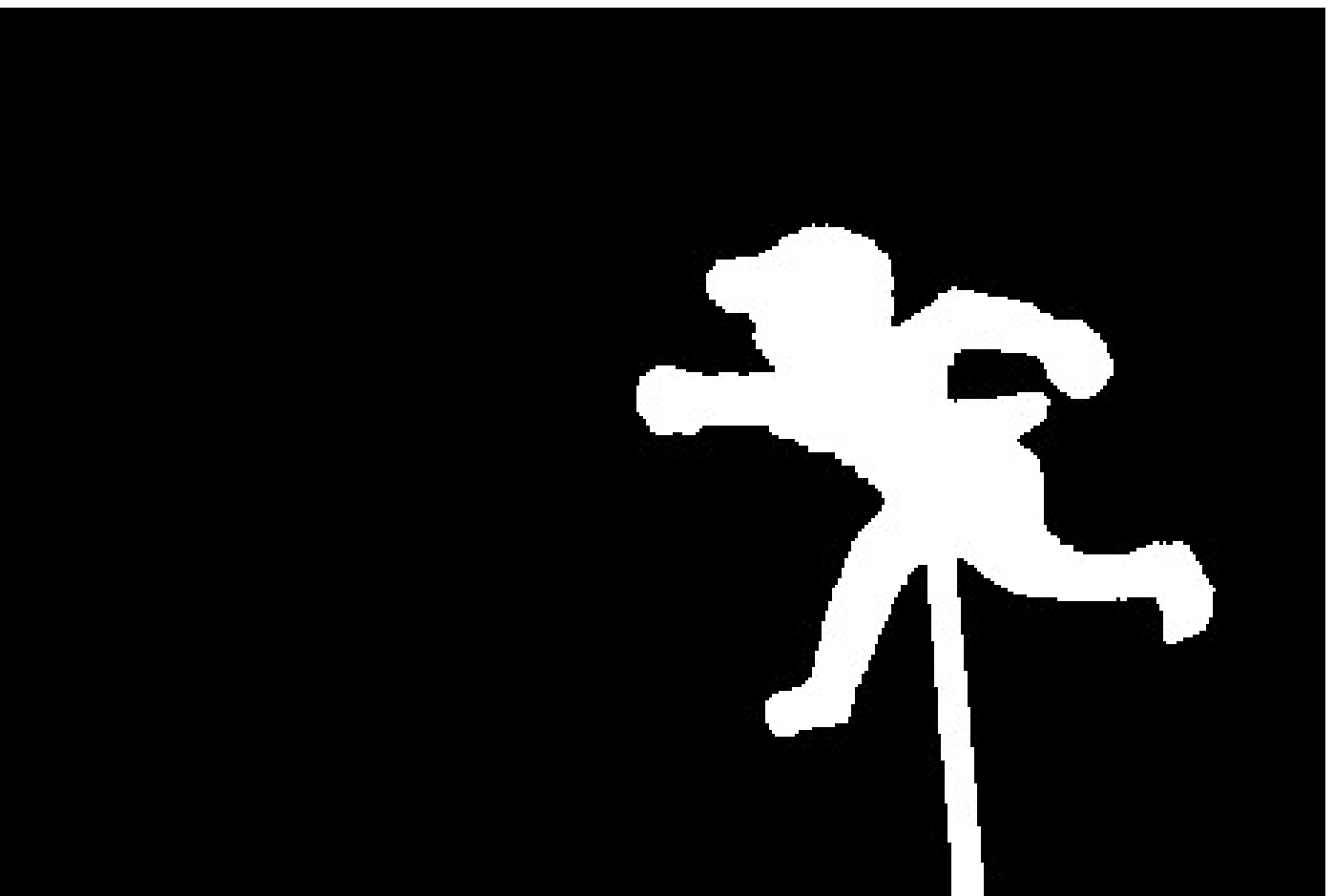} \\

&(a) & (b) & (c) & (d) & (e) \\
\end{tabular}
\caption{Experimental results on some ECSSD images. First row: IM original images; saliency maps generated using FT \cite{Achanta-09cvpr},  RC \cite{Cheng-11cvpr}, SF \cite{perazzi-cvpr12}, PCAS \cite{margolin-cvpr13}, HS \cite{Yan-cvpr13}, ST \cite{ZLiu-tip14}, our models HP and SOH,  and GT ground truth masks.} \label{fig:ECSSDimage_comp}
\end{figure}

For visual comparison, saliency maps generated by all models on MSRA and ECSSD datasets are shown in Figures \ref{fig:MSRAimage_comp} and \ref{fig:ECSSDimage_comp}, respectively.

Images selected from the MSRA set contain shadows and complex backgrounds. Again, hierarchy-based methods tend to detect complete objects, but in some cases fail to eliminate background areas (such as ST in  (b), HP in (c) or SOH in (d)). In the first example (column (a)) only SOH and ST find the complete car while the other methods highlight only the homogeneous yellow part. The last column (d) shows an example where only SOH, ST and HP manage to highlight the transparent glass but detect also fingers or part of the clouds in the background. The other methods fail to capture the complete object.

Figure~\ref{fig:ECSSDimage_comp} presents some difficult examples from the ECSSD set. We can observe images where our first method HP wrongly highlights part of the background (a,d,e) while our second method SOH produces accurate maps, extracting complete objects  and removing all the background regions. Fig.~\ref{fig:ECSSDimage_comp}(e) illustrates a case where all the models fail to detect the head of the figure.

The results of the proposed models are available at our  web page

  \url{https://imatge.upc.edu/web/resources/saliency-maps-image-hierarchies}.

Our un-optimized implementations use Matlab code from \cite{Arbelaez-pami11} to generate UCMs and \cite{UGM} for hierarchical inference. The experiments are performed on a laptop with Intel Core i7-2620M 2.7GHz CPU and 8 GB RAM. The most time-consuming task is the generation of UCM and BPT hierarchies. Once the hierarchies are created, the average run-time per image (ASD set) is 3.82s for the model based on hierarchies of partitions (6 levels, histogram model, hierarchical inference) and 3.15s for the SOH model.

\section{Conclusion}
\label{sec:conclusions}

We have presented two saliency  models for salient object segmentation based on hierarchical image segmentations.

The first model, namely Hierarchy of Partitions (HP), is a general framework for creating saliency maps by integrating the results of maps generated on a hierarchy of image partitions. Our method generalizes previous approaches based on hierarchies, working with different hierarchical segmentation techniques (BPT and gPb-UCM), number of levels, region models and fusion strategies. Experiments performed with different integration criteria have shown that despite its simplicity the $mean$ fusion method achieves the same performance as more elaborated approaches based on hierarchical inference.

The second model, Saliency Over the Hierarchy (SOH), works directly on the structure created by the segmentation algorithm. It computes saliency at each node and integrates this information in a straightforward manner into a single saliency map. 

The two models analyze saliency at several levels of detail in the hierarchy. An object that is different from its background is salient at the scale at which it is represented (completely or almost completely) with a single region in the partition (HP) or with a node in the hierarchy (SOH). The integration of the different cues allows the detection of salient objects of different sizes, with accurate boundaries.

We have shown that the two proposed models produce high quality saliency maps. The  Saliency Over the Hierarchy (SOH) outperforms the HP method in terms of precision, recall, weighted f-measure and mean absolute error, and achieves state-of-the-art performance on several benchmark datasets, being still simple and efficient.

\section*{Acknowledgement}
This work has been developed in the framework of the project BIGGRAPH-TEC2013-43935-R, financed by the Spanish Ministerio de Economía y Competitividad and the European Regional Development Fund (ERDF).

\section*{References}

\bibliography{saliency}

\end{document}